# Automatic Structure Discovery for Large Source Code

By Sarge Rogatch
Master Thesis

Universiteit van Amsterdam,
Artificial Intelligence, 2010

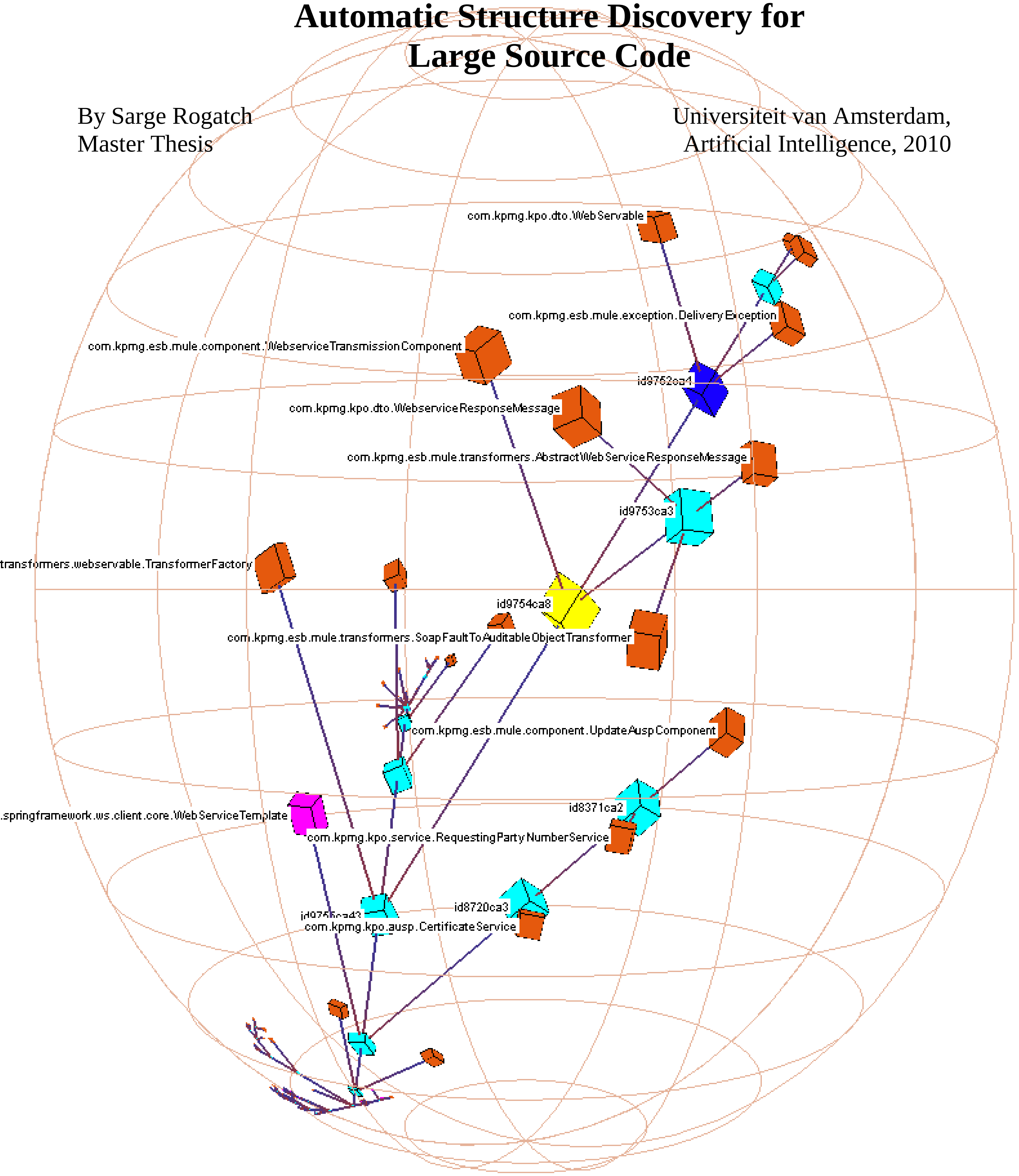

# Acknowledgements

I would like to acknowledge the researchers and developers who are not even aware of this project, but their findings have played very significant role:

- Soot developers: Raja Vall´ee-Rai, Phong Co, Etienne Gagnon, Laurie Hendren, Patrick Lam, and others.
- TreeViz developer: Werner Randelshofer
- H3 Layout author and H3Viewer developer: Tamara Munzner
- Researchers of static call graph construction: Ondˇrej Lhot´ak, Vijay Sundaresan, David Bacon, Peter Sweeney
- Researchers of Reverse Architecting: Heidar Pirzadeh, Abdelwahab Hamou-Lhadj, Timothy Lethbridge, Luay Alawneh
- Researchers of Min Cut related problems: Dan Gusfield, Andrew Goldberg, Maxim Babenko, Boris Cherkassky, Kostas Tsioutsiouliklis, Gary Flake, Robert Tarjan

# Contents

# 1 Abstract

In this work we attempt to infer software architecture from source code automatically. We have studied and used **unsupervised learning** methods for this, namely clustering. The state of the art source code (structure) analysis methods and tools were explored, and the ongoing research in software reverse architecting was studied. Graph clustering based on minimum cut trees is a recent algorithm which satisfies strong theoretical criteria and performs well in practice, in terms of both speed and accuracy. Its successful applications in the domain of Web and citation graphs were reported. To our knowledge, however, there has been no application of this algorithm to the domain of reverse architecting. Moreover, most of existing software artifact clustering research addresses legacy systems in procedural languages or C++, while we aim at modern **object-oriented** languages and the implied character of relations between software engineering artifacts. We consider the research direction important because this clustering method allows substantially larger tasks to be solved, which particularly means that we can cluster software engineering artifacts at class-level granularity while earlier approaches were only able to do clustering at package-level on real-world software projects. Given the target domain and the supposed way of usage, a number of aspects must be researched, and these are the main contributions of our work:

- extraction of software engineering artifacts and relations among them (using state of the art tools), and presentation of this information as a graph suitable for clustering
- edge weight normalization: we have developed a directed-to-undirected graph normalization, which is specific to the domain and alleviates the widely-known and essential problem of **utility** artifacts
- parameter (alpha) search strategy for hierarchical clustering and the algorithm for merging the partitions into the hierarchy in arbitrary order
- distributed version for cloud computing
- a solution for an important issue in the clustering results, namely, too many sibling clusters due to almost acyclic graph of relations between them, which is usually the case in the source code domain;
- an algorithm for computing package/namespace **ubiquity** metric, which is based on the statistics of merge operations that occur in the cluster tree

A prototype incorporating the above points has been implemented within this work. Experiments were performed on real-world software projects. The computed clustering hierarchies were visualized using state of the art tools, and a number of statistical metrics over the results was calculated. We have also analyzed the encountered remaining issues and provided the promising further work directions. It is not possible to infer similar architectural insights with any existing approach; an account is given in this paper. We conclude that our integrated approach is applicable to large software projects in object-oriented languages and produces meaningful information about source code structure.

# 2 Introduction

As the size of software systems increases, the algorithms and data structures of the computation no longer constitute the major design problems. When systems are constructed from many components, the organization of the overall system—the **software architecture** —presents a new set of design problems. This level of design has been addressed in a number of ways including informal diagrams and descriptive terms, module interconnection languages, templates and frameworks for systems that serve the needs of specific domains, and formal models of component integration mechanisms [Gar1993]. The software architecture of a program or computing system is the structure or structures of the system, which comprise software components, the externally visible properties of those components, and the relationships between them. The term also refers to documentation of a system's software architecture. Documenting software architecture facilitates communication between stakeholders, documents early decisions about high-level design, and allows reuse of design components and patterns between projects [Bass2003].

Software architecture determines the **quality attributes** exhibited by the system such as fault-tolerance, backward compatibility, extensibility, flexibility, reliability, maintainability, availability, security, usability, and other –ities. When performing Software **quality analysis**, we can split the features upon analysis into two principal categories:

- apparent: how the software behaves and looks
- latent: what is the potential of the software, what is in its source code and documentation

This is illustrated in Figure 2-1 below:

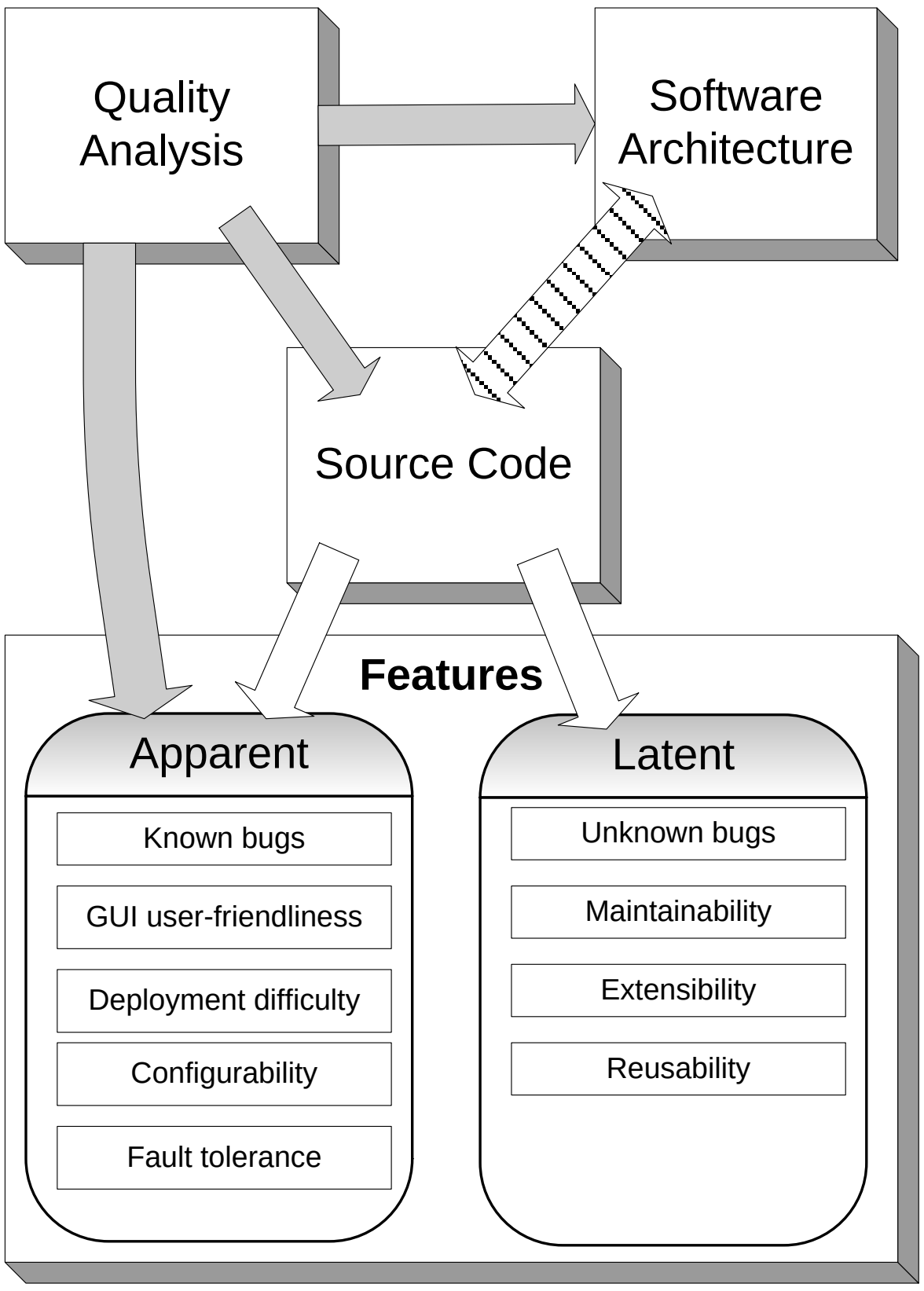

**Figure 2-1 Quality analysis options**

We can analyze the apparent features directly. But in order to analyze the latent features, we need to analyze the source code. The latter is effort-intensive if performed manually. Software architecture is a high-level view of the source code, describing the important facts and omitting the details.

- In the ideal case, the software architecture is available (in a document) and reflects the source code precisely. Then quality analysis performed only on the software architecture will give a good coverage of the latent features (perhaps, except minor unknown bugs).
- In the worst case, only the source code is available for the software, with no documentation at all, i.e. the architecture is not known. Then we can either descend to manual source code analysis, or… try to infer the software architecture from the source code automatically!
- Usually, software is not well documented: the software architecture is either too loosely described in the documentation, or only available for some parts of the software, or becomes out-of-sync with the actual source code. In this case, we can utilize the available fragments for semi-supervised inference of the software architecture from the source-code (data) and the documentation (labels).

The dashed bidirectional arrow on the picture above denotes that:

- the actual software architecture (how the source code is written) can become inconsistent with the claimed software architecture (how it is designed in the documentation). Development in a rush, time pressure, quick wins, hacks and workarounds are some of the reasons why it usually happens so;
- even when there is no explicit software architecture (no documentation), there is some implicit software architecture which is in the source code (the actual architecture).

**Software maintenance** and evolution is an essential part of the software life cycle. In an ideal situation, one relies on system documentation to make any change to the system that preserves system's reliability and other quality attributes [Pir2009]. However it has been shown in practice that **documentation** associated with many existing systems is often *incomplete, inconsistent, or even inexistent* [Let2003], which makes software maintenance a tedious and human-intensive task. This is further complicated by the fact that key developers, knowledgeable of the system's design, commonly move to new projects or companies, taking with them valuable technical and domain knowledge about the system [ACDC2000].

The objective of design and architecture recovery techniques is to recover high-level design views of the system such as its architecture or any other high-level design models from low-level system artifacts such as the source code. Software engineers can use these models to gain an overall understanding of the system that would help them accomplish effectively the maintenance task assigned to them [Pir2009].

The most dominating research area in architecture reconstruction is the inference of the structural decomposition. At the lower level, one groups global declarations such as variables, routines, types, and classes into modules. At the higher level, modules are clustered into subsystems. In the result there are flat or hierarchical modules. Hierarchical modules are often called subsystems. While earlier research focused on flat modules for procedural systems, newer research addresses hierarchical modules [Kosc2009].

## 2.1 Project Summary

All but trivial changes in software systems require a global understanding of the system to be changed. Such non-trivial tasks include migrations, auditing, application integration, or impact analysis. A global understanding cannot be achieved by looking at every single statement. The source code provides a huge amount of details in which we cannot see forest for the trees. Instead, to understand large systems, we need a more coarse-grained map - software architecture. Software architecture reconstruction is the form of reverse engineering in which architectural information is reconstructed for an existing system. [Kosc2009]

Many companies have huge repositories of source code, often in different programming languages. Automatic source code analysis tools also produce a lot of data with issues, metrics, and dependencies in the source code. This data has to be processed and visualized in order to give insightful information to IT quality experts, developers, users and managers.

There are a number source code visualization methods and tools that address this problem with different levels of success. In this project we plan to apply the Artificial Inteligence techniques to the problem of source code visualization. Applications of AI (Cluster Analysis) to collaboration, word association and protein interaction analysis [Pal2005]; social network and WWW analysis [Fla2004], where also lots of data must be processed, are well known and produce fruitful results. We hope in this project to identify similar opportunities in the software visualization domain. We further realize that our task is best characterized as **reverse architecting**, a term appearing in the literature [Riv2000]: reverse architecting is a flavour of reverse engineering that concerns with the extraction of software architecture models from the system implementation.

The known Artificial Intelligence algorithms, such as clustering of graphs, either optimize specific statistical criteria, or exploit the underlying structure or other known characteristics of the data. In our case, the data is extracted from the source code of software. The **vertices** of the graph upon analysis are software engineering artifacts, where the artifacts can be of different granularity: from instructions/operators to methods/fields and then to classes, modules, packages and libraries. The **edges** of our graph are dependencies between the artifacts, which also have different granularities in their turn: from edges of the control flow graph, to edges of the method call and field access graphs, and then to edges of the class coupling graph, the package usage and library dependency graphs.

Within the scope of this project we view the following stages:

1 **Extract the SE artifacts** and their relations, such as function/method call and field access graphs, inheritance/subtyping relations and metrics, which is a matter of pre-requisite tools. Though some uncertain decision making is needed even at this stage (e.g. polymorphism handling within static call graph extraction), we take the *state of the art* methods and do not focus on their improvement, however we try to use the best of available pre-requisites and use several of them in case they are non-dominated, i.e. none of them is better in all the aspects.

2 Devise an **automatic grouping** of the extracted artifacts in order to achieve meaningful visualizations of them. *We focus on this.*

3 Visualize the results and analyze source code quality taking into account the results of clustering. These tasks are also hard - the former involves automatic graph layout and the latter involves uncertain decision making - and thus left to the *state of the art* tools or human experts.

We implement a prototype called **InSoAr**, abbreviated from “Infer Software Architecture”. As different viewpoints specify what information should be reconstructed (in our particular case of automatic reconstruction, inferred by our program) and help to structure the reconstruction process [Kosc2009], we disambiguate the meaning in which we use “(reversed) software architecture” in the context of the goal we pursue and the major facts our program infers to **nested software decomposition**. This term is adopted in the existing works on reverse architecting ([END2004], [UpMJ2007], [Andre2007]). We decompose a set of software engineering artifacts (e.g. Java classes) according to the coupling between SE artifacts. We assume that nested software decomposition, in which artifacts serving similar purpose or acting in a composite mechanism are grouped together, is the most insightful and desirable for software engineers. This is confirmed in [Kosc2009] (see section 2 of the thesis), and we discuss further in the thesis the works that attempt to create nested software decompositions ([Ser2008], [Maqb2007], [Rays2000], [Pate2009]).

## 2.2 Global Context

Existing Software Visualization tools extract various metrics about the source code, like number of lines, comments, complexity and object-oriented design metrics as well as dependencies in the source code like call graphs, inheritance/subtyping and other relations first. As the next step they visualize the extracted data and present it to the user in an interactive manner, by allowing zooming and drill-down or expand/collapse. Examples of these tools are STAN [Stan2009], SQuAVisiT [Rou2007], DA4Java [Pin2008] and Rascal [Kli2009, also personal communication with Paul Klint].

A common problem of such tools is that there are too many SE artifacts to look at everything at once. According to [Stan2009]: "*To get useful graphs, we have to carefully select the perspectives and scopes. Otherwise we'll end up with very big and clumsy graphs. For example, we may want to look at an artifact to see how its contained artifacts interact or how the artifact itself interacts with the rest of the application*". The DA4Java tool [Pin2008], attempts to solve this problem by allowing the user to add or remove the artifacts the user wants to see in the visualization, and, also, to drill down/up from containing artifacts to the contained artifacts (e.g. from packages to classes).

We want to solve the problem of the overwhelming number of artifacts by grouping them using AI techniques such as clustering, learning and classification, so that a reasonable number of groups is presented to the user.

From the available AI techniques graph clustering is known to be applied in the software visualization domain. It seems that many clusterizers of software artifacts are employing non-MinCut based techniques, refer to [Maqb2007] for a broad review of the existing clusterizers and [Ser2008] for a recent particular clusterizer. There is some grounding for this, as according to [Ding2001] a MinCut-based clustering algorithm tends to produce skewed cuts. In the other words: a very small subgraph is cut away in each cut. However, this might not be a problem for graphs from the domain of source code analysis. Another motivation for applying MinCut-based clustering algorithms in our domain arises due to the fact that software normally has clearly-defined entry points (sources, in terms of MaxFlow-like algorithms) and less clearly-defined exit points (sinks, in terms of MaxFlow). A good choice of sink points is also a matter of research, while the current candidates in mind are: library functions, dead-end functions (which do not call any others), and the program runtime termination points.

For extraction of Software Engineering artifacts we plan to use the existing tools such as Soot (Sable group of McGill University) [Soot1999] and Rascal (CWI) [Kli2009]. Soot builds a complete model of the program upon analysis, either from the source code or from the compiled Java byte-code. The main value of this tool for our project is that it implements some heuristics for static call graph extraction.

## 2.3 Relevance for Artificial Intelligence

Many AI methods require parameters, and the performance of the methods depends on the choice of parameter values. The best choice of methods or parameter values is almost always domain-specific, usually problem-specific and even sometimes data-specific. We want to investigate these peculiarities for the domain of source code visualization and reverse engineering.

Automatic clustering algorithms have a rich history in the artificial intelligence literature, and in not so recent years have been applied to understanding programs written in procedural languages [Man1998]. The purpose of an automatic clustering algorithm in artificial intelligence is to group together similar entities. Automatic clustering algorithms are used within the context of program understanding to discover the structure (architecture) of the program under study [Rays2000]. One example of the specifics of software clustering is that we want to cluster entities based on their **unity of purpose** rather than their *unity of form*. It is not useful to cluster all four-letter variables together, even though they are similar [Rays2000]. In this project we attempt to regard relations that expose the unity of purpose, and we use the notion of similarity in this meaning.

One of long-term goals of Artificial Intelligence is creation of a self-improving program. Perhaps, this can be approached by implementing a program that does reverse engineering and then forward engineering of its own source code. In between there must be high-level understanding of the program, its architecture. It is not clear, what understanding is, but seems it has much in common with the ability to visualize, explain to others and predict behavior. This project is a small step towards automatic comprehension of software by software.

## 2.4 Problem Analysis

The particular problem of interest is inference of the architecture and different facts about software from its source code. It is desired that the high-level view on software source code is provided to human experts automatically, omitting the details that do not need human attention at this stage.

Software products often lack documentation on the design of their source code, or software architecture. Although full-fledged documentation can only be created by human designers, an automatic inference tool can also provide some high-level overview of source code by means of grouping, generalization and abstraction. Such a tool could also point the places where human experts should pay attention to. Semi-automatic inference can be used when documentation is partially available.

The key step that we make in this project is graph clustering, which splits the source code into parts, i.e. performs **grouping**. We suppose that this will help with the generalization over software artifacts and the detection of layers of abstraction within the source code.

By **generalization** we mean the detection of the common purpose which software artifacts in a group serve. One way to determine the purpose is by exploiting the linguistic knowledge found in the source code, such as identifier names and comments. This was done in [Kuhn2007], however they did not partition software engineering artifacts into structurally coupled groups prior to linguistic information retrieval. We believe that formal relations (e.g. function calls or variable accesses) should be taken into account first, and then the linguistic relations should be analyzed within the identified (e.g. by means of call graph clustering) groups, rather than doing vocabulary analysis across the whole source code.

By **abstraction** we mean the identification of abstraction layers within the source code. For example, if all indirect (i.e. mediated) calls from group A and B to groups C and D go through group E, and there are no direct calls from {A, B} to {C, D}, then it is likely that group E serves as a layer of abstraction between groups {A, B} and {C,D}.

The aforementioned decisions need *uncertain inference* and *error/noise tolerance*. Thus we think that the problem should be approached with **AI** techniques.

## 2.5 Hypotheses

In the beginning of this project we had hypotheses as listed below.

1) By applying Cluster Analysis to a graph of SE artifacts and their dependencies, we can save human efforts on some common tasks within Software Structure Analysis, namely identification of coupled artifacts and breaking down the system's complexity.

2) Partitional clustering will provide better nested software decompositions than the widely used (and to our knowledge, the only for clustering large number of SE artifacts) hierarchical clustering, which is in fact a greedy algorithm
3) Semi-supervised learning of software architecture from source code (unlabelled data) and architecture design (labeled data) can be used to improve results over usual clustering, which is unsupervised learning
   A few comments for this hypothesis:
   - Here we assume that the explicit architecture, i.e. the design documentation created by human experts, provides some partitioning of SE artifacts into groups. By means of Cluster Analysis we try to infer the implicit architecture, which is the architecture of how the source code is actually written, and we also get some partitioning of SE artifacts into groups.
   - It is obvious that this task can also be viewed as classification: for each SE artifact the learning algorithm outputs the architectural component (or, in terms of Machine Learning, the **class**, but do not confuse with SE classes) which the artifact belongs to, and perhaps also the certainty of this decision.
   - When the explicit architecture is only partially available (which is always the case, except for completely documented trivial software, where ‘completely’ stands for ‘each SE artifact’), we can think of several approaches for classifier training:
     i. Train the classifier on the documented part of this software
     ii. Train the classifier on other similar software which is better documented
     iii. Train the classifier on library functions, which are usually documented best of all (e.g. Java & Sun libraries).
4) Improvements in call graph clustering results can be achieved through integration with some of the following: class inheritance hierarchy, shared data graph, control flow graph, package/namespace structure, identifier names (text), supervision (documented architecture pieces), etc.

Evidence for the first hypothesis is provided mostly in section 7.2. The second hypothesis is discussed theoretically, mostly in sections 3.2.2.2 and 3.2.1.3.

We have only discussed hypothesis 3, as implementation and experiments would take too much time. In the resulting software decompositions we can see that library SE artifacts indeed give insight about the purpose of client-code artifacts appearing nearby. In section 5.2 we provide evidence and argue theoretically in support of hypothesis 3: proper weighting of different kinds of relations can be learnt on training data (source code with known nested decomposition) and then applied to novel software.

For hypothesis 4, empirical results show that indeed the resulting hierarchical structure looks better when multiple kinds of relations are given on input of the clustering algorithm, and the reasons are theoretically discussed in section 5.2.

## 2.6 Business Applications

Consider a company that is proposed to do maintenance for a software product. Having a visualization tool, the company can analyze the quality of the source code, so the company knows the risks associated with the software and can estimate the difficulty and expensiveness of its maintenance more accurately. To be able to do this we need to:

1 extract the architecture (even when the architecture was not designed from the very beginning, there is always the actual implicit architecture, which is how the source code actually written);
2 and in some way derive the evaluation of the source code from the result of step 1.

Both steps are problematic in the sense that they require uncertain inference and decision making, which is a task of Artificial Intelligence in case we want to do this automatically. Within the scope of this project we focus on step 1 and leave step 2 to a human expert.

In this section we consider the value of this project for potential target groups, and then for particular stakeholders of a project: administrators, managers, developers, testers and users. But first of all, below is the grounding of why reverse architecting software is valuable. According to [Riv2000]:

- Software Development domain is characterized by fast changing requirements. Developers are forced to evolve the systems very quickly. For this reason, the documentation about the internal architecture becomes rapidly obsolete. To make fast changes to the software system, developers need a clear understanding of the underlying architecture of the products.
- To reduce costs several products share software components developed in different projects. This generates many dependencies that are often unclear and hard to manage. Developers need to analyze such dependency information either when reusing the components or when testing the products. They often look for "the big picture" of the system that is a clear view of the major system components and their interactions. They also need a clear and updated description of the interfaces (and services) provided by the components.
- The developers of a particular component need to know its clients in order to analyze the impact of changes on the users. They also want to be able to inform their customers of the changes and to discuss with them the future evolution of the component.
- When defining a common architecture for the product family, architects have to identify the commonalties and variable parts of the products. This requires comparing the architectures of several products.
- The developers need a **quick** method for extracting the architecture models of the system and to analyze them. The teams often use an iterative development process that is characterized by frequent releases. The architecture models could be a valuable input for reviewing the system at the end of each iteration. This increases the need for an automatic approach to extract the architecture model.

In our view, if a visualization tool is developed, one that can reverse-architect a software from source code and provide concise accurate high-level information to its users, the effect of employing this tool will be comparable to the effect of moving from assembly language to C in the past.

Below are the benefits that a tool allowing to reverse-engineer and visualize the software architecture from the source code provides to different stakeholders. It is likely that the list is far not complete, and that it will be extended, improved and detailed in the process of development of the tool, as new facts become apparent.

| | |
|---|---|
| Administrators | Reduced expenses<br>• The development team is more productive.<br>Reduced risks<br>• Consider a company that is proposed to do maintenance for a software product. This company has a tool to analyze the quality of the software more precisely, so the company knows the risks associated with the software and can estimate the difficulty of its maintenance more accurately.<br>Increased product lifetime and safety factor<br>• The curse of complexity is alleviated.<br>Better decisions |
| | • The company can estimate the quality of software products more precisely, thus knows the situation better, and this leads to better decisions. |
| Managers | Better control of task progress<br>• Functionality + Quality = Efforts + Duration<br>Now there is a better way to check whether a task was performed in fewer efforts by means of reducing the quality.<br>New team members are trained faster<br>• Usually, developers which are new to the project spend much time studying the existing system, especially when little documentation is available.<br>Fewer obstacles to "introduce new resource" action:<br>• When while checking an ongoing project a manager determines that the project is likely not to fit the deadline, and the deadline is strict, the possible actions to alleviate this are: either shrink the functionality, or reduce the quality, or add a new developer. However, the latter action is usually problematic due to the necessity to train the developer on this particular project.<br>It is easier to recover after "reduce quality" action.<br>When producing a Work Breakdown Structure, it is easier to identify:<br>• reusage capabilities<br>• task interdependencies<br>• task workloads<br>Finally, it is easier to manage the changing requirements to a software product. |
| Developers | The tool helps developers to:<br>• take architectural decisions<br>• identify the proper usage for the existing code and the original intents of its implementers<br>• search for possible side-effects of executing some statement, introducing a fix or new feature<br>• identify the causes of a bug<br>The tool also partially relieves developers from maintaining documentation, given that the source code is good: logical, consistent and self-explanatory. |
| Testers | Testers will get a way to better<br>• identify the possible control paths<br>• determine the weak places of the software<br>When a bug reported by a client is not obvious to reproduce, taking a look at the visualization of the software can help to figure out why. |
| Users | Users are provided with better services.<br>• Support is more **prompt**:<br>o Issues are fixed faster, as it is easier for developers to find the causes and to devise the fixes.<br>o Requested new features are implemented faster, as it is easier for developers to understand the existing system, and the reusage recall is higher.<br>• More **powerful** software, because:<br>o developers can build more complex systems<br>• More **stable** software, because it is easier to |
| | ○ determine the weak parts of the software by developers and testers<br>○ determine the possible control paths and cover them with tests |

## 2.7 Thesis Outline

In this section we have introduced the state of the art in the area of Reverse Architecting and placed Artificial Intelligence techniques in this context. Further we discuss the candidate AI techniques in section 3, analyze the weaknesses of the existing approaches and give counterexamples. We also discuss state of the art in source code analysis in this section, as we need some source code analysis in order to extract input data for our approach.

Section 4 provides the background material for proper understanding of our contributions by the reader.

The theory we developed in order to implement the project is given in section 5. As we often used problem-solving approach, we are not always confident about the originality, and optimality or superiority of the solutions we devised. Certainly, we admit this as a weakness of our paper in section 9. The most-likely to be original, optimal or superior solutions are put in section 5. The solutions suspected to be non-original are discussed together with the background material in section 4. The solutions known or likely to be far from optimal are discussed together with our experiments (section 7) or implementation and specification (section 6). We do not implement to our knowledge inferior solutions, if efforts-to-value tradeoff allows given the effort limit.

The empirical evidence for the quality of clustering and meaningfulness of the produced software decompositions is given in section 7.2. We give further visual examples, evidence and proofs in the appendix. Please, note that the appendix is also an important part of the work, as we put some parts there in order not to overload the textual information of the thesis with huge visual examples and source code. We also discuss those visual examples partially in the appendix, though in a less formal way, necessary for their comprehension.

We give a list of problems that we could not solve within the limits of this project, see section 8. Finally, we summarize our contributions and discuss the approach in section 9.

# 3 Literature and Tools Survey

Architecture reconstruction typically involves three steps [Kosc2009]:

1. Extract raw data on the system.
2. Apply the appropriate abstraction technique.
3. Present or visualize the information.

Within this project we perform integrative task over the listed above steps. We select suitable state of the art methods and tools, adhering to realistic estimations on the practical needs, port the methods and tools from other domain into ours and solve the arising issues.

## *3.1 Source code analysis*

In the literature they distinguish two types of source code analysis:

- static (the subject program is NOT run)
- dynamic (the subject program is executed)

**Source code analysis** is the process of extracting information about a program from its source code or artifacts (e.g., from Java byte code or execution traces) generated from the source code using automatic tools. **Source code** is any static, textual, human readable, fully executable description of a computer program that can be compiled automatically into an executable form. To support **dynamic analysis** the description can include documents needed to execute or compile the program, such as program inputs [Bink2007]. According to [Kli2009], source code analysis is also a form of programming.

Call graphs depict the static, caller-callee relation between “functions” in a program. With most source/target languages supporting functions as the primitive unit of composition, call graphs naturally form the fundamental control flow representation available to understand/develop software. They are also the substrate on which various interprocedural analyses are performed and are integral part of program comprehension/testing [Nara2008].

In this project we consider call graph as the most important source of relations between software engineering artifacts. Thus most of our interest in source code analysis falls into call graph extraction. This is also the most difficult, and computer time- and space-consuming operation among all the extractions we perform as pre-requisites. Extraction of other relations, such as inheritance, field access and type usage is more straightforward and mostly reduces to parsing. Call graphs are commonly used as input for automatic clustering algorithms, the goal of which is to extract the high level structure of the program under study. Determining call graph for a **procedural** program is fairly simple. However, this is not the case for programs written in **object-oriented** languages, due to *polymorphism*. A number of algorithms for the *static* construction of an object-oriented program’s call graph have been developed in the compiler optimization literature in recent years. [Rays2000]

In the context of *software clustering*, we attempt to infer the unity of purpose of entities based on their relations, commonly represented in such abstractions as data dependency graphs and call graphs [Rays2000]. The latter paper experiments with 3 most common algorithms for the static construction of the call graph of an *object-oriented* program, available at that time:

- Naïve
  This algorithm assumes that the actual and the implementing types are the same as the declared type. The benefits are: no extra analysis, sufficient for the purposes of a non-optimizing compiler, and very simple.
- Class Hierarchy Analysis (CHA) [Diw1996] is a whole-program analysis that determines the actual and implementing types for each method invocation based on the type structure of the program. The whole program is not always available for analysis, not only for trivial (but common) reasons of absence of a .jar file, but also due to features such as reflection and remote method invocation. CHA is flow and context insensitive.

- Rapid Type Analysis (RTA) [Bac1996] uses the set of instantiated types to eliminate spurious invocation arcs from the graph produced by CHA. This analysis is particularly effective when a program is only using a small portion of a large library, which is often the case in Java [Rays2000]. This is also the case in our project: out of 7.5K of classes upon analysis, 6.5K are library classes. Studies have shown that RTA is a significant improvement over CHA, often resolving 80% of the polymorphic invocations to a single target [Bac1996].

At the time of [Rays2000] experiments, RTA was considered to be the best practical algorithm for call graph construction in object-oriented languages. The improved methods we are using in this project, Spark [Lho2003] and VTA [Sun1999] [Kwo2000], were under development.

The authors of [Rays2000] were only able to conclude that the choice of call graph construction algorithm does indeed affect the automatic clustering process, but not whether clustering of more accurate call graphs will produce more accurate clustering results. Assessment of both call graph and clustering accuracy is a fundamental difficulty.

### 3.1.1 Soot

Soot [Soot1999] is a framework originally aimed at optimizing Java bytecode. However, we use it first of all for parsing and obtaining structured in-memory representation of the source code upon our analysis. The framework is open-source software implemented in Java, and this is important as it gives an opportunity to modify the source code of the tool in order to tune it for our needs.

Soot supports several intermediate representations for Java bytecode analyzed with it: Baf, a streamlined representation of bytecode which is simple to manipulate; Jimple, a typed 3-address intermediate representation suitable for optimization; and Grimp, an aggregated version of Jimple suitable for decompilation [Soot1999]. Another intermediate representation implemented in the recent years is Shimple, a Static Single Assignment-form version of the Jimple representation. SSA-form guarantees that each local variable has a single static point of definition which significantly simplifies a number of analyses [EiNi2008].

Our fact extraction and the prerequisites for our analyses are built on top of the Jimple intermediate representation. The prerequisites are call graph extractors, namely, Variable Type Analysis (VTA) [Sun1999] and Spark [Lho2003], a flexible framework for experimenting with points-to analyses for Java. Soot can analyze isolated source/bytecode files, but for call graph extraction whole-program mode [EiNi2008, p.19] is required. In this mode Soot first reads all class files that are required by an application, by starting with the main root class or all the classes supplied in the directories to process, and recursively loading all classes used in each newly loaded class. The complete application means that all the entailed libraries, including java system libraries, are processed and represented in memory structurally. This causes crucial performance and scalability issues, as it was tricky to make Soot fit in 2GB RAM while processing the software projects we further used in our experiments on clustering.

As each class is read, it is converted into the Jimple intermediate representation. After conversion, each class is stored in an instance of a `SootClass`, which in turn contains information like its name, signature (fully-qualified name), its superclass, a list of interfaces that it implements, and a collection of `SootField`\`s and `SootMethod`\`s. Each `SootMethod` contains information like its (fully-qualified) name, modifier, parameters, locals, return type and a list of Jimple 3-address code instructions. All parameters and locals have declared types. Soot can produce Jimple intermediate representation directly from the Java bytecode in class files, and not only from high-level Java programs, thus we can analyze Java bytecode that has been produced by any compiler, optimizer, or other tool. [Sun1999]

### 3.1.2 Rascal

In this section we discuss recent state of the art developments in source code analysis and manipulation (SCAM) domain, placing our project and research in this context. Most of the research addresses explicit facts, while we aim at identification of the implicit facts (architecture), as there is an inference step (namely, clustering) between SE artifact relations and presentation to a user.

SCAM is a large and diverse area both conceptually and technologically. Many automated software engineering tools require tight integration of techniques for source code analysis and manipulation, but integrated facilities that combine both domains are scarce because different computational paradigms fit each domain best. Both domains depend on a wide range of concepts such as grammars and parsing, abstract syntax trees, pattern matching, generalized tree traversal, constraint solving, type inference, high fidelity transformations, slicing, abstract interpretation, model checking, and abstract state machine. Rascal is a domain-specific language that integrates source code analysis and manipulation at the conceptual, syntactic, semantic and technical level [Kli2009]. The goals of Rascal are:

- To remove the cognitive and computational overhead of integrating analysis and transformation tools
- To provide a safe and interactive environment for constructing and experimenting with large and complicated source code analyses and transformations such as, for instance, needed for refactorings
- To be easily understandable by a large group of computer programming experts

Visualization of software engineering artifacts is important. CWI/SEN1 research group is developing Rascal within The Meta-Environment, a framework for language development, source code analysis and source code transformation: http://www.meta-environment.org/ . This framework could use the results of this project, providing input and taking output. There is no call graph clustering in the framework yet. The research group is currently developing a visualization framework, which could graphically illustrate the results of this project too.

## *3.2 Clustering*

Clustering is a fundamental task in machine learning [Ra2007]. Given a set of data instances, the goal is to group them in a meaningful way, with the interpretation of grouping dictated by the domain. In the context of relational data sets – that is, data whose instances are connected by a link structure representing domain-specific relationships or statistical dependency – the clustering task becomes a means for identifying communities within networks. For example, in the bibliographic domain explored by both [Ra2007] and [Fla2004], they find networks of scientific papers. Interpreted as a graph, vertices (papers) are connected by an edge when one cites the other. Given a specific paper (or group of papers), one may try to find out more about the subject matter by pouring through the works cited, and perhaps the works they cite as well. However, for a sufficiently large network, the number of papers to investigate quickly becomes overwhelming. By clustering the graph, we can identify the community of relevant works surrounding the paper in question.

An example of value that clustering can bring into graph comprehension is illustrated in Figure 3-1 below. Both pictures on the left and on the right are adjacency matrices of the same graph. However, the vertices (which are row and column labels) in the right picture are ordered according to the cluster they belong to, so that vertices of the same cluster go subsequently. The matrix on the right is almost quasi-diagonal, thus we can look at the contracted graph of 17 vertices (one vertex per cluster) instead of the original graph of 210 vertices. The edges of the contracted graph will reflect the exceptions that prevent the adjacency matrix on the right from being strictly quasi-diagonal, and the weights of those edges reflect the cardinality of the exceptions.

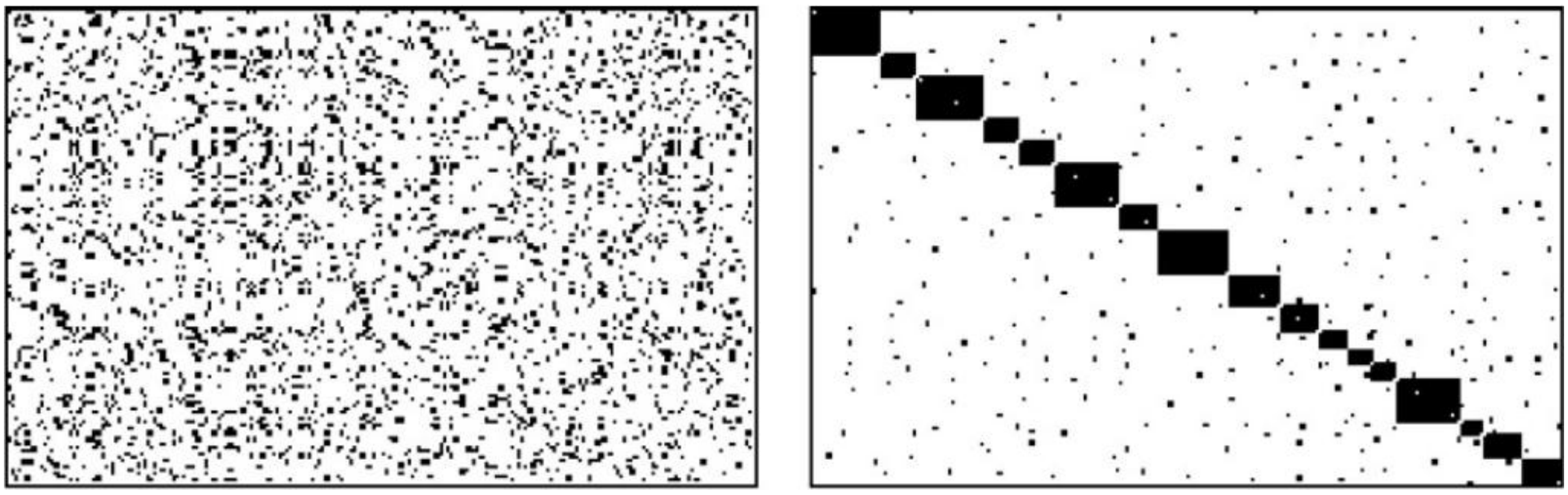

**Figure 3-1: Clustering facilitates comprehension of an adjacency matrix**

No single definition of a cluster in graphs is universally accepted [Sch2007], thus there are some intuitive **desirable cluster properties** mentioned in the literature. In the setting of graphs, each cluster should be connected: there should be at least one, preferably several paths connecting each pair of vertices within a cluster. If a vertex [u] cannot be reached from a vertex [v], they should not be grouped in the same cluster. Furthermore, the paths should be internal to the cluster: in addition to the vertex set [C] being connected in [G], the subgraph induced by [C] should be connected in itself, meaning that it is not sufficient for two vertices [v] and [u] in [C] to be connected by a path that passes through vertices in [V\C], but they also need to be connected by a path that only visits vertices included in [C]. As a consequence, when clustering a **disconnected** graph with known components, the clustering should usually be conducted on each component separately, unless some *global restriction* on the resulting clusters is imposed. In some applications, one may wish to obtain clusters of similar order and/or density, in which case the clusters computed in one component also influence the clusterings of other components. This also makes sense in the domain of software engineering artifacts clustering when we are analyzing disjoint libraries with intent to look at their architecture from the same level of abstraction.

It is generally agreed upon that a subset of vertices forms a good cluster if the induced subgraph is dense, but there are relatively few connections from the included vertices to vertices in the rest of the graph [Sch2007]. Still, there are multiple possible ways of defining density. At this point there are two things worthy to notice:

1 [Sch2007] uses the notion of cut size, c(C, V\C) to measure the sparsity of connections from cluster [C] to the rest of the graph, and this matches to the central clustering approach we use in our work: Graph Clustering based on Minimum Cut Trees [Fla2004]. Minimum cuts play central role there in both inter-cluster and intra-cluster connection density evaluation.

2 For calculation of both inter- and intra- cluster densities, in the formulas of [Sch2007, page 33] they use “maximum number of edges possible” as the denominator. However, they consider the number of edges in a *complete* graph as the maximum number of edges possible, which can be wrong due to the specific of the underlying data (not all the graph configurations are possible, i.e. the denominator must be far less than the number of edges in a complete graph), and this can cause density estimation problems and **skew the results**.

Considering the connectivity and density requirements given in [Sch2007], semantically useful clusters lie somewhere in between the two extremes: the loosest – a connected component, and the strictest – a maximal clique. Connected components are easily computed in O(|V|+|E|) time, while clique detection is NP-complete. An example of good (left), worse (middle) and bad (right) cluster is given in Figure 3-2 below.

- The cluster on the left is of good quality, dense and introvert.

- The one in the middle has the same number of internal edges, but many more edges to outside vertices, making it a worse cluster.
- The cluster on the right has very few connections outside, but lacks internal density and hence is not a good cluster.

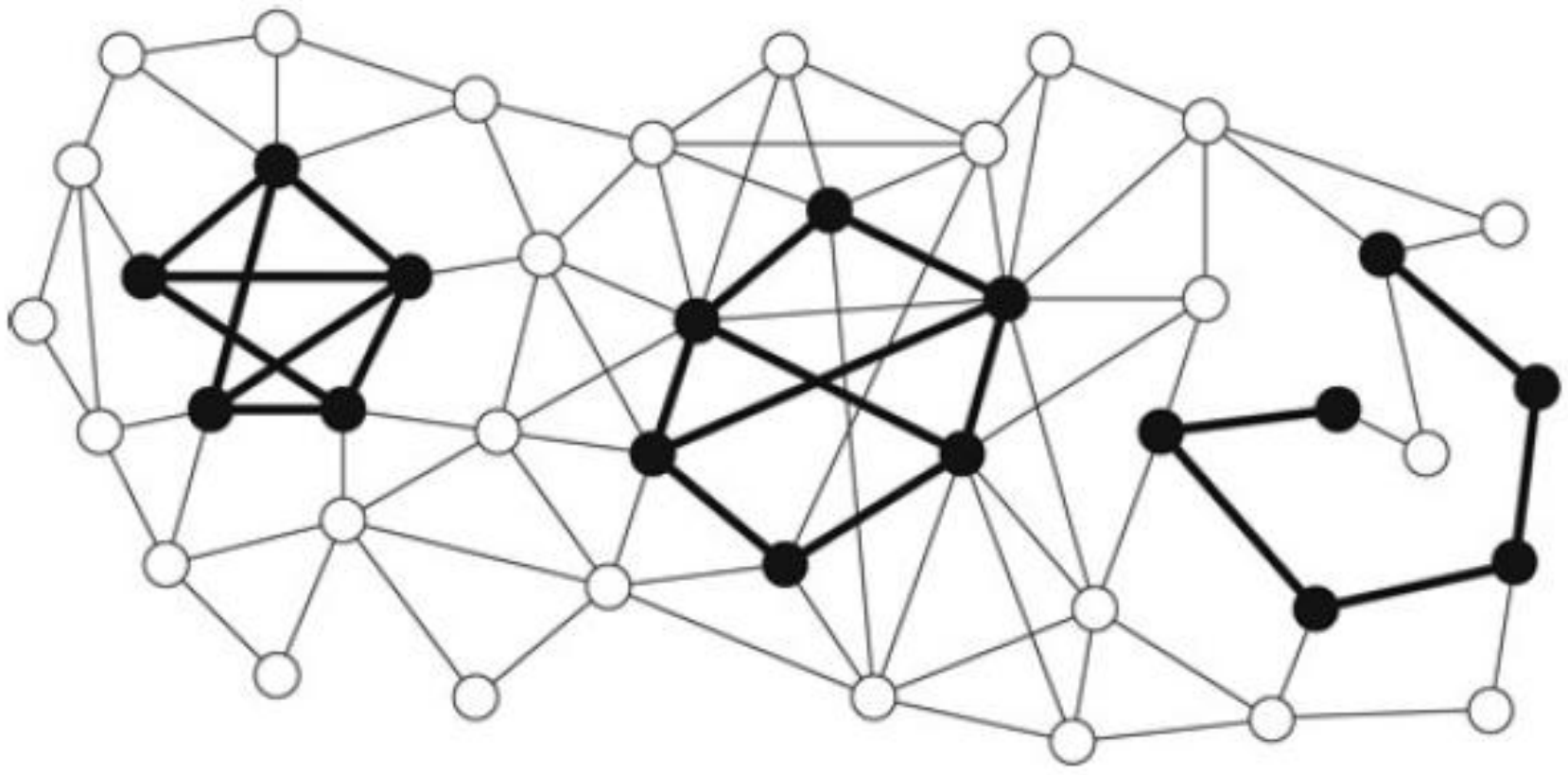

**Figure 3-2: Intuitively good (left), worse (middle) and bad (right) clusters**

It is not always clear whether each vertex should be assigned fully to a cluster or could it instead have different "levels of membership" in several clusters? [Sch2007] In Java classes clustering, such a situation is easily imaginable: a class can be converting data from XML document into a database, and hence could be clustered into "XML" with 0.3 membership, for example, and "database" with a membership level of 0.4. The coefficients can be normalized

- either per-cluster: the sum of all membership levels over all classes belonging to this cluster equals to 1.0
- or per-class: the sum of all membership levels over all the clusters which this class belongs to equals to 1.0

A solution, hierarchical disjoint clustering, would sometimes create a supercluster (parent or indirect ancestor) to include all classes related to XML and database, but the downside is that there can be classes dealing with database but having no relation to XML whatsoever. This is the solution adopted in our work; however, due to the aforementioned downside, an alternative seems interesting too: fuzzy graph clustering [Dong2006]. In a fuzzy graph, each edge is assigned a degree of presence in the graph. Different non-fuzzy graphs can be obtained by leaving only the edges with presence level exceeding a certain threshold. The algorithm of [Dong2006] exploits a connectivity property of fuzzy graphs. It first preclusters the data into subclusters based on the distance measure, after which a fuzzy graph is constructed for each subcluster and a thresholded non-fuzzy graph for the resulting graph is used to define what constitutes a cluster.

### 3.2.1 Particularly Considered Methods

Flake-Tarjan clustering, also known as graph clustering based on minimum cut trees [Fla2004], was used as the core clustering approach in this work. However, a number of other clustering methods were considered within our research too. It was concluded that all of them are either inapplicable due to our problem size (in terms of algorithmic complexity), or inferior to Flake-Tarjan clustering in terms of clustering performance (the accuracy of the results and the usefulness of the measure which the methods are aiming to optimize).

#### 3.2.1.1 Affinity Propagation

Affinity propagation [Fre2007] is a clustering algorithm that takes as input measures of similarity between pairs of data points and simultaneously considers all data points as potential exemplars. Real-valued messages are exchanged between data points until a high-quality set of exemplars and corresponding clusters gradually emerges. A derivation of the affinity propagation algorithm stemming form an alternative, yet equivalent, graphical model is proposed in [Giv2009]. The new model allows easier derivations of message updates for extensions and modifications of the standard affinity propagation algorithm.

In the initial set of data points (in our case, software engineering artifacts, e.g. Java classes), affinity propagation (AP) pursues the goal of finding a subset of **exemplar** points that best describe the data. AP associates each data point with one exemplar, resulting in a partitioning of the whole data set into clusters. The measure which AP maximizes is the overall sum of similarities between data points and their exemplars, called **net similarity**. It is important to note why a degenerate solution doesn't occur. The net similarity is not maximized when every data point is assigned to its own singleton exemplar because it is usually the case that a gain in similarity a data point achieves by assigning itself to an existing exemplar is higher than the preference value. The **preference** of point $i$, called $p(i)$ or $s(i,i)$, is the a priori suitability of point $i$ to serve as an exemplar. Preferences can be set to a global (shared) value, or customized for particular data points. High values of the preferences will cause affinity propagation to find many exemplars (clusters), while low values will lead to a small number of exemplars (clusters). A good initial choice for the preference is the minimum similarity or the median similarity [Fre2007].

Affinity propagation **iteratively** improves the clustering result (net similarity), and the time required for one iteration is asymptotically equal to the number of edges in the similarity graph. In their experiments [Fre2007] authors used some fixed number of iterations, but one can also run iterations until some pre-defined time limit is exceeded.

Normally, the algorithm takes NxN adjacency matrix on input, but we cannot allow this in our problem because such a solution is not scalable to large software projects. In a large software project there can be 0.1 millions of software engineering artifacts (e.g., Java classes), but only 1-2 millions of relations among them (method calls, field accesses, inheritance, etc), i.e. the graph upon analysis is very sparse. On medium-size software projects, consisting of no more than 10 000 artifacts at the selected granularity (e.g. classes), however, it is feasible to compute an adjacency matrix using some transitive formula for similarity of artifacts which do not have a direct edge in the initial graph of relations, thus it is worthy to mention the practical constraints for affinity propagation. The number of scalar computations per iteration is equal to a constant times the number of input similarities, where in practice the constant is approximately 1000. The number of real numbers that need to be stored is equal to 8 times the number of input similarities. So, the number of data points is usually limited by memory, because we need $N^2$ similarities for N data points.

Though affinity propagation has a sparse version, the variety of the resulting clustering configurations becomes very limited in this case. In a sparse version, the similarity between any two points not connected in the input graph is viewed as negative infinity by the algorithm. Below are two main consequences of this:

- If there is no edge between point A and point B in the input graph, then point A will never be selected as an exemplar for point B.
- If there is no path of length no more than 2 between points C and D, then affinity propagation will never assign points C and D into the same cluster

This is illustrated on Figure 3-3 below.

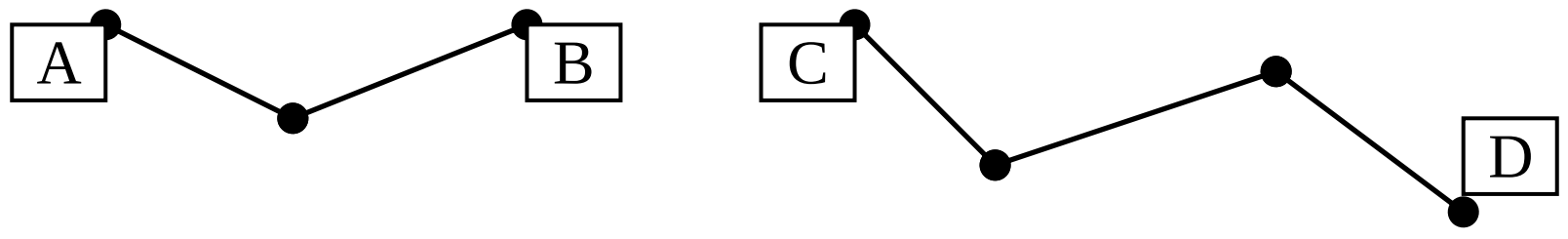

**Figure 3-3 Issues of the sparse version of affinity propagation**

When clustering software engineering artifacts, e.g. Java/C# classes, it seems reasonable that sometimes we want some classes to get into the same cluster even though there is no path of length no more than 2 between them. We conclude that affinity propagation is not applicable to our problem thus.

### 3.2.1.2 Clique Percolation Method

Cfinder is a tool for finding and visualizing overlapping groups of nodes in networks, based on the Clique Percolation Method [Pal2005]. Within this project, we used it "as is" in an attempt to cluster software engineering artifacts using state of the art tools from a different domain, namely, social network analysis. In contrast to Cfinder/CPM, other existing community finders for large networks, including the core method used in our project, find disjoint communities. According to [Pal2005], most of the actual networks are made of highly overlapping cohesive groups of nodes.

Though Cfinder is claimed to be "fast and efficient method for clustering data represented by large graphs, such as genetic or social networks and microarray data" and "very efficient for locating the cliques of large sparse graphs", our experiments showed that it is not applicable to our domain de facto, for both scalability and result usefulness issues. When our original graph, containing 7K vertices (Java classes) and 1M edges (various relations), was given on input of Cfinder, it did not produce any results in reasonable time. When we reduced the graph to client-code artifacts only, resulting in 1K classes and 10K edges, Cfinder still did not finish computations after 16 hours; however, at least it produced some results which could be visualized with Cfinder. It produced one community with several cliques in it, see Figure 10-8 in the appendix. The selected nodes belong to the same clique. Unfortunately, hardly any architectural insight can be captured from this picture even when zoomed, see Figure 10-9 in the appendix.

We suppose that the reason for such poor behavior of Clique Percolation Method in our domain, as opposed to collaboration, word association, protein interaction and social networks [Pal2005], resides in the specific of our data, namely, software engineering artifacts and relations between them. Our cliques are often huge and nested, thus the computational complexity of CPM approaches its worst case.

Certainly, we have studied Cfinder too superficially, and perhaps there is indeed a way to reduce our problem into one feasible to solve with CPM, but after spending reasonable amount of efforts on this grouping approach we conclude that either it is inapplicable, or much more efforts must be spent in order to get useful results with it.

### 3.2.1.3 Based on Graph Cut

A group of clustering approaches is based on graph cut. The problem of minimum cut in a graph is well studied in computer science. An exact solution can be computed in reasonable polynomial time. In a bipartition clustering problem, i.e. only two clusters are needed, minimum cut algorithm can be applied in order to find them. The vertices of the input graph represent the data points and the edges between them are weighted with the affinities between the data points. Intuitively, the fewer high affinity edges are cut, the better the division into two coherent and mutually different parts will be [Bie2006].

In the simplest min cut algorithm, a connected graph is partitioned into two subgraphs with the cut size minimized. However, this often results in a skewed cut, i.e. a very small subgraph is cut away [Ding2001]. This problem could largely be solved by using some cut cost functions proposed in the literature in the context of clustering, among which the average cut cost (Acut) and the normalized cut cost (Ncut). Acut cost seems to be more vulnerable to outliers (atypical data points, meaning that they have low affinity to the rest of the sample) [Bie2006]. However, skewed cuts still occur when the overlaps between clusters are large [Ding2001] and, finally, both optimizing the Acut and Ncut costs are NP-complete problems [Shi2000].

In the fully unsupervised-learning scenario, no prior information is given as to which group data points belong to. In machine learning literature these target groups are called "classes", but do not confuse with Java classes. Besides this clustering scenario, in the transduction scenario the group labels are specified for some data points. Transduction, or semi-supervised learning, received much attention in the past years as a promising middle group between supervised and unsupervised learning, but major computational obstacles were inhibiting its usage, despite the fact that many natural learning situations directly translate into a transduction problem. In graph cut approaches, the problem of transduction can naturally be approached by restricting the search for a low cost graph cut to graph cuts that do not violate the label information [Bie2006].

Fast semi-definite programs relaxations of [Bie2006] made it possible to find a better cut than the one found using spectral relaxations of [Shi2000], and the authors in their experiments were able to process graphs of up to 7K vertices and 41K edges within reasonable time and memory. However, this is still far not enough for our problem, as even a medium-size project has about 1M of relations between software engineering artifacts.

Paper [Ding2001] proposes another cut-based graph partition method, which is based on min-max clustering principle: the similarity or association between two subgraphs (cut set) is minimized, while the similarity or association within each subgraph (summation of similarity between all pairs of nodes within a subgraph) is maximized. The authors present a simple min-max cut function together with a number of theoretical analyses, and show that min-max cut always leads to more **balanced** cuts than the ratio cut [Hag1992] and the normalized cut [Shi2000]. As the optimal solution for their min-max cut function is NP-complete, the authors used a relaxed version which leads to a generalized eigenvalue problem. The second lowest eigenvector, also called the Fiedler vector, provides a linear search order (Fiedler order). Thus the min-max cut algorithm (Mcut) provides both a well-defined objective and a clear procedure to search for the optimal solution [Ding2001]. The authors report that the Mcut outperformed the other methods on a number of newsgroup text datasets. Unfortunately, the computational complexity of the algorithm is not obvious from the article, except that the computation of Fiedler vector can be done in $O(|E|+|V|)$, but the number of data points used in their experiments did not exceed 400, which is far too little for our problem.

One important detail about bipartition-based graph clustering approaches is the transition from 2-cluster to multiple-cluster solution. If this is done by means of recursive bipartition of the formerly identified clusters, then the solution is hierarchical in nature, which is good for source code structure discovery. However, this algorithm is also **greedy**, thus clustering quality can be unacceptable. Apparently, the optimal partition of a system into 3 components can differ much from the solution received by first bipartitioning the system, and then bipartitioning one of the components. This is illustrated in Figure 3-1 below.

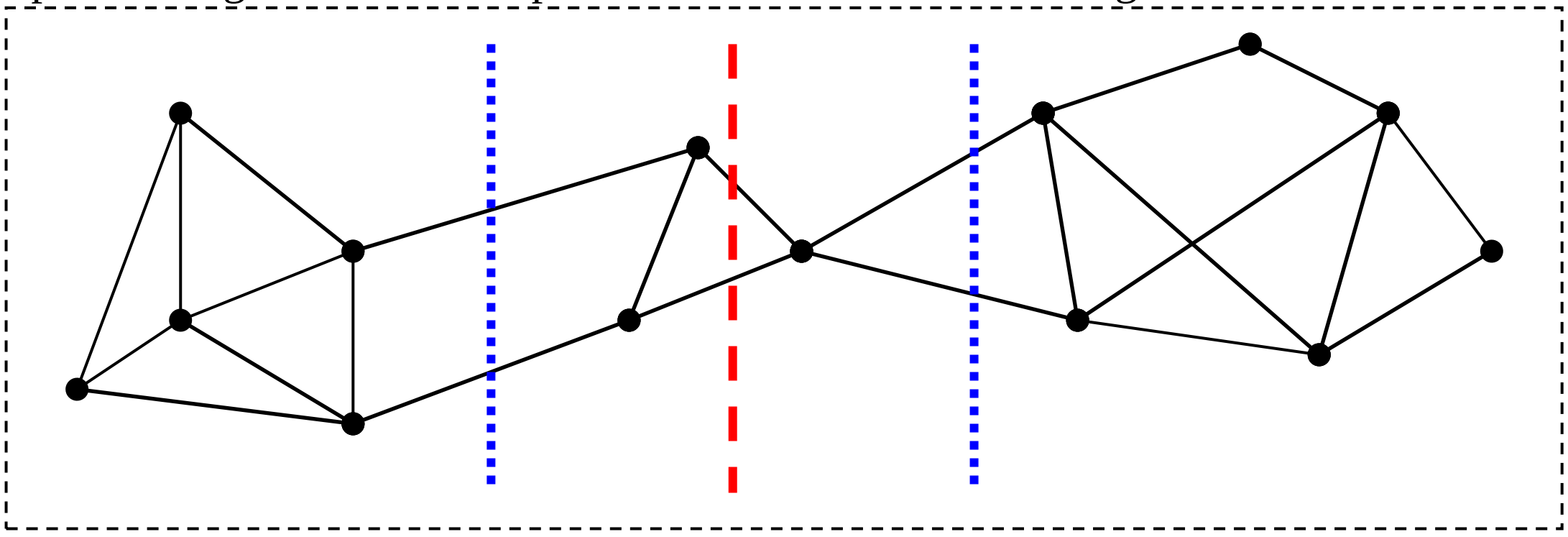

**Figure 3-4 Optimal 2-clustering (dashed) vs. 3-clustering (dotted)**

Finally, the main clustering method selected for the implementation of our project is also cut-based [Fla2004]. Though it produces hierarchical clustering, it does not suffer from the issue of greedy approaches demonstrated above. This is because the hierarchy arises due to the clustering criteria used, namely, vertices of the example graph in Figure 3-4 are sent into three sibling clusters as soon as the key parameter (alpha) is small enough, and until alpha gets even smaller to send all the vertices into one parent cluster. Depending on the input graph, there might be no value of the key parameter at which a certain amount of clusters is produced (e.g. 2 in our example), thus there can be a threshold from 3 clusters to 1 cluster incorporating all the vertices of those child clusters.

### 3.2.2 Other Clustering Methods

Other clustering methods were studied without experiments within this project. Most of the methods did not pass the early cut stage because they are not scalable to large graphs. First of all, methods that require complete adjacency matrix on input were discarded, as it implies at least $N^2$ operations while our graph is sparse. Then, greedy hierarchical clustering approaches, either **agglomerative** or **divisive**, were left out. We, however, find it important to discuss the confusion observed in the literature on software architecture recovery, e.g. the work reported in [Czi2007] and a number of hierarchical clustering for software architecture recovery approaches discussed in [Maqb2007]. While the titles say "hierarchical clustering", for the quality of results it is crucial to distinguish how the hierarchy emerges: whether it happens due to the greedy order in which clusters are identified, or it is data-driven. The clustering algorithm we use in our project falls into the latter category. We discuss those from the former category in subsection 3.2.2.2 below. The superiority of **partitional** clustering methods in the domain of software source code is confirmed in [Czi2008], where the authors improve their own earlier results of [Czi2007] by means of using partitional clustering instead of hierarchical agglomerative clustering used in their earlier work pursuing the same goal of automatic source code refactoring. The partitional clustering method used in [Czi2008] is *k-medoids* [Kau1990] with some heuristics for choosing the number of medoids (clusters) and the initial medoids, while the heuristics are domain-specific.

Another point that we consider worth discussing in a subsection is not a clustering method itself, but rather a technique that allow a series of clustering methods to work in a drastically reduced computational complexity without major precision losses, as reported in [Ra2007]. We do not use any of the clustering methods, e.g. k-medoids [Kau1990] or Girvan-Newman [Gine2002], accelerated with this network structure indices technique in our project for the following reasons:

- [Ra2007] admits the superiority of Flake-Tarjan [Fla2004] clustering method in terms of clustering result quality.
- The only argument of [Ra2007] against minimum cut tree based clustering methods is that "*they are not scalable to large graphs*". However, it seems that the authors of [Ra2007] were not aware of the actual computational complexity of Flake-Tarjan clustering, in terms of both worst-case and usual-case. There is some rationale behind this, as Flake-Tarjan clustering relies on the computation of maximum flow in a graph, which is believed to be a very hard polynomial algorithm. The widely known Dinic's algorithm for max flow in a real-valued network works in $O(|V|^2|E|)$, and the push-relabel algorithm referenced by [Fla2004] works in $O(|V|\cdot|E|\cdot\log\frac{|V|^2}{|E|})$ time. This could give the authors of [Ra2007] a wrong idea on the scalability of minimum cut tree based clustering. However, recent developments

in max flow algorithm allow computing the minimum cut in as little as $O(|E| \cdot \min(|V|^{2/3}, \sqrt{|E|}) \cdot \log \frac{|V|^2}{|E|} \cdot \log U)$, where U is the maximum capacity of the network [Gol1998]. Our practical studies shown that the actual running time of Goldberg's implementation (see section 4.1.1 of the thesis) of push-relabel based max flow algorithm is nearly $\Theta(|E|)$ in the usual case. As this algorithm requires integral arc capacities, we have developed within our work a method to approximate real-valued max flow problem with an integral one that satisfied the needs of our project, namely, the property that allowed minimum cut tree based hierarchical clustering [Fla2004] was not lost due to conversion from real-valued to integral max flow problem.

- One clustering algorithm improved in [Ra2007], namely Girvan-Newman [Gir2002], is greedy hierarchical (divisive) clustering. It is not said whether there are some non-greedy hierarchical clustering algorithms can be improved with network structure indices technique.
- In [Ra2007] the authors only worked with non-weighted graphs. It is not clear whether the technique can still handle weighted graphs, and if so, whether real-valued weights are possible.
- Maximum flow algorithms for non-weighted graphs have smaller algorithmic complexity too: e.g. Dinic blocking flow algorithm for network with unit-capacities terminates in $O(|V| \cdot \sqrt{|E|})$ [Dini1970]. Thus Flake-Tarjan clustering having Dinic's algorithm in the backend would work much faster, but this is only possible in networks with unit capacities.

Thus we conclude that Flake-Tarjan clustering algorithm [Fla2004] is fast enough, produces both better clustering quality than the rival approaches, and a data-driven hierarchical clustering that is very desired for software architecture domain.

#### 3.2.2.1 Network Structure Indices based

Simple clustering methods, like a new graphical adaptation of the k-medoids algorithm [Kau1990] and the Girvan-Newman [Gir2002] method based on edge betweenness centrality, can be effective at discovering the latent groups or communities that are defined by the link structure of a graph. However, many approaches rely on prohibitively expensive computations, given the size of relational data sets in the domain of source code analysis. Network structure indices (NSIs) are a proven technique for indexing network structure and efficiently finding short paths [Ra2007]. In the latter paper they show how incorporating NSIs into these graph clustering algorithms can overcome these complexity limitations.

The k-medoids algorithm [Kau1990] can be thought of as a discrete adaptation of the k-means data clustering method [MaQu1967]. The inputs to the algorithm are k, the number of clusters to form, and a distance measure that maps pairs of data points to a real value. The procedure is as follows: (1) randomly designate k instances to serve as "seeds" for the k clusters; (2) assign the remaining data points to the cluster of the nearest seed using the distance measure; (3) calculate the medoids of each cluster; and (4) repeat steps 2 and 3 using the medoids as seeds until the clusters stabilize. In a graph, medoids are chosen by computing the local closeness centrality [Ra2007] among the nodes in each cluster and selecting the node with the greatest centrality score. One *issue with k-medoids* approach is similar to the problem of sparse version of Affinity Propagation we discussed in 3.2.1.1: in contrast to the data-clustering counterpart k-medoids, graph distance is highly sensitive to the edges that exist in the graph. Adding a single "short-cut" link to a graph can reduce the graph diameter, altering the graph distance between many pairs of nodes. Second issue arises when graph distances are integers. In this case nodes are often equidistant to several cluster medoids. [Ra2007] resolves the latter conflicts by randomly selecting a cluster; however, this can result in clusterings that do not converge. This is further resolved by a threshold on the fraction of non-converged clusters.

The Girvan-Newman algorithm [Gir2002] is a **divisive** clustering technique based on the concept of edge betweenness centrality. Betweenness centrality is the measure of the proportion of shortest paths between nodes that pass through a particular link. Formally, betweenness is defined for each edge $e \in E$ as:

$$B(e) = \sum_{u,v \in V} \frac{g_e(u,v)}{g(u,v)}$$

, where $g(u, v)$ is the total number of geodesic paths between nodes $u$ and $v$, and $g_e(u, v)$ is the number of geodesic paths between $u$ and $v$ that pass through $e$.

A geodesic path in [Ra2007] is simply the shortest path in a graph. Note that there can be multiple shortest paths, i.e. they all have the same length but pass through different chains of edges. Also, the methods of [Ra2007] work with non-weighted graphs, i.e. each edge has length 1.

The algorithm ranks the edges in the graph by their betweenness and removes the edge with the highest score. Betweenness is then recalculated on the modified graph, and the process is repeated. At each step, the set of connected components of the graph is considered a clustering. If the desired number of clusters is known a priori (as with k-medoids), we halt when the desired number of components (clusters) is obtained.

The main problem with the two clustering algorithms described above is algorithmic complexity, and this also applies to many other approaches, but the above two were studied in [Ra2007] and accelerated dramatically with **network structure indices**. The baseline clustering algorithms are intractable for large graphs:

- For k-medoids clustering, calculation and storage of pairwise node distances can be done in $O(|V|^3)$ time and $O(|V|^2)$ space with Floyd-Warshall algorithm (can be found in e.g. [CLR2003]).
- For Girvan-Newman clustering, calculation of edge betweenness for the links in a graph is an $O(|V| \cdot |E|)$ operation.

A network structure index (NSI) is a scalable technique for capturing graph structure [Ra2007]. The index consists of a set of node annotations combined with a distance measure. NSIs enable fast approximation of graph distances and can be paired with a search algorithm to efficiently discover short (note, not the shortest!) paths between nodes in the graph. A distance to zone (DTZ) index was employed in [Ra2007]. The DTZ indexing process creates $d$ independent sets of random partitions (called **dimensions**) by stochastically flooding the graph. Each dimension consists of $z$ random partitions (called **zones**). DTZ annotations store the distances between each node and all zones across each dimension. The approximate distance between two nodes $u$ and $v$ is defined as:

$$D_{DTZ}(u,v) = \sum_d dist_d(u, zone(v)) + dist_d(v, zone(u))$$

, where $dist_d(u, zone(v))$ is the length of the shortest path between $u$ and the closest node in the same zone as $v$ . Creating the DTZ index requires $O(|E| \cdot z \cdot d)$ time and $O(|V| \cdot z \cdot d)$ space. Typically they select $z, d << |V|$ , thus DTZ index can be created and stored in a fraction of the time and space it takes to calculate the exact graph distances for all pairs of nodes in the graph. The results of empirical study of the speed improvement achieved with NSIs are illustrated in Figure 3-5 below [Ra2007]. The top line shows bidirectional breadth-first search, which can become intractable for even moderate-size graphs. The middle line shows an optimal best first search, which represents a lower bound on the run time for any search-based method. The lower line shows an NSI-based method, DTZ with 10 dimensions and 20 nodes.

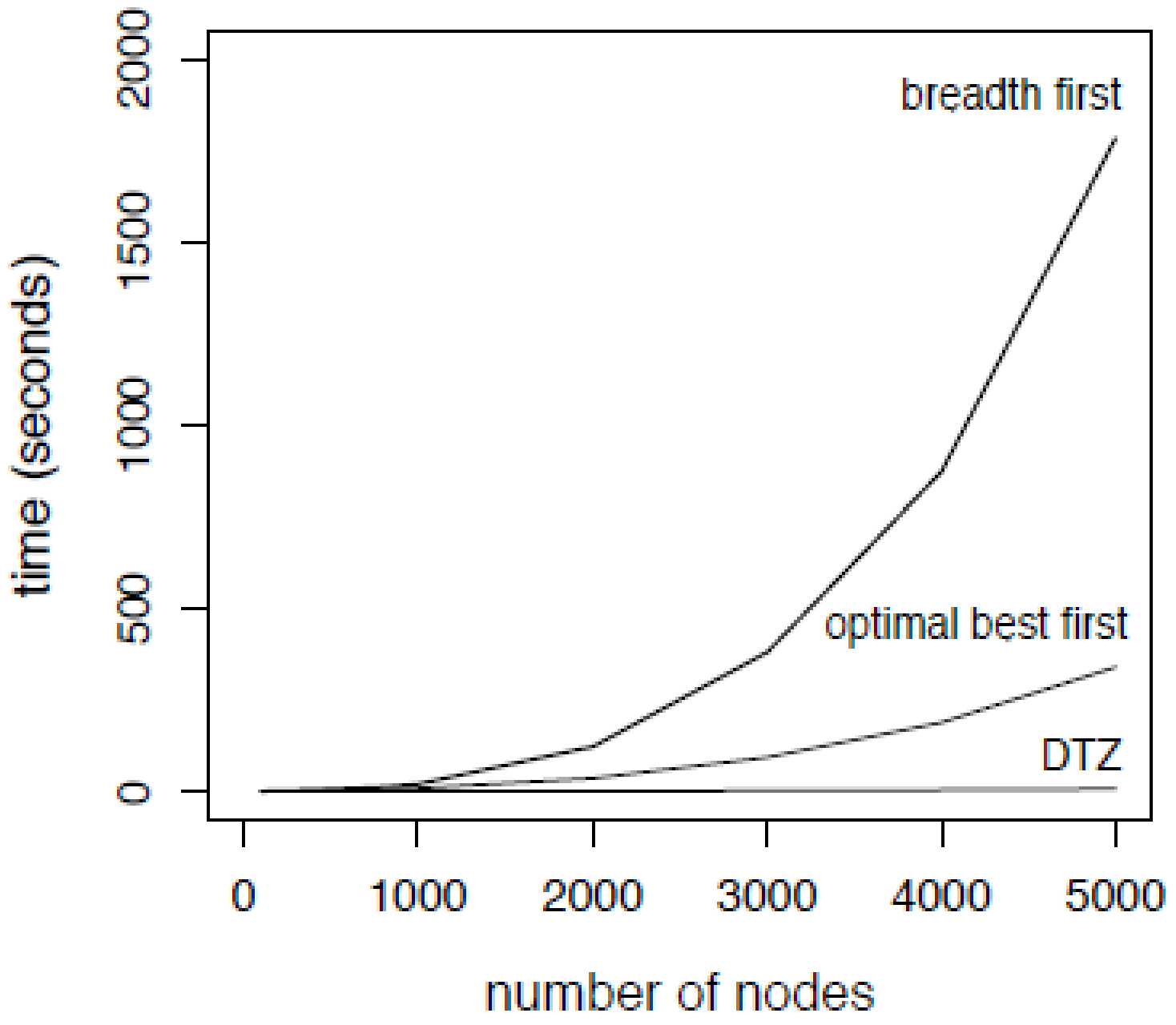

**Figure 3-5 k-medoids speed using 3 different methods of distance calculation**

### 3.2.2.2 Hierarchical clustering methods

Most clustering algorithms can be classified into two popular techniques: partitional and hierarchical clustering. Hierarchical clustering methods represent a major class of clustering techniques [Czi2007]. There are two types of hierarchical clustering algorithms: agglomerative and divisive. Given a set of $n$ objects,

- The agglomerative (bottom-up) methods begin with $n$ singletons (sets with one element), merging them until a single cluster is obtained. At each step, the most similar two clusters are chosen for merging.
- The divisive (top-down) methods start from one cluster containing all $n$ objects and split it until $n$ clusters are obtained.

The agglomerative clustering algorithms differ in the way the two most similar clusters are determined and the linkage-metric used: single, complete or average.

- Single link algorithms merge the clusters whose distance between their closest objects is the smallest.
- Complete link algorithms merge the clusters whose distance between their most distant objects is the smallest.
- Average link algorithms merge the clusters in which the average of distances between the objects from the clusters is the smallest.

In general, complete link algorithms generate compact clusters while single link algorithms generate elongated clusters. Thus, complete link algorithms are generally more useful than single link algorithms [Czi2007]. Average link clustering is a compromise between the sensitivity of complete-link clustering to outliers and the tendency of single-link clustering to form long chains that do not correspond to the intuitive notion of clusters as compact, spherical objects [Man1999].

In addition to the above mentioned issues of agglomerative clustering approaches, and the suspicious averaging of distances, the issue we discussed near Figure 3-4 still remains too, namely, the greedy nature of such algorithms. On the other hand, partitional clustering algorithms look at all the data at once, and produce a partition of the data points into some number of clusters. According to [Jain1999], the partitional techniques usually produce clusters by optimizing a criterion function defined either locally (on a subset of the patterns) or globally (defined over all of the patterns). At this point it is worthy to notice that Flake-Tarjan clustering algorithm optimizes the criteria **globally**, see section 4.3.1.

In [Jain1999] they provide taxonomy of clustering algorithms, see Figure 3-6 below. The main clustering algorithm we use in our project was not yet invented at the time the review [Jain1999] was written, and falls into "Graph Theoretic" category under "Partitional" clustering approaches. We are stressing this to prevent confusion of it with the hierarchical clustering approaches present in the literature on the basis of the fact that the clustering algorithm [Fla2004] produces clustering hierarchy too. Still, it is a partitional clustering method, well grounded theoretically and free of the disadvantages of greedy algorithms.

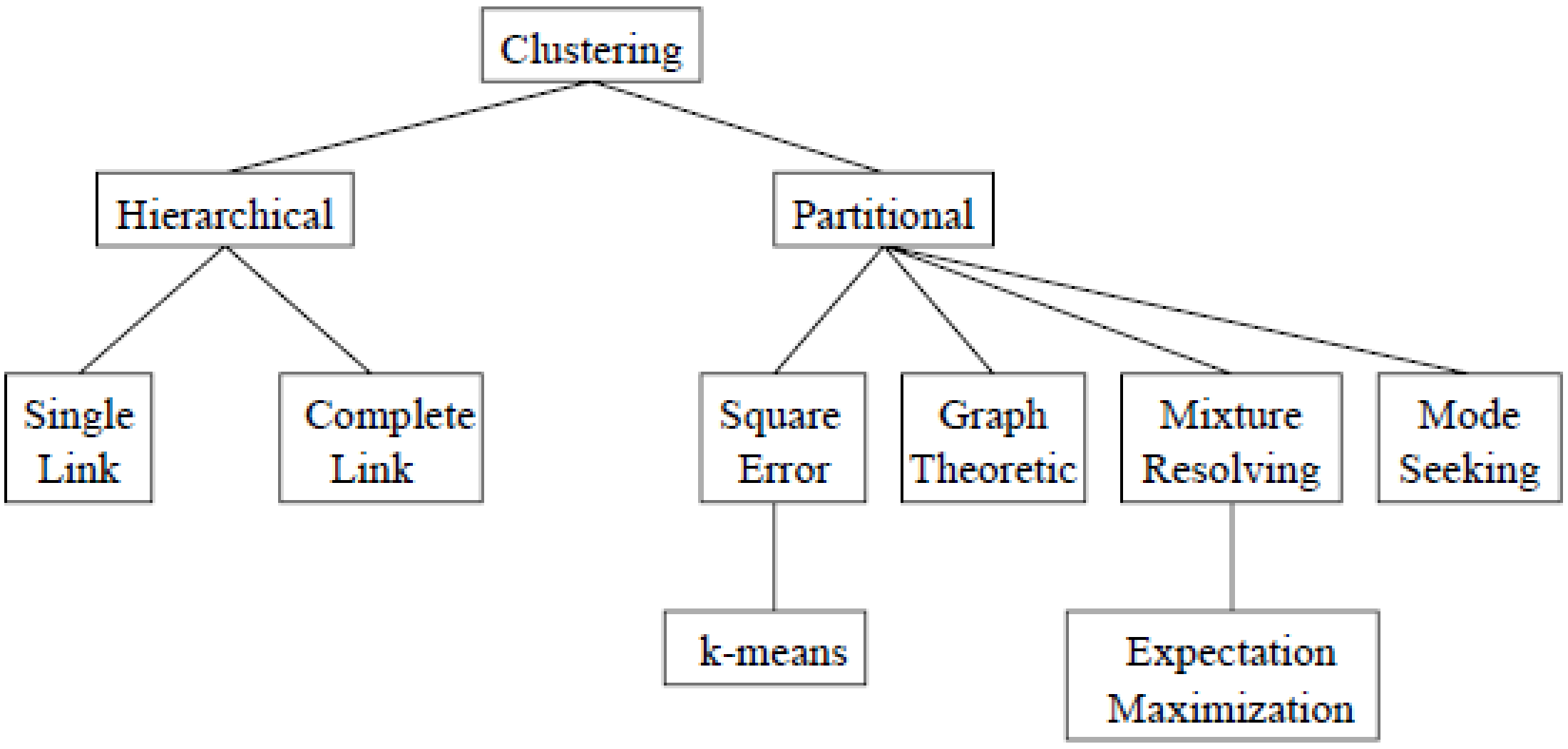

**Figure 3-6 A taxonomy of clustering approaches**

A recent review of multiple hierarchical clustering approaches applied to the domain of software architecture recovery, [Maqb2007], concluded that the performance of the state of the art algorithms is poor. The authors mention arbitrary decisions taken inside the clustering algorithms as a core source of problems. We have demonstrated another source of problems near Figure 3-4, namely, greedy nature of the algorithms. Furthermore, all the algorithms described there work in at least $O(n^2)$ algorithmic complexity, as they take $n$ by $n$ similarity matrix on input, where $n$ is the number of software engineering artifacts. In our work, we propose an approach that is scalable to large sparse graphs of relations between software engineering artifacts, and the clustering decisions are strongly grounded by the theory of [Fla2004].

# 4 Background

The pre-requisite methods and tools used in our project are described and formalized in this section. We also find it worthy to discuss the known challenges arising in the domain of software engineering artifacts clustering. In section 5 we provide the theory we devised on top of the background material given here.

## 4.1 Max Flow & Min Cut algorithm

The maximum flow problem and its dual, the minimum cut problem, are classical combinatorial problems with a wide variety of scientific and engineering applications. In a graph denoting a flow network, edge weights denote the capacities, i.e. the amount of substance that can flow through a connection between points (vertices). The task is assignment a certain amount of flow to each connection (pipe), so that the total flow from a source vertex to a sink vertex is maximized. More background about this classical problem can be found in [CLR2003]. Here we just mention the differences with the shortest-path problem that is a prevailent pre-requisite for other clustering algorithms:

- High weight is good for flow, but bad for short path.
- A path is a chain of edges, while flow can go in multiple parallel directions.
- There is no (polynomial) solution for "longest path" problem, but there are solutions for maximum flow.

Most efficient algorithms for the maximum flow are based on the blocking flow and the push-relabel methods. The shortest augmenting path algorithm, the blocking flow method, and the push-relabel method use a concept of distance, taking the length of every residual arc to be unit (one). Using a more general, binary length function [Gol1998] substantially improved the previous time bounds. As a potential further improvement direction, the authors mention considering length functions that depend on the distance labels of the endpoints of an arc in addition to the arc's residual capacity.

Within our project we do not seek to improve the speed of the max flow algorithm and take it as is with algorithmic complexity:

$$O(|E| \cdot \min(|V|^{2/3}, \sqrt{|E|}) \cdot \log \frac{|V|^2}{|E|} \cdot \log U)$$

In a typical graph for our experiments, 10K vertices and 1M edges, this amounts to 1M*464*8*32 = 119G of trivial operations in the worst case. However, the lossless heuristics [Che1997] for push-relabel based implementation of max flow (see subsection 4.1.1) kept the actual number of scalar operations to about 100M. So we use binary blocking flow for theoretical estimations of algorithmic complexity and Goldberg's implementation of push-relabel based algorithm with heuristics, which works fascinating in practice.

### 4.1.1 Goldberg's implementation

In this project we used Goldberg's implementation of push-relabel algorithm solving max flow problem, http://www.avglab.com/andrew/soft.html , see "HIPR" which is an improvement over the H_PRF version described in [Che1997]. This implementation performs much less scalar operations than the worst-case estimation due to lossless (in terms of optimality) speedup heuristics. In [Che1997] the authors point out a problem family on which all the known max flow methods have quadratic (in the number of vertices) time growth rate. However, even for this problem family their best implementation (H_PRF) processed a graph of 65K vertices and 98K edges in reasonable time

As maximum flow computation is the bottleneck in our project, it is important that the implementation is highly optimized, including low-level optimizations. This implementation is written in C and includes some heuristics that allow computing the maximum flow much faster than the worst-time complexity estimation. In our practical needs we observe that the algorithm computes max flow in a graph of 10K vertices and 1M edges in 0.02 seconds on a 1.7GHz computer with 2MB cache memory, while processor cache size is important as most of the time is spent in cache misses.

## *4.2 Min Cut Tree algorithm*

Cut trees, introduced in [GoHu1961] and also known as Gomory-Hu trees, represent the structure of all $s-t$ cuts of **undirected** graphs in a compact way. Cut trees have many applications, but in our project we use them for clustering as described in [Fla2004]. All known algorithms for building cut trees use a minimum $s-t$ cut (see 4.1 above) as a subroutine [GoTs2001]. In [GoHu1961] they showed how to solve the minimum cut tree problem using $n-1$ minimum cut computations and graph contractions, where $n$ is the number of vertices in the graph. An efficient implementation of this algorithm is non-trivial due to subgraph contraction operations used [GoTs2001]. Gusfield [Gus1990] proposed an algorithm that does not use graph contraction; all $n-1$ minimum $s-t$ cut computations are performed on the input graph. We use this algorithm in our project for one more reason in addition to the above mentioned: it is possible to apply the community heuristic, specific to the purpose for which we need computation of minimum cut tree [Fla2004]. Both Gusfield algorithm and the community heuristic are described in the subsections.

The input to the cut tree problem is an undirected graph $G=(V,E)$, in which edges have capacities, each denoting the maximum possible amount of flow through an edge. We say that an edge crosses the cut if its two endpoints are on different sides of the cut. Capacity of a cut is the sum of capacities of edges crossing the cut. For $s,t \in V$, an $s-t$ cut is a cut such that $s$ and $t$ are on different sides of it. A *minimum* $s-t$ cut is an $s-t$ cut of minimum capacity. A (*global*) minimum cut is a minimum $s-t$ cut over all $s,t$ pairs. A cut tree is a weighted tree $T$ on $V$ with the following property. For every pair of distinct vertices $s$ and $t$, let $e$ be a minimum weight edge on the unique path from $s$ to $t$ in $T$. Deleting $e$ from $T$ separates $T$ into two connected components, $X$ and $Y$. Then $(X,Y)$ is a minimum $s-t$ cut. Note that $T$ is *not a subgraph* of $G$, i.e. edges of $T$ do not need to be in $G$.

### 4.2.1 Gusfield algorithm

In [Gus1990] they provide a simple method for implementing minimum cut tree algorithm, which does not involve graph contraction [GoHu1961] and works in the same algorithmic complexity. Below is the pseudo code of Gusfield algorithm:

```
For all vertices, i=2…n, set prev[i]=1;
For all vertices, s=2…n do
    t = prev[s];
    Calculate a minimal cut (S,T) and the value [w] of a maximal flow in the graph,
      using [s] as the source and [t] as the sink
    Add edge (s,t) with weight [w] to the resulting tree
    For all vertices [i] from S /* the source-side vertices after the cut */
        If i > s and prev[i]==t then
            prev[i] = s;
        End;
    End;
End;
```

Apart from simplicity of the algorithm, we also use it because the community heuristic [Fla2004] (also described in section 4.2.2 of the thesis) can be applied thus reducing the required number of max flow computations substantially.

### 4.2.2 Community heuristic

The running time of the basic cut clustering algorithm [Fla2004] is equal to the time to calculate the minimum cut tree, plus a small overhead for extracting the subtrees under the artificial sink $t$. But calculating the min-cut tree can be equivalent to computing $n-1$ maximum flows in the worst case for [GoHu1961], and always for [Gus1990] which we provided in section 4.2.1 and use in our project. Fortunately, [Fla2004] proves a property that allows to find clusters much faster, in practice, usually in time equal to the *total number of clusters* times the time to compute max flow.

The gist of the community heuristic follows. If the cut between some node $v$ and $t$ yields the community $S$ (vertices on the source-side of the cut), then we do not use any of the nodes in $S$ as subsequent sources to find minimum cuts with $t$, since according to a lemma proved in [Fla2004] their communities would be subsets of $S$. Instead, we mark the vertices of $S$ as being in community $S$, and later if $S$ becomes part of a larger community $S'$ we mark all nodes of $S$ as being part of $S'$.

The heuristic relies on the order in which we iterate over the vertices of the graph, as opposed to the baseline Gusfield algorithm (section 4.2.1) which passes the vertices in arbitrary order. It is desired that the largest clusters are identified first. As proposed in [Fla2004], we sort all nodes according to the sum of the weights of their adjacent edges, in decreasing order.

## *4.3 Flake-Tarjan clustering*

In [Fla2004] they introduce simple graph clustering methods based on minimum cuts within the graph. The cut clustering methods are general enough to apply to any kind of graph but, according to the authors of the paper, are well-suited for graphs where the link structure implies a notion of reference, similarity or endorsement. The authors experiment with Web and citation graphs in their work.

Given an undirected graph $G(V,E)$ and a value of parameter $\alpha$, the basic clustering algorithm of [Fla2004], which we call “Alpha-clustering” (see section 4.3.1) due to the presence of parameter $\alpha$, finds a community for each vertex with respect to an artificial sink $t$ added to the graph $G$. The artificial sink is connected to each node of $G$ via an undirected edge of capacity $\alpha$. The community of vertex $s$ with respect to vertex $t$ is the set of vertices on the source-side of the minimum cut between vertex $s$ as the source, and vertex $t$ as the sink.

In the hierarchical version of Flake-Tarjan clustering algorithm, we can observe that the algorithm does not depend on parameter $\alpha$ in case all the breakpoints of parametric max flow [Bab2006] have been considered, where parametric edges are those connecting the artificial sink to the rest of the graph. Thus given the input graph, a hierarchical clustering is produced on output. It is very important to stress that there are **no parameters to tune**, unlike parameter $k$ in k-medoids clustering [Kau1990] or exemplar preferences in affinity propagation [Fre2007]. The resulting clustering is completely data-driven.

### 4.3.1 Alpha-clustering

Parameter $\alpha$ serves as an upper bound of the **inter-cluster** edge capacity and a lower bound of the **intra-cluster** edge capacity, according to a theorem proven in [Fla2004]:

Let $G(V,E)$ be an undirected graph, $s \in V$ a source, and connect an artificial sink $t$ with edges of capacity $\alpha$ to all nodes. Let $S$ be the community of $s$ with respect to $t$. For any non-empty $P$ and $Q$, such that $P \cup Q = S$ and $P \cap Q = \{\}$, the following bounds always hold:

$$\frac{c(S, V-S)}{|V-S|} \le \alpha \le \frac{c(P,Q)}{\min(|P|,|Q|)}$$

The left side of the inequality bounds the inter-community edge-capacity, thus guaranteeing that communities will be relatively disconnected. Here $c(S, V-S)$ is the cut size (the sum of the capacities of edges going from the left set of vertices to the right set) between the vertices in $S$ and the rest of the graph.

The right side of the inequality means that for any cut inside the community $S$, even the minimum one, its value (i.e. the sum of edges crossing the cut) will be at least $\alpha$ times the minimum of the cardinalities over the two sides of the cut. In the other words:

- If we want to separate 1 vertex from a cluster (containing at least 2 vertices), we have to cut away edges with total weight at least $\alpha$
- If we want to separate 2 vertices from a cluster (containing at least 4 vertices), we have to cut away edges with total weight at least $2\alpha$
- If we want to separate 3 vertices from a cluster (containing at least 6 vertices), we have to cut away edges with total weight at least $3\alpha$
- And so on.

As $\alpha$ goes to 0, the cut clustering algorithm will produce only one cluster, namely the entire graph $G$, as long as $G$ is connected. On the other extreme, as $\alpha$ goes to infinity, there will be $n$ trivial clusters, all singletons. When a particular number of clusters is needed, say $k$, we can apply binary search in order to determine the value of $\alpha$ that produces the number of cluster closest to $k$. When a hierarchy of clusters is needed (see section 4.3.2), the results of clustering using multiple values of $\alpha$ must be merged.

The basic clustering algorithm, as in [Fla2004], is shown in Figure 4-1 below.

```
CutClustering_Algorithm(G(V, E), α)
    Let V' = V ∪ t
    For all nodes v ∈ V
        Connect t to v with edge of weight α
    Let G'(V', E') be the expanded graph after connecting t to V
    Calculate the minimum-cut tree T' of G'
    Remove t from T'
    Return all connected components as the clusters of G
```

**Figure 4-1 Cut clustering algorithm**

### 4.3.2 Hierarchical version

The hierarchical cut clustering algorithm provides a means to look at graph $G$ in a more structured, multi-level way [Fla2004]. In contrast to the greedy hierarchical clustering algorithms, discussed in 3.2.2.2 and used in the state of the art reverse architecting approaches ([Maqb2007], [Czi2007]), the hierarchality of clusters produced by Flake-Tarjan clustering algorithm follows from the nesting property proven in [Fla2004], namely, clusters produced using lower values of $\alpha$ are always supersets of clusters produced at higher values of $\alpha$ in the basic cut clustering algorithm

The hierarchical cut clustering algorithm of [Fla2004] is given in Figure 4-2 below. The authors propose to contract clusters produced with higher values of $\alpha$ before running the algorithm on smaller values of $\alpha$. However, this puts a constraint on the order in which we can try different values of $\alpha$. If we want to try smaller values of $\alpha$ first, e.g. because we are limited in time and want to get more high-level views on the software system first, instead of contracting the input graph we should rather be able to merge clustering obtained at an arbitrary $\alpha$ into a globally maintained clustering hierarchy, as devised within our project and described in section 5.4 of the thesis.

```
HIERARCHICAL_CUTCLUSTERING(G(V, E))
  Let G^0 = G
  For (i = 0; ; i++)
    Set new, smaller value α_i /* possibly parametric */
    Call CutCluster_Basic(G^i, α_i)
    If ((clusters returned are of desired number and size) or
        (clustering failed to create nontrivial clusters))
          break
    Contract clusters to produce G^{i+1}
  Return all clusters at all levels
```

**Figure 4-2 Hierarchical cut clustering algorithm**

## 4.4 Call Graph extraction

A dynamic call graph is a record of an execution of the program, e.g., as output by a profiler. Thus, a dynamic call graph can be exact, but only describes one run of the program. A static call graph is a call graph intended to represent every possible run of the program. The exact static call graph is undecidable, so static call graph algorithms are generally overapproximations. That is, every call relationship that occurs is represented in the graph, and possibly also some call relationships that would never occur in actual runs of the program. Below are some examples of the difficulties encountered when generating call graph from source code (static):

- polymorphism: depending on the class of object assigned to a variable of the base class, different methods are called
- invariants: if in the code below x >= 0 always, then the call to func2() actually never occurs:
  - if(x < 0) { func1(x); } else { func2(x); }
- contextuality: in the example above, we can consider the reasons for x to be negative or non-negative, and mark in the call graph the fact that either func1() or func2() can be called from the current function depending on the context.

So, both dynamic and static call graph generation have drawbacks:

- static: the call graph is imprecise
- dynamic: we need many runs to ensure that the source code is covered enough

In our prototype there is a point at which the program does not care whether static or dynamic call graph is supplied. In principle, we accept any graph of relations on input without binding to a programming language or static/dynamic kinds of analysis. In the experiments within this project, however, we used static call graph extracted from source code in Java using Soot [Soot1999] and the approaches for virtual method call resolution available within the framework: [Sun1999], [Lho2003].

## 4.5 The Problem of Utility Artifacts

Not all component dependencies have the same level of **importance**. This applies particularly to utility components which tend to be *called by many* other components of the system, and as such they encumber the structure of the system without adding much value to its understandability [Pir2009]. A research about the properties of utility artifacts [HaLe2004] concluded that:

- Utilities can have different scope, i.e. not only at the system level.
- Utilities are often packaged together, but not necessarily
- Utilities implement general design concepts at a lower level of abstraction than those design concepts

The common practice for detecting utilities is to use heuristics that are based on computing a component's fan-in and fan-out. The rationale behind this is that [HaLe2004]:

- something that is called from many places is likely to be a utility, whereas
- something that itself makes many calls to other components is likely to be too complex and too highly coupled to be considered a utility.

An exhaustive review of the existing reverse architecting approaches based on clustering and the ways they detect and remove utilities is given in [Pir2009]. Among these approaches are [Man1998] / [Bunch1999], [Mull1990], [Wen2005], [Pate2009]. In [ACDC2000] they used somewhat different approach. As in the first phase of ACDC algorithm they simulate the way software engineers group entities into subsystems, the authors observed and used the fact that software engineers tend to group components with large fan-in into one cluster: support library cluster containing the set of utilities of the system.

To our knowledge, all the existing reverse architecting approaches that address the problem of utility artifacts at all, detect and remove utility artifacts from further analysis. In our project we devise and implement weighting of relations according to their chance to be utility calls/dependencies, and the theory is given in section 5.1. In [Roha2008] they do use weighting according to utility measures developed within that work, however, that weighting applies to components (vertices of the graph) in contrast to edges (relations) in our project. Furthermore, they do not run clustering after weight assignment.

The major technique used for detection of utility artifacts is **fan-in** analysis, where the variations are based on the exploration of the component dependency graph built from static analysis of the system [Pir2009]. Dependencies include method calls, generalization, realization, type usage, field access, and others. Some approaches represent the cardinality of dependencies with weights on the edges of the dependency graph, e.g. [Stan2009]. The rationale behind using fan-in analysis as indication of the extent to which an artifact can be considered utility is as follows: the more calls a component has from different places (i.e. the more incoming edges in the static component graph), then the more purposes it likely has, and hence the more likely it is to be a utility, and the researchers currently converge on this rationale [Mull1990], [Pate2009], [HaLe2006], [Roha2008].

The weak points of the approaches that attempt to solve the problem of utility **detection** are listed in [Pir2009]. Those approaches use evidently more complicated fan-in analysis than we do in this project, sometimes they even combine fan-in with fan-out analysis. However, the strong point of our approach is that the solution of utility artifacts problem is **shared** between pre-clustering phase and the clustering itself, see section 5.1. At the pre-clustering phase we estimate "**utilityhood**" of software engineering artifacts and relations. Then the clustering phase smoothes the likely utility connections because those connections are assigned low weight in the pre-clustering (normalization) phase.

Though [HaLe2006] experimented with combination of fan-in and **fan-out** analysis in order to determine the extent to which a component can be considered a utility, their metrics were only able to detect system-scope utilities. We argue that this issue was encountered because the authors were trying to solve the problem of *detection*, thus they had to introduce a **threshold** in order to make a decision. But thresholds differ for the whole system and for utilities in local subsystems; furthermore, local utilities do not necessary have the same decision threshold across different subsystems.

A counterexample against fan-in analysis alone was given in [Roha2008], we show it too in Figure 4-3 below. It is arguably whether C2 is a utility indeed, but C3 apparently is, according to utility rationale discussed above. However, functions usually call only functions at lower level levels of abstraction, thus, a utility function either does not call any others or calls mostly utility functions [HaLe2006], [HaLe2004]. Thus, C2 is likely to be a utility too. However, fan-in analysis alone would not detect it as such.

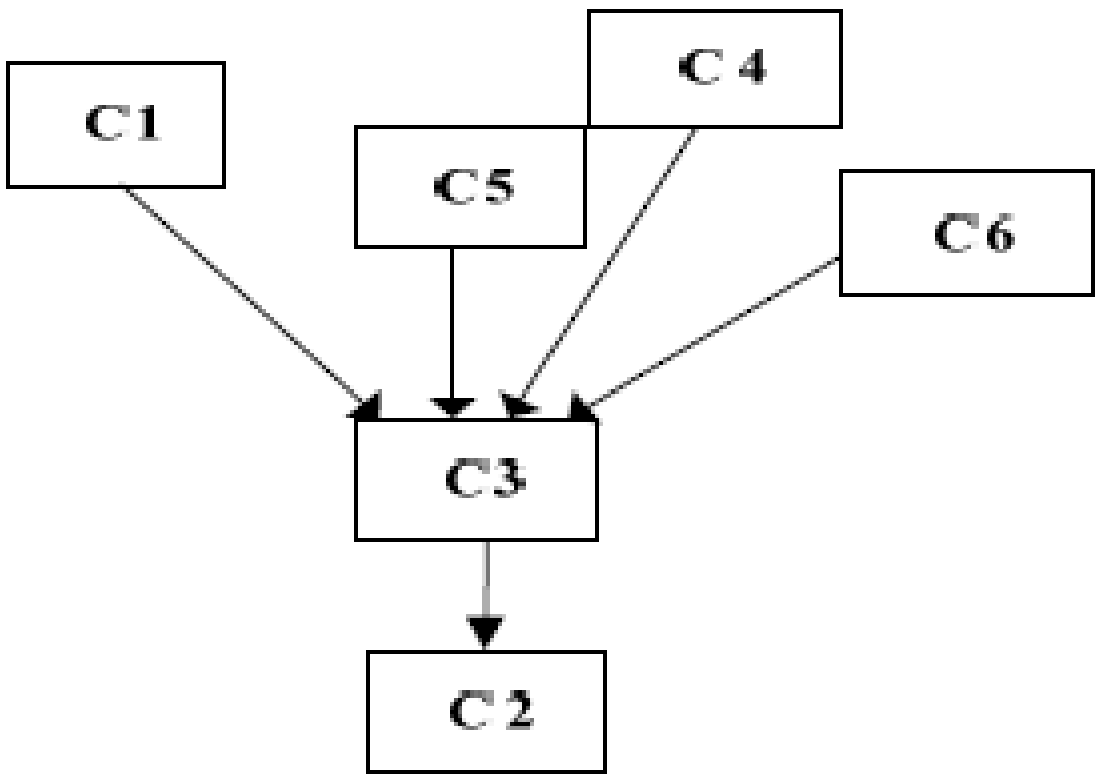

**Figure 4-3 Likely utility C2, but with low fan-in**

The above example is not a problem for our approach, as the connection weight between C3 and C2 stays strong (see section 5.1), thus C3 will be clustered with C2 first (in the bottom of the clustering hierarchy or, in other words, at a high value of parameter alpha, see [Fla2004] and section 4.3.2). Thus if some magic oracle (which we do not have explicitly in our approach) deems C3 to be a utility, then C2 will be the closest to it in terms of unity of purpose, according to the clustering results. Afterwards it is not hard to infer that C2 is a utility too. In our work we also give a **counterargument** *against fan-**out*** analysis, which arises in practice due to impreciseness of call graph extraction (discussed in sections 3.1 and 4.4) and the kind of errors the state of the art call graph analyses make, namely, due to polymorphism there appear excessive calls to multiple derived classes (subtypes) in the call graph that never occur in practice. We observed this in our experiments (appendix 10.3.2, also 10.3.1) and provide our argument in section 5.1.3 below.

Finally, it makes sense to mention that [HaLe2004] identify (in their reasoning, not automatically) different kinds of utilities:

- Utilities derived from the usage of a particular programming language. An example is a class that implements `Enumeration` interface in Java.
- Utilities derived from the usage of a particular programming paradigm. For example, accessor methods or initializing functions
- Utilities that implement data structures (inserting, removing, sorting)
- Mathematical functions
- Input/Output Operations

Our results show (see appendix 10.1.7) that not only utility artifact problem has been alleviated, but also utility artifacts are categorized according to their purpose, likewise the other artifacts.

## 4.6 Various Algorithms

A number of classical algorithms in computer science were used in order to implement this project, and thus appear in this paper. The most important of them were discussed earlier in this section. Below we give a short list and remarks about the rest. A reader that needs more background can refer to [CLR2003], [AHU1983] and [Knu1998].

| Algorithm | Remarks |
|---|---|
| Breadth & Depth First Searches | |
| Priority Queue | |
| Priority Blocking Queue | Java |
| Minimum Spanning Tree | |
| Tree Traversals, Metrics & Manipulations | Height, depth, cardinality, etc. |
| Lowest Common Ancestor | |
| Disjoint-set data structure / union-find algorithm | |
| Reindexing Techniques | Graph contraction<br>Subgraph/subset processing |
| Reusable full-indexing map | Insertion/Removal: O(1)<br>Creation: O(nIndices)<br>Listing: O(nStoredItems) |
| Dynamic Programming | For the statistics |
| Suffix Tree | For the statistics |

# 5 Theory

Within this work we have devised and used some theory needed in order to:

- Apply the clustering method of [Fla2004] to the source code analysis domain
- Solve the issues encountered during the application, namely, excessive number of sibling nodes (aka alpha-threshold). This happens due to the specifics of the domain, namely, software system are usually nearly-hierarchical, i.e. there are few (ideally, no) cycles.
- Optimize the search direction in order to get the most important solutions as early as possible during the iterative runtime of the hierarchical clustering algorithm
- Allow parallel computation, as the clustering process still takes considerable time

The following subsections provide this theory in the amount necessary to implement our system. Some proofs and empirical evaluations require considerable efforts and are thus left out of our scope.

We represent the source code of software as a directed graph $G(V,E)$, with $|V| = n$ vertices and $|E| = m$ edges. Each vertex corresponds to a software engineering artifact (usually, a class of object-oriented languages, e.g. Java class) and each directed edge to a relation between software engineering artifacts, e.g. method call or class inheritance. Also, we usually assume that $G$ is connected, as otherwise each component can be analyzed separately unless some global restrictions on clustering granularity are posed.

## 5.1 Normalization

In section 4.5 we have discussed the problem of utility artifacts. Our practical experiments have confirmed that with Flake-Tarjan clustering algorithm also produces degenerate results in case we cluster the graph of relations “as is”, i.e. in case each relation corresponds to an edge of weight 1 in the input graph for clustering.

Moreover, the graph of relations between software engineering artifacts is directed, however Flake-Tarjan clustering is restricted to **undirected** graphs due to the underlying minimum cut tree algorithm [GoHu1961], which is only known for undirected graphs even though its own underlying algorithm, max flow [Gol1998], is available for both directed and undirected graphs.

Extending Flake-Tarjan clustering algorithm to work with directed graphs is both hard theoretical and risky task (see section 8). Thus within this project we decided to convert directed graph into undirected by means of normalization. The authors of [Fla2004] in their experiments used normalization similar to the first iteration of HITS [HITS1999] and to PageRank [Brin1998]. Each node distributes a constant amount of weight over its out-bound edges. The fewer pages a node points to, the more influence it can pass to its neighbors that it points to. In their experiment with CiteSeer citations between documents [Fla2004], the authors normalize over all outbound edges for each node (so that the total sums to unity), remove edge-directions, and combine parallel edges. Parallel edges resulting from two directed edges are resolved by summing the combined weight.

However, in the domain of software source code, it seems more reasonable to normalize over the incoming arcs rather than outgoing, so that each node receives a constant (or logarithmic) amount of weight from its incoming edges. This is grounded in [Pir2009], as they review many works on utility artifact detection and point the fact that utility functions are called by many other components as a main property. In the literature on reverse architecting the exploitation of this property is called **fan-in** analysis, discussed in section 4.5 of the thesis too.

The crucial difference in how our approach addresses the problem of utility artifacts, compared to other existing approaches (see section 4.5), is in the following. The existing approaches focus on *detection* of utility artifacts with the goal of further removal prior to clustering. Our solution for the utility artifacts issue is split between pre-clustering and clustering phases, thus we are **not** concerned with the problem of detection, which would entail further binary decision on whether to remove an artifact before clustering.

### 5.1.1 Directed Graph to Undirected

Our conversion from directed to undirected graph works as follows. We discard arcs that loop a vertex to itself. Apparently, we lose some information in this step, namely, the fact that the corresponding SE artifact references (calls, uses, etc) itself. However, we do not see a way to make use of this information without damaging clustering quality. The latter was observed in our experiments.

For each vertex $j$, its fan-in is calculated as the sum of weights of all the incoming arcs in the initial graph:

$$S_j = \sum_i w_{i,j}$$

Each arc in the graph is then replaced with a normalized arc having weight $\tilde{w}_{i,j}$, which without leverage (section 5.1.2 below) amounts to:

$$\tilde{w}_{i,j} = \frac{w_{i,j}}{S_j}$$

In the target undirected graph, an edge between vertices $i$ and $j$ receives weight $u_{i,j}$ equal to the sum of the weights of the opposite-directed arcs:

$$u_{i,j} = u_{j,i} = \tilde{w}_{i,j} + \tilde{w}_{j,i}$$

Let's define $U_j$ as the total weight of edges adjacent to vertex $j$ in the target undirected graph (adjacent weight):

$$U_j = \sum_i u_{i,j} = \sum_i \left( \frac{w_{i,j}}{S_j} + \frac{w_{j,i}}{S_i} \right)$$

The following properties can be observed:

- Each vertex $j$ receives a constant (C=1) amount of weight via arcs $\tilde{w}_{i,j}$
- The total weight of edges adjacent to a vertex in the undirected graph can be both more or less than $C$
- If a vertex has at least one incoming arc in the directed graph, it will have adjacent weight at least $C$ in the undirected graph
- In practice, there are seldom SE artifacts that only use others, but are not used from any place thus do not have an incoming arc. Thus in most cases: $U_j \geq 1$
  - Exclusions: artifacts that are called externally (e.g. thread entry points or contexts launched from Spring framework) may not have any incoming arcs when the input graph does not contain relations of the whole program (e.g. the code of system libraries or Spring framework is not available). For such artifacts, it can happen that $U_j < 1$
  - It seems that $U_j \geq 1$ (or equivalent for the leveraged counterpart from section 5.1.2) can be the (partial) cause of alpha-threshold issue observed, see section 5.6

If vertices are SE methods, and edges are method calls, then a vertex has high adjacent weight when the corresponding method calls many methods infrequently called from other places. Frequency is calculated by the number of occurrences in the source code.

Obviously, such a conversion can be performed in $O(|E|\log|V|)$ scalar operations by means of two passes through all the arcs of the graph: first, calculate the values of $S_j$; second, calculate the weight $\tilde{w}_{i,j}$ for each arc and combine it with $\tilde{w}_{j,i}$ (if this opposite arc is present) using balanced trees of incident vertices for each vertex. Practically, we use hash maps here.

### 5.1.2 Leverage

In the previous section we gave formulas that force vertices to receive constant amount of weight, i.e. for each vertex $j$:

$$\sum_i \tilde{w}_{i,j} = 1 = C$$

However, it seems not reasonable to discard the cardinality of references to an SE artifact completely. Thus we use logarithmic leverage of the bound on the weight that a vertex can receive from the incoming arcs. In this case, the values of the discounted arc weight $\tilde{w}_{i,j}$ and the adjacent weight for a vertex $U_j$ are instead calculated as follows:

$$\tilde{w}_{i,j} = \frac{w_{i,j}}{S_j} \cdot \log S_j$$

$$u_{i,j} = u_{j,i} = \tilde{w}_{i,j} + \tilde{w}_{j,i}$$

$$U_j = \sum_i u_{i,j} = \sum_i \left( \frac{w_{i,j}}{S_j} \log S_j + \frac{w_{j,i}}{S_i} \log S_i \right)$$

In our view, the usage of leveraged estimation of connection strength, as described above, pursues (and, empirically, achieves) the following objectives:

- Alleviate the problem of utility artifacts (also characteristic for the normalization described in section 5.1.1)
- Regard the scale of connectedness rather than the magnitude. E.g., the scale of difference between 2 and 4 connections is the same as the one between 100 and 200. The former distinguishes between more and less coupled high-level (in other interpretation, specific) SE artifacts. The latter distinguishes between more and less omnipresent utility (general-purpose) artifacts.

By looking at the clusterings produced with and without leverage, and comparing some inherent indicators, namely

- the range of parameter alpha between single-cluster and all-singleton-clusters results of partitional clustering [Fla2004]
- the number of excessive sibling clusters in the clustering hierarchy due to alpha-threshold (also, see section 5.6)

Though we are not able to provide a comparison in percentage, as evaluation of clustering quality is not a straightforward task in itself, from the experiments, indicators as described above and subjective evaluation of the resulting clustering hierarchy we conclude that leverage improves clustering quality for this algorithm [Fla2004] in software source code domain.

Comparing to the literature, we can observe that some kind of logarithmic leverage is used in other reverse architecting approaches [HaLe2006], [Roha2008]. The utilityhood metric of [HaLe2006] consists of two factors multiplied: one is the fan-in based ratio (straight division of the fan-in cardinality by the number of artifacts), another is based on fan-out with logarithmic leverage. Their rationale is that fan-in is much more important than fan-out, however, fan-out should also play role in the utilityhood of an artifact (we have discussed this too in section 4.5). [Roha2008] approves this rationale and adopts a derived approach for estimating the impact of component modification in their TWI (Two Way Impact) metric. However, the logarithmic multiplier still stays in the part responsible for fan-out (in the terms of [Roha2008], it is Class Efferent Impact) and fan-in is still represented by a direct ratio of cardinalities, in contrast to our approach.

### 5.1.3 An argument against fan-out analysis

In the existing reverse architecting approaches, e.g. [Roha2008] and [HaLe2006], they use fan-out analysis in addition to fan-in. By doing this they attempt to make use of the second part of the rationale for utility detection (we discussed it in section 4.5), namely: something that itself makes many calls to other components is likely to be too complex and too highly coupled to be considered a utility [HaLe2004], [Pir2009].

However, in case the underlying data for fan-in/fan-out analysis is a call graph extracted from source code of a program in object-oriented language, we can observe that such a relation graph has excessive outgoing arcs, which are noise (illustrated in section 10.3.2). This happens due to impreciseness of call graph extraction (section 4.4). Though the existing heuristics for call graph extraction ([Sun1999], [Bac1996], [Lho2003]) can alleviate this problem, they cannot eliminate it and we are still getting vertices with excessive fan-out.

Thus in practice we argue against fan-out analysis. Though we agree that a component that makes many calls is likely to be complex and highly coupled, thus utilityhood of such a component should be discounted with respect to a metric inferred from pure fan-in analysis, we can only do this when our graph of relations does not have excessive outgoing arcs, i.e. in theory or in dynamic analysis. In practice of static analysis, for each polymorphic call site there is usually only a single or a few calls to some most specific subtypes that actually occur and are designed by software engineers, and the rest is noise. Thus, by discounting the utilityhood for the components containing such call sites due to high fan-out, we would propagate the mistake. We suppose to achieve more **noise-tolerant** solution by not using fan-out analysis (at least, in the form of ratio or logarithmic multiplier).

### 5.1.4 Lifting the Granularity

The input graph contains relations between SE artifacts of different granularity. There are method to method, method to field, method to class and class to class relations. Analyzing Java programs, we generalize any less than class-level artifacts as members (nested classes do not fall into this category), see section 5.2. In this project we experimented with only class-level artifact clustering. Thus we have to lift the relations involving less than class-level artifacts to the class-level, i.e. lift the granularity to class level. An alternative is given in section 5.1.5. For the approaches and a discussion about lifting the component dependencies in general one can refer to [Kri1999].

In couple with the normalization that we are discussing in section 5.1, we see two principal options for lifting the granularity:

1 Before normalization
2 After normalization

Adopting the first option, we would first aggregate all the arcs in the initial directed graph $\boxed{\tilde{G}}$ which connect members of the same class, and connect the vertices (which represent SE classes) of a derived graph $\boxed{G}$ with arcs having the aggregated weights. This corresponds to the lifting of [Kri1999]. In the next step we would normalize the directed graph $\boxed{G}$ as described in our previous sections.

Adopting the second option, we attempt to tolerate the noise in the input graph $\boxed{\tilde{G}}$ and improve the quality of our heuristic addressing utility artifact problem. An alternative solution pursuing the same goal is proposed in 5.1.5. Below is the rationale for the heuristic that show in this section and choose to implement in our project. An error in utilityhood estimation for a single member artifact is smoothed by utilityhood estimations for the rest of artifacts which are members of the same SE class, in case lifting to class-level occurs after normalization. We empirically observed that, indeed, option 2 leads to better clustering results than option 1. This is further confirmed by the indicators intrinsic to the clustering method used: alpha-threshold and excessive number of sibling clusters (see section 5.6).

Formally, consider

- $\tilde{V}$ is the initial set of vertices where a vertex can correspond to both a member-level and a class-level SE artifact,
- the heterogeneous relations between SE artifacts in the initial directed graph $\tilde{G}$ constitute its set of arcs $\tilde{E}$ whereas arc from vertex $i$ to vertex $j$ has weight $a_{i,j}$,
- $V$ is a subset of $\tilde{V}$ consisting of class-level SE artifacts, and a class-level artifact is never also a member-level artifact,
- the membership relations are defined by mapping $M : \tilde{V} \to V$, where membership relations are only defined from a member-level artifact to a class-level artifact and, for convenience, each class-level artifact maps to itself,

If we lift the granularity prior to normalization, we get undirected graph $G_1$ with edge weights $\overset{(1)}{u}_{i,j}$ and properties (section 5.1.1) as follows:

$$\overset{(1)}{w}_{i,j} = \sum_{\forall k,l | M(k)=i, M(l)=j} a_{k,l} \; ; \quad \overset{(1)}{S}_j = \sum_{i \in V} \overset{(1)}{w}_{i,j} \; ; \quad \overset{(1)}{\tilde{w}}_{i,j} = \frac{\overset{(1)}{w}_{i,j}}{\overset{(1)}{S}_j} \; ;$$

$$\overset{(1)}{u}_{i,j} = \overset{(1)}{u}_{j,i} = \overset{(1)}{\tilde{w}}_{i,j} + \overset{(1)}{\tilde{w}}_{j,i}$$

Merging the formulas in order to demonstrate the intuition about the resulting weights in the undirected graph, we get:

$$\overset{(1)}{u}_{i,j} = \overset{(1)}{u}_{j,i} = \frac{\sum_{k \in M^{-1}(i), l \in M^{-1}(j)} a_{k,l}}{\sum_{i \in V} \sum_{k \in M^{-1}(i), l \in M^{-1}(j)} a_{k,l}} + \frac{\sum_{k \in M^{-1}(i), l \in M^{-1}(j)} a_{l,k}}{\sum_{j \in V} \sum_{k \in M^{-1}(i), l \in M^{-1}(j)} a_{l,k}}$$

**Figure 5-1 Lift, then normalize**

In the above formula, $M^{-1}(i)$ is the inverse for mapping $M$. Namely, $M^{-1}(i)$ is the set of members of SE class $i$. Formally:

$$M^{-1}(i) = \{k | M(k) = i\}$$

It is easy to notice that the iteration in both denominators in Figure 5-1 occurs over all the vertices of graph $\tilde{G}$, thus:

$$\overset{(1)}{u}_{i,j} = \overset{(1)}{u}_{j,i} = \frac{\sum_{k \in M^{-1}(i), l \in M^{-1}(j)} a_{k,l}}{\sum_{k \in \tilde{V}, l \in M^{-1}(j)} a_{k,l}} + \frac{\sum_{k \in M^{-1}(i), l \in M^{-1}(j)} a_{l,k}}{\sum_{k \in M^{-1}(i), l \in \tilde{V}} a_{l,k}}$$

**Figure 5-2**

Now let's regard the formulas for the second option, where we first normalize and then lift the granularity. We get undirected graph $G_2$ with edge weights $\boxed{\overset{(2)}{u}_{i,j}}$ and properties as follows:

$$S_l = \sum_{k \in \tilde{V}} a_{k,l} \;; \quad \tilde{a}_{k,l} = \frac{a_{k,l}}{S_l} \;; \quad b_{k,l} = b_{l,k} = \tilde{a}_{k,l} + \tilde{a}_{l,k}$$

$$\overset{(2)}{\tilde{w}}_{i,j} = \sum_{\forall k,l \mid M(k)=i, M(l)=j} \frac{a_{k,l}}{S_l} = \sum_{l \in M^{-1}(j)} \frac{\sum_{k \in M^{-1}(i)} a_{k,l}}{\sum_{k \in \tilde{V}} a_{k,l}}$$

$$\overset{(2)}{u}_{i,j} = \sum_{\forall k,l \mid M(k)=i, M(l)=j} b_{k,l}$$

**Figure 5-3 Normalize, then lift**

To compare the outcome over both options, $\boxed{\overset{(1)}{u}_{i,j}}$ and $\boxed{\overset{(2)}{u}_{i,j}}$, let's bring the formula in Figure 5-3 into a similar to Figure 5-2 presentation:

$$\overset{(2)}{u}_{i,j} = \sum_{\forall k,l \mid M(k)=i, M(l)=j} \left( \frac{a_{k,l}}{\sum_{k \in \tilde{V}} a_{k,l}} + \frac{a_{l,k}}{\sum_{l \in \tilde{V}} a_{l,k}} \right) \Rightarrow$$

$$\overset{(2)}{u}_{i,j} = \overset{(2)}{u}_{j,i} = \sum_{l \in M^{-1}(j)} \frac{\sum_{k \in M^{-1}(i)} a_{k,l}}{\sum_{k \in \tilde{V}} a_{k,l}} + \sum_{k \in M^{-1}(i)} \frac{\sum_{l \in M^{-1}(j)} a_{l,k}}{\sum_{l \in \tilde{V}} a_{l,k}}$$

**Figure 5-4**

Whereas formula for the first option, $\boxed{\overset{(1)}{u}_{i,j}}$, from Figure 5-2 can be rewritten as:

$$\overset{(1)}{u}_{i,j} = \overset{(1)}{u}_{j,i} = \frac{\sum_{l \in M^{-1}(j)} \sum_{k \in M^{-1}(i)} a_{k,l}}{\sum_{l \in M^{-1}(j)} \sum_{k \in \tilde{V}} a_{k,l}} + \frac{\sum_{k \in M^{-1}(i)} \sum_{l \in M^{-1}(j)} a_{l,k}}{\sum_{k \in M^{-1}(i)} \sum_{l \in \tilde{V}} a_{l,k}}$$

**Figure 5-5**

If we fix $\boxed{i}$ and let $c_n = \sum_{k \in M^{-1}(i)} a_{k,l_n}$ and $C_n = \sum_{k \in \tilde{V}} a_{k,l_n}$, where $\{l_n\} = M^{-1}(j)$,

we can observe that the left summands of formulas Figure 5-4 and Figure 5-5 (and by analogy, the right summands too) relate to each other as:

$$\overset{(1)}{\tilde{w}}_{i,j} \vee \overset{(2)}{\tilde{w}}_{i,j} \Leftrightarrow \frac{c_1 + c_2 + \ldots + c_N}{C_1 + C_2 + \ldots + C_N} \vee \frac{c_1}{C_1} + \frac{c_2}{C_2} + \ldots + \frac{c_N}{C_N}$$

**Figure 5-6 Comparison of the undirected weights**

We hope this can further be used for formal study of the effect of **noise**, but leave this out of the scope of this paper. Call graph extraction methods put some noise ([Bac1996], [Sun1999], [Lho2003], also section 3.1) into the resulting graph, either by adding calls which never occur, or drop some calls which however may occur (section 4.4). In general, we can designate this noise as $0 \le \varepsilon_{k,l} \le 1$ for each arc $a_{k,l}$ in the input graph of relations, meaning that there are $(1-\varepsilon_{k,l}) \cdot a_{k,l}$ "true" calls and there are $\varepsilon_{k,l} \cdot a_{k,l}$ "false" calls.

### 5.1.5 An Alternative

In principle we could, without lifting the granularity, normalize and then run Flake-Tarjan hierarchical clustering algorithm over heterogeneous graph consisting of both member- and class-level SE artifacts as vertices, and heterogeneous relations between SE artifacts as edges. We could try to lift the granularity from member-level to class-level after clustering of this graph has been performed.

We argue that this solution can produce a better clustering hierarchy, in terms of how well it reflects the actual decomposition of the software system, because, compared to the solution of section 5.1.4, less information is lost prior to clustering. Namely, the information loss occurs in the following:

- By aggregating edge weights over all the members of a SE class, we get a single (if any) edge (relation) between any two SE classes.
- A SE class becomes connected to other SE classes with edges, where for each edge its weight represents the connection strength between the two classes.
- However, some members of an SE class, vertex $v$, in the initial graph might be more connected with members of one SE class, vertex $v_1$, and the other members of that class $v$ might be more connected with members of another SE class $v_2$.
- An example of two cases which clustering will not be able to distinguish due to this information loss is illustrated in Figure 5-7 below.

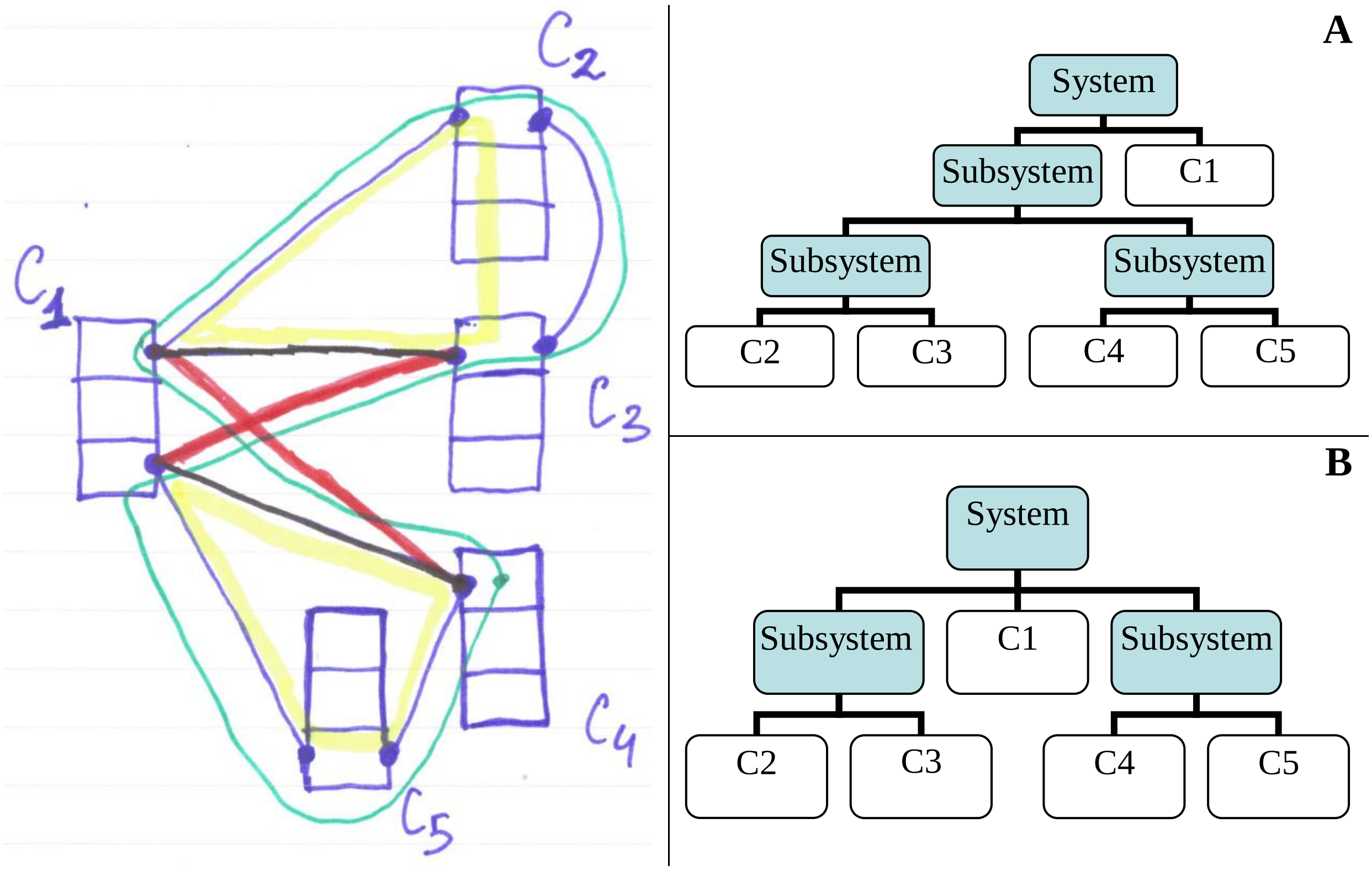

**Figure 5-7 A counterexample to lifting the granularity prior to clustering**

Consider a system of 5 classes (rectangles) containing 3 members (adjacent squares) each. The member-level relations (edges), which should be considered present:

- in both cases are drawn in **blue**;
- only in case A – in **black**;
- only in case B – in **red**.

It is obvious that in case A the graph contains 2 separate cycles, drawn in yellow, however, in case B there is a single cycle traversing all the five SE classes, drawn in **green**. The above is drawn on the left side of Figure 5-7. It is reasonable that case A and case B determine different decompositions of the system, illustrated in the top and in the bottom of the right side of Figure 5-7 correspondingly. The difference is that in case **B** (single cycle, bottom diagram) the classes $\{C_2, C_3, C_4, C_5\}$ do not constitute a subsystem without class $C_1$, even though $\{C_2, C_3\}$ and $\{C_4, C_5\}$ constitute subsystems of which the whole system can be composed by adding $\{C_1\}$. On the other hand, in case **A** (two cycles, upper diagram) classes $\{C_2, C_3, C_4, C_5\}$ constitute a subsystem, which is a combination of two disjoint subsystems. In practice, fact "disjoint" will be replaced with "loosely-connected" (in terms of connection density, i.e. do not confuse with weakly connected components in a directed graph), and instead of the criterion of a "connected component", the criterion of a "cluster" will be used.

An apparent disadvantage of member-level clustering is the computational complexity. In our experiments we observed 14.5 times more members than classes usually. However, in addition to this disadvantage, it is not clear on how to lift the granularity to class-level after clustering at member-level. Each class contains several members, each its member will appear somewhere in the member-wise clustering hierarchy. How to arrange the classes into a hierarchy then, having the data on where their members appear in the member-wise hierarchy?

One approach is (weighted) voting. However, a problem arises: member-wise partitional clusterings will have the nesting property ([Fla2004], also section 4.3.2), but after voting it is most likely to be lost at class-level. At this point many *options* arise for solving this problem, e.g.

1. For each pair of classes, count how many times their members form pairs through appearing together and at each level of the member-wise hierarchy. We get a sparse matrix of counts, perhaps also weighted by depth/height of the node, at which pairs were encountered. We can now run a clustering algorithm on this new matrix as edge weights, perhaps with some normalization. This solution seems to be also vulnerable to the issue displayed in Figure 5-7, though less than a solution that losses member-wise relation information in the very beginning.
2. Start building a new tree. Let each class node to appear at the position where the (weighted) majority of its members has appeared in a subtree of the member-wise hierarchy. This solution is prone to non-deep hierarchies with excessive number of sibling nodes, and the latter would hinder comprehensibility.

Due to practical difficulties (risen computational complexity) and many reasonable options without a single good theoretical option, we did not develop this alternative within the current project further.

## *5.2 Merging Heterogeneous Dependencies*

The phase of extraction of relations between software engineering artifacts can produce various kinds of relations. In this project we used:

1. Method-to-method calls
2. Class-to-class inheritance
3. Method-to-class field access
4. Method-to-class type usage: a method has statements with operands of that class)
5. Method-to-class parameter & return values: a method takes parameters or returns values of which are instances of a certain class

Note that the kind 5 is not exhausted by kind 4, as e.g. methods in an interface do not have bodies, thus do not have statements.

Now the question is how to consider the various kinds of relations for the inference of software structure. We see the following principal options.

**Option 1:** Clusterize the graphs of homogenous relations separately, i.e. one graph per one kind of relation. Then combine the resulting multiple hierarchies into a single one. This solution has the same root disadvantage as the one discussed in 5.1.5, namely, it is not clear on how to merge the hierarchies.

The challenge of nearly-hierarchical input data for clustering in software engineering domain is discussed in section 5.6. Thus, another disadvantage arises from the fact that a graph representing a single kind of relation is even more nearly-hierarchical than a graph combining multiple relations (dependencies) between SE artifacts. For example, inheritance always forms a directed acyclic graph (DAG) of relations. In Java programming language, if we consider only classes (not interfaces, i.e. only “extends” but not “implements” kind of inheritance), it is always a tree. In C++ it can still be a DAG.

**Option 2:** Combine the multiple graphs into a single prior to clustering. This has the same disadvantage comparing to option 1 as discussed in section 5.1.5 (obviously, a similar counterexample can be given by analogy), namely, loss of information about the kind of relation which an edge in the input graph for clustering represents. On the other hand, an strong side of this option is in the fact that a graph combining multiple kinds of relations is less likely to be nearly a tree, thus this solution alleviates the issue discussed in section 5.6.

In this project we implement option 2, and point out option 1 together with the similar alternative discussed in section 5.1.5 as a direction for further research. We use equal weight for one relation of each type. An improved approach could try to learn the optimal weights by means of training on systems, for which authoritative decompositions are available, comparing its performance using an appropriate metric for nested software decompositions (see [UpMJ2007], [END2004]), and then use the same weights for merging the relations of novel software.

## 5.3 Alpha-search

Basic cut clustering algorithm (section 4.3), given some value of the parameter alpha, produces a partition of vertices into groups, i.e. flat decomposition of the system upon analysis. For smaller values of the parameter alpha, there are fewer groups. For higher value of alpha, there are more groups. The groups have nesting property [Fla2004], i.e. they naturally form a hierarchy. The **exact** hierarchy can be computed by the hierarchical clustering algorithm (section 4.3.2), but this requires running the basic cut clustering algorithm (section 4.3.1) over all the values of parameter alpha producing different number of clusters. There are can be many flow breakpoint alpha-s that can be found fast [Bab2006], of them no more than $|V|-2$ produce different number of clusters.

In our experiments, calculation of clustering for a single alpha was taking 4.5 minutes for 7K vertices, thus it is not feasible to do this operation $|V|-2$ times. In order to produce as much as possible result within limited time, we perform the most important probes first. We used a binary search tree approach, described in the subsections.

### 5.3.1 Search Tree

An initial interval $[\alpha_{\min};\alpha_{\max}]$ is chosen as the root of the tree, such that $\alpha_{\min}$ yields a single cluster and $\alpha_{\max}$ yields many singleton clusters. These bounds can be found with binary search, as proposed in [Fla2004], but in practice we just use values that produce small enough and large enough number of clusters correspondingly.

Each child node in the search tree corresponds to a half of the interval of its parent, so that node $[\alpha_l;\alpha_r]$ will have children $\left[\alpha_l;\frac{\alpha_l+\alpha_r}{2}\right]$ and $\left[\frac{\alpha_l+\alpha_r}{2};\alpha_r\right]$. It is convenient to view the space of alpha values as a tree because in the search algorithm we can then maintain the following invariant:

- At each iteration, there is a tree of alpha-values for which probes (runs of basic cut clustering algorithm) have been already performed
- We can use **any** leaf node as the base for the next probe

Thus the search tree does not have to be balanced. We can do more probes in a more interesting interval (where more fine-grained decomposition of the system will say more to a software engineer), and less probes in another. An illustration of alpha space and search tree is given in Figure 5-8 below. The alpha-interval for each node is denoted with a block arrow.

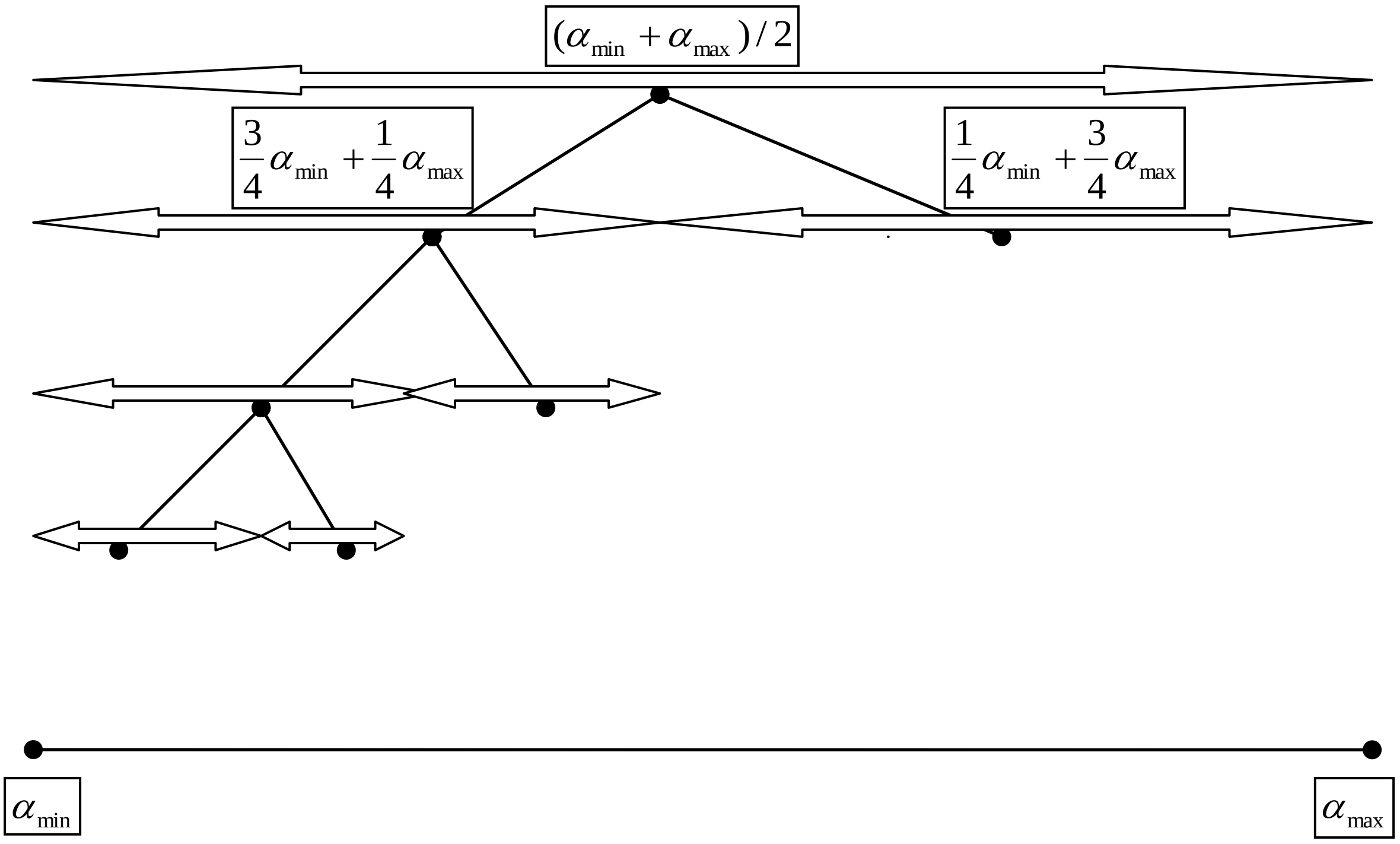

**Figure 5-8 Search Tree and Alpha Space**

Each iteration is an attempt to improve the clustering hierarchy, consisting of the following steps:

- Select a leaf node, and without loss of generality consider its interval is $[\alpha_l; \alpha_r]$
- Calculate the alpha value for the next probe: $\alpha_m = \frac{\alpha_l + \alpha_r}{2}$
- Run the basic cut clustering algorithm using $\alpha_m$
- Add two child nodes into the search tree, corresponding to intervals $[\alpha_l; \alpha_m]$ and $[\alpha_m; \alpha_r]$

A node does not add a left child (or a right child, by analogy) into the search tree in case $k(\alpha_l) = k(\alpha_m)$, where $k(\alpha)$ is the number of clusters produced by the basic cut clustering algorithm using this value of parameter $\alpha$.

All the above gives a base for prioritization described in the following section.

### 5.3.2 Prioritization

In the beginning we put the root node corresponding to the whole interval $[\alpha_{\min}; \alpha_{\max}]$ into a priority queue. It is also a leaf node at this moment, as no child nodes have been added. In the previous section we showed that any leaf node can be taken at each iteration for the next probe. Thus we can maintain invariant that there are only leaf nodes in the priority queue, and chose a reasonable priority function.

Let $k(\alpha)$ be the number of clusters that the basic cut clustering algorithm produces using parameter $\alpha$. Consider we have performed a probe for $\alpha = \alpha_m$ and are now going to push children of node $[\alpha_l;\alpha_r]$, which are $[\alpha_l;\alpha_m]$ and $[\alpha_m;\alpha_r]$, into the queue. Then for child $[\alpha_l;\alpha_m]$ (and by analogy, for child $[\alpha_m;\alpha_r]$) we set the following priority:

$$P_{l,m} = \log(k(\alpha_m) - k(\alpha_l))^2 + \log(\alpha_m - \alpha_l) + \min\{k(\alpha_m) - k(\alpha_l), k(\alpha_r) - k(\alpha_m)\} + 1/\sqrt{\frac{\alpha_l + \alpha_m}{2}}$$

Below is the motivation for each of the summands constituting $P_{l,m}$:

- $\log(k(\alpha_m) - k(\alpha_l))^2$ forces the intervals spanning large number of clusters (from $k(\alpha_m)$ to $k(\alpha_l)$) to be considered earlier.
- $\log(\alpha_m - \alpha_l)$ forces large intervals to be considered earlier. Here "large" refers to the difference of $\alpha$, in contrast to the previous point where the difference of the number of clusters is regarded.
- $\min\{k(\alpha_m) - k(\alpha_l), k(\alpha_r) - k(\alpha_m)\}$ forces the more balanced intervals to be considered earlier. This summand contributes most of all into the priority, as it is a big improvement when we e.g. split an interval of 2000 clusters into parts of 1000 and 1000, rather than 1998 and 2.
- $1/\sqrt{\frac{\alpha_l + \alpha_m}{2}}$ forces to make probes for small values of alpha earlier. The probes at small value of alpha yield decisions about the upper (closer to the root) levels of the clustering hierarchy.

Of multiple priority functions considered, the one given in this section demonstrated the best value-for-time in our experiments.

## 5.4 Hierarchizing the Partitions

The way we merge the partitions produced by the basic cut clustering algorithm differs from the simple hierarchization method described and employed in [Fla2004] because we do not pass all the alpha-s from the highest till the smallest determined by parametric max flow algorithm as flow breakpoints (see [Bab2006]), but instead we run the basic cut clustering using the "most-desired" alpha as determined by our prioritization heuristic (section 5.3). Thus we must be able to merge the outcome of basic cut clustering algorithm (a partition of vertices into clusters) into the globally maintained clustering hierarchy for arbitrary alpha. In this way we allow arbitrary order of passing through the values of parameter alpha.

The need for this ability is further motivated by the intent to compute in parallel (section 5.5). Different processors may compute single-alpha clustering (one run of the basic cut clustering algorithm) with different speed, not only due to the difference in computational power, but also because the running time of basic cut clustering algorithm is, in practice, proportional to the number of clusters in the resulting partition. As we discussed in section 4.2.2, this happens due to the community heuristic described in [Fla2004].

We solve the following problem: given the global clustering tree, and a result of basic cut clustering for $\alpha$ which is not yet in the tree, transform the tree so that it reflects the result of clustering for this new $\alpha$. Formally:

- Let $T$ be the global clustering tree, in which leaf nodes denote SE artifacts and inner nodes denote clusters at different levels of the hierarchy, and the height of the tree is $height(T)$
- Let $par_v$ be the parent node for node $v$ in $T$

- For an inner node $v$, let $chi_v$ be the set of its children nodes in $T$, and let $alpha(v)$ be the value of parameter alpha at which the basic cut clustering algorithm united all the descendants of node $v$ into a single cluster, thus introducing node $v$ into $T$
- In order for each node have a parent, introduce a fake root $fr$ in $T$ having $alpha(fr) < \alpha_{\min}$, where $\alpha_{\min}$ is defined as in section 5.3.1
- Let $C(\alpha) = \{C_i(\alpha)\}$ be the partition of SE artifacts into clusters produced by the basic cut clustering algorithm using the novel value $\alpha$.

We remind that basic cut clustering is a partitional clustering algorithm and produces clusters that have nesting property, i.e. for $\alpha_1 < \alpha_2$ each cluster produced with $\alpha_2$ contains a set of SE artifacts which is a subset of some cluster produced with $\alpha_1$. Formally:

$$\forall \alpha_1 < \alpha_2 \forall C_i(\alpha_2) \exists C_j(\alpha_1) \mid v \in C_i(\alpha_2) \Rightarrow v \in C_j(\alpha_1)$$

**Figure 5-9 Cluster nesting property**

Note, that the above formula forbids a case when there are two vertices $u, v \in C_i(\alpha_2)$ and $v \in C_j(\alpha_1)$ while $u \notin C_j(\alpha_1)$.

Our task is: integrate the clustering result $C(\alpha)$ into the global clustering tree $T$. Now we can define it formally. For each $C_i(\alpha) \in C(\alpha)$, find vertex $p$ in $T$ such that there exists $C_j(alpha(p))$ having $C_i(\alpha)$ as its subset, i.e. $C_i(\alpha) \subset C_j(alpha(p))$, but none of the nodes in the subtree of $p$ satisfies this requirement. Taking into account the formula in Figure 5-9, this task amounts to finding node $p$ such that:

$$alpha(p) < \alpha < alpha(v), \forall v \in chi_p$$

**Figure 5-10 The place to insert new cluster**

The latter can be done in $O(height(T))$ operations by simply scanning the nodes of $T$, starting from any node $u \in C_i(\alpha)$ and testing for match to the criteria in Figure 5-10 above. By noticing that nodes in the path from any node $u$ to the root of $T$ are sorted by alpha, i.e. $alpha(p) < alpha(v), \forall v \in chi_p$, we can apply a binary-search-like approach used in some algorithms for Lowest Common Ancestor (section 4.6). Thus we can reduce this subtask to $O(\log(height(T)))$ operations.

It is now obvious that the algorithmic complexity of merging a novel partition (clustering) into the global cluster tree is:

$$O(|V| + |C(\alpha)| \log(height(T)))$$

In the above formula, $|C(\alpha)|$ is the number of clusters produced by the basic cut clustering algorithm using this value of parameter $\alpha$.

## 5.5 Distributed Computation

In the previous sections, 5.3 and 5.4, we have devised a ground for distributed computing of the hierarchical clustering tree. This is an **improvement** over the hierarchical clustering algorithm of [Fla2004] (also discussed in section 4.3.2), which is limited to sequential processing due to contraction while passing from the larger alpha-s down to the smaller.

The idea is in running multiple basic cut clusterings in parallel, processing one $\alpha$ at a processor. We can notice that basic cut clusterings (partitions) can be computed independently for different alpha-s. After the result for some $\alpha$ has been computed, we must merge it into the clustering hierarchy. In order for the cluster tree to remain consistent, we need synchronization during this merge operation. Then the released processor can take another "most interesting" alpha from the priority queue (section 5.3), and synchronization is required here again in order for the queue and the search tree to remain consistent. We can do distributed computation on as many processors as the number of leaves in the search tree. The number of leaves in the search tree grows fast, as processing of a node usually adds two new leaf nodes for further search. This was implemented within our project, see section 6.

Note that we are splitting each alpha-interval in the search tree into 2 child interval, half of the parent each. We could, however, split the parent interval into 3 or more child intervals, thus producing 3 or more child leaf nodes in the search tree. This makes sense to do when there are very many processors (e.g. a network of computers), thus we want the search tree to grow fast in order for as many as possible processor to get their tasks earlier.

## 5.6 Perfect Dependency Structures

A specific property of data that arises in the domain of software engineering, is nearly-acyclic structure of dependencies among software engineering artifacts. In case this structure stays (locally) acyclic even after conversion from directed to undirected graph (see section 5.1), the clustering algorithm receives on input a tree, which is a **degenerate** case for graph clustering. According to [Sch2007], "*There should be at least one, preferably several paths connecting each pair of vertices within a cluster*". But in a tree there is exactly one path between each pair of vertices.

In case of Flake-Tarjan clustering [Fla2004], a phenomenon undermining clustering quality was observed. We call it **alpha-threshold**, which is in the following:

- Often there is no way to get a certain amount of clusters, say more or less close to $K$.
- Using the notation of section 5.3.1, we formalize this as:

  $$k(\alpha_t - \varepsilon) << K << k(\alpha_t + \varepsilon), \forall \varepsilon \to 0$$

- In the other words, any alpha less than $\alpha_t$ yields a significantly smaller number of clusters than $K$, while any alpha greater than $\alpha_t$ yields a significantly larger number of clusters than $K$.

Let $\alpha_l$ be the greatest alpha yielding a number of clusters smaller than $K$, and $\alpha_r$ be the smallest alpha yielding a number of clusters larger than $K$. Then we can rewrite the phenomenon as:

$$\left.\begin{array}{l} k(\alpha_t - \varepsilon) << K << k(\alpha_t + \varepsilon), \forall \varepsilon \to 0 \\ \alpha_l < \alpha_t < \alpha_r \end{array}\right\} \Rightarrow$$

$$k_l = k(\alpha_l) << K << k(\alpha_r) = k_r$$

**Figure 5-11 Alpha-threshold**

It is now easy to notice that in the cluster tree (section 5.4) alpha-threshold can imply a parent node, i.e. a cluster produced by the basic cut clustering at $\alpha_l$, having an excessive number of children, while every child corresponds to a cluster produced at $\alpha_r$. All the $k_r - k_l$ child clusters do not need to have the same parent however, as demonstrated in a counterexample, see Figure 5-12 below:

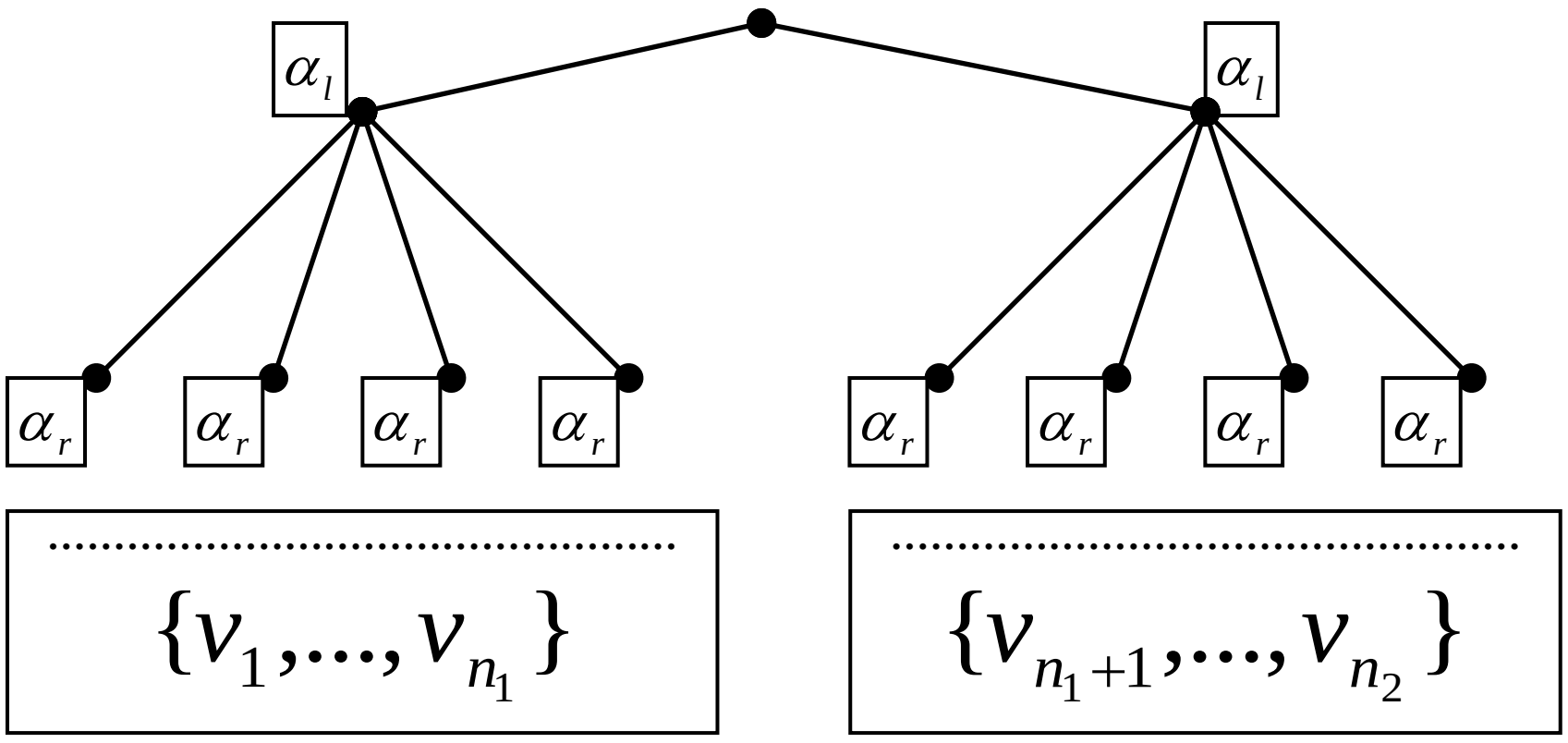

**Figure 5-12 Excessive clusters, but over different leaves**

In practice, some nodes in the cluster tree do indeed have an excessive number of children. In our source code domain experiments we observed that there is always one alpha-threshold entailing a single node with many children. For example, an alpha-threshold from $k_l = 433$ to $k_r = 3839$ clusters while all 3406 child clusters appear under the same parent in the cluster tree. We observed similar effect using any options for:

- normalization (sections 5.1.1 and 5.1.2), including the case of **no** normalization (just summing up the weights of the opposite directed arcs),
- or granularity lifting (section 5.1.4),
- or production of the input graph from the various dependencies between SE artifacts (5.2),
- or software project upon analysis and the set of libraries included (section 7.1).

Thus we conclude that the phenomenon is **intrinsic** to the domain of software source code, to the best of our knowledge and empirical evidence. Apparently, this phenomenon hinders comprehension of nested software decompositions produced with hierarchical Flake-Tarjan clustering algorithm.

While further study of this phenomenon is a hard theoretical task (but see section 8.4), we make a reasonable assumption that the phenomenon occurs due to a specific property of the underlying data, namely, almost perfectly hierarchical structure of dependencies is a common practice, while software engineers do their best to achieve this.

### 5.6.1 Maximum Spanning Tree

Consider the issue of an excessive number of children (due to alpha-threshold, section 5.6) occurred for some node in the cluster tree, thus its cluster has many nested clusters at the immediately next level, i.e. the decomposition is flat. A flat decomposition containing many items is not nearly as comprehensible as if we hierarchize the items so that a kind of divide-n-conquer approach is applicable for comprehension of the subsystem. Thus let us hierarchize the flat decomposition.

Let $C_p$ be the parent cluster containing $n_p$ (excessive number of) child clusters $C_{p,1}, C_{p,2}, ..., C_{p,n_p}$, thus $C_p = C_{p,1} + C_{p,2} + ... + C_{p,n_p}$. For a cluster $C$ let $V(C)$ be the set of vertices of the input graph (they are also the leaf nodes of the cluster tree, and they are also SE artifacts like Java classes) which constitute the cluster $C$.

First of all, we create graph $G_p(V_p, E_p)$, where $V_p$ contains $n_p$ vertices, i-th vertex stands for i-th cluster $C_{p,i}$, and each edge in $E_p$ has weight $e_{i,j}$ equal to the aggregated weight over all the edges of the input (SE artifact relation) graph connecting a vertex from cluster $C_i$ to a vertex from cluster $C_j$.

Second, we assume that there is an almost perfect hierarchy in $G_p$, and the rest is noise. Thus our task is to filter "signal" from "noise". The hierarchy is the signal, and the cycles in $G_p$ are noise. "Hierarchy" can be formalized as the subset of $T_p \subseteq E_p$ being a tree, in which an edge from parent to a child denotes the decomposition intended by software engineers (e.g. reduction from a task to subtask, or from general to specific, etc). Noise is the rest of edges, namely $E_p - T_p$, and each of them is either a violation of the architecture (e.g. a "hack" written by a software engineer), or the noise propagated from call graph construction (section 4.4), or a minor relation between SE artifacts.

It is now obvious that a reasonable solution for our task of filtering signal from noise is Maximum Spanning Tree, which filters a graph from cycles so that the sum of edge weights in the resulting tree is the maximum over all the possible trees spanning graph $G_p$.

At this point it is important to notice, that graph $G_p$ is connected, as otherwise clusters $C_{p,1}, C_{p,2}, \ldots, C_{p,n_p}$ would not become children of the same parent cluster $C_p$. Thus, there is always a tree spanning the whole graph $G_p$.

Usually, the problem of *minimum* spanning tree appears in the literature. For the convenience of the reader, we show here how the problem of *maximum spanning tree can be reduced* to the problem of minimum spanning tree. In the graph $G_p$, let $B = 1 + \max_{i,j} e_{i,j}$. Then we replace each edge of weight $e_{i,j}$ with an edge of weight $B - e_{i,j}$, solve the problem of minimum spanning tree with any of the efficient algorithms (section 4.6) and return the edge weights back, in both $T_p$ and $G_p$.

At this point, we have filtered signal from noise in the graph induced by the excessive children, and constructed a tree (hierarchy) that spans them. However, an unexpected question arises: what should be selected as the root of the tree?

### 5.6.2 Root Selection Heuristic

Proper selection of the root of the maximum spanning tree is crucial for understanding of the hierarchy. We illustrate this in Figure 5-13 below:

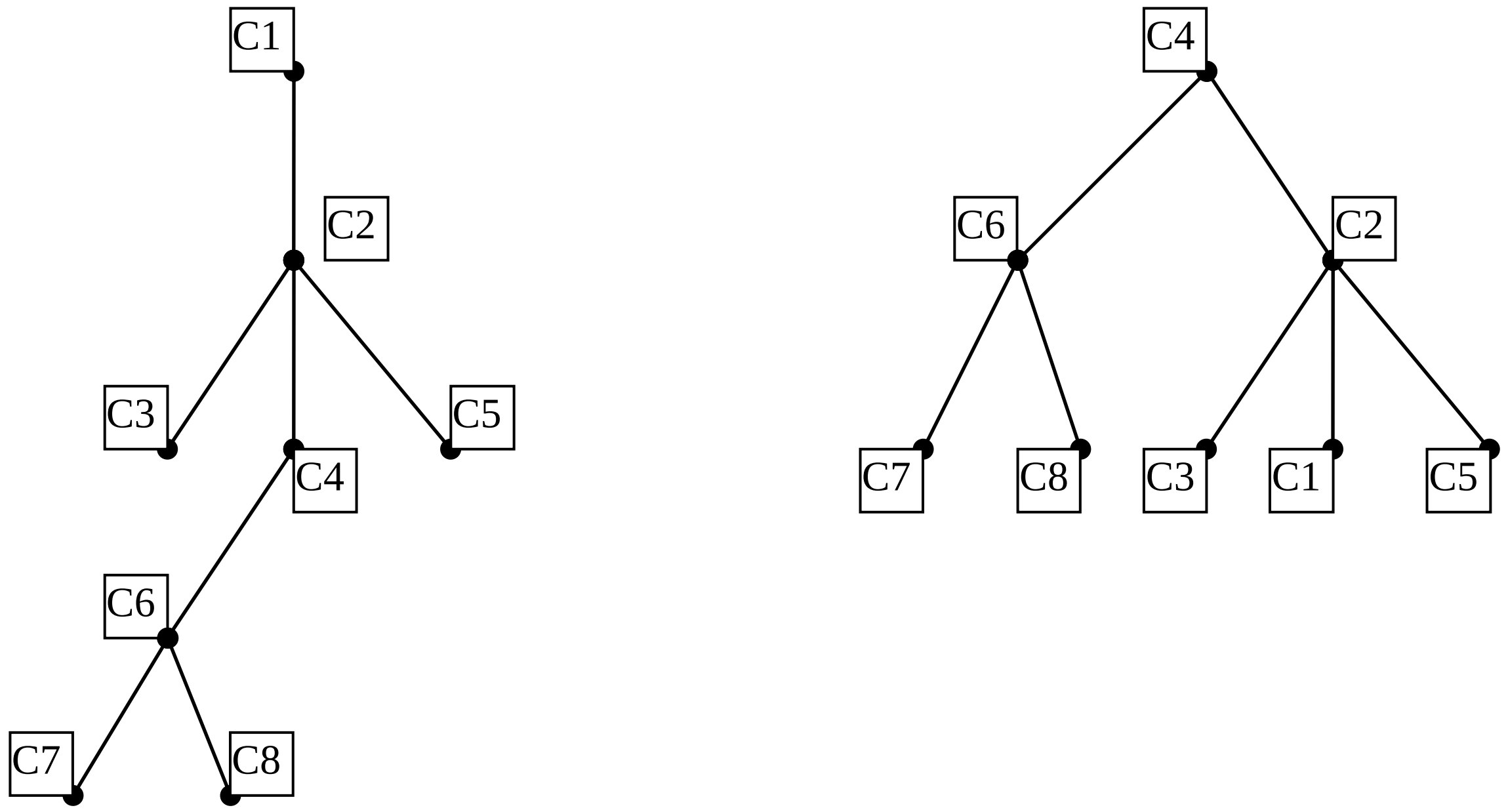

**Figure 5-13 Node C1 vs. node C4 as the root**

Obviously, the same maximum spanning *tree* is illustrated both in the left and in the right of the above picture. However, the understanding about which SE artifact is more high-level/general or low-level/specific totally differs.

Of the considered options for selection of the root, two seemed reasonable and we did experiments with them.

**Option 1**. The intention is to select the root in such a way, that heavy cycles (the noise removed from $G_p$) appear as far as possible from the root, i.e. closer to the leaves. The algorithm for this option is given below:

```
(1) Sort all the edges of graph G_p in the order of their weights, heavier first
(2) Let U be the set of disjoint subsets of the vertices of G_p
(3) Passing the edges of G_p from the heavier to the lightest, do
(4)  (1) If the edge is in the tree T_p, unite in U its incident vertices
(5)  (2) Else, find the path between its incident vertices through only the edges in T_p,
    and unite in U all the vertices encountered on the path
(6)  (3) If U has become a single subset, stop the passing of edges.
(7) End.
(8) The last united vertex (or the weighted middle of the path, if multiple), becomes the
    root of T_p
```

Disjoint-set data structure and union-find algorithm was used for $U$, see section 4.6. The algorithmic complexity of the root selection is: $O\left(\left|E_p\right|\log\left|E_p\right| + \left|V_p\right| \cdot \alpha k\left(\left|V_p\right|\right)\right)$ where $\alpha k(n)$ is the inverse Ackermann function.

**Option 2**. The intention is to select the central node, while the selection is prioritized by the weights of the edges in the tree only (i.e. not in $G_p$). The algorithm in this case is prioritized breadth-first search starting from the leaves. Initially, all the leaves are put into the priority queue. When a vertex is removed from the queue, we decrease the "to go" counter for its single adjacent vertex. If "to go" counter becomes 1, this adjacent vertex is put into the queue with priority equal to the weight of the incident edge (the more weight, the earlier will be removed). "To go" counter for a vertex denotes the number of adjacent vertices which have not yet been regarded, and is initially equal to its degree. The last vertex pushed into the queue becomes the root of our maximum spanning tree $T_p$. The algorithmic complexity of this root selection option is: $O\left(\left|V_p\right| \cdot \log\left|V_p\right|\right)$.

In practice, the second root selection option is producing empirically much better hierarchies. The results we are showing throughout the paper are processed with this heuristic after hierarchical clustering. We can see (sections 7 and 10) that indeed, the problem of excessive child clusters has been alleviated, and SE artifacts are still grouped according to their unity of purpose.

# 6 Implementation and Specification

We implemented parallel computation of hierarchical Flake-Tarjan clustering within this project as multiple OS-processes on our double-core processor, each working in separate directory. Changing the prototype to working on multiple computers amounts to sharing the parent directory over the network and launching remote processes rather than local.

The choice of programming language was driven by whether we need speed of implementation or runtime speed of the program. Most of InSoAr, 14K lines of code, is implemented in Java: the source code is 434KB in size, and it was all written by one programmer, the author of the thesis, within the short time period of this project. Some state-of-the-art source code metrics over InSoAr are produced with STAN ([Stan2009]) and demonstrated in Figure 6-1 to the right.

The bottleneck part, minimum cut tree algorithm (section 4.2.1) using the community heuristic (section 4.2.2), is implemented in C, and uses Goldberg's implementation of maximum flow algorithm (section 4.1.1) modified for our needs. We used all possible including low-level optimizations for the bottleneck part.

A visualization of InSoAr at *package*-level (not the class-level InSoAr operates) with, to our knowledge, the best state-of-the art **structure** analysis tool STAN [Stan2009] is given in appendix 10.4.1, and a zoomed-out version in Figure 6-2 below. The shadow is the sliding window visible in full size.

| Category/Metric | Value |
|---|---|
| Count | |
| Libraries | 1 |
| Packages | 35 |
| Units | 122 |
| Units / Package | 3.49 |
| Classes / Class | 0.22 |
| Methods / Class | 5.51 |
| Fields / Class | 4.49 |
| ELOC | 14001 |
| ELOC / Unit | 114.76 |
| Complexity | |
| CC | 2.83 |
| Fat - Libraries | 0 |
| Fat - Packages | 112 |
| Fat - Units | 413 |
| Tangled | 21.62% |
| Tangled - Libraries | 0.00% |
| ACD - Library | 0.00% |
| ACD - Package | 22.35% |
| ACD - Unit | 6.65% |
| Robert C. Martin | |
| D | -0.39 |
| \|D\| | 0.42 |
| Chidamber & Kemerer | |
| WMC | 15.6 |
| DIT | 0.98 |
| NOC | 0.03 |
| CBO | 5.45 |
| RFC | 10.69 |
| LCOM | 6.9 |

**Figure 6-1 Metrics over InSoAr**

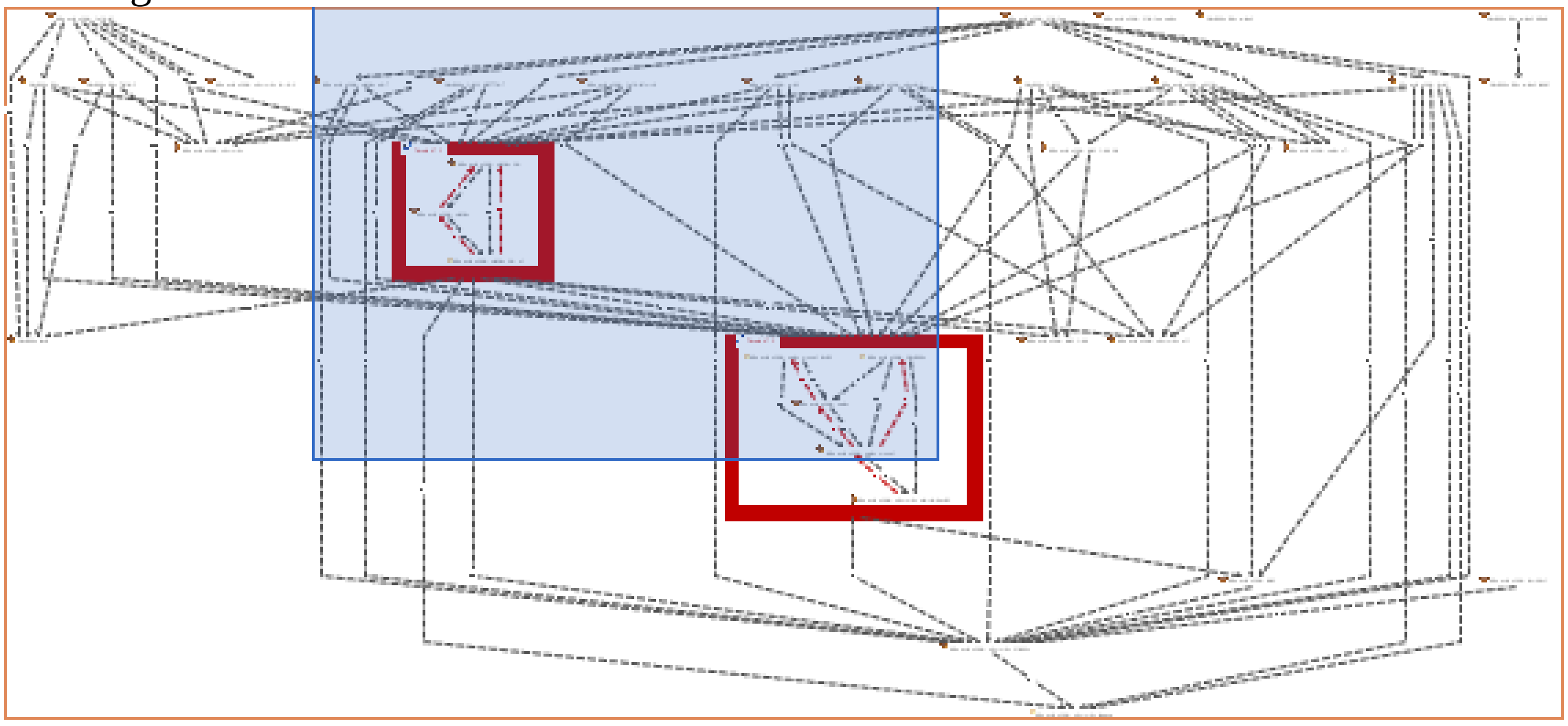

**Figure 6-2 InSoAr at package-level visualized with STAN**

## 6.1 Key Choices

Most of the key choices are theoretical, thus described under sections 3, 4 and 5. We do not provide a blow by blow description of InSoAr due to the nature of the paper, limit in pages and size of the system. Below are the most important, though applied aspects.

### 6.1.1 Reducing Real- to Integer- Weighted Flow Graph

After normalization (section 5.1) we get an undirected graph with real-valued edge weights. Flake-Tarjan clustering algorithm (section 4.3) also uses real-valued parameter "alpha" in order to prepare a minimum cut tree task (section 4.2). The algorithm solving the min cut tree problem relies on computations of maximum flow in a graph (section 4.1). Though there are algorithms solving maximum flow problem for real-valued edge capacities, however, they are much slower. Both the fastest known max flow algorithm (we use it for theoretical bounds on the worst-case complexity, section 4.1) and the best known implementation of another max flow algorithm (section 4.1.1) we used in practice, require *integer* arc or edge capacities. Thus we must convert from real- to integer-weighted graph.

For each vertex in the graph we calculate the sum of weights of the adjacent edges. Then we adjust the weights proportionally, so that they have the largest possible integer values, taking into account the limitations of 32-bit and 64-bit integers. The latter two are used as edge capacities and excess flow in Goldberg's implementation of push-relabel max flow (section 4.1.1). Our experiments have shown that max flow solution never became suboptimal due to this conversion.

### 6.1.2 Results Presentation

The result of hierarchical clustering is a tree (more precisely, a forest, when there are multiple disjoint components in the software artifact dependency graph), where

- Leaves are classes of the software upon analysis and its libraries.
- Inner nodes are clusters at different levels.
- There is at least one root per disjoint component.
- Multiple roots per disjoint component appear in case the selected lower bound of alpha was not low enough to unite all the nodes of that component into a single cluster.

As it is not trivial to present the results in a comprehensible form, some aspects of the used presentation approaches are described further. Our main representation of the results is in text format. Not going into the details of each value, we see 3 principal ways to represent a tree:

1. Indented by Depth
2. Indented by Height
3. Bracketed

The first is more convenient to view, as nested clusters (or SE artifacts, if leaves) appear under their parents. An example of this presentation is in Figure 7-3. However, this presentation takes a lot of space on hard drive. The second presentation has an advantage that SE artifacts (the nodes that have labels) always appear in the beginning of a line, as they are leaves thus have height 0. However, effective comprehension of this presentation needs some training, see appendix 10.4.3.1. This presentation also takes less space, as nodes are often at large depth, but rarely at large height. The third, bracketed presentation, aims to show much more labels (leaf nodes) on a limited space. Inner nodes do not take a line each, but are grouped in one line and represented as brackets. An example is in Figure 10-10.

## 6.2 File formats

A number of file formats is used at various stages of the software engineering artifacts extraction and clustering pipeline. It does not make sense to describe them in detail at this stage. Thus we give a short list of the formats:

- Text (identifiers like “call graph” etc. appear for historical reasons, indeed they contain heterogeneous relations):
  - Literal graph of relations (user friendly form): litCallGraph.txt
  - Computer-friendly form of the graph of relations: callGraph.txt, ccgClasses.txt, ccgMembership.txt, ccgMethods.txt)
  - Cluster tree: ctHier.txt, perfTree.txt, treeXXXX.txt, hitPerfClu.txt (height-indented), ctBracketed.txt (bracketed presentation)
  - Inputs for Cfinder (list of arcs)
  - Inputs for H3Viewer: h3reduced2.lvlist, h3sphere.lvlist
  - Inputs/outputs for a process performing basic cut clustering: passOrder.txt, intGraph.txt (DIMACS format), ver2node.txt, ftClusters.txt, ftcConOut.txt
- XML:
  - Cluster Tree XML
  - Per-package statistics in XML
  - Input for TreeViz: perfTv*.XML

For example, below is a short description of the cluster tree XML format. Several XML representations were considered, e.g. an XML element corresponding to a cluster tree node could contain properties like “alpha” and “heads” as nested XML elements along with an XML element “children” which would list all the child nodes and their subtrees. However, we attempted to choose a representation that is easier to view by a human, and this should be the one that contains only child nodes as the child XML elements for a node, i.e. homogenous.
The root node looks like below:

```
<clusterTree vertexCount="7474" nodeCount="7721" rootCount="6476" disjointCount="2">
```

Below is an example of an inner node (cluster), “alb” is the alpha at which the cluster was produced, “djComp” is the number of its disjoint component:

```
<node id="7477" childCount="2" alb="0.01780273437500000000" heads="5287, 7710" djComp="1">
```

Below is an example of a leaf node, i.e. a SE artifact (Java class in this case):

```
<node id="4578" label="net.sf.freecol.client.control.InGameInputHandler" djComp="1" />
```

## 6.3 Visualization

Pure XML or HTML formats, GraphViz and FreeMind tools were considered. However, we chose the following visualization tools because they perform well at large trees:

- H3Viewer: http://graphics.stanford.edu/~munzner/h3/download.html
  - This tool can draw large trees in 3D hyperbolic space
- TreeViz: http://www.randelshofer.ch/treeviz/index.html
  - This tool supports 7 different presentations for large trees

## 6.4 Processing Pipeline

The runtime of the analyzer is divided into stages, where outputs from a preceding stage are inputs to a succeeding stage. Outputs are flushed into files. This allows reusing the results of a stage without re-running it, as well as substituting different implementations of a stage, e.g. Java or C#, static or dynamic call graph extractors. Below is a diagram of the present stages:

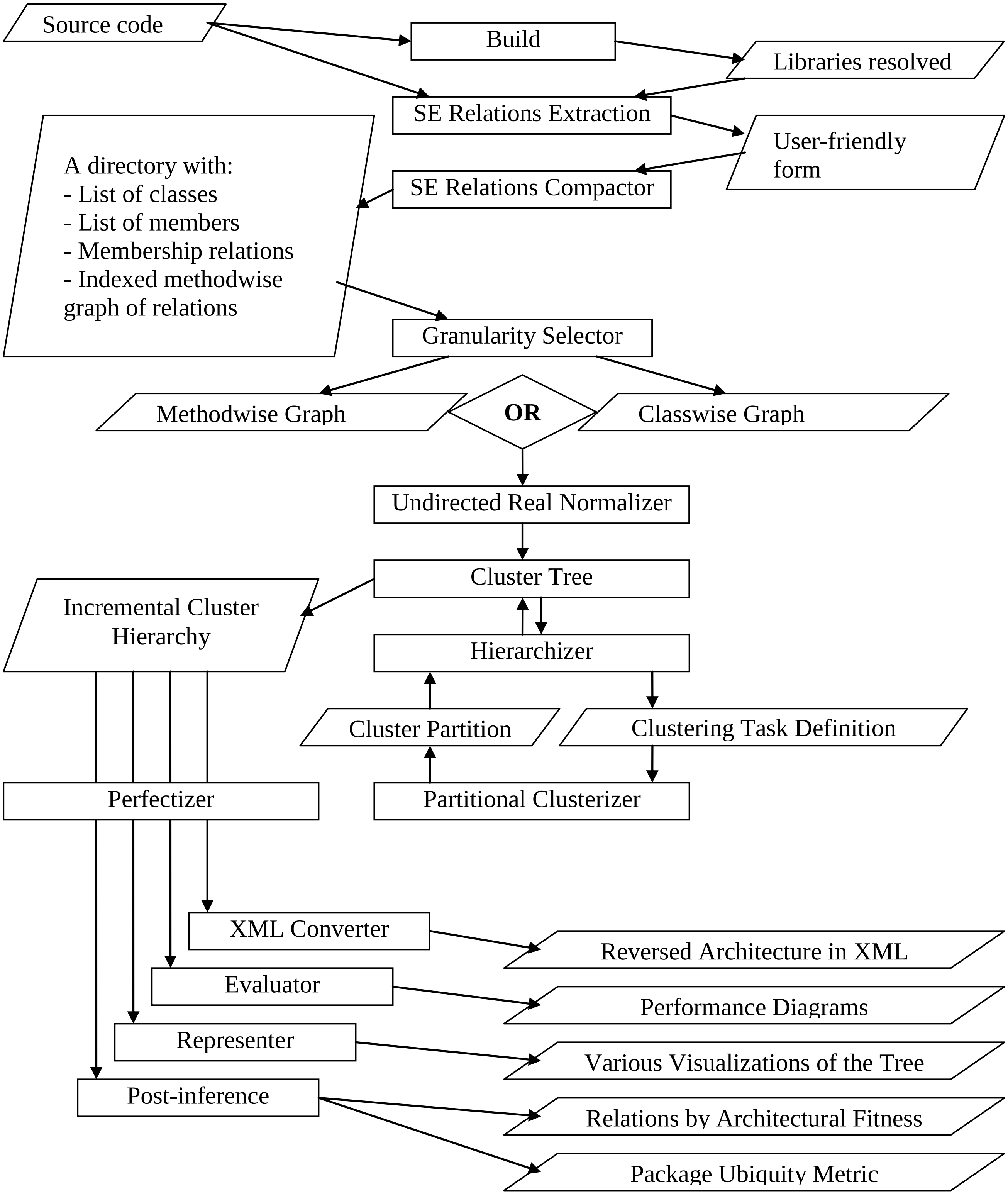

**Figure 6-3 InSoAr Processing Pipeline**

In Figure 6-3, processing stages are drawn as rectangles, while inputs or results are drawn as parallelograms. The pipeline takes *source code* on input. Source code should be *built*, in order to resolve library dependencies. Then call graph and other relations between software engineering artifacts must be extracted. In the current implementation, using Soot to process Java programs, we produce the graph of relations in user-friendly form. Java classes are on the outer level, inside are methods and fields, and for each method there is a list of relations with other member- or class-level SE artifacts. If something can produce such a graph of relations from other programming language, e.g. C# or C++, we do not depend on programming language since this point. A graph of relations produced by means of dynamic analysis is also an option here. Then we run a stage called "`SE Relations Compactor`", which converts the graph of relations into a computer-friendly form. Though it is also a text format, it occupies substantially less space and is easier to read into memory. "`SE Relations Compactor`" also performs some reindexing, so that further stages a released from these operations after input.

Implementation of the further stages follows the theory we gave in section 5. After loading the graph of relations, a stage called "`Granularity Selector`" allows to choose whether we are going to clusterize at class- or method-level, and can be used to lift the granularity prior to clustering. Its output is a directed graph of relations between SE artifacts. "`Undirected Real Normalizer`" converts a directed graph to undirected, normalizes and lifts the granularity to class-level, if necessary. It holds a graph, from which the initial "`Cluster Tree`" can be built. The initial cluster tree contains all the SE artifacts as leaf nodes, which are children of one fake root, even if in different disjoint components. "`Cluster Tree`" is updated incrementally by "`Hierarchizer`". The latter maintains the alpha search tree, prepares a new task for basic cut clustering processor, receives the result and merges it into the global cluster tree. "`Partitional Clusterizer`" is a separate, probably remote, process that performs a single flat clustering using Flake-Tarjan algorithm, taking the input from a file and producing output into a file. There is always an option to use named pipes instead of files here, so that slow hard drive is not needed.

The current cluster tree is flushed every certain amount of minutes to disk. This is "`Incremental Cluster Hierarchy`". We can force the pipeline to stop by creating file "shutdown.sig". Then the latest hierarchy is also saved to disk. "`Perfectizer`" addresses the issue and computes solution we discussed in section 5.6. It takes the result of hierarchical clustering on input, and produces perfected result in the same format. This step is, of course, optional. The rest of the pipeline after "`Perfectizer`" addresses various presentations, evaluation and post-inference.

# 7 Evaluation

The main premise for high quality of a produced clustering hierarchy is the theoretical grounding of the clustering algorithm we used: the quality of the produced clusters is bounded by strong minimum cut and expansion criteria [Fla2004]. We consider *cut size* a rational criterion in the domain of software engineering because the sum of edge weights reflects the amount of interaction (relations) between SE artifacts (e.g. Java classes), which a software engineer needs to study in the source code in order to understand coupling between either two SE artifacts, or two groups (communities, clusters) of SE artifacts. This matches the main idea behind max-flow/min-cut clustering technique, according to [Fla2004]: *"to create clusters that have small inter-cluster cuts (i.e. between clusters) and relatively large intra-cluster cuts (i.e. within clusters). This guarantees strong connectedness within the clusters and is also a strong criterion for a good clustering, in general."*

Assuming from the above that the clustering algorithm performs well, we should study whether this quality has not been lost due to the adaptations we used for the clustering to work in the domain of software source code, see section 5. These adaptations also include extraction of the call graph and other relations between SE artifacts, which is data, specific to the domain. We stress that not only the quality of the clustering method is important, but it is also important that its input data is adequate and of high quality, see sections 4.4 and 3.1.1.

Another theoretical premise for high-quality of the reconstructed architecture is that we have incorporated a solution (section 5.1, which follows state of the art best practices discussed in section 4.5) to the main problem for clustering in source code domain, according to the literature [Pir2009], – utility artifacts.

## *7.1 Experiments*

In the largest of our experiments we processed software containing

- 2.07M (2 070 645) graph edges (relations) over heterogeneous set of vertices (SE artifacts) containing
- 11.2K (11 199) Java classes,
- 163K (163183) of class member level artifacts (methods and fields)

This is a real-world medium-size project provided by KPMG for our experiments during the internship. The client code contains about 500 classes, thus the remaining 10.7K classes are in libraries, which include Java system libraries, Spring framework, Hibernate, Apache commons, JBPM, Mule, Jaxen, Log4j, Dom4j, and others. Together with libraries the project becomes 22 times bigger and falls into category of large software.

Note that our conception of a medium-sized project differs significantly from the claimed in other scientific works. In [Pate2009] they analyze (mostly, clusterize) a project containing 147 classes in 10 packages. In practical software engineering this project must be classified as small or even above-tiny. In contrast, just the client code of our medium-sized projects contains 500-1000 Java classes. The total number of Java classes we clustered hierarchically is 11 199 in the largest experiment, and 7000-7500 in the usual experiments.

Furthermore, we suspect that the input data of related works analyzing only a part of the program (e.g. only client code) was far not as precise as ours, because advanced call graph extraction techniques (VTA, Spark in Soot) require analysis of the whole program with libraries, and even simulate native calls of the Java Virtual Machine [Lho2003]

### 7.1.1 Analyzed Software and Dimensions

FreeCol is an *open source* game similar to Civilization or Colonization. Its source code is in Java and available here: http://www.freecol.org/download.html . The project is medium-size, containing about 1000 of client-code classes. Together with libraries it becomes about 7.5K classes. Thus we used it in our experiments. The extracted graph of relations contains 1M edges for this project.

"dem0" project is a web application that also provides web services, uses Spring framework and works with database through Hibernate. It is not open-source, thus we are only showing the parts for which we received permission from KPMG. This project contains about 500 classes of client code and many classes in libraries. In order for Soot to fit in 2GB memory limit during VTA (variable type analysis) call graph construction, we had to limit the number of library classes to 6.5K. In the largest experiment we used RTA (rapid type analysis) for call graph construction, thus it was possible to process all 10.7K library classes with Soot. In the former case, the graph of relations contained 0.5M edges, while in the latter there were 2M edges.

| | | |
|---|---|---|
| 1 | **InSoAr processing** | |
| 1.1 | Clustering hierarchy we demonstrate in this paper | 72 hours, 0.6GB RAM |
| 1.2 | Acceptable results (differences are visible empirically, conclusions need statistical studies) | 1-2 hours |
| 1.3 | The largest experiment (11.2K classes, 2M relations) | 1.3GB RAM, 120 hours |
| 2 | **Call graph construction** (and other relations with Soot) | |
| 2.1 | VTA in usual experiments, 7.5K classes: | 2GB RAM 0.5 hour |
| 2.2. | RTA in the largest experiment, 11.2K classes: (VTA gets out of memory in this case) | 2GB RAM, 2 hours |
| 3 | **Basic cut clustering** (one alpha, in a separate process) | |
| 3.1 | In the usual experiments | 4.5 minutes |
| 3.2 | In the largest experiment | 20 minutes |
| 3.3 | Memory requirement, no more than | 35MB |

**Figure 7-1 Actual time and space requirements**

The actual significant time and space requirements are given in Figure 7-1 above. Note, that we are using far not optimal implementation of the prototype (Java). E.g., basic cut clustering implemented in C/C++ requires only 35MB in the largest experiment. We use double-core, 1.7GHz each, machine with 2GB RAM and 2MB L2 cache. When 2 basic cut clusterizers are run in parallel, the duration is 6.5 minutes instead of expected the same 4.5 minutes due to cache misses (cache memory is shared between the 2 cores).

## *7.2 Interpretation of the Results*

Altogether, our nested software decomposition (hierarchical tree) shows SE artifacts from general (in the top, closer to the root) to specific (in the bottom). SE artifacts are grouped according to their unity of purpose, so that a group of artifacts serving similar purpose or collaborating for a composite purpose (act together) constitutes a subtree.

More precisely, the hierarchical tree reflects the strength of coupling. There is some noise in the input data (extracted call graph) and some uncertainty on how to combine (coefficients, etc) different kinds of relations between SE artifacts prior to clustering. However, the clustering algorithm further decomposes the vertices (SE artifacts) into hierarchical communities **strictly**, using a bound between inter-cluster and intra-cluster connection strengths (SE artifacts coupling).

One interpretation of the latter paragraph is that, in the second approximation:

- In the top (i.e. near root) of the decomposition appear artifacts, which are:
  - less coupled to the rest of the program, or
  - more general (general-purpose)
- In the bottom (far from the root) appear artifacts, which are:
  - closer to the core of the program (more coupled with the rest of the program), or
  - more specific (complex)

For a cluster node (which is non-leaf, inner node), not only the depth (distance from the root), but also the height (distance to the remotest leaf in its subtree) should be considered.

### 7.2.1 Architectural Insights

In the subsequent sections we provide an account of particular facts, which become apparent to a sufficiently experienced software engineer by browsing the (various presentations of) the results produced with our prototype. Mining these facts with state of the art tools is either not possible, or requires immense efforts, e.g. browsing and interpreting manually many lines of source code. In general, we call the inferred facts “architectural insights”, as they help the viewer to, at least, get a first impression of the source code, and mostly comprehend the decomposition of the software system into subsystems. Taking 10M lines of source code on input, InSoAr produces only about 10K nodes of cluster tree on output. The gain in comprehension is 1000 times, which is, roughly, calculated from the number of items necessary to scan in order to get a global understanding of the system, see section 2.1.

Having a nested software decomposition provided by InSoAr, a software engineer can effectively apply divide&conquer approach for software comprehension (appendix 10.1.6), or detect cross-package subsystems implementing complicated logic (appendix 10.1.5). One can also observe some metrics calculated after architecture reconstruction, and we give some examples in appendix 10.2. These metrics can give idea of how ubiquitous a package is (i.e. how broad in the architecture the classes of this package are spread, appendix 10.2.1), and how well couplings between SE artifacts fit the implicit architecture (appendix 10.2.2). Often, insights not only about architecture, but also about implementation can be captured. We give such examples in appendix 10.1.4.

Certainly, the list cannot be exhaustive as these are only example architectural insights we could think of and describe within limited time and pages. We invite the reader to browse the hierarchy on his/her own by downloading the clustering hierarchy of a demo project and the H3 sphere visualizer from the internet. Below are the links:

- Data files. Leaf nodes of the trees correspond to Java classes of libraries and client code. Inner nodes correspond to clusters at different levels in the hierarchy. Client source code (the application, i.e. non-libraries) is in package `com.dem0.*`
  - In XML: http://zrobim.info/InSoAr/Demo/ds0/clusterTree.xml
  - In H3Viewer format: http://zrobim.info/InSoAr/Demo/ds0/h3sphereCT.lvhist
  - In TreeViz format: http://zrobim.info/InSoAr/Demo/ds0/treevizCT.xml
- H3Viewer: please, download it from the website of its developer: http://graphics.stanford.edu/~munzner/h3/download.html
- TreeViz website: http://www.randelshofer.ch/treeviz/index.html
  - BUT: we have tuned TreeViz within our project, so that it shows client-code artifacts in green (and the rest is in orange), and lists descendants of a subtree upon mouse hover, when no more than 100. Download the archive and unpack the two files into the same directory before running.
  - Tuned TreeViz: http://zrobim.info/InSoAr/Demo/ds0/TreeVizCliFi.zip

### 7.2.2 Class purpose from library neighbors

Library neighbors can tell an experienced software engineer a lot about the purpose of client code classes, see Figure 7-2 below. This follows directly from the criteria for clustering: dense interaction (many calls, field accesses, type usages) between SE classes within a cluster and relatively loose interaction between classes from different clusters.

The *crucial advantage* that software engineers acquire having software structure inferred with InSoAr is in the following. In order to figure out the purpose of library classes, as well as other facts like requirements, constraints and limitations, one usually can read the *documentation*. Application classes, on the other hand, are not well documented (section 2), thus software engineers would have to scan and interpret manually the source code of the class. However, having our clustering hierarchy, a software engineer can simply read the documentation for library classes which are coupled with the application classes upon analysis.

It is obvious what is meant by purpose. Below are examples of other facts that can be read from the documentation of a library class (instead of the source code of an application class):

- Requirement: an open database connection
- Limitation: usage of 128-bit encryption, which is not strong enough for certain purposes

Basing on these facts, violations of the constraints can be identified easier.

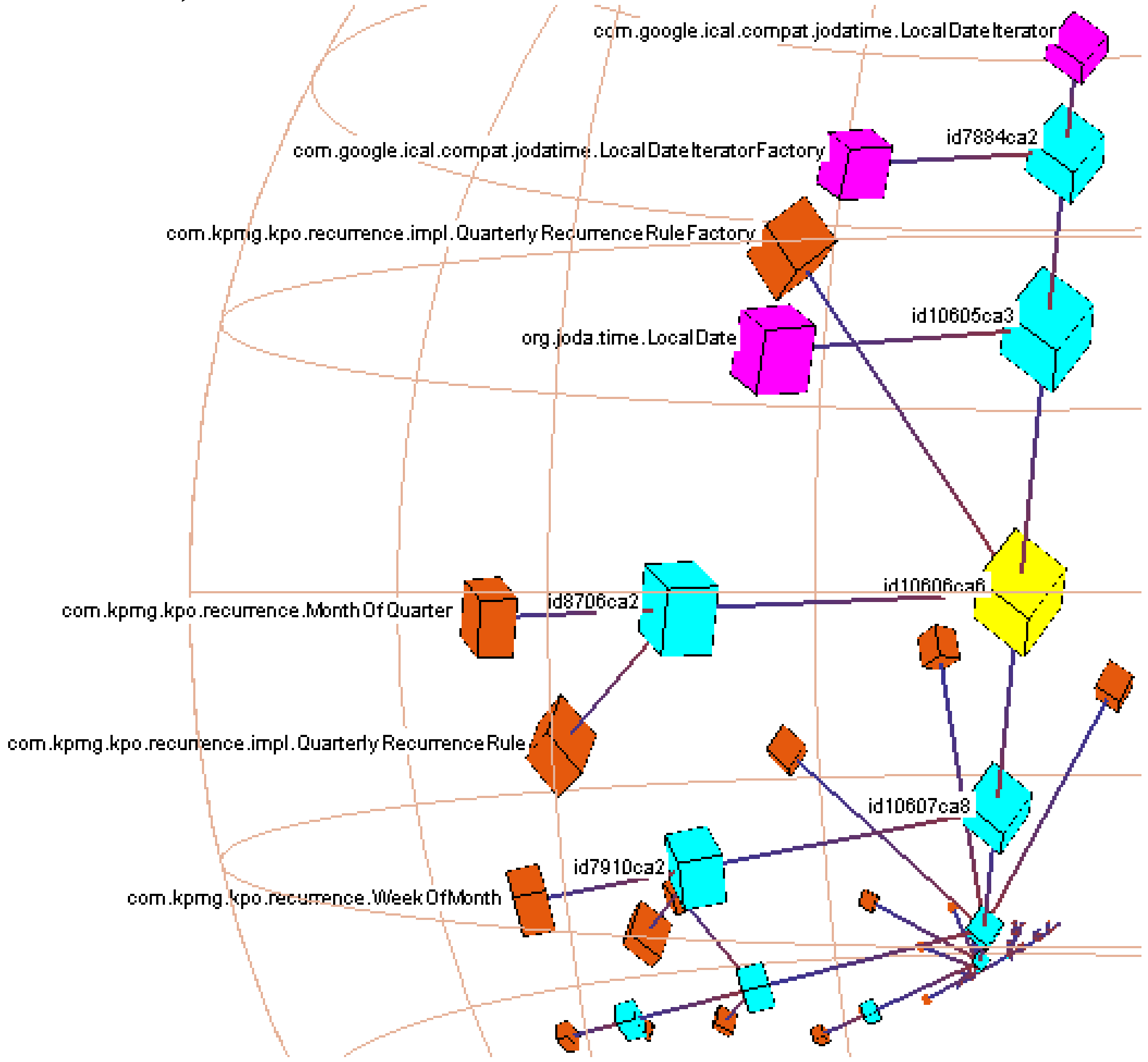

**Figure 7-2 Library classes are in pink, application classes are in orange, clusters are in light-blue**

In the above figure, we see a subtree with classes serving the same purpose, as can be understood from their names. One fact that we easily infer is that the application's subsystem for time and scheduling relies on `JodaTime` library rather than inferior Java system library for time.

### 7.2.2.1 Obvious from class name

One can argue that the clustering hierarchy does not bring any value about the purpose of a SE class when the class appears near similarly named, sometimes library classes, because class purpose was already obvious from class name, as in Figure 7-3 below. In this figure we see that application class `com.kpmg.kpo.web.security.EmployeeUserDetailsService` and others appear coupled (descendants of cluster #8604) with library classes `org.springframework.security.userdetails.UserDetailsService` and `org.springframework.security.userdetails.UserDetails`. However, the point is:

- The fact that these similarly named classes got into the same cluster tells us about good architectural style: classes with similar names serve a similar purpose. The purpose of the library classes is known from the documentation.
- Application class "…EmployeeUserDetailsService" is most coupled with library classes which are supposed to serve this purpose, and not with something else, which would be architecture violation
- Good quality of our clustering hierarchy is confirmed by such an occurrence!

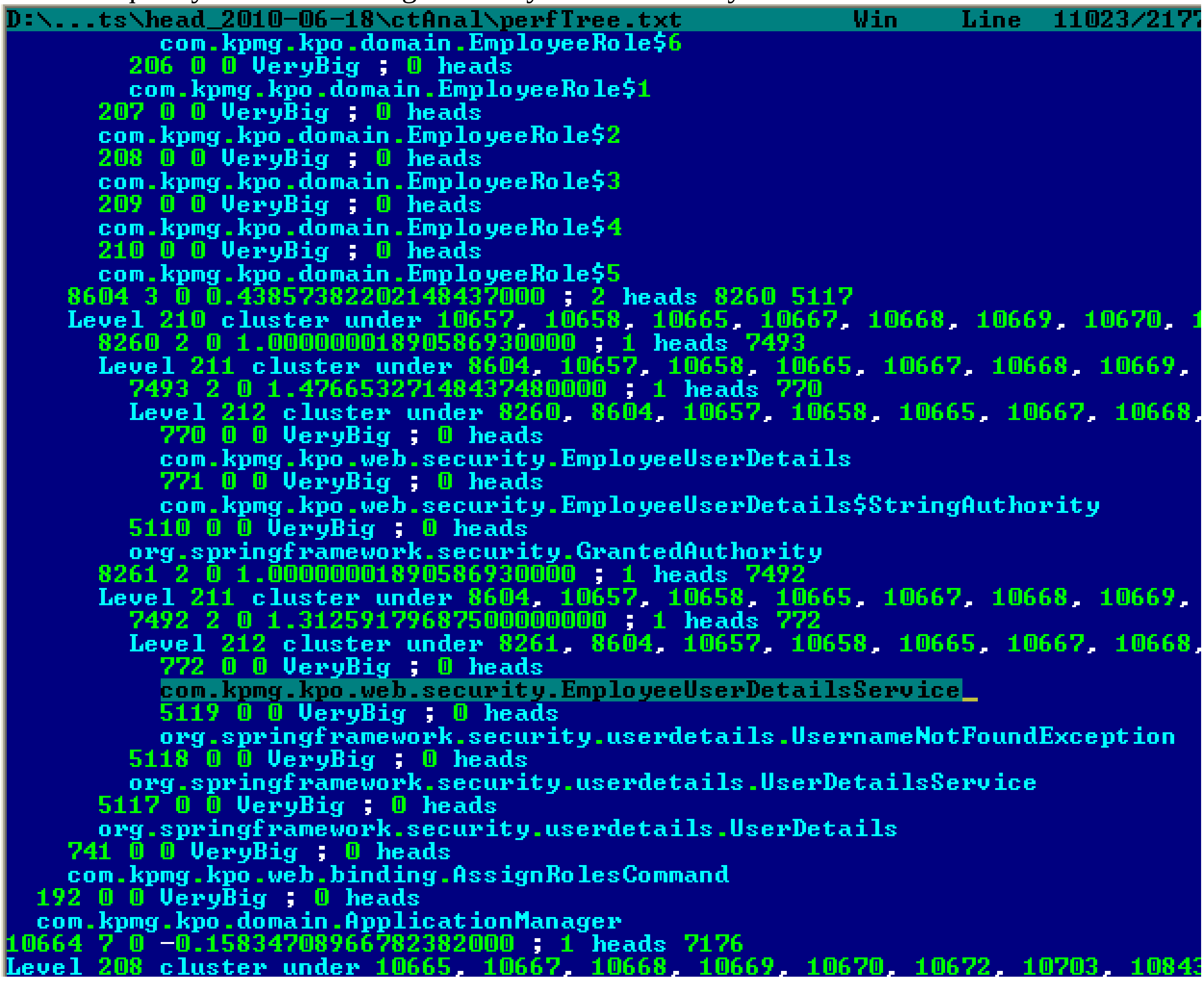

```
D:\...ts\head_2010-06-18\ctAnal\perfTree.txt          Win     Line   11023/2177
          com.kpmg.kpo.domain.EmployeeRole$6
        206 0 0 VeryBig ; 0 heads
        com.kpmg.kpo.domain.EmployeeRole$1
      207 0 0 VeryBig ; 0 heads
      com.kpmg.kpo.domain.EmployeeRole$2
      208 0 0 VeryBig ; 0 heads
      com.kpmg.kpo.domain.EmployeeRole$3
      209 0 0 VeryBig ; 0 heads
      com.kpmg.kpo.domain.EmployeeRole$4
      210 0 0 VeryBig ; 0 heads
      com.kpmg.kpo.domain.EmployeeRole$5
    8604 3 0 0.43857382202148437000 ; 2 heads 8260 5117
    Level 210 cluster under 10657, 10658, 10665, 10667, 10668, 10669, 10670, 1
      8260 2 0 1.00000001890586930000 ; 1 heads 7493
      Level 211 cluster under 8604, 10657, 10658, 10665, 10667, 10668, 10669,
        7493 2 0 1.47665327148437480000 ; 1 heads 770
        Level 212 cluster under 8260, 8604, 10657, 10658, 10665, 10667, 10668,
          770 0 0 VeryBig ; 0 heads
          com.kpmg.kpo.web.security.EmployeeUserDetails
          771 0 0 VeryBig ; 0 heads
          com.kpmg.kpo.web.security.EmployeeUserDetails$StringAuthority
        5110 0 0 VeryBig ; 0 heads
        org.springframework.security.GrantedAuthority
      8261 2 0 1.00000001890586930000 ; 1 heads 7492
      Level 211 cluster under 8604, 10657, 10658, 10665, 10667, 10668, 10669,
        7492 2 0 1.31259179687500000000 ; 1 heads 772
        Level 212 cluster under 8261, 8604, 10657, 10658, 10665, 10667, 10668,
          772 0 0 VeryBig ; 0 heads
          com.kpmg.kpo.web.security.EmployeeUserDetailsService_
          5119 0 0 VeryBig ; 0 heads
          org.springframework.security.userdetails.UsernameNotFoundException
        5118 0 0 VeryBig ; 0 heads
        org.springframework.security.userdetails.UserDetailsService
      5117 0 0 VeryBig ; 0 heads
      org.springframework.security.userdetails.UserDetails
    741 0 0 VeryBig ; 0 heads
    com.kpmg.kpo.web.binding.AssignRolesCommand
  192 0 0 VeryBig ; 0 heads
  com.kpmg.kpo.domain.ApplicationManager
10664 7 0 -0.15834708966782382000 ; 1 heads 7176
Level 208 cluster under 10665, 10667, 10668, 10669, 10670, 10672, 10703, 10843
```

**Figure 7-3 Class EmployeeUserDetailsService and neighbors**

In addition to the above points, nearby we also see classes with very different names and from very different packages, e.g.

- `GrantedAuthority` from library package `org.springframework.security`,
- `AssignRolesCommand` from `com.kpmg.kpo.web.binding`,
- `ApplicationManager` from `com.kpmg.kpo.domain`
- anonymous nested classes of `EmployeeRole` from `com.kpmg.kpo.domain`

As a result, a human software engineer is provided with an insight about the **subsystem**, which:

> *…manages user details, where the users are most likely employees, and there is a dedicated service for this, which is based on the standard service of Spring framework* *addressing this purpose. When a user becomes authorized by the subsystem, a corresponding security token is issued (class* …`GrantedAuthority`*), which is a string (look at nested class* `StringAuthority` *under …*`EmployeeUserDetails`*). When authorization fails (perhaps, only for the reason that there is no such user/employee),* …`UsernameNotFoundException` *is thrown. The latter is a standard exception from Spring framework, thus it is likely that the client code (application classes) does not handle this exception at all or in full, but rather relies on the standard facilities of Spring framework, otherwise a more specific exception inheriting* …`UsernameNotFoundException` *would be implemented in the application and appear nearby in the clustering hierarchy. The set of business entities which an employee can access is determined through assignment of roles, application class* …`EmployeeRole`*, and roles are assigned using* `com.kpmg.kpo.web.binding.AssignRolesCommand`*, which is likely to occur when a privileged user takes the corresponding action from web UI.*

We wrote the above paragraph without looking at a single line of source code of either of the mentioned classes, even more, having almost no experience with Spring framework, just principal understanding of programming concepts.

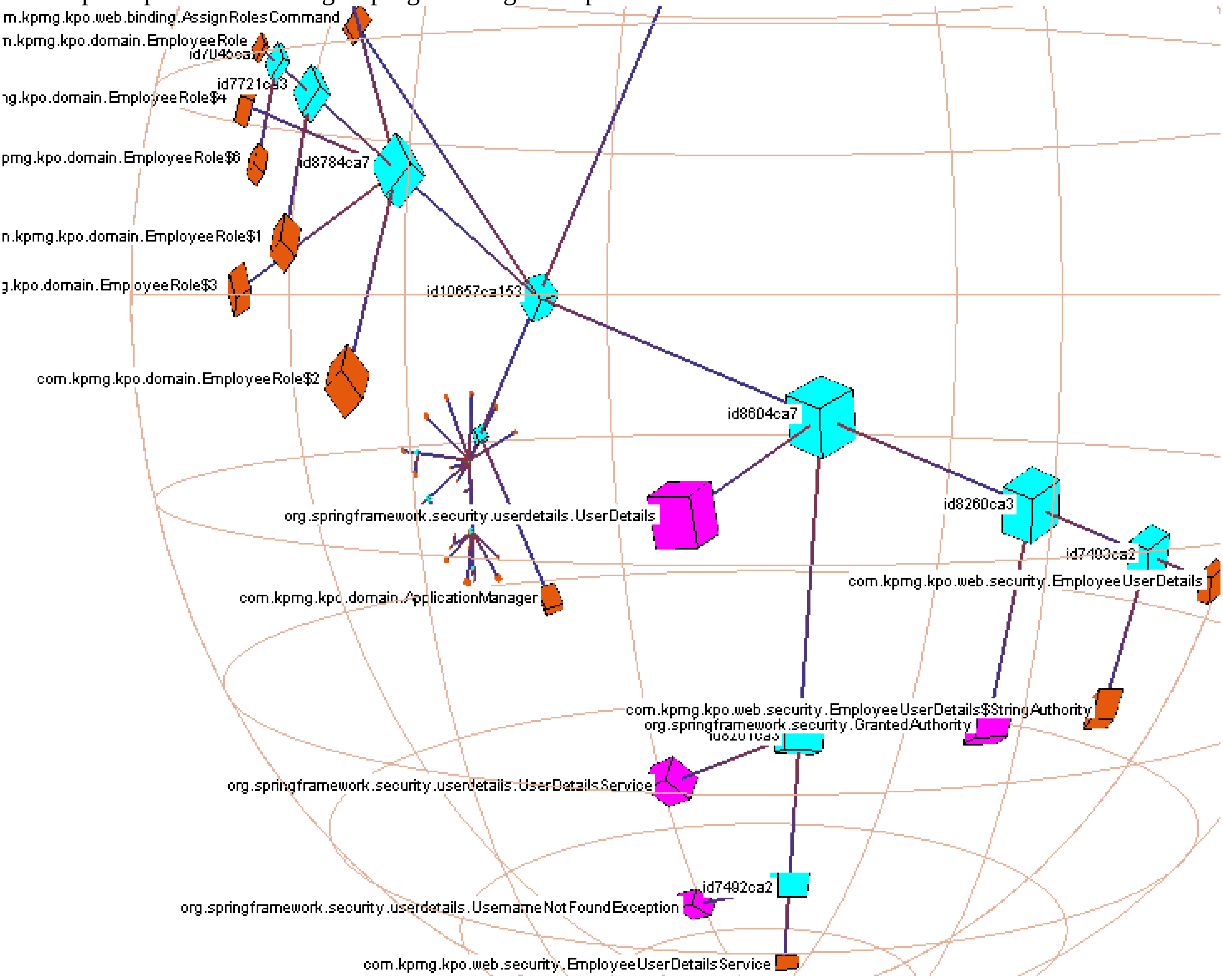

### 7.2.2.2 Hardly obvious from class name

In contrast to the previous example, it is not that easy to realize the purpose of a class called `com.kpmg.kpo.generated.jaxws.crm.CrmSOAP`. CRM is likely to stand for Customer Relations Management and SOAP is the well known (otherwise, it is as easy as a search in Google) Simple Object Access Protocol for exchanging structured information for web services. The latter two potential concepts are pretty distant from one another. Its situation in the clustering hierarchy makes things much more clear, namely, the following facts becomes apparent to a human software engineer:

- `…CrmSOAP` is much more about SOAP than CRM, because it is clustered together with SOAP-related library classes.
- If the software engineer was not familiar with SOAP, after seeing the clustering hierarchy he/she can realize that XML underlies SOAP, because the neighbor library classes are in `javax.xml` package

See Figure 7-4 below and a 3D view on the same part of the cluster tree in appendix 10.4.8.1.

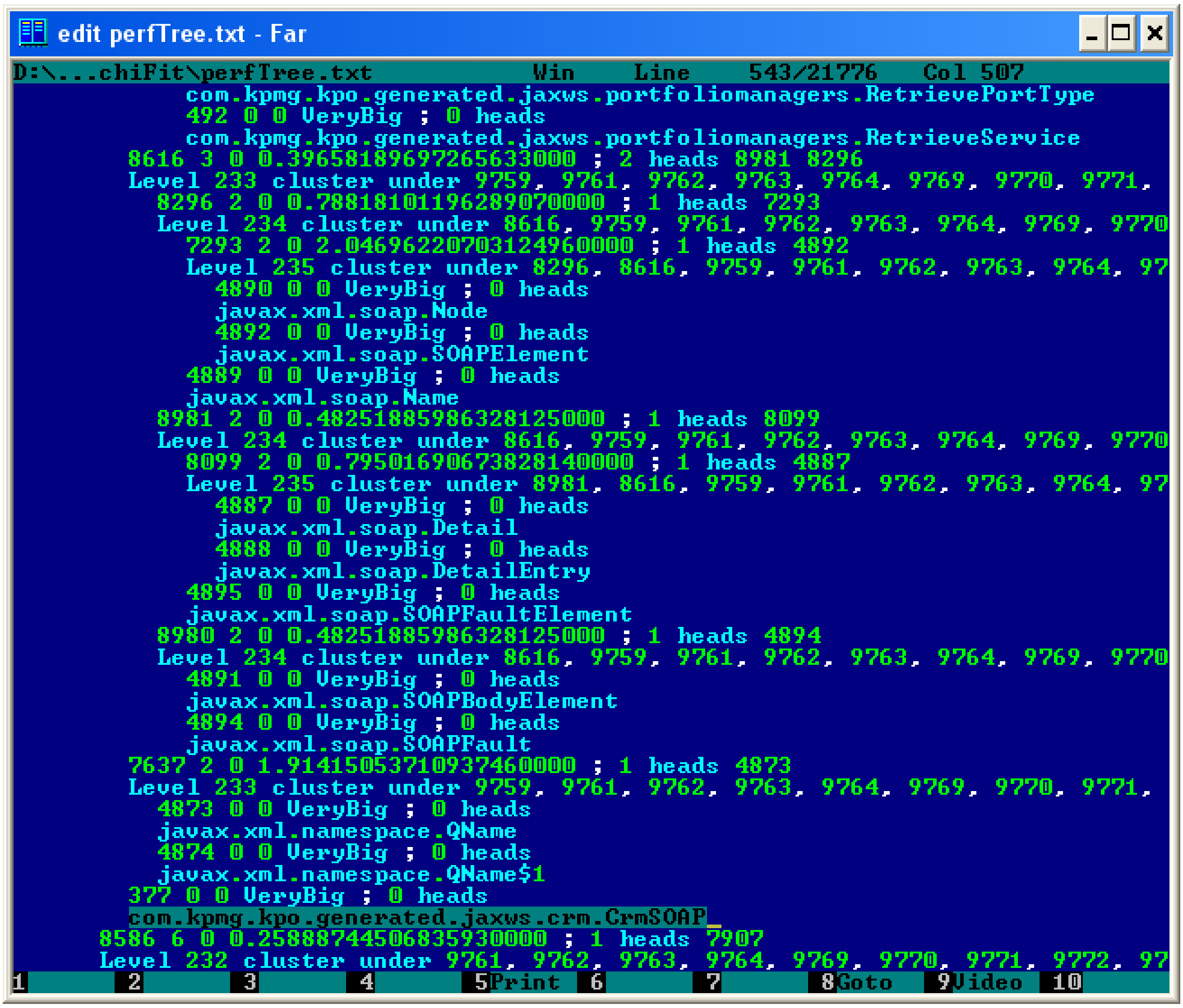

```
edit perfTree.txt - Far
D:\...chiFit\perfTree.txt          Win     Line     543/21776    Col 507
            com.kpmg.kpo.generated.jaxws.portfoliomanagers.RetrievePortType
            492 0 0 VeryBig ; 0 heads
            com.kpmg.kpo.generated.jaxws.portfoliomanagers.RetrieveService
        8616 3 0 0.3965818969726563300 ; 2 heads 8981 8296
        Level 233 cluster under 9759, 9761, 9762, 9763, 9764, 9769, 9770, 9771,
          8296 2 0 0.788181011962890700000 ; 1 heads 7293
          Level 234 cluster under 8616, 9759, 9761, 9762, 9763, 9764, 9769, 9770
            7293 2 0 2.04696220703124960000 ; 1 heads 4892
            Level 235 cluster under 8296, 8616, 9759, 9761, 9762, 9763, 9764, 97
              4890 0 0 VeryBig ; 0 heads
              javax.xml.soap.Node
              4892 0 0 VeryBig ; 0 heads
              javax.xml.soap.SOAPElement
            4889 0 0 VeryBig ; 0 heads
            javax.xml.soap.Name
          8981 2 0 0.48251885986328125000 ; 1 heads 8099
          Level 234 cluster under 8616, 9759, 9761, 9762, 9763, 9764, 9769, 9770
            8099 2 0 0.79501690673828140000 ; 1 heads 4887
            Level 235 cluster under 8981, 8616, 9759, 9761, 9762, 9763, 9764, 97
              4887 0 0 VeryBig ; 0 heads
              javax.xml.soap.Detail
              4888 0 0 VeryBig ; 0 heads
              javax.xml.soap.DetailEntry
            4895 0 0 VeryBig ; 0 heads
            javax.xml.soap.SOAPFaultElement
          8980 2 0 0.48251885986328125000 ; 1 heads 4894
          Level 234 cluster under 8616, 9759, 9761, 9762, 9763, 9764, 9769, 9770
            4891 0 0 VeryBig ; 0 heads
            javax.xml.soap.SOAPBodyElement
            4894 0 0 VeryBig ; 0 heads
            javax.xml.soap.SOAPFault
        7637 2 0 1.91415053710937460000 ; 1 heads 4873
        Level 233 cluster under 9759, 9761, 9762, 9763, 9764, 9769, 9770, 9771,
          4873 0 0 VeryBig ; 0 heads
          javax.xml.namespace.QName
          4874 0 0 VeryBig ; 0 heads
          javax.xml.namespace.QName$1
        377 0 0 VeryBig ; 0 heads
        com.kpmg.kpo.generated.jaxws.crm.CrmSOAP
      8586 6 0 0.25888744506835930000 ; 1 heads 7907
      Level 232 cluster under 9761, 9762, 9763, 9764, 9769, 9770, 9771, 9772, 97
1      2      3      4      5Print 6      7      8Goto  9Video 10
```

**Figure 7-4 CrmSOAP and neighbors**

### 7.2.2.3 Not obvious from class name

In this example a class is called `…AuditEntryDTO` which says nothing about its purpose, unless we know that the software project is heavily related to Auditing business and lookup DTO in Wikipedia: http://en.wikipedia.org/wiki/Data_transfer_object . After the above two steps we know still do not know why it is “Entry”, i.e. entry of what?

However, a glance at the clustering hierarchy makes things clear; perhaps even replacing the need for the two aforementioned steps, see Figure 7-5 below. Apparently, there is some logging (classes containing `"Trail"`, which is a synonym for logging, and `ConsoleAuditLogImpl`). That is why “Entry” – it is an entry of some log (namely, audit log). And the logging is implemented as a service, transferring `AuditEntryDTO` objects between software application subsystems.

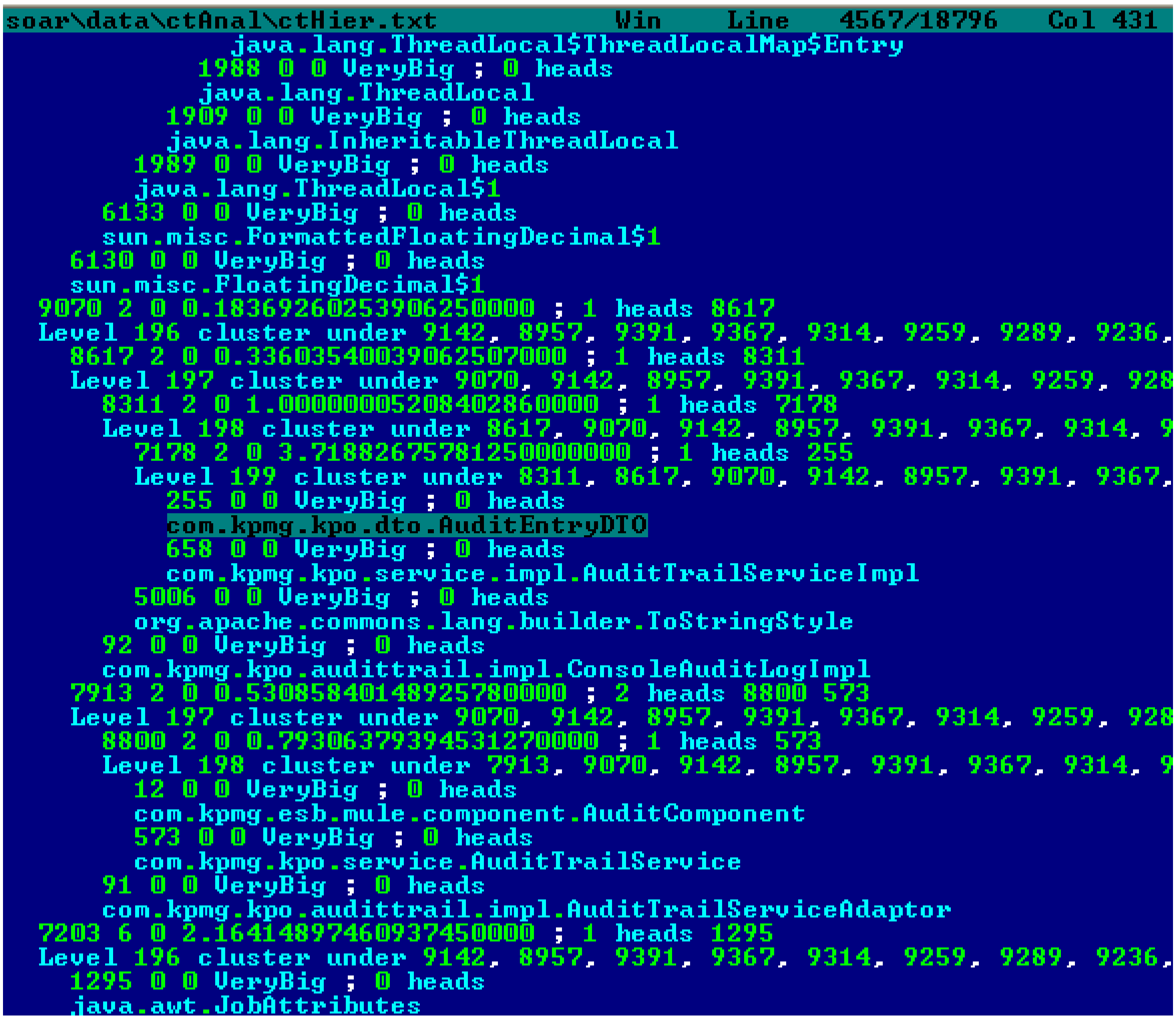

**Figure 7-5 Class AuditEntryDTO**

#### 7.2.2.4 Class name seems to contradict the purpose

While browsing through the clustering hierarchy we encountered an example where class name seems to contradict the purpose. Though a class is named `com.kpmg.kpo.action.GenericHandler`, it appears in the cluster that addresses Java Regular Expressions and expression evaluation in JBPM, http://www.jboss.org/jbpm . This is a strong claim involving also doubts about clustering quality, thus we looked into the source code of the class (provided in section 10.3.3), which is obviously confirming the result of clustering.

Let us look at this case closer in terms of software quality. The fact that class name contradicts the purpose *does* **not** *just mean* that the class is named incorrectly, which can seem a minor defect. Indeed, it means that the designers of the architecture *saved efforts* (i.e. took "**reduce quality**" action, as we discussed in section 2.6) at some point during software development cycle. What can be the reason for not naming a class properly? Most likely, it happened because *the purpose was not identified* properly, and identification of purpose constitutes significant amount of design efforts.

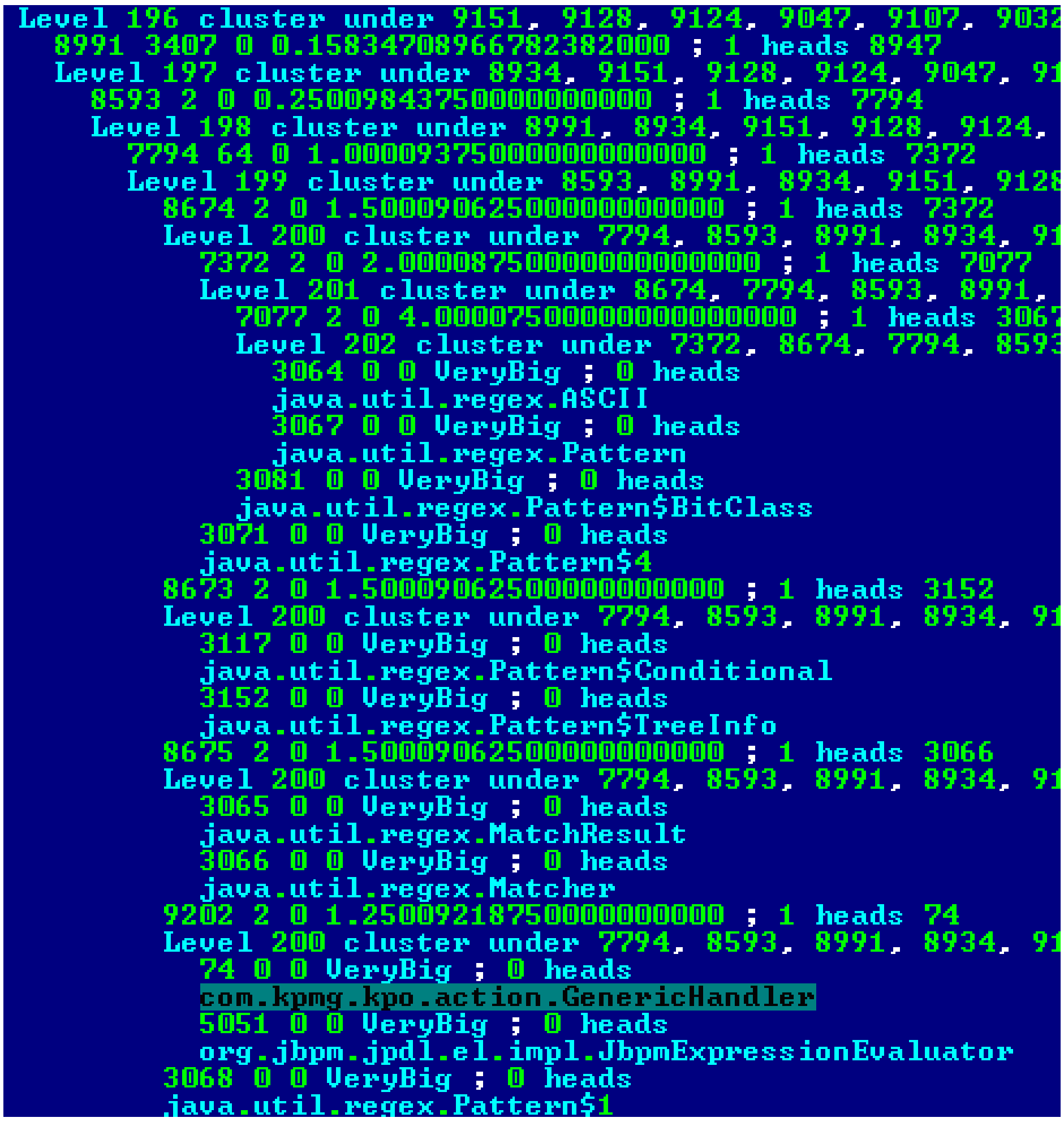

**Figure 7-6 GenericHandler contradicting the purpose**

We identify two consequence of such fact, for programmers and for business:

- Programmers (developers working directly with source code) get a wrong idea about the purpose of the class when considering its reusage, e.g. through inheritance, modification (adding/removing/changing methods) or simply usage from another place in the software
- Companies that buy or take for outsourcing services the source code containing such architectural violations, get less value than they may think they get, as at some point the earlier saved design efforts will "pay-off" with unexpected expenses

We studied the software upon analysis further in order to provide evidence for the claim that this architectural violation propagates into the rest of source code, if not fixed timely. Indeed, class `…GenericHandler` is inherited by 4 loosely related classes, we give their names below: `com.kpmg.kpo.action.{AbstractDocumentHandler`, `PrintOutAction`, `SendFile` and `SendNotification}`, while other "action" classes in the package do not. In general, the package `com.kpmg.kpo.action` is suspected to have low quality.

### 7.2.3 Classes that act together

In our view, the most valuable inference InSoAr makes is detection of sets of classes that act together. This follows directly from the property of clustering and the data we analyze: coupling of SE classes within a cluster is higher than coupling between clusters. To our knowledge, there is no means to identify efficiently (in terms of human efforts) such groups with any existing static or dynamic code analysis tools. As was discussed in section 2.2, state of the art tools either allow user to select a set of SE artifacts for which the user wants to see their couplings, or to drill down from packages to subpackages, classes and methods ([Stan2009], [Rou2007], [Pin2008]). In contrast, we do this globally, for all the SE classes at once. Less coupled classes get into a group only after more coupled classes have been sent into that group, where the former stands for higher levels of the clustering hierarchy (closer to the root), and the latter stands for lower levels (closer to the leaves).

Below is an example a piece of XML output that demonstrates the claim.

```
<node id="8837" childCount="2" alb="0.84384472656250000000" heads="7179, 8897">
  <node id="8897" childCount="2" alb="0.89071943359375000000" heads="7179">
    <node id="7179" childCount="2" alb="2.00008750000000000000" heads="241">
      <node id="9104" childCount="2" alb="2.12508671874999950000" heads="241">
        <node id="241" label="com.kpmg.kpo.domain.TaskInstance" />
        <node id="654" label="com.kpmg.kpo.service.impl.AbstractTaskInstanceService" />
      </node>
      <node id="790" label="com.kpmg.kpo.web.view.TaskInstanceView" />
    </node>
    <node id="550" label="com.kpmg.kpo.jbpm.AssignToEmployee" />
  </node>
  <node id="8898" childCount="2" alb="0.93759414062500000000" heads="219">
    <node id="219" label="com.kpmg.kpo.domain.PeerStatusType" />
    <node id="712" label="com.kpmg.kpo.usertypes.PeerStatusTypeUserType" />
  </node>
</node>
```

**Figure 7-7 Classes that act together (XML)**

In the figure above we see 6 classes from 5 different packages under `com.kpmg.kpo` are indeed a single subsystem, according to the **implicit** architecture, while package structure can be viewed as a kind of *explicit* architecture. Modern integrated development environments (IDEs), e.g. Eclipse or Microsoft Visual Studio, can easily show all the classes/files in a package/namespace, telling a software engineer about the explicit architecture. However, there is no way in these leading IDEs to show what we have shown in Figure 7-7. At present, software engineers can only get such diagrams from explicit software architecture, e.g. a subsystem or coupling documented in Software Design Document.

As this is the central inference in which InSoAr specializes, we provide further evidence for the quality of hierarchical clustering and meaningfulness of the results as a number of images showing different parts of the system, see appendix 10.4.8.2 and across the paper. Though it is hard to prove that this property also holds at global level due to large visualizations required, we claim that this result is not local and not random, i.e. parts which are not shown in our pictures, look fine and reflect the implicit architecture too. We kindly ask an unconvinced reader to download the samples from the internet (section 7.2) and try them himself/herself.

### 7.2.3.1 Coupled classes are in different packages

Detection of class coupling across packages/namespaces is important for the reasons discussed throughout the paper (implicit architecture without scanning millions of lines of source code manually), we just give a few examples below as the evidence that InSoAr does grouping of classes together according to their unity of purpose, which can be validated from the names of the classes.

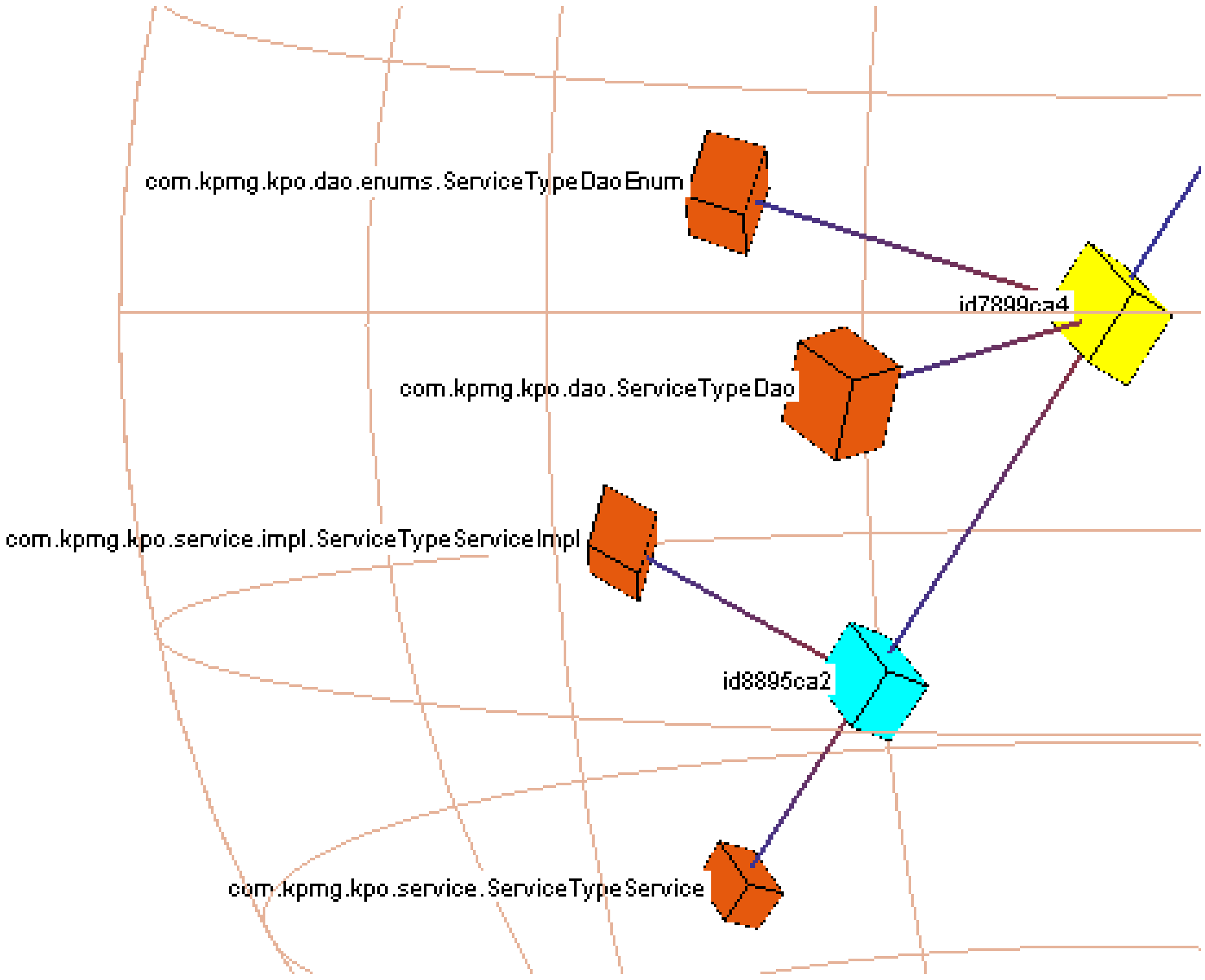

```
8703 2 0 0.218360354614257820000 ; 2 heads 9013 194
Level 196 cluster under 9142, 8957, 9391, 9367, 9314, 9259,
  8374 2 0 0.288184136962890640000 ; 1 heads 152
  Level 197 cluster under 8703, 9142, 8957, 9391, 9367, 9314
    7892 2 0 0.769626440429687500000 ; 1 heads 114
    Level 198 cluster under 8374, 8703, 9142, 8957, 9391, 93
      8896 2 0 0.995699246215820300000 ; 1 heads 657
      Level 199 cluster under 7892, 8374, 8703, 9142, 8957,
        572 0 0 VeryBig ; 0 heads
        com.kpmg.kpo.service.AuditLevelService
        657 0 0 VeryBig ; 0 heads
        com.kpmg.kpo.service.impl.AuditLevelServiceImpl
      114 0 0 VeryBig ; 0 heads
      com.kpmg.kpo.dao.AuditLevelDao
    152 0 0 VeryBig ; 0 heads
    com.kpmg.kpo.dao.enums.AuditLevelDaoEnum
  9013 2 0 0.234473535156250000000 ; 1 heads 194
  Level 197 cluster under 8703, 9142, 8957, 9391, 9367, 9314
    194 0 0 VeryBig ; 0 heads
    com.kpmg.kpo.domain.AuditLevel
    707 0 0 VeryBig ; 0 heads
    com.kpmg.kpo.usertypes.AuditLevelUserType
8732 2 0 0.226661083984375000000 ; 1 heads 8602
Level 196 cluster under 9142, 8957, 9391, 9367, 9314, 9259,
```

### 7.2.3.2 Coupled classes are in the same package

As software engineers often put most coupled classes in the same package/namespace, in addition to naming them similarly, the fact that classes from the same package appear nearby in the clustering hierarchy can serve as validation for the clustering results. We can observe a match of the explicit and implicit architecture in this case. A useful fact that becomes apparent after looking at a cluster of classes from the same package, differentiation of coupling, is discussed in section 7.2.5.

In Figure 7-8 below we can see a number of classes that act together. Class `PortfolioManagerComponent` is from package `com.kpmg.esb.mule.component`, while the rest of classes are from package `com.kpmg.service.portfoliomanager`. We can infer that, most likely, `PortfolioManagerComponent` is a high level class that operates the simple classes in its cluster. Classes `PortfolioManagerComponent`, `PortfolioManagerType` and `ServiceType` (cluster #8964) are the **most coupled** among the group displayed in the figure. The second highly coupled group consists of classes `RetrieveFault` and `RetrieveFault_Exception`, cluster #9408. The latter two groups, together with two more classes, `PortfolioManagersType` and `RetrieveResult`, form a larger group #9409. Only afterwards the rest of classes displayed in the picture (except `RetrievePortType`) attach to this group, and thus to all the classes in it. This happens in cluster #7445, which is at a higher level of clustering hierarchy than cluster #8964, #9408 or #9409 and the interaction density (coupling) is lower among the classes of group #7445.

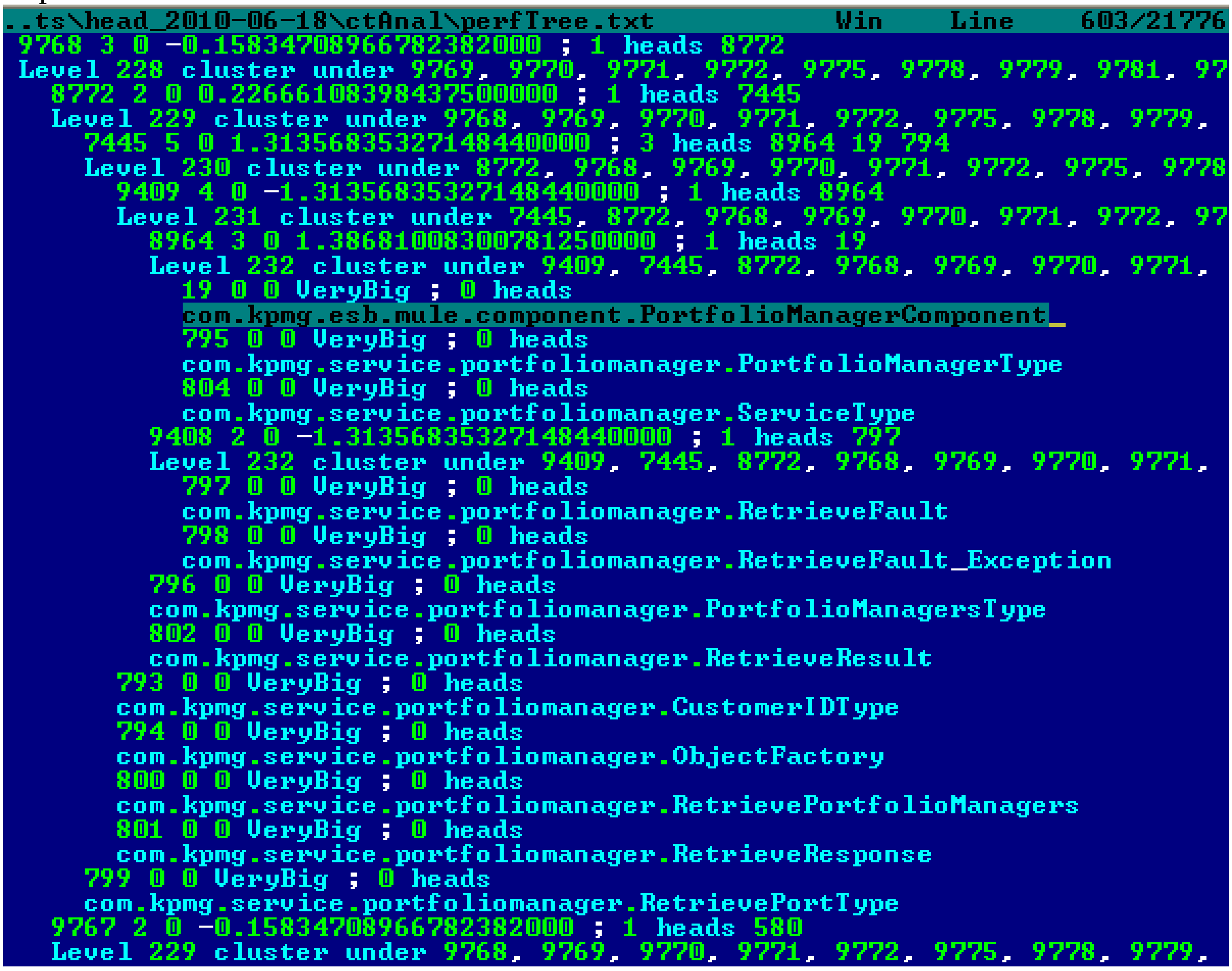

**Figure 7-8 Except PortfolioManagementComponent, coupled classes from the same package**

### 7.2.4 Suspicious overuse of a generic artifact

In this example we see a class called `GenericComponentException` which is, however, coupled with class `DocumentComponent` from a different package, see Figure 7-9. We rather mention the fact that the classes are from different packages for convenience of the reader, in order not to forget that state of the art tools cannot help. However, the observation that helps to discover an issue here is in the fact that the class representing the exception is called "Generic…" while in the cluster tree we can see that it is coupled and thus serves error-handling for a specific class `DocumentComponent`.

We can guess (without looking at the source code, thus saving efforts 1000 times) that this happened

- either because the purpose of `GenericComponentException` was not well identified while designing the architecture, and it should rather be called `DocumentComponentException` (or something even more specific – a study of the source code is needed),
- or because even though the purpose of `GenericComponentException` was well identified and at some places in the source code it indeed serves as a generic artifact (e.g. as the base for inheritance to more specific exceptions), during the evolution of software it happened that this "generic" artifact was *too heavily used* in class `DocumentComponent`.

In the second case, a suggested improvement of the architecture is to create another exception class specific to `DocumentComponent`, e.g. "`DocumentComponentException`" and **refactor** the source code of `DocumentComponent` to make it using this dedicated specific artifact.

With the two points above we have exhausted the possible cases, i.e. there is no reason to call an exception-class **`Generic`**`ComponentException` while it mostly serves (and is mostly coupled with) a class called `DocumentComponent`. Thus the source code is not optimal, while detecting such a defect in novel source code (i.e. when there is no programmer that knows about it) is not possible with state of the art tools, except that by scanning all the source code line by line. The benefit of clustering is obvious: 10M lines of source code vs. 10K nodes in the cluster tree.

At any rate, the detected architecture violation says about saved efforts during the design of the software, and will result in unexpected expenses later, by analogy to what we discussed in section 7.2.2.4.

```
:\...PMG\[redacted]\cg-spark-mfir\strat2\final.txt          Win     Line    4495/18796    Col 452
       javax.security.auth.Destroyable
     2407 0 0 VeryBig ; 0 heads
     java.security.KeyStore$ProtectionParameter
   7246 2 0 2.6719583007812506000 ; 1 heads 2400
   Level 197 cluster under 8717, 9142, 8957, 9391, 9367, 9314, 9259, 9289, 9236, 9387, 93
     2400 0 0 VeryBig ; 0 heads
     java.security.KeyStore$Builder$FileBuilder
     2401 0 0 VeryBig ; 0 heads
     java.security.KeyStore$Builder$FileBuilder$1
 8781 2 0 0.2061533996582031200 ; 1 heads 7854
 Level 196 cluster under 9142, 8957, 9391, 9367, 9314, 9259, 9289, 9236, 9387, 9333, 9357
   7854 2 0 1.0098593139648437000 ; 2 heads 8923 7171
   Level 197 cluster under 8781, 9142, 8957, 9391, 9367, 9314, 9259, 9289, 9236, 9387, 93
     8923 2 0 1.4571221435546873000 ; 1 heads 7171
     Level 198 cluster under 7854, 8781, 9142, 8957, 9391, 9367, 9314, 9259, 9289, 9236,
       7171 2 0 2.9532065429687497000 ; 1 heads 13
       Level 199 cluster under 8923, 7854, 8781, 9142, 8957, 9391, 9367, 9314, 9259, 9289
         13 0 0 VeryBig ; 0 heads
         com.kpmg.esb.mule.component.DocumentComponent
         259 0 0 VeryBig ; 0 heads
         com.kpmg.kpo.dto.DocumentMessage$Builder
       33 0 0 VeryBig ; 0 heads
       com.kpmg.esb.mule.exception.GenericComponentException_
     22 0 0 VeryBig ; 0 heads
     com.kpmg.esb.mule.component.TestComponent
   7450 3 0 1.1407178710937500000 ; 1 heads 73
   Level 197 cluster under 8781, 9142, 8957, 9391, 9367, 9314, 9259, 9289, 9236, 9387, 93
     73 0 0 VeryBig ; 0 heads
     com.kpmg.kpo.action.DocumentCommand
     80 0 0 VeryBig ; 0 heads
     com.kpmg.kpo.action.PutDocumentsInFile
     83 0 0 VeryBig ; 0 heads
     com.kpmg.kpo.action.SendDocumentForApproval
 9061 2 0 0.1739270385742187600 ; 1 heads 8709
 Level 196 cluster under 9142, 8957, 9391, 9367, 9314, 9259, 9289, 9236, 9387, 9333, 9357
```

**Figure 7-9 GenericComponentException serving mostly DocumentComponent**

### 7.2.5 Differentiation of coupling within a package

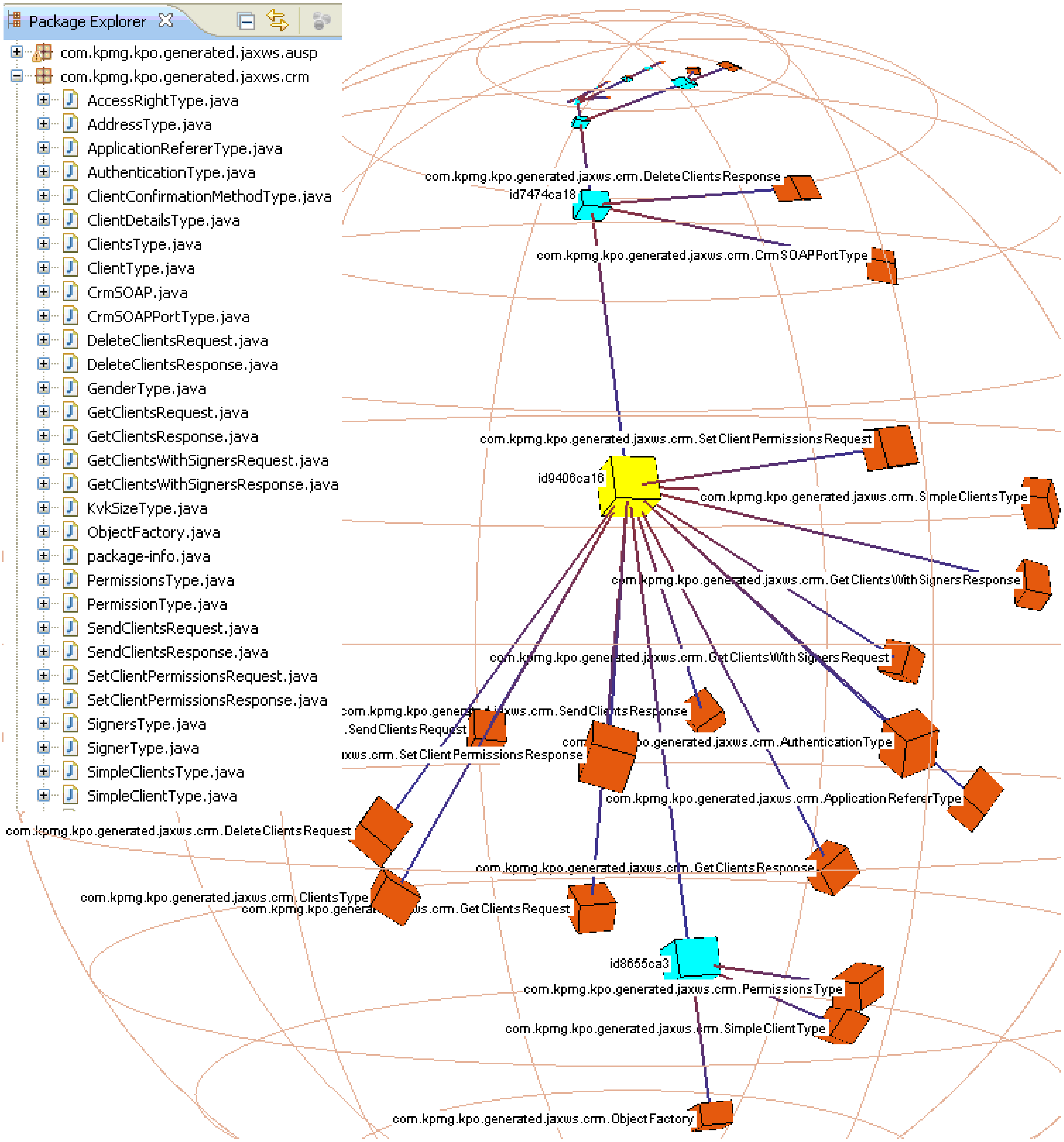

**Figure 7-10 Many classes at the same package in IDE**

Often there are too many classes in one package, which hinders comprehension for a software engineer looking at the package/namespace explorer in IDE[1]. In Figure 7-10 above we demonstrate such an example, how it looks in a popular Java IDE (Eclipse) and how it looks after computing the clustering hierarchy with our approach. A package containing even more classes with very different purposes (in contrast to what we observe here) is demonstrated in section 7.2.6. In principle, InSoAr differentiated the coupling within that package too, but there is a separate fact to be discussed because the classes appeared scattered across the system.

[1] IDE – Integrated Development Environment

Describing more extensively, Figure 7-10 shows two representations of a set of classes, while not all the classes from the left side have to be present in the right side: the rest can appear somewhere else in the cluster tree. An alphabetical list of classes in a package is on the left, and this is what a software engineer sees with state of the art tools (IDEs). A subtree containing many of the classes from the list is displayed on the right, and this is what we can see in the cluster tree produced by InSoAr. The task here is that a user needs to infer the **purpose** of these classes or **how they are related** to each other, including a global understanding, i.e. ***not** just pairwise relations*. Our argument is that this is much (in this context, our "much" usually means 1000 times across the paper) easier to do having the cluster tree.

When the class names in the package are not very meaningful, accomplishing of this task for a human expert amounts to scanning the source code of the classes, which is usually 1000 lines per class. Even after scanning the source code, there is a comprehensional difficulty in taking into account thousands of the observed facts at once (for humans). To alleviate this, human experts need some diagrams to be drawn, which is a mechanical difficulty. In the remaining case when the class names are very meaningful, the user can pick out the groups from the list, which is debatably $\boxed{O(N \cdot \log N)}$ operations in the mind of the user (if the user follows sorting based on pairwise similarity comparison, and disjoint subsets unification algorithms), where N is the number of classes in the package: the classes are sorted alphabetically, however the first token is not necessary the one that gives the user an idea about proper grouping, think of `GetClientsResponse` and `SendClientsResponse` in Figure 7-10 above. Even in this rare case of very meaningful names of classes in a package containing many classes, obviously, a software engineer benefits from having the cluster tree.

From Figure 7-10, as well as Figure 10-10 (appendix 10.4.3.2) and Figure 7-11 demonstrating more or less the same fragment of the clustering hierarchy, we can see that coupling of classes differs even though they are in the same package and appear as a plain list in IDE. We claim that this *differentiation is an important feature* that facilitates program comprehension by a software engineer. For example, we immediately see that class `ObjectFactory` has a different nature than `GetClientsRequest`, or `GetClientsResponse`, or others from that upper group in Figure 7-10. The bottom group is the most coupled within itself, and then to the rest of classes than any other class shown in the figure. Without looking at a single line of source code, we guess that `ObjectFactory` is some manager-class, while `Permission`**`s`**`Type` (note "s" after "Permission") and `SimpleClientType` are the most thorough watched by it.

On the other hand, we also see in Figure 7-11 evidence for correctness of grouping. Classes `PermissionType` (note the absence of "s" after "Permission") and `AccesRightType` got clustered together, and guess from the names that this is semantically true.

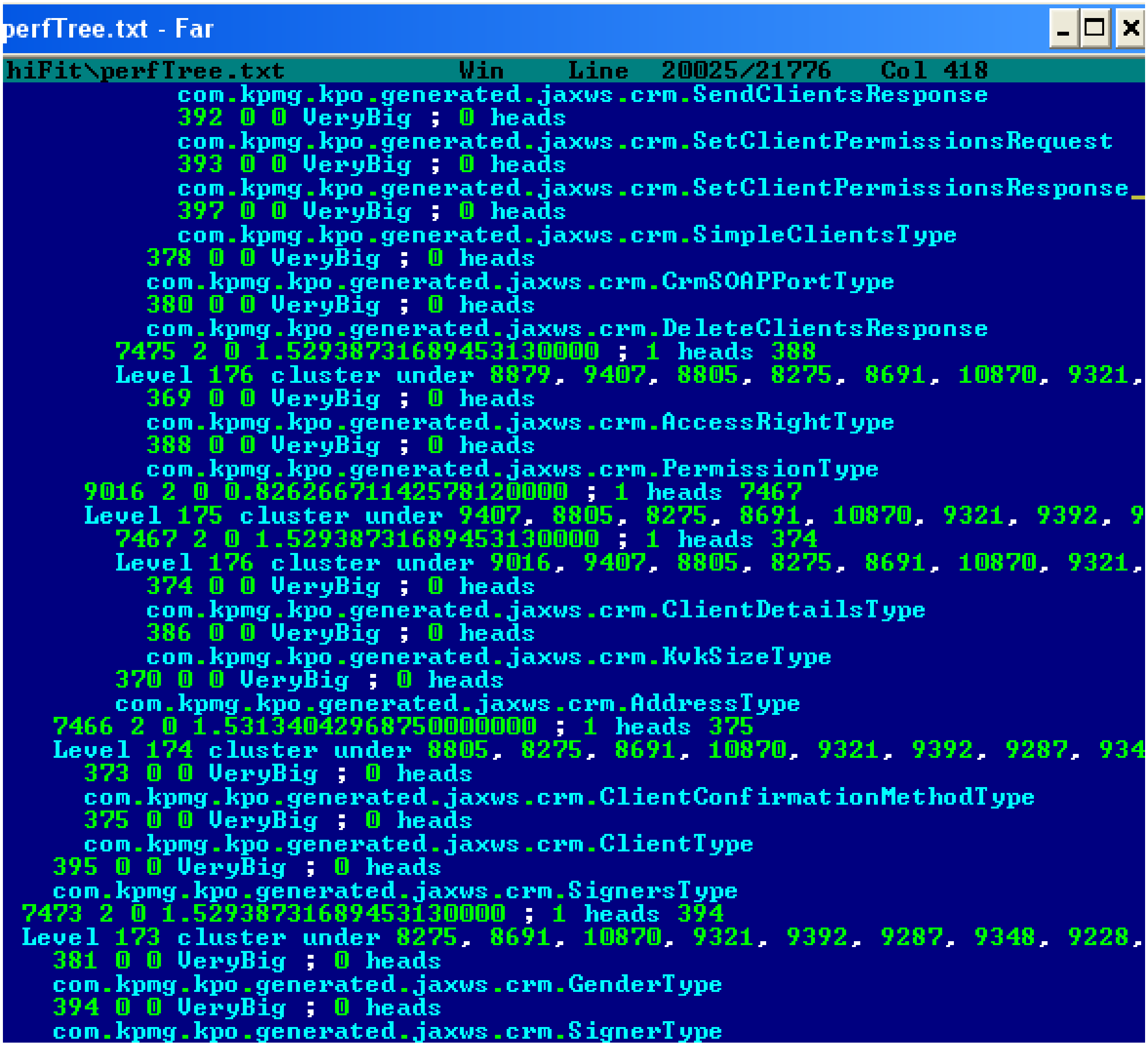

**Figure 7-11 Cluster tree indented by node depth**

Apparently, representational power differs across the three our textual representations of cluster tree, in terms of number of labels (only leaf nodes have them) that can be shown to a user within limited space and the easiness of interpretation of the presented information by the user (software engineer). The bracketed approach, Figure 10-10 in appendix 10.4.3.2, is the most powerful in terms of the number of labels (classes, leaf nodes) that can be displayed within the same space. However, efficient comprehension of this representation needs some training and familiarity with nested structures, i.e. trees where only leaf nodes have labels.

### 7.2.6 A package of omnipresent classes

Another example is essential for understanding of our endeavor and the advance over state of the art tools. In section 7.2.5 we have shown that InSoAr can differentiate coupling within a package and thus facilitate comprehension of the package and classes in it. However, the classes from that package were devoted to more or less the same purpose.

In this section we show a principally distinct case, where even though classes are in the same package, they serve different purposes. Figure 7-12 shows on the left how such a package looks in Integrated Development Environment (IDE), and splitting classes into packages is an instance of explicit architecture being declared and implemented by software engineers. However, according to how the source code was written (the implicit architecture), each (well, almost each) of these classes is coupled with a distinct group of classes from other packages (serves a distinct purpose), and this is shown on the right of Figure 7-12. Further evidence is provided in appendix 10.1.1. Thus we conclude that developers grouped classes into `com.kpmg.kpo.domain` package according to some more high-level purpose, e.g. because the classes are omnipresent (and our decomposition of the software system says that they are indeed omnipresent).

On the other hand, we can also see that package `com.kpmg.kpo.domain` is 28th (out of 98) in ubiquity among client-code container artifacts (packages, or classes that have nested classes) and ranked as 338th (out of 1235) among all the containers including libraries, see Figure 10-3. Its average merge height (the second column) is 20, which is not high relatively. This means that the classes of this package become united into a single subsystem (containing classes from other packages too) more or less soon, not too far from the bottom of the hierarchy. Thus we conclude that there is still another high-level purpose, except omnipresence discussed in the previous paragraph.

Without looking at a single line of the source code of either of these classes, we will not be surprised in case their mission is to support Object-to-Relational Mapping (ORM)[2], where a class is also a table in the database, and instances of this class are also rows of that table[3]. We conclude this from the following facts (and of course, InSoAr gave us those facts):

- class `com.kpmg.kpo.domain.DomainEntity` got clustered with `com.kpmg.kpo.dao.Dao`[4] and `com.kpmg.kpo.dao.DaoFactory` (see the top-right part of Figure 7-12)
- class `com.kpmg.kpo.domain.AuditLevel` got clustered with `com.kpmg.kpo.dao.AuditLevelDao` (see the bottom-right part of Figure 7-12)

Furthermore, we see that this ORM is supplied by Hibernate[5] technology, as in Figure 10-1 (appendix 10.1.1) class `com.kpmg.kpo.domain.ArchiveEntry` got clustered with `com.kpmg.kpo.dao.hibernate.HibernateArchiveDao` class.

To recap, just by looking at the hierarchy produced with InSoAr, we realized:

- The high-level purpose a group of SE artifacts (Java classes here) serves
- The lower-level purpose for each SE artifact, by looking at the classes with which it is coupled

No other tool can to this extent facilitate comprehension of software system by humans.

[2] ORM: http://en.wikipedia.org/wiki/Object-relational_mapping

[3] In software engineering, it is proper to speak about **instances** of classes here, because an object is an instantiated class.

[4] DAO – Data Access Object, http://en.wikipedia.org/wiki/Data_access_object#Advantages

[5] Hibernate – is an ORM library for Java, http://en.wikipedia.org/wiki/Hibernate_%28Java%29

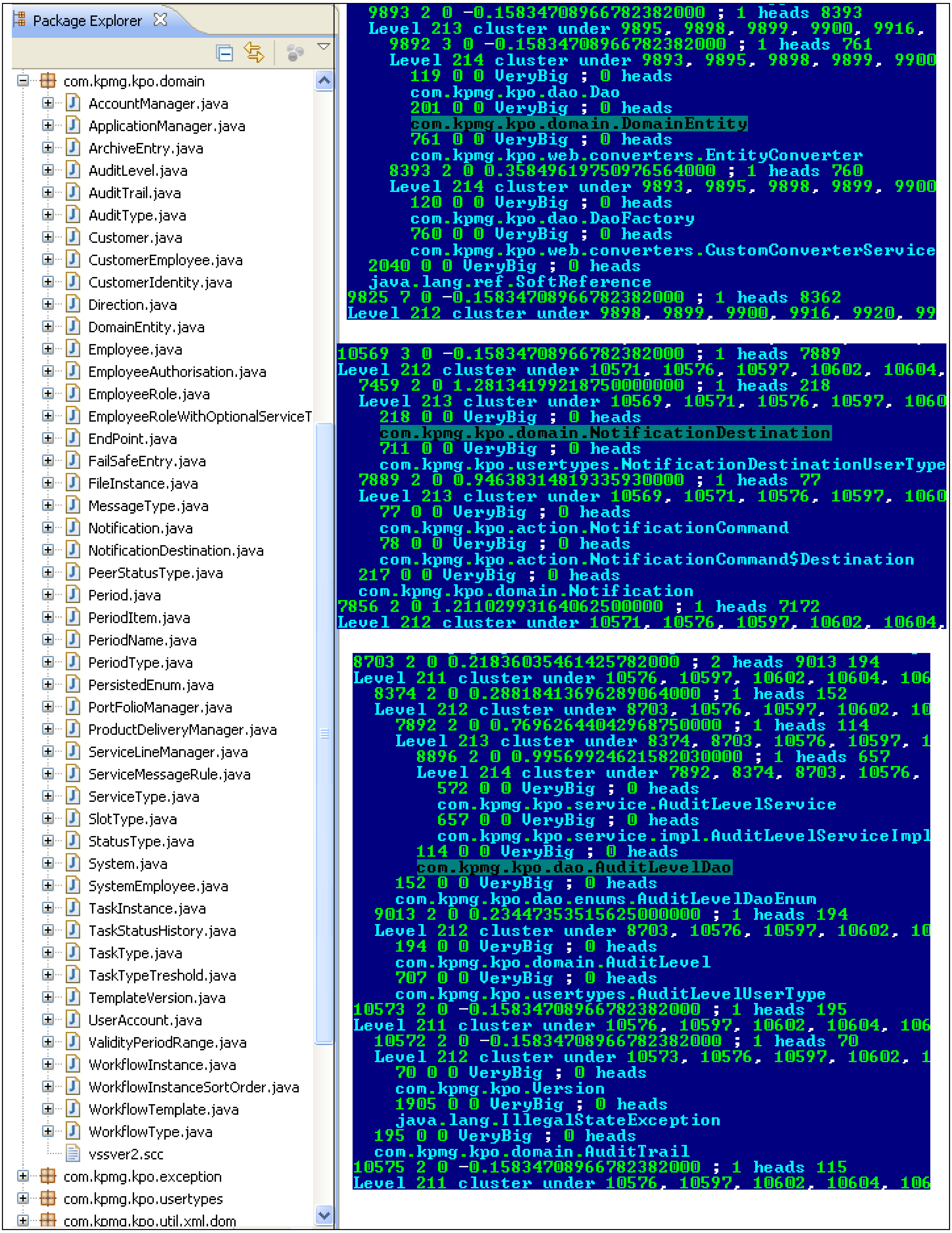

**Figure 7-12 Single package in IDE, but multiple differen logical subsystems**

### 7.2.7 Security attack & protection

Security attack and protection is usually a dual task, like cryptography and cryptanalysis. Thus discussing one we usually mean both. In terms of software protection, many schemes rely on incomprehensibility of the protection mechanism for an attacker. An example is injection of serial key/license checking code (instructions, subroutines, classes) all across the program being protected. Apparently, this leads to coupling of SE artifacts in the program onto the security mechanism, and the participants of security mechanism itself will be clustered together, as we discussed in general for classes that act together (section 7.2.3). In the above scenario (injection all across the program), the security mechanism becomes a group of utility artifacts, and we will observe the same effect as in appendix 10.1.7 or section 7.2.6 for utility or omnipresent artifacts in general. Thus, an attacker is able to **identify and circumscribe** the security mechanism and study its couplings efficiently using the general techniques we discuss in this paper, subsections of 7.2 and appendixes 10.1, 10.4.3 – 10.4.8.

This approach works even when only binaries are available (Soot extracts relations from binary code too) and when the binary code is obfuscated[6]. Security mechanism will get clustered together anyway, and non-obfuscated neighboring artifacts (library classes or at least lower-level OS subroutines) will discover its purpose. In the rare case when everything is obfuscated down to machine code level, i.e. I/O ports and interrupts, clustering of dynamically extracted graph of relations can be used for efficient discovery of the security mechanism.

Pursuing the goal of protection, one can do the same: study how the security mechanism looks after clustering, whether it is easy to identify and circumscribe, and whether the latter information provides an attacker with sufficient means for breaking the security. There are some nuances, however: even though the security mechanism may seem strong to a defender using one parameter set when clustering, another parameter set (e.g. by selecting some other options from discussed in sections 5.1, 5.2, 5.3.2 and 5.6.2) may still exhibit the weaknesses of the security mechanism.

The evidence – (a part of) security mechanism identified in “dem0” project – is provided in appendix 10.1.2:

- Even if we did not know the purpose of client-code classes `EmployeeUserDetailsService`, `EmployeeUserDetails` and `StringAuthority` (nested in `EmployeeUserDetails`), e.g. due to obfuscation, we could determine it from the library neighbors from package `org.springframework.security`.
- If we know weaknesses of classes coupled with the security mechanism (and thus clustered together), we can attempt to exploit them to compromise the security mechanism, even though the latter is strong itself. Examples of such potential targets visible in the picture in appendix 10.1.2 are:
  - `WebservableObjectInToAuditableObjectTransformer`: how is it about boundary cases?
  - `org.apache.log4j.Logger`: can we inject our code into this class, or substitute it entirely by changing the CLASSPATH on the server upon attack?

InSoAr is not an ultimate tool for compromising security. Structural security is vulnerable. Algorithmic (e.g. Petri networks) or mathematical (e.g. factorization) methods will sustain.

### 7.2.8 How implemented, what does, where used

In the two figures below we see the classes serving time and scheduling clustered together in a subtree. We see that the time & scheduling subsystem has recurrence rules, rule factories and a rule service, Figure 7-13. There is a base class `com.kpmg.reccurrence.RecurrenceRule`, while weekly and monthly recurrence rules inherit from it. A more specific part of this subsystem is shown in appendix 7.2.8. We see that there is also quarterly recurrence rule and factory. The more specific part, Figure 10-2, contains also classes for representing the days of week and week of month. On the other hand, the more general part further gets joined with classes `TaskInstanceCommand`, `CreateWorkflowInstanceAction`, `CreateInstanceCommand` and `CreateInstanceValidator`, see Figure 7-13. The latter mentioned group of classes, obviously, uses the time and scheduling subsystem, e.g. `com.kpmg.web.action.CreateWorkflowInstanceAction` allows user to define some scheduled workflow action. Note that the classes are from different packages. We cannot figure out this configuration using any state of the art tools.

[6] Obfuscated source or binary code is the one that has been made difficult to understand, by e.g. replacing Java class names with some meaningless strings. http://en.wikipedia.org/wiki/Obfuscated_code

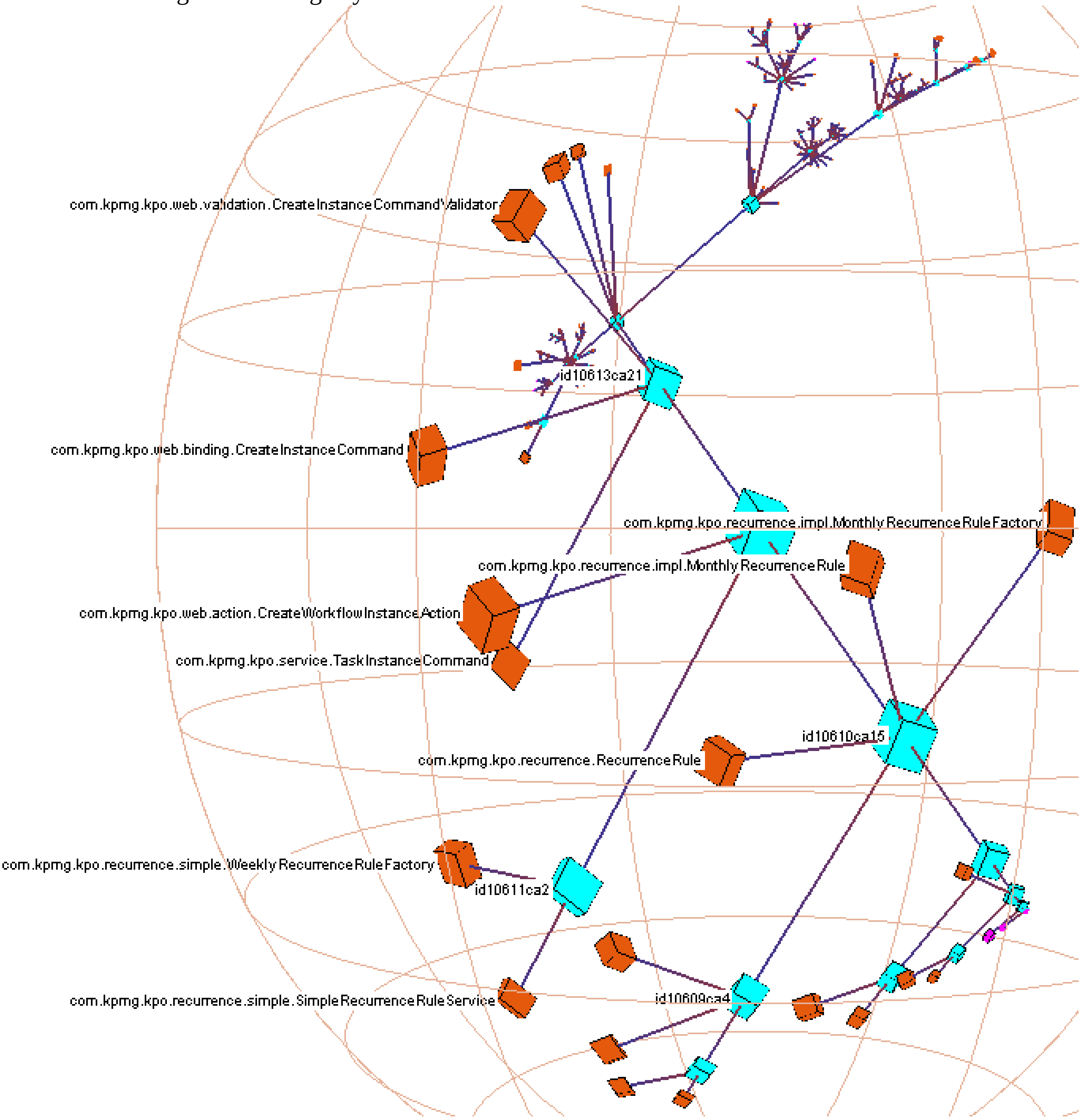

**Figure 7-13 Time & Scheduling subsystem**

# 8 Further Work

## 8.1 A Self-Improving Program

In order to make a program that improves itself we first need to make a program that improves programs and point it to itself. In its turn, prior to making a program that improves programs, we need a program that understands programs, at least in the way humans do. Obviously, ability to understand requires ability to analyze as a prerequisite. State of the art source code analysis tools exist, but they do processing without understanding. In this project we have implemented a program that infers structure of source code to facilitate its further comprehension by humans. A further work direction is implementing a program that attempts to comprehend the structure without humans and then does some forward-engineering of improved source code.

## 8.2 Cut-based clustering

As we have discussed in section 5.1, Flake-Tarjan clustering [Fla2004] is only available for undirected graphs. This restriction is posed by minimum cut tree algorithm [GoHu1961], but maximum flow algorithms are available for *both* directed and undirected graphs. To satisfy this restriction, in our project we were converting directed graph of software engineering artifact relations into undirected using normalization akin to the one described in [Fla2004] and PageRank [Brin1998]. Though the clustering demonstrated good results, it is obvious that important information is lost during the conversion from directed to undirected graph. Within our project we have also tried to eliminate the constraint posed by minimum cut tree, as we do not need a full-fledged cut tree, but only the edges separating the artificial sink from the rest of the tree [Fla2004]. However, this is a hard theoretical task being too risky given the nature of our project (master thesis).

Thus, as a direction for further work we propose eliminating the requirement of undirectedness from Flake-Tarjan clustering, thus devising a version that takes a directed graph on input, uses directed max flow algorithm in the backend, and somehow workarounds minimum cut tree exploiting the fact that we need clustering of a directed graph, but not the entire correct minimum cut tree of an undirected graph.

## 8.3 Connection strengths

Our normalization, motivated in section 4.5 and provided in section 5.1, lets clustering produce good results (section 7) alleviating the problem of utility artifacts. It is interesting to investigate, whether edge weighting considering more properties (fan-out, graph-wide facts) can result in even better clustering hierarchy.

## 8.4 Alpha-threshold

We observed this phenomenon during the adaptation of Flake-Tarjan clustering into the domain of source code, and discussed it in 5.6 proposing an ad-hoc solution that alleviates the issue. However, it is still interesting to analyze and formalize the cases when this phenomenon occurs, and its extent, in terms of the properties of the input graph. In our intuition, the following two theoretical facts should lead to a sound theoretical conclusion:

1 The central theorem of [Fla2004], discussed in section 4.3.1:

$$\frac{c(S,V-S)}{|V-S|} \le \alpha \le \frac{c(P,Q)}{\min(|P|,|Q|)}$$

2 The formalization of alpha-threshold phenomenon we provided in section 5.6:

$$k(\alpha_t - \varepsilon) << K << k(\alpha_t + \varepsilon), \forall \varepsilon \to 0$$

In the further work one should investigate, why there exists alpha $\alpha_t$ such that for any small epsilon, there is a community $S$ such that there is no partition $S = S_1 + ... + S_q$ in which each $S_i$ can satisfy the bicriterion (fact 1) using some alpha from the epsilon-range of $\alpha_t$, i.e. $\alpha \in [\alpha_t - \varepsilon; \alpha_t + \varepsilon]$.

# 9 Conclusion

In the whole, we conclude that the solution we used in this project is practical. The software processed by our prototype is large, real-world and typical (see section 7.1). The clustering hierarchy produced is meaningful for software engineers (section 7.2, also 10.4), correlates with the known explicit architecture (appendix 10.2.1) and reflects the existing implicit architecture (sections 7.2, 10.2.2 and 10.3.3) providing valuable facts for software engineers which can not be observed using state of the art tools except that by scanning millions of lines of source code manually.

We conclude that the method we devised (section 5.1) for alleviating the problem of utility artifacts (section 4.5) and directed-to-undirected conversion of the relation graph in the domain of software source code, works well in practice. The empirical proof is provided in section 10.1.7. Not only utility artifacts did not confuse the clustering results, but also they were clustered together reflecting the unity of purpose. This can be viewed as (perhaps, a prerequisite for) the categorization concluded to be desirable in [HaLe2004].

In section 5.1.5, a disadvantage of lifting SE artifact granularity prior to clustering was investigated, namely, information loss. The possible solutions for lifting the granularity to class level after member-level clustering were given. We concluded not to adopt any of the alternative solutions due to practical reasons (computational complexity) and lack of a reasonably grounded solution.

The alternative solutions discussed in section 5.1.5 and 5.2 (option 1) have the following fact in common: they both rely on the **merge** operation for two nested decompositions inferred using different features, either different members of a SE class or different kinds or relations between SE artifacts. The strong point of both solutions is reduced information loss, if compared to the solution we implemented in this project. Thus we conclude that by researching a suitable merge operation, one has a possibility to make two improvements at once.

One crucial contribution is the scale at which our reverse-architecting approach can operate. While the existing approaches are only able to process small or tiny software (e.g. [Pate2009] takes 147 Java classes on input), whereas we process up to 11 199 classes in our experiments and bump into the limits of the tool that provides input relations for our prototype, namely, call graph extraction ([Soot1999], [Bac1996], [Sun1999], [Lho2003]) exhausts 2GB of memory on a 32-bit machine. This is not a problem for a 64-bit machine, and companies who have huge projects also have appropriate hardware (we do not mean supercomputers by the latter). We speculate that a 64-bit machine with 100GB of RAM and 32 processors should be sufficient for analyzing **any** *real-world software project* with our tool and, less surely, the prerequisite tools in reasonable time.

We stress the importance of operational scale in reverse architecting, as reverse architecting is mostly to address the issue of overwhelming complexity, which arises in large software projects and causes incomprehensibility. Certainly, speed of processing does not bring any value without **quality** of the results. The hierarchical clustering technique [Fla2004] underlying our reverse-architecting approach is well grounded *theoretically*, giving the premises for claims about the resulting quality, even though there is no unbiased indicator of software clustering quality available (except the unlikely case of the exact match) because such an estimation is a subjective task for human software engineers. Even though there are some metrics for software clustering quality proposed in the literature ([UpMJ2007], [END2004], [MoJo1999]), not only their adequacy but also their *scale is in question*: e.g. in [UpMJ2007] their experiments with the devised metric UpMoJo are limited to hierarchical decompositions of no more than 346 artifacts and average height 4. Research and experiments with automatic metrics of clustering quality would require large efforts which we can not allow within this project, thus we leave this as a direction for the future work. Apart from strongly grounded clustering algorithm of [Fla2004], we have studied the literature on reverse engineering topics, incorporated the best practices and ideas from there (see sections 3 and 4), and developed our own theory necessary for adaptation of the clustering algorithm into the domain of software source code and containing our ideas for improvement as well (section 8). In addition to the theoretical premises for high quality of the resulting clustering hierarchies, our experiments confirmed that software artifacts in the results are indeed grouped according to their unity of purpose (as motivated in section 2) and, apparently, the visualized hierarchies (section 10) are comprehensible and meaningful for software engineers.

[Nara2008] makes the following note on how graph topology could influence software processes:

> Understanding call graph structure helps one to construct tools that assist the developers in comprehending software better. For instance, consider a tool that extracts higher-level structures from program call graph by grouping related, lower-level functions. Such a tool, for example, when run on a kernel code base, would automatically decipher different logical subsystems, say, networking, file system, memory management or scheduling. Devising such a tool amounts to finding appropriate similarity metric(s) that partition the graph so that nodes within a partition are “more” similar compared to nodes outside. Understandably, different notions of similarities entail different groupings.

Such a tool has been implemented in our project. In the results section above we show that different logical subsystems are indeed identified. The similarity metric we use is the amount of interaction between software engineering artifacts. However, a desirable ability which we do not yet have in our tool is inference of cluster labels. This direction appears in [Kuhn2007], however, in their turn the authors propose to combine linguistic topics with formal application concepts as a future work.

Finally, we give an account of the weak sides of our project. We are limited in available efforts, and the nature of the project (master thesis) constrains us to certain decisions and strategy, such as avoiding risky research directions (in an attempt to invent, e.g. see section 8) and preference of breadth (multiple approaches; extraction, format conversion, clustering, presentation, import/export, visualization, statistics; clustering and software engineering literature review, comparison to rival approaches, implementation, specification, experiments) rather than depth (single best approach; devising new theory, implementation according to best SE practices).

- Our claims about the resulting clustering quality, also in comparison to the other approaches, are mostly theoretical and empirical.
- Our statistical proof (section 10.2.1) exploits the assumption that more high-level Java classes have shorter package prefixes, in terms of token count (name depth). This is not always true, as there can be higher- and lower- level classes within one package (e.g. see section 7.2.5), as well as there are low-level classes having short prefixes (think of `java.lang.String`).
- Statistical comparison to other reverse architecting approaches is desired, using the same experimental setting (the same software upon analysis). However, this is very effort-consuming without bringing much value into the result of the project in terms of clustering quality and speed.
- More illustrative statistical proof for theory and claims is desired.

- Our theory should be checked for originality. Though we tried our best to review the existing approaches, it is not possible to prove that something does not exist, perhaps in different terms or in a different domain.
- We often use problem-solving approach: given an intention, define the problem, solve it and implement the solution. Thus we do not always know whether someone else has already solved the same problem, and if so, how our solution relates to his/her in terms of precision, speed, advantages and disadvantages. This saves huge amount of efforts, up to 99% in our view, thus letting us to implement more solutions although having less evidence for their originality and optimality or superiority.

## 9.1 Major challenges

Applicability of the clustering algorithm and the success of all this endeavor of integration were not obvious since the beginning. The following subsections list the major challenges were identified prior to the start or during the project.

### 9.1.1 Worst-case complexity

The theoretical estimations on the worst-case complexity seemed prohibiting. Flake-Tarjan clustering uses minimum cut tree algorithm, which uses maximum flow algorithm up to $|V|$ times in the worst case. Hierarchical version of the clustering algorithm adds factor of $|V|$ further. Thus, the total algorithmic complexity is:

$$O(|V|^2 \cdot |E| \cdot \min(|V|^{2/3}, \sqrt{|E|}) \cdot \log \frac{|V|^2}{|E|} \cdot \log U)$$

For the source code of a typical medium-size software project of (for example) 10K classes and 1M relations, the number operations is $11.9 \cdot 10^{18}$. This could take a thousand of years on the usual computer on which the results we showed in the thesis were indeed produced in 72 hours within our project. This was achieved due to careful choice of the implementations and the underlying heuristics.

The software projects we used in our experiments are **typical**, as most of their classes are Java library classes. Thus we conclude that the clustering method is applicable in general.

### 9.1.2 Data extraction

Analyzing real-world software projects is a challenge even for tools that extract the data we use on input of our tool. In order to make Soot to extract the call graph we had to study its design and implementation, tune the parameters carefully and even change the source code of Soot in order to allow call graph extraction (and effectively, whole-program analysis) without providing and analyzing all the libraries on which the software upon analysis depends.

### 9.1.3 Noise in the input

Call graph extraction is far from precise. By manually analyzing the extracted call graph and the original source code, we observe that many call relations are absent from the call graph though the source code clearly states the presence, and vice versa, there are many calls in the call graph which never occur during program execution and are not designed to occur by the developers.

Another cause of noise in the input for architecture reconstruction algorithm is the mistakes made by software engineers due to lack of global understanding of the system. These mistakes violate the implicit architecture, or even sometimes the explicit one. The fact of violation of the latter can be proved when the documentation or any other form of explicit archiecture, e.g. packaging structure, is available.

Thus formal relation graph's "comprehension" of the source code differs a lot from the comprehension of developers who wrote the source code. As can be seen from the resulting clustering, this noise has been successfully tolerated.

### 9.1.4 Domain specifics

The data configurations specific to the domain are omnipresent utility artifacts and almost perfectly hierarchical structure of software. We have discussed these issues within the thesis, and devised and implemented solutions, which also constitute the major theoretical contributions of our work.

### 9.1.5 Evaluation of the results

It was challenging to evaluate the reversed architecture due to the common problems in Artificial Intelligence and other relevant fields:

- Human to machine intelligence gap: while a machine can only calculate some measure over the results, humans can find them meaningful, useful, easy to comprehend, etc.
- Lack of objective criteria: even when software engineers discuss some architecture (either the currently documented, or prospective architectural decisions), arguments often bump into the philosophy of software engineering. Different experts adopt different approaches, or they just like some decisions more than other.
- Lack of labelled data: real-world (at least, non-trivial) software projects never have documentation of hierarchical architecture till SE class-level, i.e. "target" nested software decomposition which we could train on or compare with. Furthermore, architectural documentation is usually not a hierarchy.
- Lack of adequate measures: a counterexample to [END2004] is provided in [UpMJ2007], while the latter is not scalable to large nested software decompositions.

What we did manually is a brute-force evaluation of the reversed architecture by looking at subtrees of the cluster tree and arguing for the useful and adequate (reflecting the actual architecture) facts that a software engineer can see in those subtrees (section 7). There is an advantage in this kind of evidence too: we provide realistic evaluations as usually concluded by humans, rather than abstract measures that might not reflect what humans want to see.

## *9.2 Gain over the state of the art*

### 9.2.1 Practical contributions

The output of our prototype needs human analysis in the end. However, we stress the *gain in comprehensibility*: instead of scanning and manually interpreting millions of lines of source code, human software engineers need to look at a few thousands of nodes in the clustering hierarchy to get architectural insights, e.g. those described in 7.2. The latter section contains the typical actions the humans should take for this, though mainly it is a matter of experience and natural intelligence.

To summarize, leaf node labels (i.e. the names of SE classes) are heavily used for both validation of the architecture (e.g. class name must not confuse a software engineer about its purpose) and validation of the quality of clustering. The latter is possible because our approach *does **not** use textual information* at any stage of inference, either identifier (type, variable) names, keywords, comments or whatever. InSoAr's inference is purely based on *formal* relations between software engineering artifacts. The fact that SE classes having similar labels (the same textual features, e.g. the words composing package names or class names) appear nearby in the clustering hierarchy (under the same parent, in the same subtree) says both about the quality of the architecture (decoupled, class purposes are well-defined) and the quality of our result.

In Figure 9-1 below we give a trace of software comprehensibility gain in numbers from an experiment with FreeCol open-source project. See section 5.2 for the details on what relations were extracted and how they were merged into a single input graph for clustering. So we conclude that there is nearly-1000 times gain in comprehensibility of software: from 7.5M lines of source code to 11K nodes in the cluster tree (or, debatabely, 3.9K subsystems – inner nodes). Although there are 7.5K classes in the software, they are not comprehensible if presented as a plain list (section 7). The same for non-perfectized cluster tree: though there are 9.6K nodes, the hierarchy is less comprehensible (section 5.6.1) than 11.4K nodes of a perfectized counterpart due to the issue of an excessive number of children (section 5.6).

| | | |
|---|---|---|
| 1 | Estimated number of lines of source code | **7 500 000** |
| 2 | Number of formal relations extracted | 822 353 |
| 2.1 | - Number of edges in the call graph | *527 555* |
| 2.2 | - Number of field accesses | 143 033 |
| 2.3 | - Thus, number of other relations | 151 765 |
| 3 | Number of SE artifacts | |
| 3.1 | - Classes | 7 474 |
| 3.2 | - Fields | 29 834 |
| 3.3 | - Methods | 70 257 |
| 4 | Number of items in the cluster tree | |
| 4.1 | - Before perfectization | 9 682 |
| 4.1.1 | - - Of them, inner nodes (subsystems) | 2 208 |
| 4.2 | - After perfectization (section 5.6) | **11 351** |
| 4.2.1 | - - Of them, inner nodes (subsystems) | ***3 877*** |
| 4.2.2 | - - Labelled leaf nodes (SE classes) | 7 474 |

**Figure 9-1 Software comprehensibility gain, in numbers**

From the above table we also see that the gain in comprehensibility *over* ***non****-clustered* graph of extracted relations, calculated as the ratio of the item numbers, is nearly-100 times: 822K edges in the input graph vs. 11.4K nodes in the cluster tree. Note that the graph of relations is **not** *just the call graph*:

- 2/3 are indeed, method call relations (call graph)
- 1/3 are other relations (field access, inheritance, type usage, parameter and return types)

The examples in section 7 illustrate that the determined clusters make it possible for a software engineer to infer the purpose of SE classes from the names of these classes and the neighbouring classes in the cluster tree. This is useful even in case the purpose of a class is obvious from its name, as its position in the cluster tree validates its proper naming, assuming that the quality of the clustering is high, which was also concluded.

The central inference our tool does is identification of hierarchical groups of classes that act together (section 7.2.3). In composition with identified purposes (the paragraph above), this can be used for obtaining overviews of systems that lack documentation, or documenting subsystems (including the private case of a single-class subsystem). Along the way it simplifies detection of anomalies in the software system by a software engineer, e.g. overcomplicated coupled groups as in appendix 10.1.5.

Differentiation of coupling within a package (section 7.2.5) presents a software engineer with a structure while the explicit architecture shows a plain list, which can be much harder to comprehend in case there are many SE classes in the list. On the other hand, when classes belonging to different subsystems appear in the same package due to some more high-level property (section 7.2.6), a software engineer can observe the actual implicit subsystem for each class in the cluster tree. These facts particularly help a software engineer in refactoring and identification of subsystems affected by a change: as we see in section 7.2.6, the subsystem affected by a change in a class from that package is not the package, but the implicit subsystem with which it is coupled. And that is the one inferred by our tool.

Apparently, security mechanism is a private case of a subsystem, section 7.2.7. Thus with our tool software engineers can inspect structural vulnerabilities of a software system. As shown in section 7.2.8 and appendix 10.1.4, sometimes also insight on the implementation can be captured.

The ultimate goal is in allowing Divide&Conquer approach to software comprehension, however it is hard to support such claim, as visualizational problems are encountered in the upper levels (near the root) of the cluster tree, namely: there become too many leaves (labelled nodes) in a subtree, thus some inference of cluster labels is needed. We provide the evidence we currently have in appendix 10.1.5. Still, software engineers can start labelling subsystems from the bottom level up. From Figure 9-1 above we see that for 7.5 million lines of source code, there are only 3.9K inner nodes (i.e. subsystems) in the cluster tree. Labelling these nodes manually for the sake of Divide&Conquer opportunity can be a reasonable task, given that our tool provides a mean to identify the purpose of subsystems near leaf nodes cheaply.

To recapitulate, in section 7.2 we discussed the practical facts that can be inferred automatically using the approach we devised. These facts can not be inferred with any other state of the art software engineering tool. This list is not exhaustive, as there are only facts we could think of and discuss illustratively. Altogether, we characterize these facts as architectural insights with practical applications in reverse engineering, software quality and security analysis.

### 9.2.2 Scientific contributions

An efficient algorithm for high-quality clustering of large software has been invented.

The algorithm is based on Flake-Tarjan clustering (sections 4.3) thus inherits its intrinsic hierarchical property, optimization of graph cut criteria, and a premise for high-quality as reported in [Fla2004] in general. Our contributions on top of Flake-Tarjan clustering algorithm are provided in section 5. Of them, the following are specific to the domain of source code:

- Edge weight normalization (section 5.1): incorporates the recent conclusions in the literature on Reverse Engineering (section 4.5) about the main domain-specific problem – utility artifacts – and proposes directed-to-undirected graph conversion (as Flake-Tarjan algorithm requires undirected graph on input) based on utilitihood rationale from the literature (fan-in analysis).
- Perfectization (section 5.6): makes adjustments to the hierarchical results of Flake-Tarjan clustering, so that a specific property of data (namely, nearly-perfect hierarchical structure of software as a good practice) does not confuse the clustering result.

The following our contributions are improvements over hierarchical Flake-Tarjan clustering algorithm in general:

- Distributed version (section 5.5, using the contributions of sections 5.3 and 5.4): motivated by the need for hierarchical clustering of **large** software, a distributed version allows running multiple basic (section 4.3.1) Flake-Tarjan clustering probes in parallel, one processor per one value of parameter alpha. The results are then merged into a single hierarchy, as described in section 5.4.
- Prioritized alpha-search (section 5.3): in the absence of time to compute the result of basic Flake-Tarjan clustering for each necessary (i.e. potentially producing different number of clusters) value of parameter alpha, it allows taking the most important probes first, so that the more important decisions about the clustering hierarchy are taken earlier.

A purely theoretical, within our work, contribution is given in section 5.2: it discusses the potential solutions for considering multiple kinds of formal relations between SE artifacts during clustering, however we did not have time to implement semi-supervised learning proposed there. It currently serves as evidence for our hypothesis 3. In our current clustering algorithm we use the same merge-weight (equal to 1) for each kind of SE relations.

Minor contributions include:

- Reduction of real- to integer- valued flow graph (section 6.1.1): this allows substitution of integer-capacities max flow algorithms into Flake-Tarjan clustering, instead of more computationally expensive real-capacities max flow algorithms.
- A review of state of the art clustering and source code analysis methods and tools under section 3, and pre-requisites from clustering, source code analysis and reverse engineering in section 4.

# 10 Appendices

## 10.1 Evidence

This section contains evidence for the claims about properties and quality of the resulting clustering hierarchy. This does not include evidence involving source code demonstration: such evidence is listed in section 10.3 below.

### 10.1.1 A package of omnipresent client artifacts

Here we continue the evidence for the claim discussed in 7.2.6.

```
10555 2 0 -0.15834708966782382000 ; 1 heads 7177
Level 210 cluster under 10560, 10563, 10604, 10614, 10
  7177 2 0 3.75007656250000000000 ; 1 heads 228
  Level 211 cluster under 10555, 10560, 10563, 10604,
    228 0 0 VeryBig ; 0 heads
    com.kpmg.kpo.domain.ServiceMessageRule
    754 0 0 VeryBig ; 0 heads
    com.kpmg.kpo.web.binding.ServiceMessageRuleCommand
  7905 2 0 0.59189317626953130000 ; 1 heads 216
  Level 211 cluster under 10555, 10560, 10563, 10604,
    216 0 0 VeryBig ; 0 heads
    com.kpmg.kpo.domain.MessageType
    752 0 0 VeryBig ; 0 heads
    com.kpmg.kpo.web.binding.MessageTypeCommand
```

```
...chiFit\perfTree.txt                Win    Line   3439/21776   Col 443
      Level 221 cluster under 9574, 9578, 9579, 9611, 9612, 9793, 980
        8109 2 0 1.00000001890586930000 ; 1 heads 5112
        Level 222 cluster under 8398, 9574, 9578, 9579, 9611, 9612, 9
          5111 0 0 VeryBig ; 0 heads
          org.springframework.security.context.SecurityContext
          5112 0 0 VeryBig ; 0 heads
          org.springframework.security.context.SecurityContextHolder
        769 0 0 VeryBig ; 0 heads
        com.kpmg.kpo.web.security.EmployeeProviderSpringSecurity
      103 0 0 VeryBig ; 0 heads
      com.kpmg.kpo.auth.LoggedInEmployeeProvider
    8514 2 0 0.48349541625976560000 ; 1 heads 6988
    Level 220 cluster under 9578, 9579, 9611, 9612, 9793, 9807, 9808,
      6988 2 0 13.50001562500000000000 ; 1 heads 734
      Level 221 cluster under 8514, 9578, 9579, 9611, 9612, 9793, 980
        734 0 0 VeryBig ; 0 heads
        com.kpmg.kpo.web.ajax.TaskInstancesPopupController
        737 0 0 VeryBig ; 0 heads
        com.kpmg.kpo.web.ajax.beans.TaskInstanceBean
      579 0 0 VeryBig ; 0 heads
      com.kpmg.kpo.service.DomainTaskInstanceService
    9573 2 0 -0.15834708966782382000 ; 1 heads 242
    Level 220 cluster under 9578, 9579, 9611, 9612, 9793, 9807, 9808,
      111 0 0 VeryBig ; 0 heads
      com.kpmg.kpo.comparator.TaskStatusHistoryComparator
      242 0 0 VeryBig ; 0 heads
      com.kpmg.kpo.domain.TaskStatusHistory
  8568 3 0 0.49130786743164060000 ; 2 heads 8012 2590
  Level 219 cluster under 9579, 9611, 9612, 9793, 9807, 9808, 9813, 9
    8012 2 0 0.69150192871093760000 ; 1 heads 2589
    Level 220 cluster under 8568, 9579, 9611, 9612, 9793, 9807, 9808,
      2589 0 0 VeryBig ; 0 heads
      java.text.DateFormat
      3955 0 0 VeryBig ; 0 heads
      javax.swing.JTable$DateRenderer
    2590 0 0 VeryBig ; 0 heads
    java.text.DateFormat$DateFormatGetter
    2634 0 0 VeryBig ; 0 heads
    java.text.spi.DateFormatProvider
9610 3 0 -0.15834708966782382000 ; 1 heads 7560
Level 218 cluster under 9611, 9612, 9793, 9807, 9808, 9813, 9898, 989
```

```
10592 2 0 -0.15834708966782382000 ; 1 heads 8773
Level 210 cluster under 10597, 10602, 10604, 10614, 10615, 10669, 1067
  10591 2 0 -0.15834708966782382000 ; 1 heads 7918
  Level 211 cluster under 10592, 10597, 10602, 10604, 10614, 10615, 10
    7918 2 0 0.71298616943359370000 ; 1 heads 697
    Level 212 cluster under 10591, 10592, 10597, 10602, 10604, 10614,
      8801 2 0 1.00000001890586930000 ; 1 heads 113
      Level 213 cluster under 7918, 10591, 10592, 10597, 10602, 10604,
        113 0 0 VeryBig ; 0 heads
        com.kpmg.kpo.dao.ArchiveDao
        157 0 0 VeryBig ; 0 heads
        com.kpmg.kpo.dao.hibernate.HibernateArchiveDao
      697 0 0 VeryBig ; 0 heads
      com.kpmg.kpo.service.impl.SimpleArchiveService
    193 0 0 VeryBig ; 0 heads
    com.kpmg.kpo.domain.ArchiveEntry
  8773 2 0 0.18808710632324220000 ; 1 heads 7181
  Level 211 cluster under 10592, 10597, 10602, 10604, 10614, 10615, 10
    7181 2 0 2.08602446289062460000 ; 1 heads 603
    Level 212 cluster under 8773, 10592, 10597, 10602, 10604, 10614, 1
      603 0 0 VeryBig ; 0 heads
      com.kpmg.kpo.service.auditdecorator.ArchiveServiceAuditingImpl
      604 0 0 VeryBig ; 0 heads
      com.kpmg.kpo.service.auditdecorator.ArchiveServiceAuditingImpl$1
    571 0 0 VeryBig ; 0 heads
    com.kpmg.kpo.service.ArchiveService
10590 3 0 -0.15834708966782382000 ; 1 heads 260
Level 210 cluster under 10597, 10602, 10604, 10614, 10615, 10669, 1067
```

Figure 10-1 HibernateArchiveDao clustered with ArchiveEntry

```
10058 2 0 -0.15834708966782382000 ; 1 heads 180
Level 206 cluster under 10062, 10063, 10079, 10120, 101
  8701 2 0 0.21884863281250000000 ; 1 heads 172
  Level 207 cluster under 10058, 10062, 10063, 10079, 1
    7900 2 0 0.76962644042968750000 ; 1 heads 134
    Level 208 cluster under 8701, 10058, 10062, 10063,
      8888 2 0 0.99569924621582030000 ; 1 heads 700
      Level 209 cluster under 7900, 8701, 10058, 10062,
        595 0 0 VeryBig ; 0 heads
        com.kpmg.kpo.service.SlotTypeService
        700 0 0 VeryBig ; 0 heads
        com.kpmg.kpo.service.impl.SlotTypeServiceImpl
      134 0 0 VeryBig ; 0 heads
      com.kpmg.kpo.dao.SlotTypeDao
    172 0 0 VeryBig ; 0 heads
    com.kpmg.kpo.dao.hibernate.HibernateSlotTypeDao
  180 0 0 VeryBig ; 0 heads
  com.kpmg.kpo.dao.hibernate.SlotTypeComparator
106 0 0 VeryBig ; 0 heads
com.kpmg.kpo.comparator.CodeBasedTemplateVersionCompara
107 0 0 VeryBig ; 0 heads
com.kpmg.kpo.comparator.CustomerComparator
110 0 0 VeryBig ; 0 heads
com.kpmg.kpo.comparator.SEDTemplateVersionComparator
253 0 0 VeryBig ; 0 heads
com.kpmg.kpo.domain.WorkflowTemplate$1
275 0 0 VeryBig ; 0 heads
com.kpmg.kpo.exception.ActionNotAllowedException
2658 0 0 VeryBig ; 0 heads
java.util.Arrays
3442 0 0 VeryBig ; 0 heads
javax.management.openmbean.CompositeDataSupport
6712 0 0 VeryBig ; 0 heads
sun.security.util.Password
043 2 0 -0.15834708966782382000 ; 1 heads 1417
vel 205 cluster under 10063, 10079, 10120, 10121, 10138
```

### 10.1.2 Security mechanism: identify and circumscribe

See section 7.2.7 for the discussion.

```
edit ctBracketed.txt - Far
D:\...ws2\Insoar\data\ctAnal\ctBracketed.txt        Win     Line     2354/9413    Col 57
 197 2412 "java.security.KeyStoreSpi"
h3<9082> p9142 <8717>h3 <8441>h2 <7998>h1
 199 2405 "java.security.KeyStore$PasswordProtection"
 199 3606 "javax.security.auth.Destroyable"
h1<7998> p8441 <7554>h1
 199 2402 "java.security.KeyStore$CallbackHandlerProtection"
 199 3634 "javax.security.auth.callback.CallbackHandler"
h1<7554>
 198 2407 "java.security.KeyStore$ProtectionParameter"
h2<8441> p8717 <7246>h1
 198 2400 "java.security.KeyStore$Builder$FileBuilder"
 198 2401 "java.security.KeyStore$Builder$FileBuilder$1"
h1<7246> h3<8717> p9142 <8604>h3 <8260>h2 <7493>h1
 199  770 "com.kpmg.kpo.web.security.EmployeeUserDetails"
 199  771 "com.kpmg.kpo.web.security.EmployeeUserDetails$StringAuthority"
h1<7493>
 198 5110 "org.springframework.security.GrantedAuthority"
h2<8260> p8604 <8261>h2 <7492>h1
 199  772 "com.kpmg.kpo.web.security.EmployeeUserDetailsService"
 199 5119 "org.springframework.security.userdetails.UsernameNotFoundException"
h1<7492>
 198 5118 "org.springframework.security.userdetails.UserDetailsService"
h2<8261>
 197 5117 "org.springframework.security.userdetails.UserDetails"
h3<8604> p9142 <8622>h3 <8196>h2 <7085>h1
 199 3344 "javax.imageio.stream.ImageInputStreamImpl"
 199 3346 "javax.imageio.stream.ImageOutputStreamImpl"
h1<7085>
 198 3347 "javax.imageio.stream.MemoryCache"
 198 3348 "javax.imageio.stream.MemoryCacheImageInputStream"
 198 3349 "javax.imageio.stream.MemoryCacheImageInputStream$StreamDisposerRecord"
 198 3350 "javax.imageio.stream.MemoryCacheImageOutputStream"
h2<8196>
 197 3341 "javax.imageio.stream.FileImageOutputStream"
h3<8622> p9142 <9068>h3 <8487>h2 <8979>h1
 199   57 "com.kpmg.esb.mule.transformers.WebservableObjectInToAuditableObjectTransformer"
 199 5018 "org.apache.log4j.Logger"
h1<8979>
 198   51 "com.kpmg.esb.mule.transformers.PersistentObjectToPersistentMessageTransformer"
h2<8487> p9068 <7886>h1
 198   17 "com.kpmg.esb.mule.component.MessageLogger"
```

### 10.1.3 Subsystems

In Figure 10-2 below we continue illustrating the time & scheduling subsystem, as appeared in the clustering hierarchy and discussed in section 7.2.8. This part is closer to the bottom of the cluster tree, as can be seen from the non-branching nodes on the right.

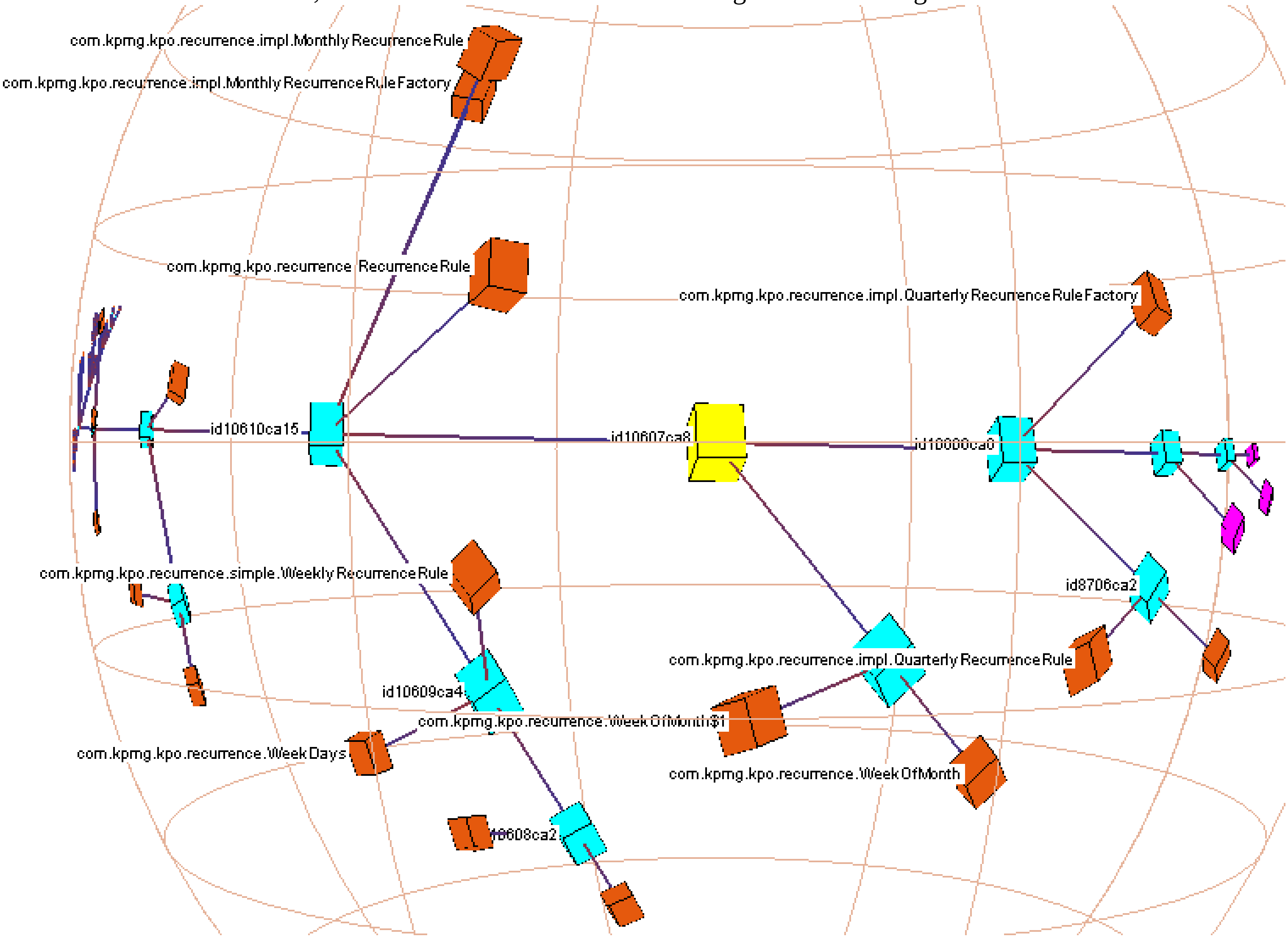

**Figure 10-2 Time&Scheduling subsystem (continued)**

### 10.1.4 Insight on the implementation

Below we show obvious examples of insights that the reversed architecture gives to software engineers.

```
8237 2 0 0.87118830566406260000 ; 2 heads 7642 8871
Level 196 cluster under 9142, 8957, 9391, 9367, 9314, 9259,
  8871 2 0 0.87997731323242190000 ; 2 heads 7642 9003
  Level 197 cluster under 8237, 9142, 8957, 9391, 9367, 9314
    9003 2 0 1.00000001890586930000 ; 1 heads 7642
    Level 198 cluster under 8871, 8237, 9142, 8957, 9391, 93
      7642 2 0 1.89852563476562470000 ; 1 heads 5030
      Level 199 cluster under 9003, 8871, 8237, 9142, 8957,
        5026 0 0 VeryBig ; 0 heads
        org.hibernate.criterion.Criterion
        5030 0 0 VeryBig ; 0 heads
        org.hibernate.criterion.Restrictions
      5028 0 0 VeryBig ; 0 heads
      org.hibernate.criterion.LogicalExpression
    177 0 0 VeryBig ; 0 heads
    com.kpmg.kpo.dao.hibernate.HibernateWorkflowTemplateDao
  5031 0 0 VeryBig ; 0 heads
  org.hibernate.criterion.SimpleExpression
```

```
7917 2 0 0.61728364257812510000 ; 2 heads 686 8912
Level 196 cluster under 9142, 8957, 9391, 9367, 9314, 9259,
  8912 2 0 0.71298616943359370000 ; 1 heads 686
  Level 197 cluster under 7917, 9142, 8957, 9391, 9367, 9314
    8803 2 0 1.00000001890586930000 ; 1 heads 131
    Level 198 cluster under 8912, 7917, 9142, 8957, 9391, 93
      131 0 0 VeryBig ; 0 heads
      com.kpmg.kpo.dao.PeriodTypeDao
      170 0 0 VeryBig ; 0 heads
      com.kpmg.kpo.dao.hibernate.HibernatePeriodTypeDao
    686 0 0 VeryBig ; 0 heads
    com.kpmg.kpo.service.impl.PeriodServiceImpl
  8802 2 0 0.75009531250000000000 ; 1 heads 129
  Level 197 cluster under 7917, 9142, 8957, 9391, 9367, 9314
    129 0 0 VeryBig ; 0 heads
    com.kpmg.kpo.dao.PeriodDao
    168 0 0 VeryBig ; 0 heads
    com.kpmg.kpo.dao.hibernate.HibernatePeriodDao
```

For humans it is now easy to make a note that, e.g. client code class `HibernateWorkflowTemplateDao` works with `Criterion`, `Restrictions` and `LogicalExpression` of Hibernate library; in its turn, it is likely to be used by `SimpleExpression`

```
\data\ctAnal\ctHier.txt                    Win     Line    3379/18796    Col 397
    2621 0 0 VeryBig ; 0 heads
    java.text.RBCollationTables
    2629 0 0 VeryBig ; 0 heads
    java.text.RuleBasedCollator
    6855 0 0 VeryBig ; 0 heads
    sun.text.IntHashtable
  8778 4 0 0.20493270416259768000 ; 1 heads 7922
  Level 196 cluster under 9142, 8957, 9391, 9367, 9314, 9259, 9289, 9
    7922 4 0 1.00000001890586930000 ; 1 heads 766
    Level 197 cluster under 8778, 9142, 8957, 9391, 9367, 9314, 9259,
      766 0 0 VeryBig ; 0 heads
      com.kpmg.kpo.web.filter.AddHeaderFilter
      3638 0 0 VeryBig ; 0 heads
      javax.servlet.Filter
      3639 0 0 VeryBig ; 0 heads
      javax.servlet.FilterChain
      3640 0 0 VeryBig ; 0 heads
      javax.servlet.FilterConfig
    7931 3 0 1.00000001890586930000 ; 1 heads 768
    Level 197 cluster under 8778, 9142, 8957, 9391, 9367, 9314, 9259,
      768 0 0 VeryBig ; 0 heads
      com.kpmg.kpo.web.security.CustomAuthenticationEntryPoint
      5108 0 0 VeryBig ; 0 heads
      org.springframework.security.AuthenticationException
      5115 0 0 VeryBig ; 0 heads
      org.springframework.security.ui.AuthenticationEntryPoint
    3643 0 0 VeryBig ; 0 heads
    javax.servlet.ServletRequest
    3644 0 0 VeryBig ; 0 heads
    javax.servlet.ServletResponse
  7365 2 0 2.41414741210937500000 ; 1 heads 7040
  Level 196 cluster under 9142, 8957, 9391, 9367, 9314, 9259, 9289, 9
    7040 3 0 14.25001093750000000000 ; 1 heads 6969
    Level 197 cluster under 7365, 9142, 8957, 9391, 9367, 9314, 9259,
      6969 6 0 16.00000000000000000000 ; 1 heads 1859
      Level 198 cluster under 7040, 7365, 9142, 8957, 9391, 9367, 931
        1859 0 0 VeryBig ; 0 heads
        java.lang.Character
        1863 0 0 VeryBig ; 0 heads
        java.lang.CharacterData00
        1864 0 0 VeryBig ; 0 heads
```

```
D:\...\Dev\ws2\Insoar\data\ctAnal\ctHier.txt                Win     Line    3743/18796    Col 492
  1924 0 0 VeryBig ; 0 heads
  java.lang.NoSuchFieldException
7855 2 0 1.17978012695312500000 ; 2 heads 7173 8666
Level 196 cluster under 9142, 8957, 9391, 9367, 9314, 9259, 9289, 9236, 9387, 9333, 9357,
  8666 2 0 1.52938731689453130000 ; 2 heads 7173 9194
  Level 197 cluster under 7855, 9142, 8957, 9391, 9367, 9314, 9259, 9289, 9236, 9387, 9333
    9194 3 0 1.53134042968750000000 ; 1 heads 7173
    Level 198 cluster under 8666, 7855, 9142, 8957, 9391, 9367, 9314, 9259, 9289, 9236, 93
      7173 2 0 2.84383222656250000000 ; 2 heads 9165 61
      Level 199 cluster under 9194, 8666, 7855, 9142, 8957, 9391, 9367, 9314, 9259, 9289,
        7043 2 0 7.81255117187500000000 ; 1 heads 62
        Level 200 cluster under 7173, 9194, 8666, 7855, 9142, 8957, 9391, 9367, 9314, 9259
          62 0 0 VeryBig ; 0 heads
          com.kpmg.esb.mule.transformers.webservable.DocumentMessageToSendDocumentsRequest
          333 0 0 VeryBig ; 0 heads
          com.kpmg.kpo.generated.jaxws.approval.FileFormatType
        9165 2 0 2.90633183593750000000 ; 1 heads 61
        Level 200 cluster under 7173, 9194, 8666, 7855, 9142, 8957, 9391, 9367, 9314, 9259
          61 0 0 VeryBig ; 0 heads
          com.kpmg.esb.mule.transformers.webservable.DocumentMessageToSendDocumentsRequest
          326 0 0 VeryBig ; 0 heads
          com.kpmg.kpo.generated.jaxws.approval.DocumentType
      317 0 0 VeryBig ; 0 heads
      com.kpmg.kpo.generated.jaxws.approval.AnnualReportStatusType
      342 0 0 VeryBig ; 0 heads
      com.kpmg.kpo.generated.jaxws.approval.NonPublishingPartFileFormatType
    344 0 0 VeryBig ; 0 heads
    com.kpmg.kpo.generated.jaxws.approval.PlaceAnnualReportType
  346 0 0 VeryBig ; 0 heads
  com.kpmg.kpo.generated.jaxws.approval.SendDocumentsRequest
8285 2 0 1.00000001890586930000 ; 1 heads 7407
Level 196 cluster under 9142, 8957, 9391, 9367, 9314, 9259, 9289, 9236, 9387, 9333, 9357,
```

The part of cluster tree demonstrated below tells software engineers multiple architectural facts.

- Coupling structure
  - JMSUtil
  - AbstractDocumentHandler
- Purpose from library neighbors
- How documents are handled
  - via JMS
  - Messages, Sessions, Connections

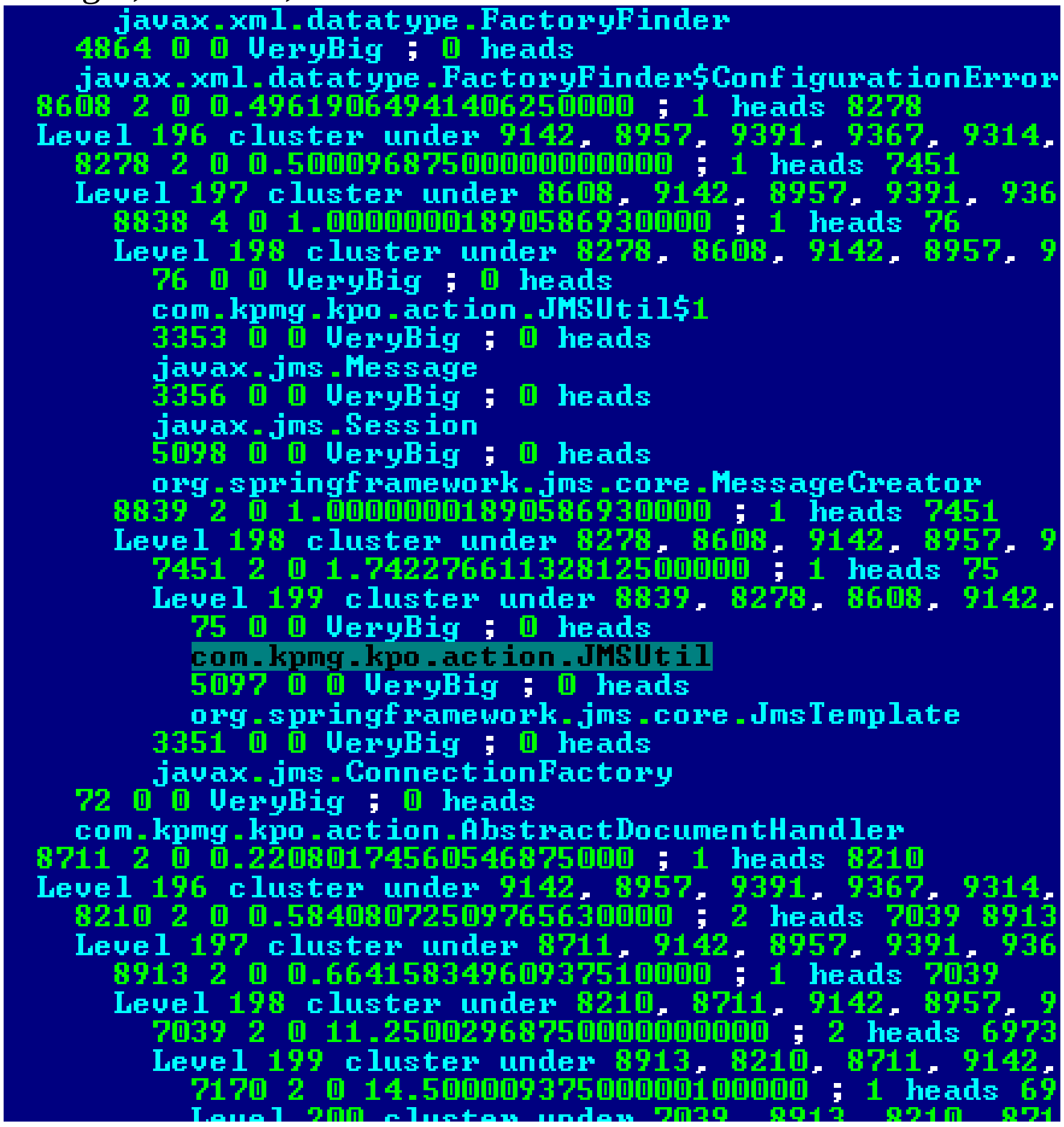

### 10.1.5 An overcomplicated subsystem

In the picture below we see a number of classes with, concluding from the names, different purposes acting as a single mechanism. The classes are also from different packages, thus we cannot detect this coupling efficiently using state of the art tools. However, when changing the software, it is important to identify the extent of subsystem to be changed. By circumscribing a subsystem subject for change, we narrow down the search space of side effects.

This is a fragment of cluster tree for project FreeCol, which is an open-source game. The nature of task the classes are performing, AI player, validates the intuition about complexity of the subsystem and the clustering result.

```
...010-05-11 with libs\trees\tree0203.txt           Win     Line    1096/18306    Col 5
_       9103 2 1 1.211621093749999800000 ; 2 heads 8992 9129
        Level 6 cluster under 9054, 8022, 8427, 9059, 8695, 7474
         9129 4 1 1.250634765624999900000 ; 1 heads 8992
         Level 7 cluster under 9103, 9054, 8022, 8427, 9059, 8695, 7474
          8992 2 1 1.484716796874999700000 ; 1 heads 8542
          Level 8 cluster under 9129, 9103, 9054, 8022, 8427, 9059, 8695, 7474
           8542 4 1 1.554941406249999800000 ; 2 heads 7743 9035
           Level 9 cluster under 8992, 9129, 9103, 9054, 8022, 8427, 9059, 8695,
            9035 2 1 1.867050781249999700000 ; 2 heads 7743 9077
            Level 10 cluster under 8542, 8992, 9129, 9103, 9054, 8022, 8427, 905
             9077 2 1 1.960683593749999800000 ; 1 heads 7743
             Level 11 cluster under 9035, 8542, 8992, 9129, 9103, 9054, 8022, 8
              7743 2 1 3.973789062500000000000 ; 1 heads 7544
              Level 12 cluster under 9077, 9035, 8542, 8992, 9129, 9103, 9054,
               7544 3 1 4.254687499999999900000 ; 1 heads 5216
               Level 13 cluster under 7743, 9077, 9035, 8542, 8992, 9129, 910
                5216 0 1 VeryBig ; 0 heads
                net.sf.freecol.common.model.Map
                5219 0 1 VeryBig ; 0 heads
                net.sf.freecol.common.model.Map$3
                5220 0 1 VeryBig ; 0 heads
                net.sf.freecol.common.model.Map$4
               5221 0 1 VeryBig ; 0 heads
               net.sf.freecol.common.model.Map$5
              5443 0 1 VeryBig ; 0 heads
              net.sf.freecol.server.ai.mission.Mission$2
             5301 0 1 VeryBig ; 0 heads
             net.sf.freecol.common.model.pathfinding.CostDeciders$1
            5396 0 1 VeryBig ; 0 heads
            net.sf.freecol.server.ai.ColonialAIPlayer$1
            5407 0 1 VeryBig ; 0 heads
            net.sf.freecol.server.ai.EuropeanAIPlayer$1
            5419 0 1 VeryBig ; 0 heads
            net.sf.freecol.server.ai.StandardAIPlayer$2
           5436 0 1 VeryBig ; 0 heads
           net.sf.freecol.server.ai.mission.CashInTreasureTrainMission$1
          5297 0 1 VeryBig ; 0 heads
          net.sf.freecol.common.model.pathfinding.AvoidBlockingUnitsCostDecider
          5298 0 1 VeryBig ; 0 heads
          net.sf.freecol.common.model.pathfinding.BaseCostDecider
          5300 0 1 VeryBig ; 0 heads
          net.sf.freecol.common.model.pathfinding.CostDeciders
         5420 0 1 VeryBig ; 0 heads
         net.sf.freecol.server.ai.StandardAIPlayer$3
        5454 0 1 VeryBig ; 0 heads
        net.sf.freecol.server.ai.mission.UnitSeekAndDestroyMission$1
       5397 0 1 VeryBig ; 0 heads
       net.sf.freecol.server.ai.ColonialAIPlayer$2
       5408 0 1 VeryBig ; 0 heads
       net.sf.freecol.server.ai.EuropeanAIPlayer$2
      5452 0 1 VeryBig ; 0 heads
      net.sf.freecol.server.ai.mission.TransportMission$1
     5057 0 1 VeryBig ; 0 heads
     net.sf.freecol.client.gui.panel.SelectDestinationDialog$1
    5449 0 1 VeryBig ; 0 heads
    net.sf.freecol.server.ai.mission.ScoutingMission$1
    5450 0 1 VeryBig ; 0 heads
    net.sf.freecol.server.ai.mission.ScoutingMission$2
```

### 10.1.6 Divide & Conquer

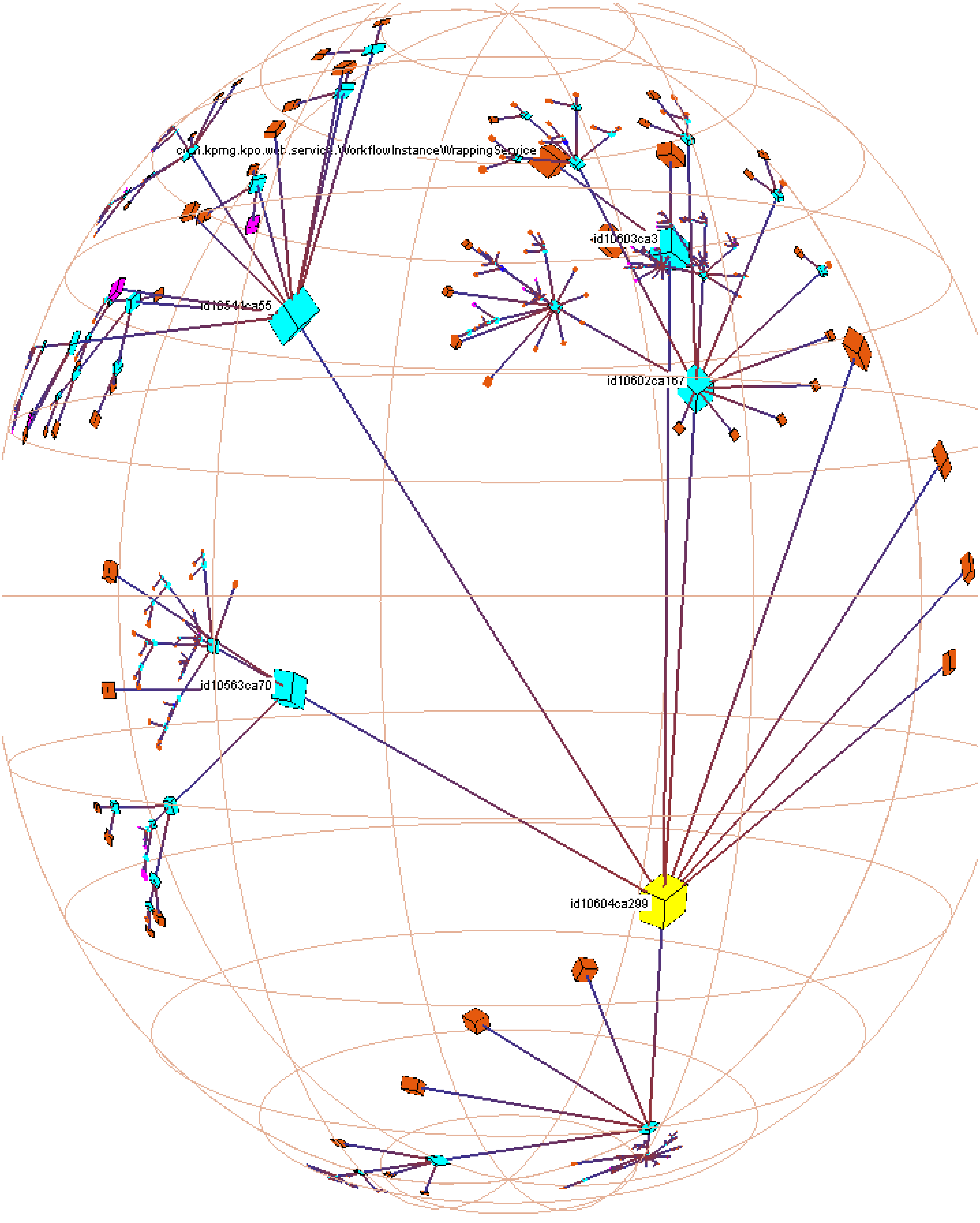

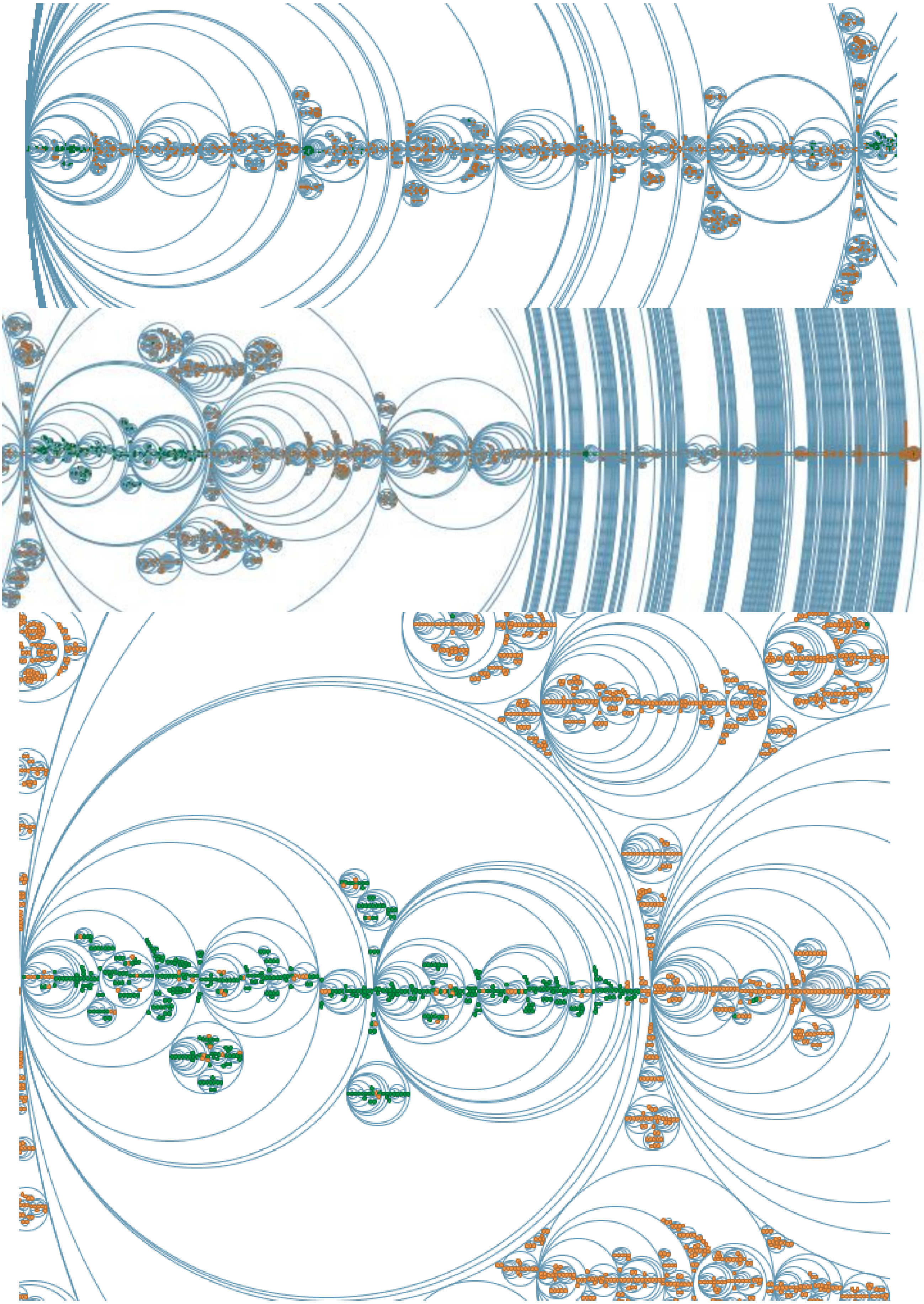

### 10.1.7 Utility Artifacts

In the figure below we can see that utility artifacts have been indeed identified and clustered together, even fairly exhibiting their unity of purpose. `StringBuilder` and `StringBuffer` are unarguably utility classes in Java. Descendants of cluster `id9792` are these 2 classes together with others serving a similar purpose, except the subtree of `id9791`. The latter subtree contains the rest of the program, and the artifacts there can usually be viewed as more high-level or specific (in the meaning opposed to general-purpose artifacts).

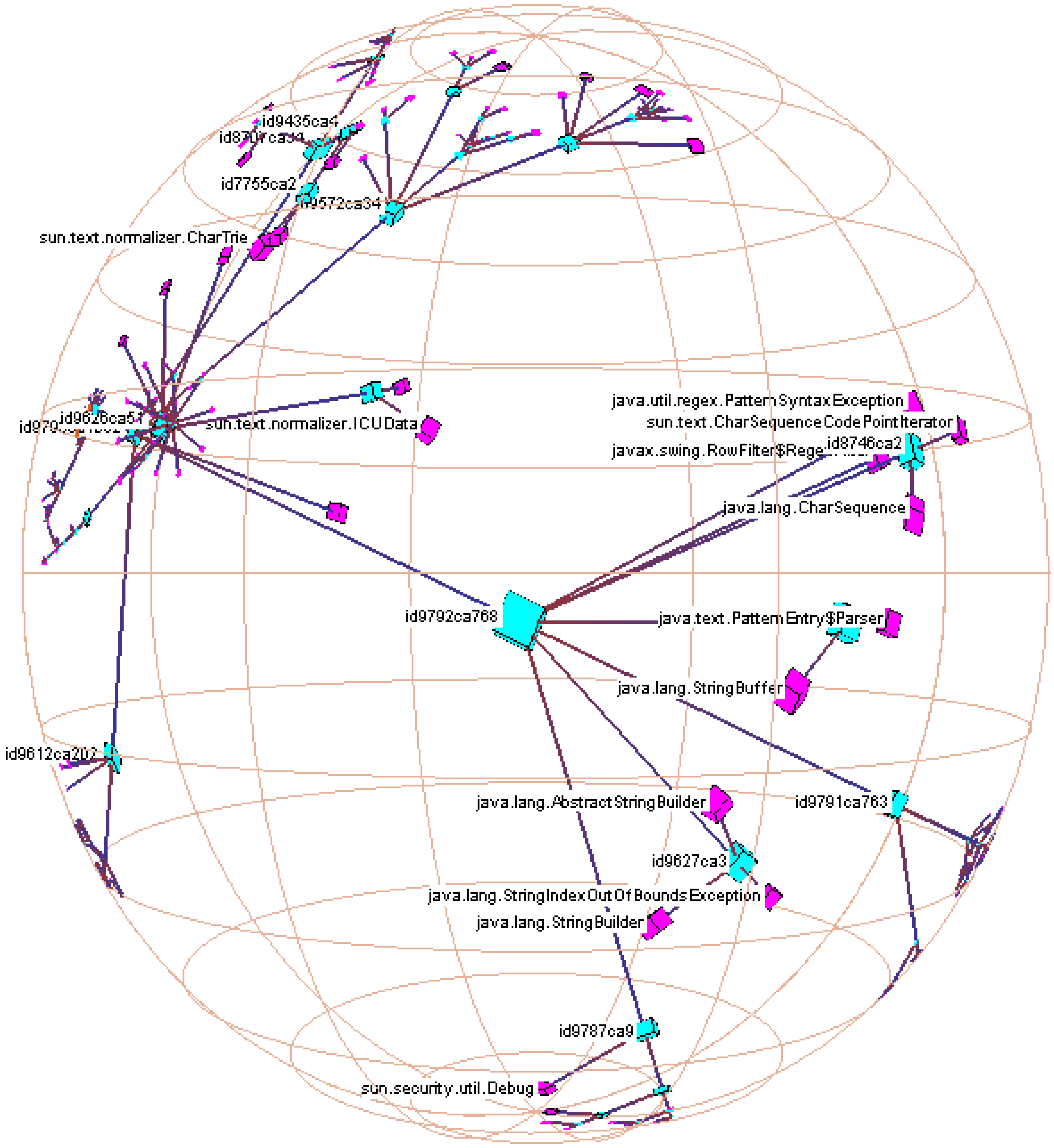

## 10.2 Statistical Measures

### 10.2.1 Packages by ubiquity

In these statistical measures we investigated how class name length (in tokens[7]) correlates with the position in the cluster tree. For each node in the cluster tree we calculated how many token-wise suffixes of each name are matched (i.e. have the same token-wise prefix) in the subtree of the node. Suffix Tree data structure and Dynamic Programming (section 4.6) allowed computing it fast (subquadratic complexity). Then we computed multiple averages (per token-wise prefix):

- average match depth: the average depth of cluster tree node at which the suffixes of this prefix matched
- average match height: the same, but height is averaged
- average number of nodes in the subtree: the same, but the number of nodes in the subtree of a cluster tree node is averaged

Then we applied ranking approach: comparison across each dimension adds +1/-1 to the sum. Afterwards, we sorted the tokenwise name prefixes according to this rank, the figure below demonstrates the result. We call this rank "package ubiquity".

D:\...0-06-18\ctEval\perf\clientUbiquity.txt Win Line 29/99 Col 41

| rank | avgMH | avgMD | avgNU | \|Suff.\| | Prefix |
|---|---|---|---|---|---|
| 2 | 201.000 | 38.0000 | 6687.00 | 2 | com.kpmg.kpo.dto.PortfolioManagerServiceMessageOut |
| 124 | 52.3289 | 179.002 | 3552.29 | 62 | com.kpmg.esb |
| 130 | 50.2879 | 178.242 | 2967.27 | 12 | com.kpmg.esb.mule.exception |
| 135 | 49.6095 | 181.248 | 3399.22 | 15 | com.kpmg.esb.mule.component |
| 152 | 44.1871 | 187.064 | 3436.16 | 20 | com.kpmg.kpo.dto |
| 158 | 42.6056 | 188.310 | 3343.50 | 60 | com.kpmg.esb.mule |
| 160 | 41.0000 | 198.000 | 6183.00 | 2 | com.kpmg.kpo.auth |
| 160 | 41.0000 | 198.000 | 6183.00 | 2 | com.kpmg.kpo.domain.WorkflowTemplate |
| 171 | 37.6233 | 195.893 | 3615.07 | 25 | com.kpmg.esb.mule.transformers |
| 182 | 35.5175 | 200.900 | 3468.13 | 194 | com.kpmg.kpo.generated |
| 182 | 35.5175 | 200.900 | 3468.13 | 194 | com.kpmg.kpo.generated.jaxws |
| 186 | 33.5211 | 197.355 | 3246.39 | 799 | com.kpmg |
| 187 | 32.5333 | 202.067 | 4309.33 | 6 | com.kpmg.kpo.audittrail.impl |
| 197 | 32.3370 | 193.917 | 2200.85 | 24 | com.kpmg.kpo.dao.hibernate |
| 204 | 32.1333 | 199.200 | 3879.33 | 6 | com.kpmg.kpo.dao.enums |
| 210 | 31.7556 | 192.289 | 1620.00 | 10 | com.kpmg.kpo.usertypes |
| 211 | 31.6315 | 198.902 | 3168.09 | 724 | com.kpmg.kpo |
| 214 | 31.6190 | 201.190 | 3970.38 | 7 | com.kpmg.kpo.comparator |
| 216 | 30.5714 | 202.190 | 4072.00 | 8 | com.kpmg.kpo.audittrail |
| 219 | 28.8791 | 197.183 | 2063.89 | 78 | com.kpmg.kpo.dao |
| 247 | 28.0444 | 208.600 | 3442.38 | 10 | com.kpmg.esb.mule.transformers.webservable |
| 248 | 27.6667 | 208.667 | 4122.67 | 3 | com.kpmg.kpo.audittrail.impl.DelegatingAuditLogImpl |
| 251 | 24.6405 | 203.248 | 2822.97 | 18 | com.kpmg.kpo.action |
| 263 | 23.3333 | 202.533 | 2777.40 | 6 | com.kpmg.kpo.web.security |
| 266 | 25.3649 | 202.679 | 2688.68 | 76 | com.kpmg.kpo.web |
| 278 | 20.2308 | 203.352 | 1688.14 | 14 | com.kpmg.kpo.web.action |
| 280 | 20.8222 | 203.711 | 1613.04 | 10 | com.kpmg.kpo.web.validation |
| 338 | 19.9954 | 203.975 | 1308.08 | 64 | com.kpmg.kpo.domain |
| 362 | 19.7866 | 202.939 | 1111.20 | 135 | com.kpmg.kpo.service |
| 367 | 19.2692 | 205.029 | 1450.69 | 41 | com.kpmg.kpo.exception |
| 383 | 8.57635 | 174.520 | 451.298 | 30 | com.kpmg.kpo.generated.jaxws.crm |
| 385 | 18.7579 | 204.458 | 1193.55 | 20 | com.kpmg.kpo.web.binding |
| 386 | 18.3576 | 204.422 | 1104.15 | 51 | com.kpmg.kpo.service.impl |
| 393 | 16.3333 | 205.000 | 485.000 | 3 | com.kpmg.kpo.web.service |
| 394 | 15.6000 | 204.667 | 448.200 | 6 | com.kpmg.kpo.web.view |
| 397 | 16.0000 | 206.000 | 180.000 | 2 | com.kpmg.kpo.domain.EmployeeAuthorisation |
| 398 | 15.5000 | 213.000 | 806.500 | 4 | com.kpmg.kpo.web.exceptionhandling |
| 407 | 13.0000 | 207.000 | 299.000 | 3 | com.kpmg.kpo.domain.WorkflowInstance |
| 407 | 13.0000 | 207.000 | 299.000 | 2 | com.kpmg.kpo.dto.DocumentMessage |
| 423 | 11.6044 | 225.385 | 441.538 | 15 | com.kpmg.kpo.generated.jaxws.serviceprocessor |

**Figure 10-3 Packages by ubiquity**

[7] For example, java.lang.String has 3 tokens: java, lang and String

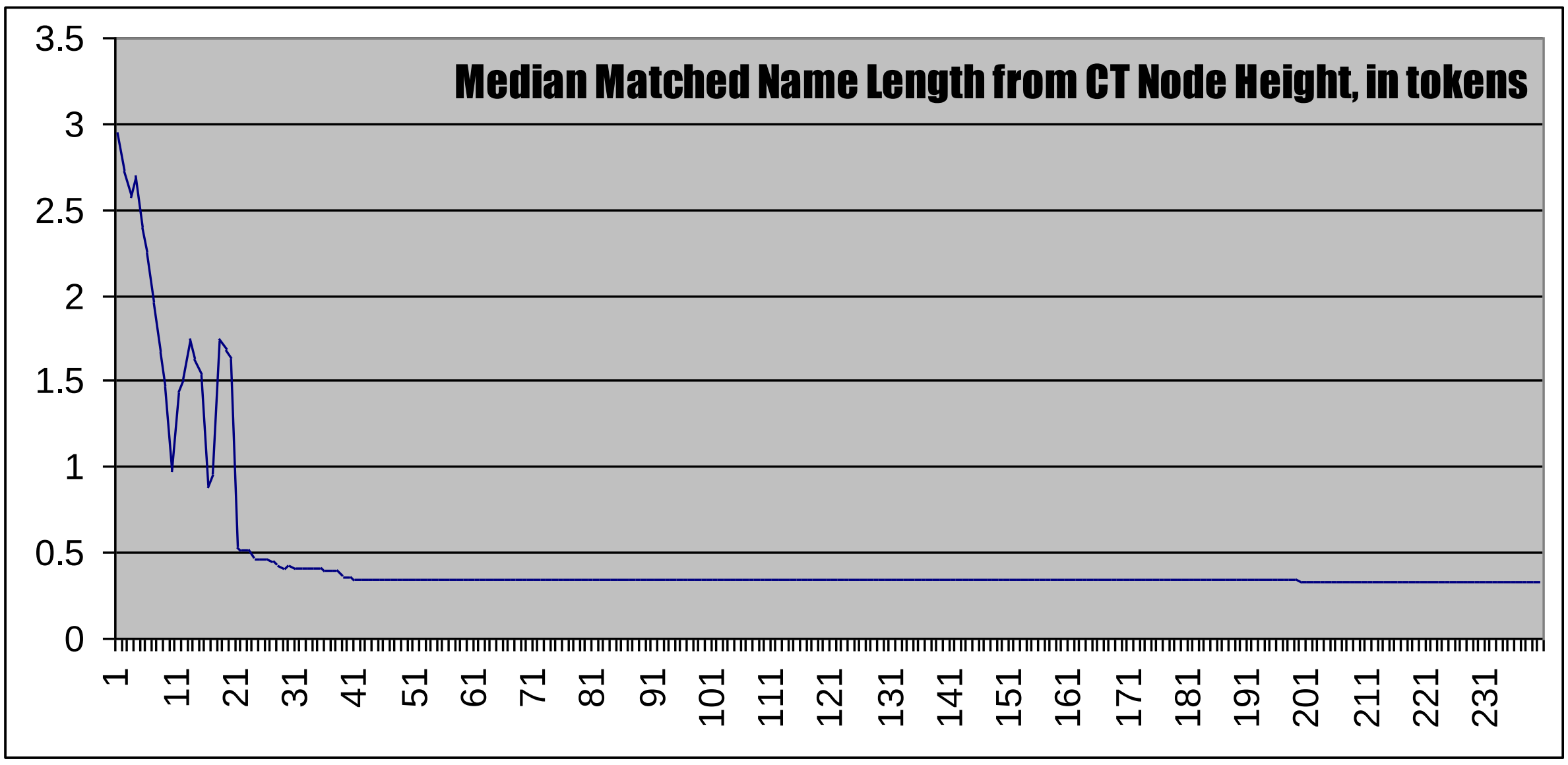

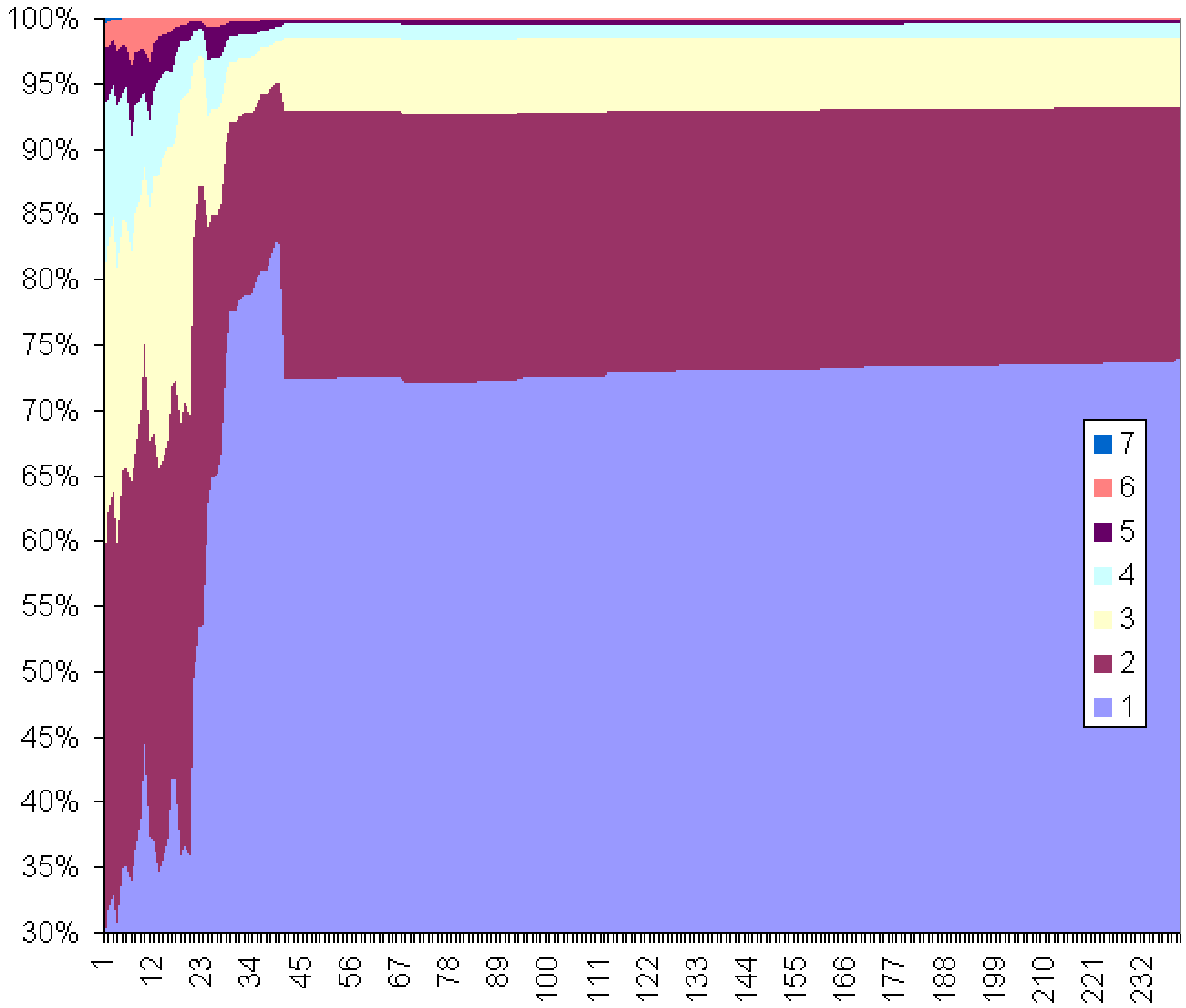

**Figure 10-4 Shares of package name depth over cluster tree node height**

### 10.2.2 Architectural Fitness of SE Class couplings

In the Figure 10-5 below we see that 10% of relations between SE artifacts violate the implicit architecture, while 90% fit it very well. In the next diagram, Figure 10-6, we see that 80% of weighted misfits is constituted by 16% of relations (couplings).

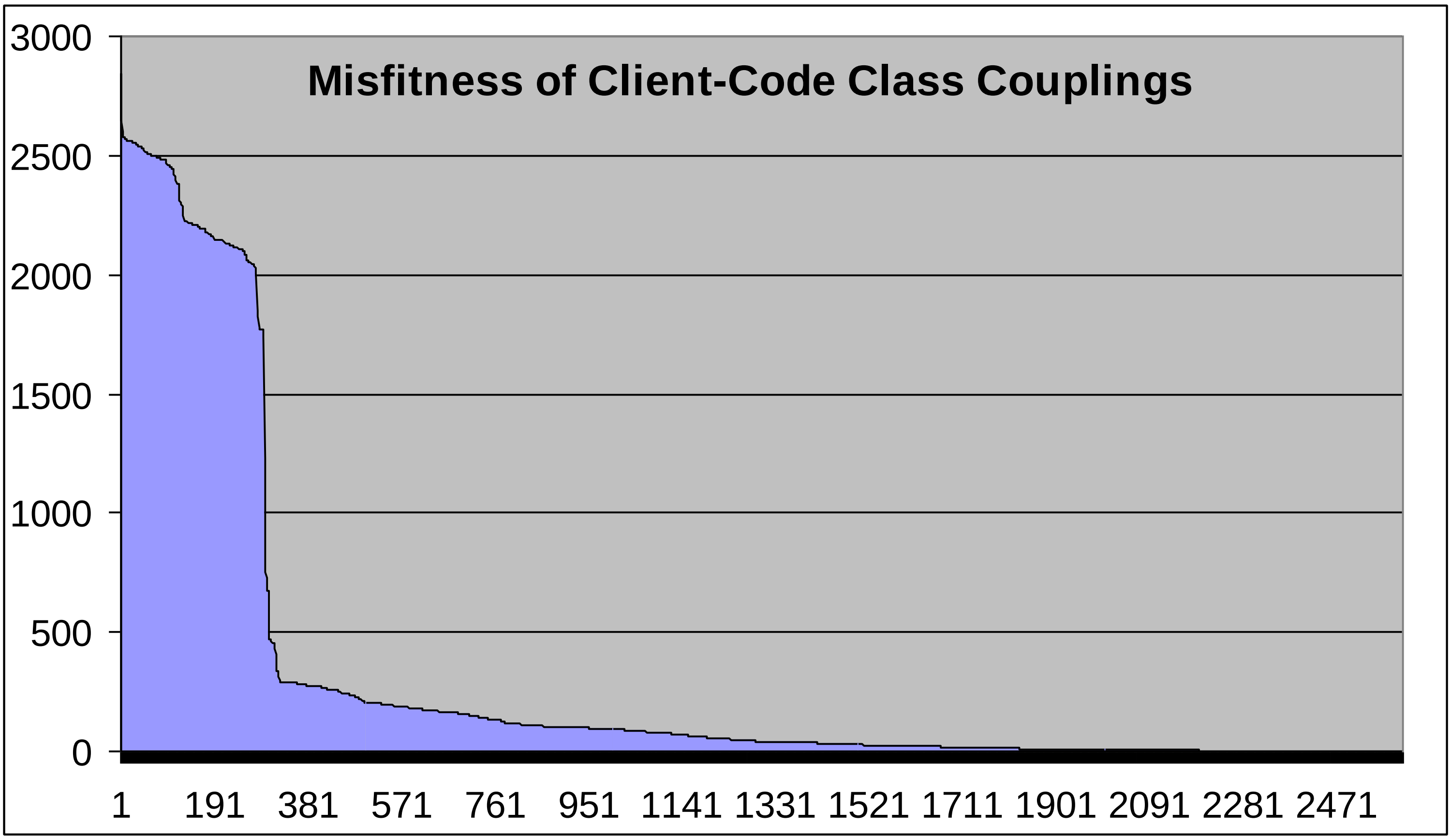

**Figure 10-5 Architectural Violation Extent over Sorted Ordinal**

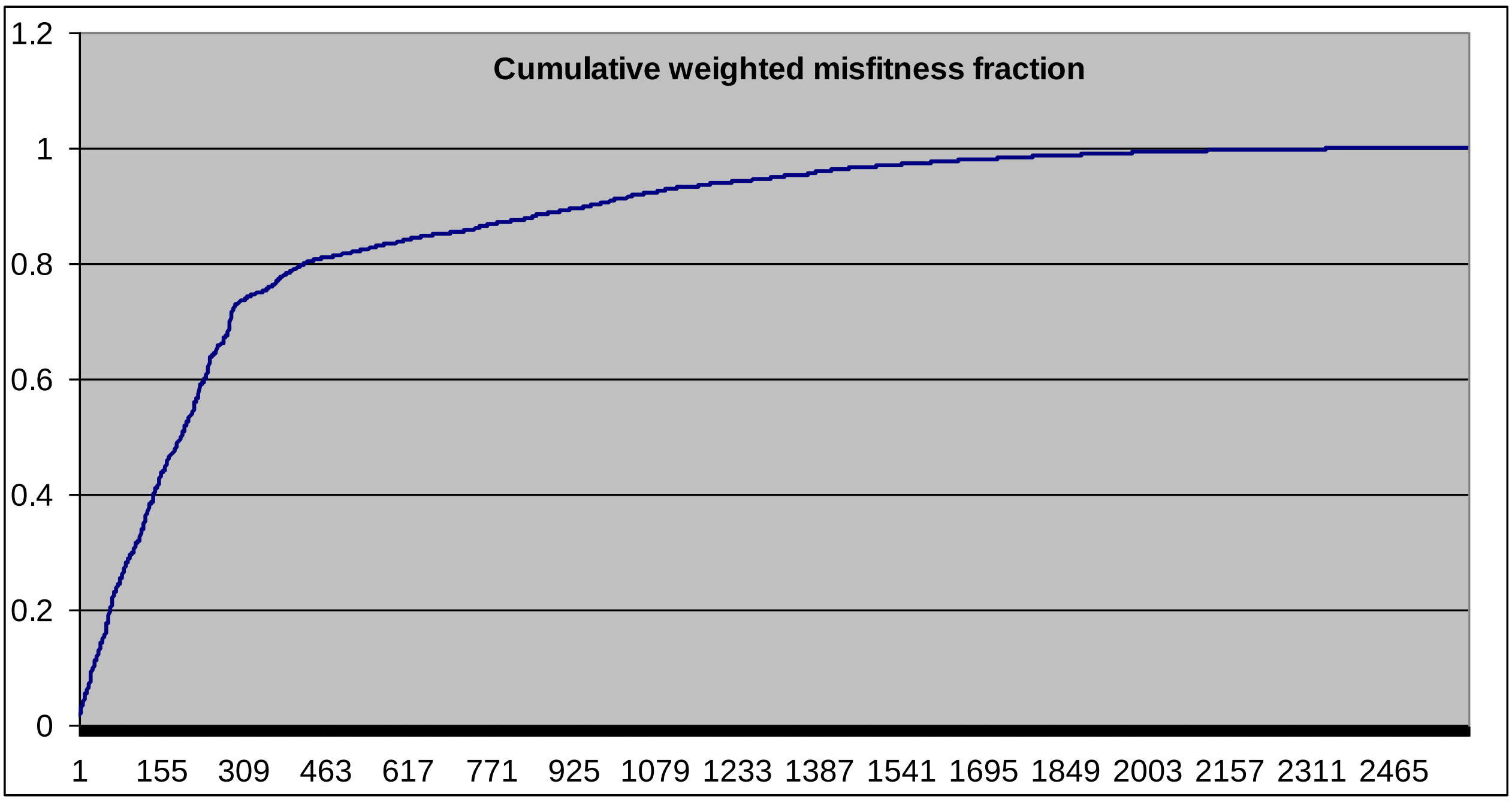

**Figure 10-6 Sum of misfitness times weight, from most misfitting on**

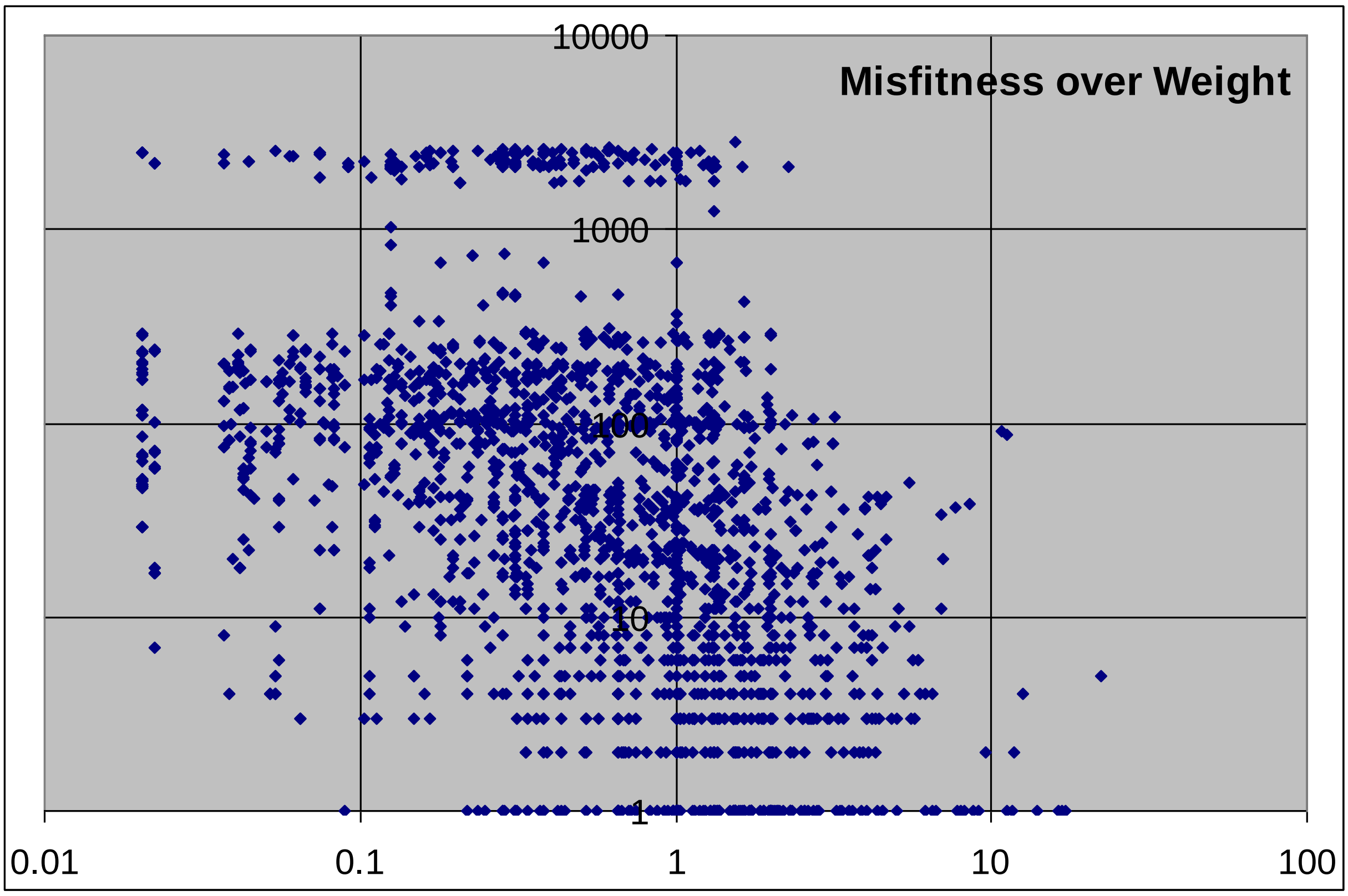

**Figure 10-7 Architectural Violation over Coupling Strength: Note that the scales are logarithmic**

To draw the diagram below, relations between SE artifacts were sorted by weiht, from the most strongly coupled to the least coupled.

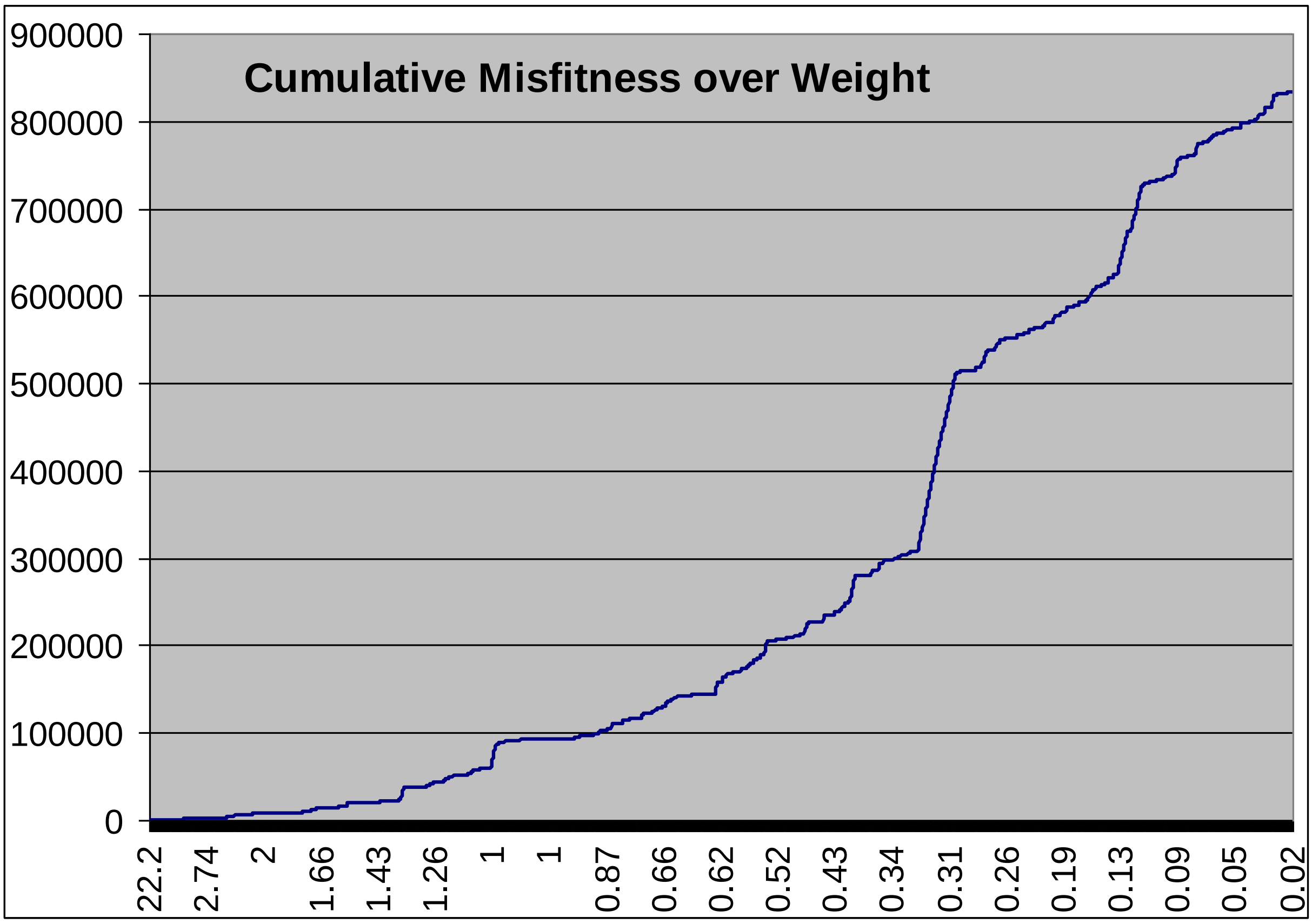

## 10.3 Analyzed Source Code Examples

### 10.3.1 Dependency analysis

#### 10.3.1.1 A Java class that does not use calls or external field accesses

```
package com.kpmg.esb.mule.component;

import com.kpmg.kpo.service.ServiceMessageRuleService;

/**
 * Common abstract super class that provides injection mechanism for services
 *
 */
public abstract class AbstractPersistableComponent {

     /**
      * serviceMessageRuleService injected by Spring
      */
     private ServiceMessageRuleService serviceMessageRuleService;

     public ServiceMessageRuleService getServiceMessageRuleService() {
            return serviceMessageRuleService;
     }

     /**
      * Accessor method.
      *
      * @param serviceMessageRule
      */
     public void setServiceMessageRule(ServiceMessageRuleService serviceMessageRule) {
            this.serviceMessageRuleService = serviceMessageRule;
     }

}
```

#### 10.3.1.2 Some classes whose dependencies are not specific at all

```
package com.kpmg.kpo.action;

import org.jbpm.graph.def.ActionHandler;
import org.jbpm.graph.exe.ExecutionContext;

public class SendFile extends GenericHandler implements ActionHandler {

     private static final long serialVersionUID = -5704117378931708811L;

     private String messageType;

     @Override
     public void execute(ExecutionContext executionContext) throws Exception {
            System.out.println(messageType);
     }

     public void setMessageType(String messageType) {
            this.messageType = messageType;
     }

}
```

Perhaps objects of this class are used as items in array or linked data structures.

Firstly, the type of an item (CompleteTaskCommand in this case) plays a considerable role in decision making regarding the further handling of the item and further program flow. Conditional flow branches basing on **instanceof** operator result.
Secondly, specific features of an instance of this class are stored in String fields, <comment> and <transition> in our example.

```
Package com.kpmg.kpo.web.binding;

import java.io.Serializable;
import com.kpmg.kpo.domain.TaskInstance;

public class CompleteTaskCommand implements Serializable {
	/**
	 * SerialVersionUID, required by Serializable.
	 */
	private static final long serialVersionUID = 1L;

	private TaskInstance task;
	private String comment;
	private String transition;

	/**
	 * @return the task
	 */
	public TaskInstance getTask() {
		return task;
	}
	/**
	 * @param task the task to set
	 */
	public void setTask(TaskInstance task) {
		this.task = task;

		//Reset other fields if we choose a new Task.
		This.comment = null;
		this.transition = null;
	}
	/**
	 * @return the comment
	 */
	public String getComment() {
		return comment;
	}
	/**
	 * @param comment the comment to set
	 */
	public void setComment(String comment) {
		this.comment = comment;
	}
	/**
	 * @return the transition
	 */
	public String getTransition() {
		return transition;
	}
	/**
	 * @param transition the transition to set
	 */
	public void setTransition(String transition) {
		this.transition = transition;
	}
}
```

#### 10.3.1.3 Dependencies lost at Java compile time

Certain relations are lost while compiling .java files into .class files. This occurs for an example class below:

```
package com.kpmg.kpo;
/**
 * Global variables for the worklfow
 */
public final class WorkFlowVariables {
      /**
       * Class cannot be instantiated.
       */
      private WorkFlowVariables() {
      }
      /**
       * Constant name used to store the transition map.
       */
      public static final String TRANSITIONS = "transitions";
      /**
       * Constant name use to identify the domain peer in jBPM.
       */
      public static final String PEER = "peer";
      /**
       * The comma-separated list of undo-actions.
       */
      public static final String UNDO_ACTIONS = "undo_actions";
      /**
       * Due-date for a specific Task.
       */
      public static final String DUE_DATE = "dueDate";
      /**
       * Warning start date for a specific task.
       */
      public static final String WARNING_START_DATE = "warningStartDate";
      /**
       * Key to store the default transition (if any) under.
       */
      public static final String DEFAULT_TRANSITION = "default_transition";
}
```

Whenever a string constant from WorkFlowVariables class is used in java source code, e.g. WorkFlowVariables.PEER, its value is substituted into the binary code ("peer" in our example) rather than a reference to field PEER of type WorkFlowVariables. See the example that uses WorkFlowVariables.PEER below. As a consequence of this fact, vertex WorkFlowVariables gets no adjacent edges in the relations graph and thus becomes an orphan, i.e. the single vertex in a disjoint component.

```
Package com.kpmg.kpo.action;

import java.io.Serializable;

import org.jbpm.context.exe.ContextInstance;
import org.jbpm.graph.def.ActionHandler;
import org.jbpm.graph.exe.ExecutionContext;

import com.kpmg.kpo.WorkFlowVariables;
import com.kpmg.kpo.domain.MessageType;
import com.kpmg.kpo.domain.WorkflowInstance;

/**
 * Generic action handler for action that need a {@link MessageType} and the id
 * of a {@link WorkflowInstance}
 * <p>
 * The taskName may be provided. It can be derived from the task that is
 * executed but a designer may need to choose a different taskName as a
 * different task in essence is responsible for firing the event. For example a
```

```
 * reset task does execute this handler but the task that executed the reset
 * task itself is the taskName we want to provide.
 *
 */
public abstract class AbstractDocumentHandler extends GenericHandler implements ActionHandler
{

	private static final long serialVersionUID = 7525089866530362953L;

	private String messageType;
	private String taskName;

	@Override
	public void execute(ExecutionContext executionContext) throws Exception {
		ContextInstance jbpmContext = executionContext.getContextInstance();

		final WorkflowInstance peer = (WorkflowInstance) jbpmContext
				.getVariable(WorkFlowVariables.PEER);

		if ( taskName == null || taskName.equals("")) {
			taskName = executionContext.getEventSource().getName();
		}

		Serializable command = getCommandObject(peer.getId(), taskName, messageType,
executionContext);
		JMSUtil.getInstance().sendByJMS(getQueueName(), command);
	}

	public void setMessageType(String messageType) {
		this.messageType = messageType;
	}

	abstract String getQueueName();

	abstract Serializable getCommandObject(Long workflowInstanceId,
			String taskName, String messageType,
			ExecutionContext context);
}
```

### 10.3.1.4 Problematic cases

```
package com.kpmg.kpo.audittrail.impl;

import java.util.UUID;
import java.util.concurrent.ExecutorService;
import java.util.concurrent.Executors;
import java.util.concurrent.ThreadFactory;

import com.kpmg.kpo.audittrail.AuditLog;
import com.kpmg.kpo.dto.AuditEntryDTO;

/**
 * AuditLog implementation that delegates to another AuditLog running in a
 * separate thread. This fire-and-forget approach helps in keeping audit trail
 * logging fast, yet the code invoking these log statements does not know about
 * success or failure of storing the log entry in the database (nor about
 * validation results).
 * <p/>
 * As a side-effect, this delegating AuditLog generates a GUID returned by the
 * log method (and sets that GUID on the AuditEntryData instance passed to the
 * delegate).
 * <p/>
 * The assumption is that using "unmanaged" threads in Jboss is OK (within
```

```
 * certain bounds, of course).
 */
public final class DelegatingAuditLogImpl implements AuditLog {

     private AuditLog delegate;
     private ExecutorService executorService;

     public DelegatingAuditLogImpl(AuditLog delegate,
                  ExecutorService executorService) {
            this.delegate = delegate;
            this.executorService = executorService;
     }

     public DelegatingAuditLogImpl(AuditLog delegate) {
            // The following ExecutorService is guaranteed to execute the log calls
sequentially
            this(delegate, Executors.newSingleThreadExecutor(new ThreadFactory() {

                  public Thread newThread(Runnable r) {
                         Thread t = Executors.defaultThreadFactory().newThread(r);
                         t.setName("audittrail-" + t.getName());
                         return t;
                  }
            }));
     }

     public String log(AuditEntryDTO data) {
            String guid = UUID.randomUUID().toString();
            final AuditEntryDTO dataWithGuid = data.withGuid(guid);

            // This is fire-and-forget. If the delegate's log call in the separate thread
throws an exception,
            // it will not affect this thread.

            This.executorService.execute(new Runnable() {
                  public void run() {
                         DelegatingAuditLogImpl.this.delegate.log(dataWithGuid);
                  }
            });

            return guid;
     }
}
```

### 10.3.2 Call Graph Extraction

The class `StatusType` is an Enum (Java). Obviously, it does not call any methods from `com.sun.imageio` package indeed, and the source code confirms that. However, call graph extraction adds noise which we can see in the figure below. The figure demonstrates relations between SE artifacts lifted to class-level. The number "238" in the left-top corner stands for the number of other classes with whom class `StatusType` has relations.

```
edit finalEd.txt - Far
..KPMG\    \cg-cha-mfir\trees\finalEd.txt          Win     Line    18384/1557249    Col 43
    8312 3.4059862969764293000e-01
238 com.kpmg.kpo.domain.StatusType
    82 8.3785451354258180000e-01 com.kpmg.kpo.action.ResetAction
    176 1.0839695307160290000e+00 com.kpmg.kpo.dao.hibernate.HibernateWorkflowInstanceDao
    232 2.6186920515673130000e+00 com.kpmg.kpo.domain.StatusType$1
    233 1.6186920515673129000e+00 com.kpmg.kpo.domain.StatusType$2
    234 1.6186920515673129000e+00 com.kpmg.kpo.domain.StatusType$3
    235 1.6186920515673129000e+00 com.kpmg.kpo.domain.StatusType$4
    236 1.6186920515673129000e+00 com.kpmg.kpo.domain.StatusType$5
    237 1.6186920515673129000e+00 com.kpmg.kpo.domain.StatusType$6
    238 1.6186920515673129000e+00 com.kpmg.kpo.domain.StatusType$7
    224 2.6355009617787870000e-01 com.kpmg.kpo.domain.PersistedEnum
    1903 1.2559628619977468000e-02 java.lang.IllegalArgumentException
    1930 2.5254945709878303000e-02 java.lang.Object
    1965 2.1457469942358720000e-02 java.lang.String
    1895 3.0118283048264290000e-01 java.lang.Enum
    854 2.4461120645361198000e-02 com.sun.imageio.plugins.jpeg.DHTMarkerSegment$Htable
    853 2.4281827519883340000e-02 com.sun.imageio.plugins.jpeg.DHTMarkerSegment
    856 2.4281827519883340000e-02 com.sun.imageio.plugins.jpeg.DQTMarkerSegment$Qtable
    855 2.4281827519883340000e-02 com.sun.imageio.plugins.jpeg.DQTMarkerSegment
    862 2.4281827519883340000e-02 com.sun.imageio.plugins.jpeg.JFIFMarkerSegment$ICCMarkerS
    864 2.4193248576394103000e-02 com.sun.imageio.plugins.jpeg.JFIFMarkerSegment$JFIFExtens
    865 2.4193248576394103000e-02 com.sun.imageio.plugins.jpeg.JFIFMarkerSegment$JFIFThumb
    866 2.4281827519883340000e-02 com.sun.imageio.plugins.jpeg.JFIFMarkerSegment$JFIFThumbJ
    860 2.4281827519883340000e-02 com.sun.imageio.plugins.jpeg.JFIFMarkerSegment
    881 2.4281827519883340000e-02 com.sun.imageio.plugins.jpeg.JPEGMetadata
    882 2.3760687367118564000e-02 com.sun.imageio.plugins.jpeg.MarkerSegment
    884 2.4281827519883340000e-02 com.sun.imageio.plugins.jpeg.SOFMarkerSegment$ComponentSp
    883 2.4371115269888378000e-02 com.sun.imageio.plugins.jpeg.SOFMarkerSegment
    886 2.4281827519883340000e-02 com.sun.imageio.plugins.jpeg.SOSMarkerSegment$ScanCompone
    885 2.4371115269888378000e-02 com.sun.imageio.plugins.jpeg.SOSMarkerSegment
    896 2.4922231212397946000e-02 com.sun.imageio.plugins.png.PNGMetadata
    1169 2.4371115269888378000e-02 java.awt.BufferCapabilities
    1283 2.4281827519883340000e-02 java.awt.GridBagConstraints
    1291 2.4371115269888378000e-02 java.awt.ImageCapabilities
    1293 2.4018182561718605000e-02 java.awt.Insets
    1295 2.4371115269888378000e-02 java.awt.JobAttributes
    1345 2.4371115269888378000e-02 java.awt.PageAttributes
    1367 2.4018182561718605000e-02 java.awt.RenderingHints
    1430 2.4371115269888378000e-02 java.awt.datatransfer.DataFlavor
    1437 2.4281827519883340000e-02 java.awt.datatransfer.MimeType
```

The aforementioned class-level noise results from the underlying method-level noise. CHA call graph extraction encountered a call to `Object.clone()` from `StatusType.values()` and concluded many calls to different classes derived from `Object` possible, however those calls never occur indeed and were by no means intended by software engineers designing and implementing the source code. The data containing noise calls is illustrated in the figure below. The number "145" stands for the number of destinations which method StatusType.values() may call, according to this call graph extraction approach. This number is also the number of outgoing arcs from the corresponding vertex in our input graph we would get in case we use this call graph extraction approach.

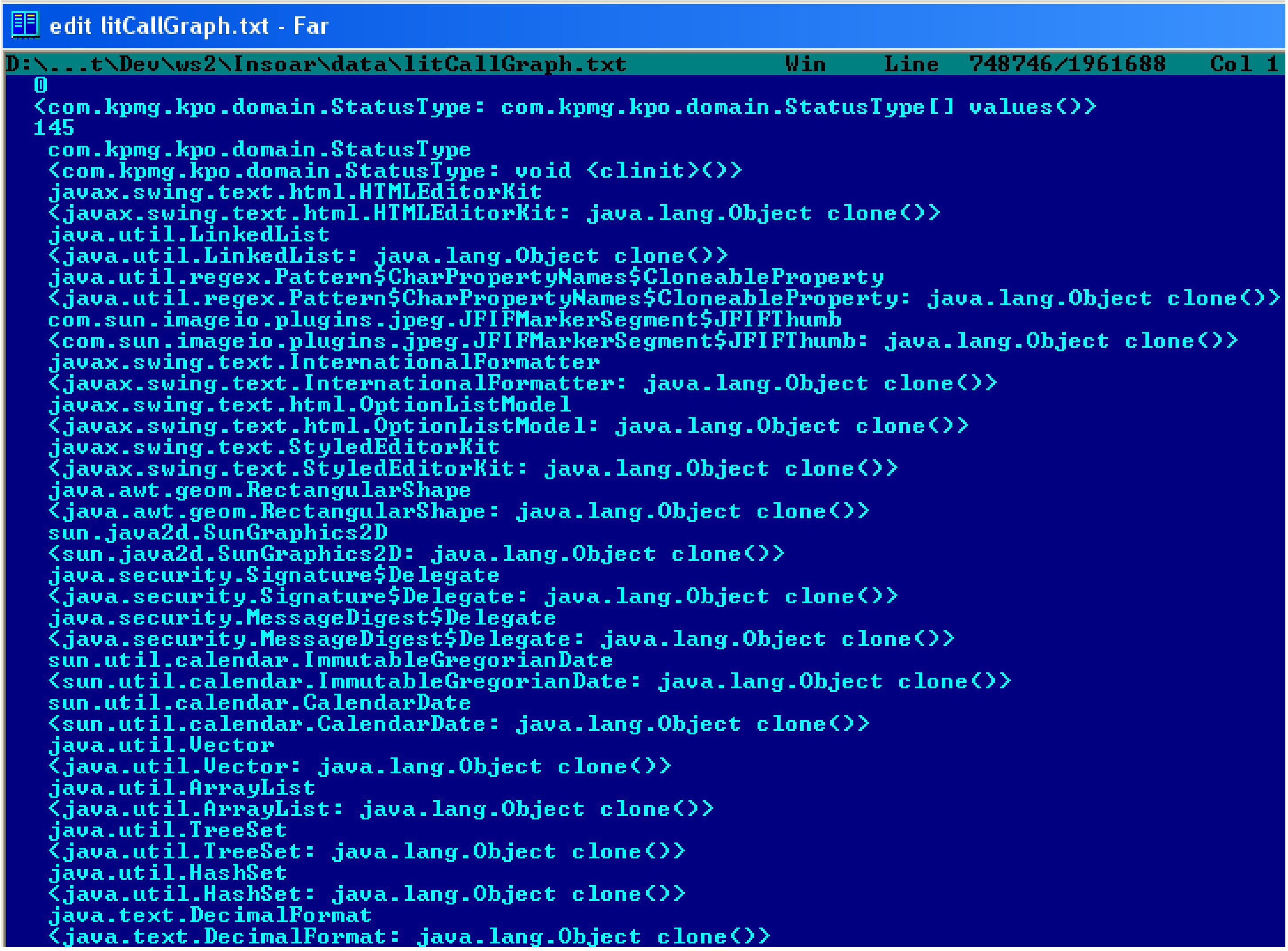

```
edit litCallGraph.txt - Far
D:\...t\Dev\ws2\Insoar\data\litCallGraph.txt          Win     Line  748746/1961688    Col 1
  0
  <com.kpmg.kpo.domain.StatusType: com.kpmg.kpo.domain.StatusType[] values()>
  145
   com.kpmg.kpo.domain.StatusType
   <com.kpmg.kpo.domain.StatusType: void <clinit>()>
   javax.swing.text.html.HTMLEditorKit
   <javax.swing.text.html.HTMLEditorKit: java.lang.Object clone()>
   java.util.LinkedList
   <java.util.LinkedList: java.lang.Object clone()>
   java.util.regex.Pattern$CharPropertyNames$CloneableProperty
   <java.util.regex.Pattern$CharPropertyNames$CloneableProperty: java.lang.Object clone()>
   com.sun.imageio.plugins.jpeg.JFIFMarkerSegment$JFIFThumb
   <com.sun.imageio.plugins.jpeg.JFIFMarkerSegment$JFIFThumb: java.lang.Object clone()>
   javax.swing.text.InternationalFormatter
   <javax.swing.text.InternationalFormatter: java.lang.Object clone()>
   javax.swing.text.html.OptionListModel
   <javax.swing.text.html.OptionListModel: java.lang.Object clone()>
   javax.swing.text.StyledEditorKit
   <javax.swing.text.StyledEditorKit: java.lang.Object clone()>
   java.awt.geom.RectangularShape
   <java.awt.geom.RectangularShape: java.lang.Object clone()>
   sun.java2d.SunGraphics2D
   <sun.java2d.SunGraphics2D: java.lang.Object clone()>
   java.security.Signature$Delegate
   <java.security.Signature$Delegate: java.lang.Object clone()>
   java.security.MessageDigest$Delegate
   <java.security.MessageDigest$Delegate: java.lang.Object clone()>
   sun.util.calendar.ImmutableGregorianDate
   <sun.util.calendar.ImmutableGregorianDate: java.lang.Object clone()>
   sun.util.calendar.CalendarDate
   <sun.util.calendar.CalendarDate: java.lang.Object clone()>
   java.util.Vector
   <java.util.Vector: java.lang.Object clone()>
   java.util.ArrayList
   <java.util.ArrayList: java.lang.Object clone()>
   java.util.TreeSet
   <java.util.TreeSet: java.lang.Object clone()>
   java.util.HashSet
   <java.util.HashSet: java.lang.Object clone()>
   java.text.DecimalFormat
   <java.text.DecimalFormat: java.lang.Object clone()>
```

Points-to analysis techniques (RTA, VTA, Spark) help to alleviate this problem substantially, though, at the cost of some mistakenly dropped calls too. See the picture below.

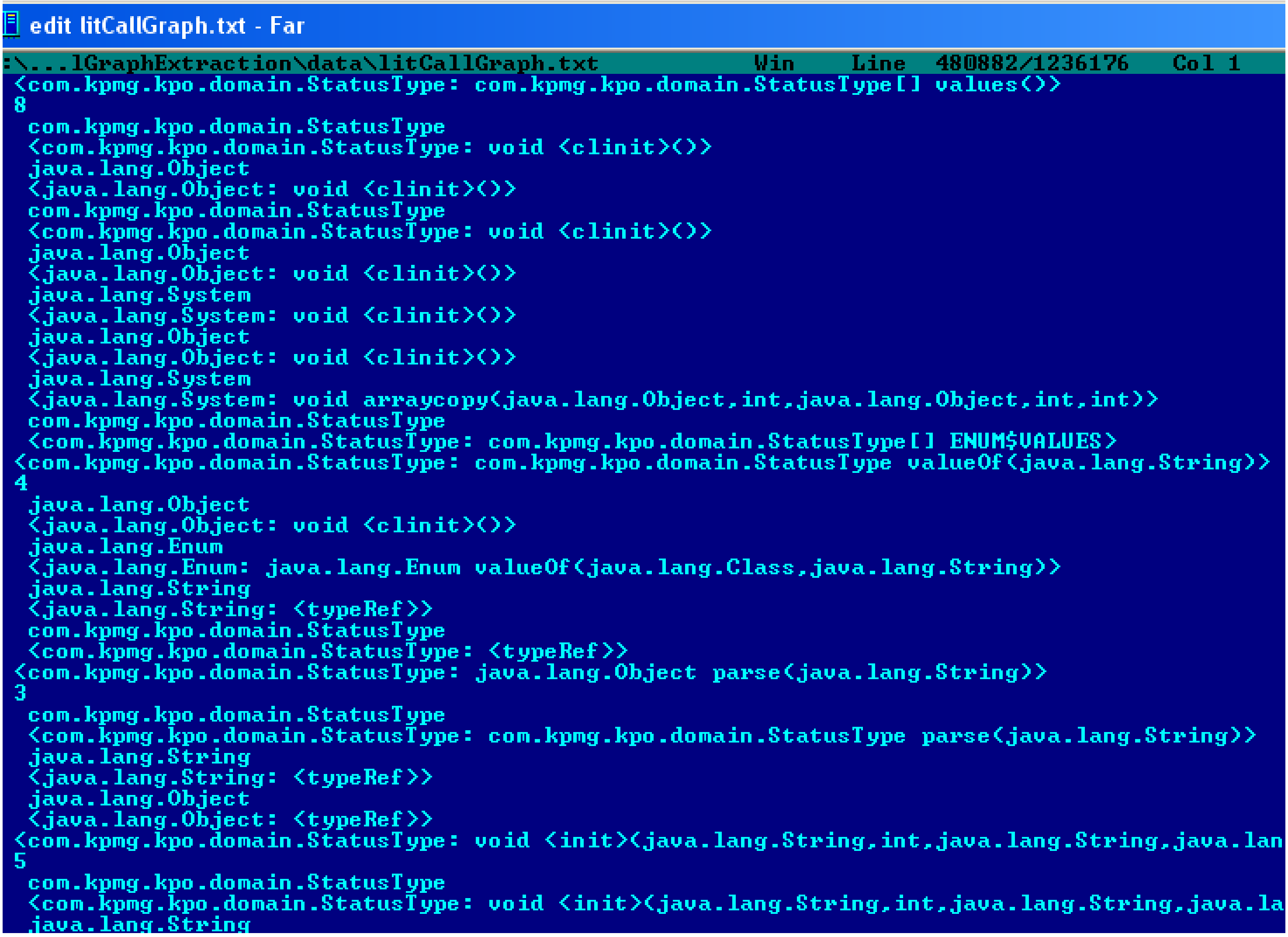

```
edit litCallGraph.txt - Far
:\...lGraphExtraction\data\litCallGraph.txt          Win     Line   480882/1236176    Col 1
<com.kpmg.kpo.domain.StatusType: com.kpmg.kpo.domain.StatusType[] values()>
8
 com.kpmg.kpo.domain.StatusType
 <com.kpmg.kpo.domain.StatusType: void <clinit>()>
 java.lang.Object
 <java.lang.Object: void <clinit>()>
 com.kpmg.kpo.domain.StatusType
 <com.kpmg.kpo.domain.StatusType: void <clinit>()>
 java.lang.Object
 <java.lang.Object: void <clinit>()>
 java.lang.System
 <java.lang.System: void <clinit>()>
 java.lang.Object
 <java.lang.Object: void <clinit>()>
 java.lang.System
 <java.lang.System: void arraycopy(java.lang.Object,int,java.lang.Object,int,int)>
 com.kpmg.kpo.domain.StatusType
 <com.kpmg.kpo.domain.StatusType: com.kpmg.kpo.domain.StatusType[] ENUM$VALUES>
<com.kpmg.kpo.domain.StatusType: com.kpmg.kpo.domain.StatusType valueOf(java.lang.String)>
4
 java.lang.Object
 <java.lang.Object: void <clinit>()>
 java.lang.Enum
 <java.lang.Enum: java.lang.Enum valueOf(java.lang.Class,java.lang.String)>
 java.lang.String
 <java.lang.String: <typeRef>>
 com.kpmg.kpo.domain.StatusType
 <com.kpmg.kpo.domain.StatusType: <typeRef>>
<com.kpmg.kpo.domain.StatusType: java.lang.Object parse(java.lang.String)>
3
 com.kpmg.kpo.domain.StatusType
 <com.kpmg.kpo.domain.StatusType: com.kpmg.kpo.domain.StatusType parse(java.lang.String)>
 java.lang.String
 <java.lang.String: <typeRef>>
 java.lang.Object
 <java.lang.Object: <typeRef>>
<com.kpmg.kpo.domain.StatusType: void <init>(java.lang.String,int,java.lang.String,java.lan
5
 com.kpmg.kpo.domain.StatusType
 <com.kpmg.kpo.domain.StatusType: void <init>(java.lang.String,int,java.lang.String,java.la
 java.lang.String
```

### 10.3.3 Class name contradicts the purpose

We have to support a strong claim about proper clustering result from section 7.2.2.4 with source code of the class whose name contradicts the purpose. We can see that indeed the source code works mostly with regular expressions and JBPM expression evaluation. Thus, the clustering was correct, while the name of the class is deceiptive.

```
package com.kpmg.kpo.action;
import java.util.regex.Matcher;
import java.util.regex.Pattern;
import org.jbpm.graph.exe.ExecutionContext;
import org.jbpm.jpdl.el.impl.JbpmExpressionEvaluator;

public abstract class GenericHandler {
    //Prepare to identify any EL expressions, #{…}, regex: "#\{.*?\}".
    Private static final String EL_PATTERN_STRING = "#\\{.*?\\}";
    //Turn the pattern string into a regex pattern class.
    Private static final Pattern EL_PATTERN = Pattern.compile(EL_PATTERN_STRING);

    //Evaluate the input as a possible EL expression.
    Protected Object evaluateEL(String inputStr, ExecutionContext ec) {
        if (inputStr == null) { return null; }
```

```
        Matcher matcher = EL_PATTERN.matcher(inputStr);
        if (matcher.matches()) { //input is one big EL expression
            return JbpmExpressionEvaluator.evaluate(inputStr, ec);
        } else {
            return inputStr;
        }
    }

    /* Treats input as a possible series of EL expressions and concatenates what is found.  */
    protected String concatenateEL(String inputStr, ExecutionContext ec) {
        if (inputStr == null) { return null; }
        Matcher matcher = EL_PATTERN.matcher(inputStr);
        StringBuffer buf = new StringBuffer();
        while (matcher.find()) {
            // Get the match result
            String elExpr = matcher.group();
            // Evaluate EL expression
            Object o = JbpmExpressionEvaluator.evaluate(elExpr, ec);
            String elValue = "";
            if (o != null) {
                elValue = String.valueOf(JbpmExpressionEvaluator.evaluate(elExpr, ec));
            }
            // Insert the calculated value in place of the EL expression
            matcher.appendReplacement(buf, elValue);
        }
        matcher.appendTail(buf);
        // Deliver result
        if (buf.length() > 0) {
            return buf.toString();
        } else { return null; }
    }

    /* Returns true if the value is a String which contains the pattern delineating an EL
expression.  */
    protected boolean hasEL(Object value) {
        if (value instanceof String) {
            Matcher matcher = EL_PATTERN.matcher((String) value);
            return matcher.find();
        }
        return false;
    }

    /* Returns true if the value is a String which in its entirety composes one EL expression.
*/
    protected boolean isEL(Object value) {
        if (value instanceof String) {
            Matcher matcher = EL_PATTERN.matcher((String) value);
            return matcher.matches();
        }
        return false;
    }
}
```

## 10.4 Visualizations

### 10.4.1 State of the art tool STAN

Below is visualization of a part (which fits a sheet) of InSoAr at package-level with a state of the art tool STAN:

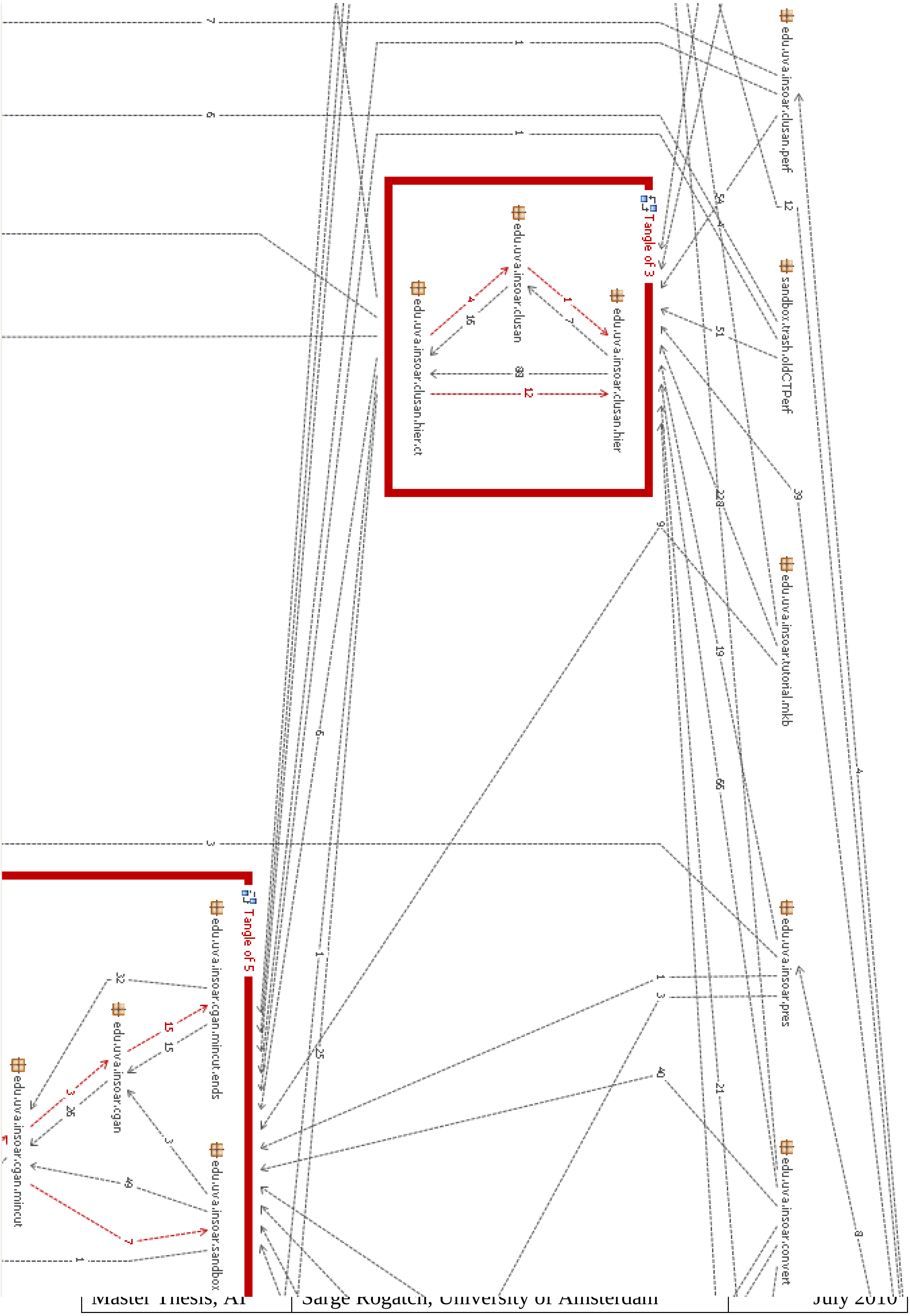

### 10.4.2 Cfinder (Clique Percolation Method)

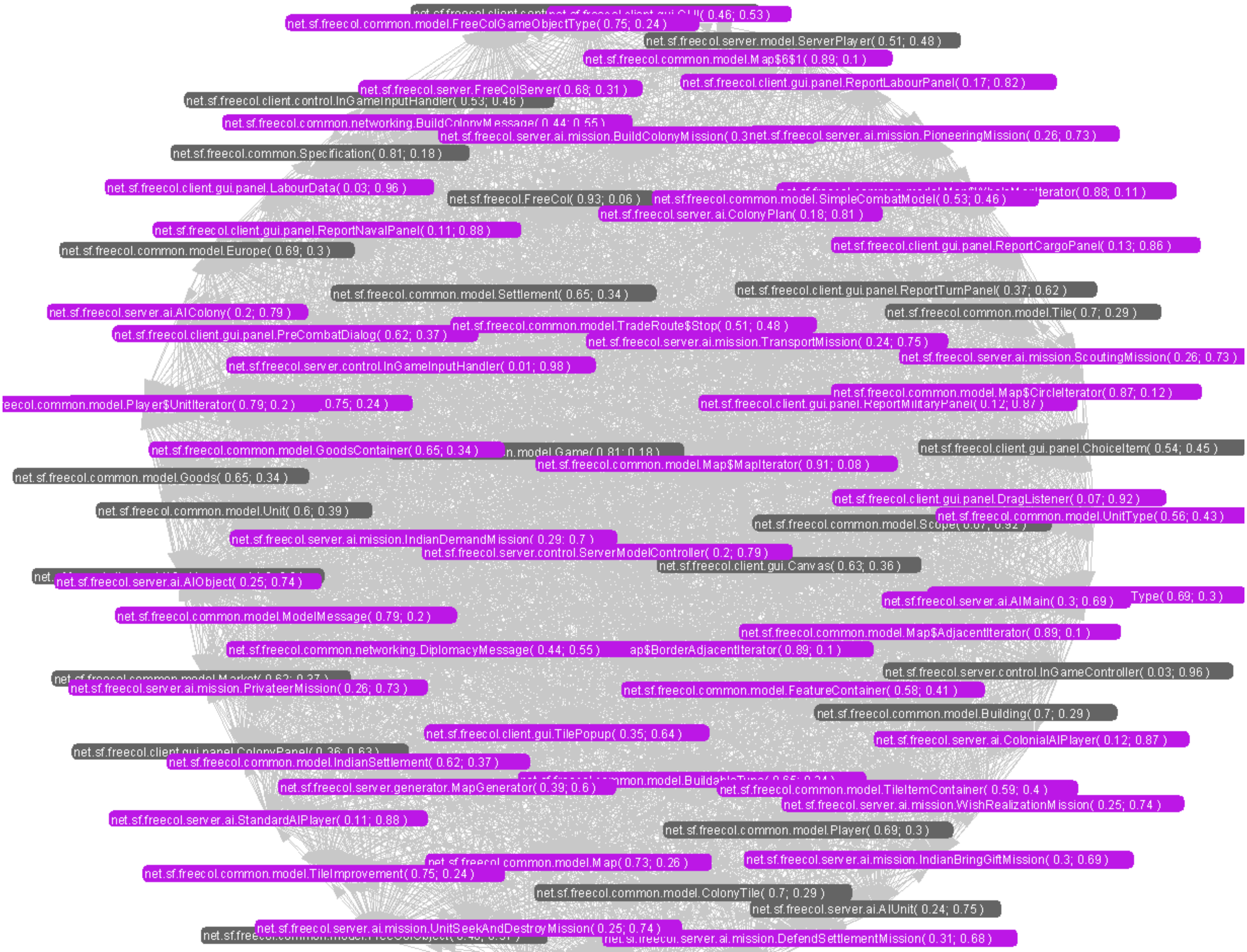

**Figure 10-8 Cliques found by Cfinder**

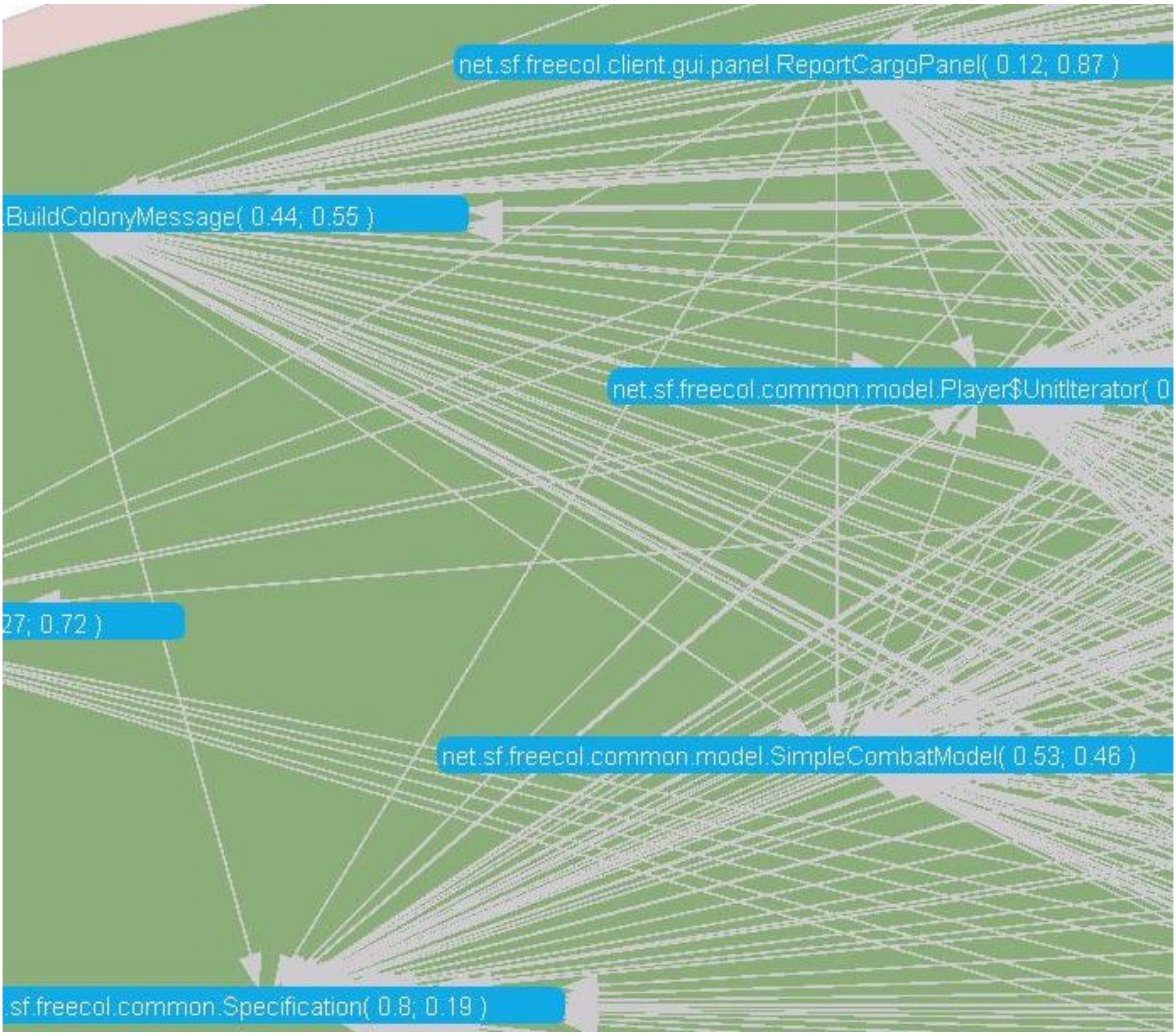

**Figure 10-9 Zoomed in visualization of Cfinder results**

### 10.4.3 Cluster Tree in Text

#### 10.4.3.1 Indentation by Height

```
 8392 2 0 0.40439434814453123000 ; 1 heads 749
 Level 219 cluster under 9587, 9611 ...
749 0 0 VeryBig ; 0 heads
com.kpmg.kpo.web.binding.MaintainAuditTrailsCommand
781 0 0 VeryBig ; 0 heads
com.kpmg.kpo.web.validation.MaintainAuditTrailsCommandValidator
2661 0 0 VeryBig ; 0 heads
java.util.Calendar
       9579 2 0 -0.15834708966782382000 ; 1 heads 8568
       Level 218 cluster under 9611, 9612 ...
      9578 4 0 -0.15834708966782382000 ; 1 heads 8514
      Level 219 cluster under 9579, 9611 ...
     9577 3 0 -0.15834708966782382000 ; 1 heads 8774
     Level 220 cluster under 9578, 9579 ...
    9576 2 0 -0.15834708966782382000 ; 1 heads 7046
    Level 221 cluster under 9577, 9578 ...
   9575 2 0 -0.15834708966782382000 ; 1 heads 8395
   Level 222 cluster under 9576, 9577 ...
  8395 2 0 0.49912031860351563000 ; 1 heads 731
  Level 223 cluster under 9575, 9576 ...
 8106 2 0 1.00000001890586930000 ; 1 heads 5001
 Level 224 cluster under 8395, 9575 ...
5000 0 0 VeryBig ; 0 heads
net.sf.json.JSON
5001 0 0 VeryBig ; 0 heads
net.sf.json.JSONSerializer
731 0 0 VeryBig ; 0 heads
com.kpmg.kpo.web.ajax.AjaxView
3647 0 0 VeryBig ; 0 heads
javax.servlet.http.HttpServletResponse
 7046 2 0 4.87506953125000000000 ; 1 heads 732
 Level 222 cluster under 9576, 9577 ...
732 0 0 VeryBig ; 0 heads
com.kpmg.kpo.web.ajax.CustomerController
735 0 0 VeryBig ; 0 heads
com.kpmg.kpo.web.ajax.beans.CustomerBean
  8774 2 0 0.24277426452636716000 ; 1 heads 8140
  Level 221 cluster under 9577, 9578 ...
 8140 2 0 1.00000001890586930000 ; 1 heads 5129
 Level 222 cluster under 8774, 9577 ...
5129 0 0 VeryBig ; 0 heads
```

### 10.4.3.2 Bracketed presentation

```
edit ctBracketed.txt - Far
D:\...ws2\Insoar\data\ctAnal\ctBracketed.txt          Win    Line    8305/9413    Col 19
h37<9313>
 171 1719 "java.beans.PropertyVetoException"
h38<9386> p8691 <8275>h5 <8805>h4 <8879>h3 <7474>h2 <8655>h1
 175  387 "com.kpmg.kpo.generated.jaxws.crm.ObjectFactory"
 175  389 "com.kpmg.kpo.generated.jaxws.crm.PermissionsType"
 175  396 "com.kpmg.kpo.generated.jaxws.crm.SimpleClientType"
h1<8655>
 174  371 "com.kpmg.kpo.generated.jaxws.crm.ApplicationRefererType"
 174  372 "com.kpmg.kpo.generated.jaxws.crm.AuthenticationType"
 174  376 "com.kpmg.kpo.generated.jaxws.crm.ClientsType"
 174  378 "com.kpmg.kpo.generated.jaxws.crm.CrmSOAPPortType"
 174  379 "com.kpmg.kpo.generated.jaxws.crm.DeleteClientsRequest"
 174  380 "com.kpmg.kpo.generated.jaxws.crm.DeleteClientsResponse"
 174  382 "com.kpmg.kpo.generated.jaxws.crm.GetClientsRequest"
 174  383 "com.kpmg.kpo.generated.jaxws.crm.GetClientsResponse"
 174  384 "com.kpmg.kpo.generated.jaxws.crm.GetClientsWithSignersRequest"
 174  385 "com.kpmg.kpo.generated.jaxws.crm.GetClientsWithSignersResponse"
 174  390 "com.kpmg.kpo.generated.jaxws.crm.SendClientsRequest"
 174  391 "com.kpmg.kpo.generated.jaxws.crm.SendClientsResponse"
 174  392 "com.kpmg.kpo.generated.jaxws.crm.SetClientPermissionsRequest"
 174  393 "com.kpmg.kpo.generated.jaxws.crm.SetClientPermissionsResponse"
 174  397 "com.kpmg.kpo.generated.jaxws.crm.SimpleClientsType"
h2<7474> p8879 <7475>h1
 174  369 "com.kpmg.kpo.generated.jaxws.crm.AccessRightType"
 174  388 "com.kpmg.kpo.generated.jaxws.crm.PermissionType"
h1<7475> h3<8879> p8805 <9016>h2 <7467>h1
 174  374 "com.kpmg.kpo.generated.jaxws.crm.ClientDetailsType"
 174  386 "com.kpmg.kpo.generated.jaxws.crm.KvkSizeType"
h1<7467>
 173  370 "com.kpmg.kpo.generated.jaxws.crm.AddressType"
h2<9016> p8805 <7466>h1
 173  373 "com.kpmg.kpo.generated.jaxws.crm.ClientConfirmationMethodType"
 173  375 "com.kpmg.kpo.generated.jaxws.crm.ClientType"
h1<7466>
 172  395 "com.kpmg.kpo.generated.jaxws.crm.SignersType"
h4<8805> p8275 <7473>h1
 172  381 "com.kpmg.kpo.generated.jaxws.crm.GenderType"
 172  394 "com.kpmg.kpo.generated.jaxws.crm.SignerType"
h1<7473> h5<8275> h39<8691> p9321 <8534>h4 <8969>h3 <8843>h2 <7582>h1
 173 3321 "javax.imageio.spi.FilterIterator"
 173 3335 "javax.imageio.spi.ServiceRegistry$Filter"
```

**Figure 10-10 Bracketed presentation of clustering hierarchy**

### 10.4.4 Sunray Representation

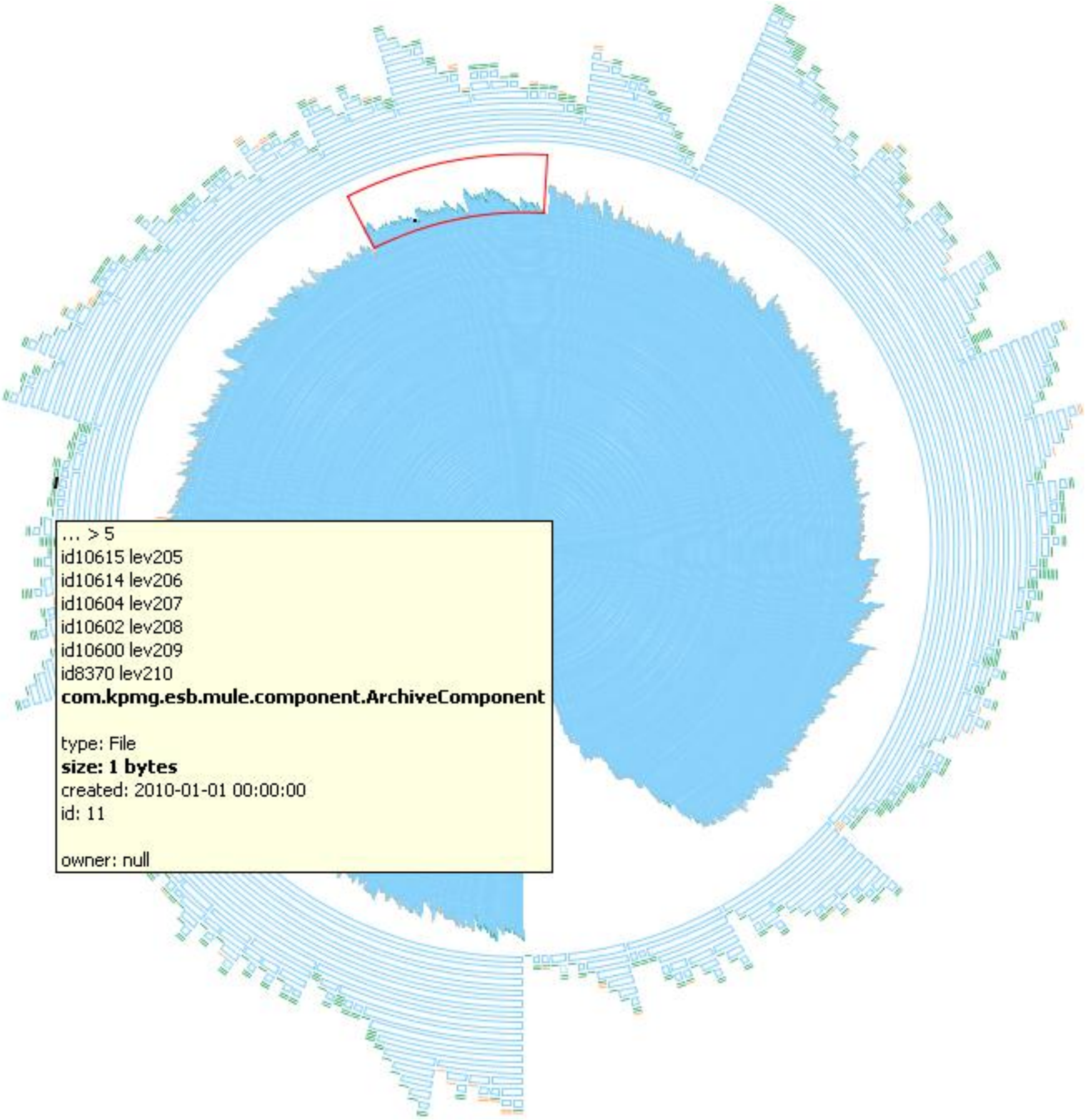

### 10.4.5 Sunburst

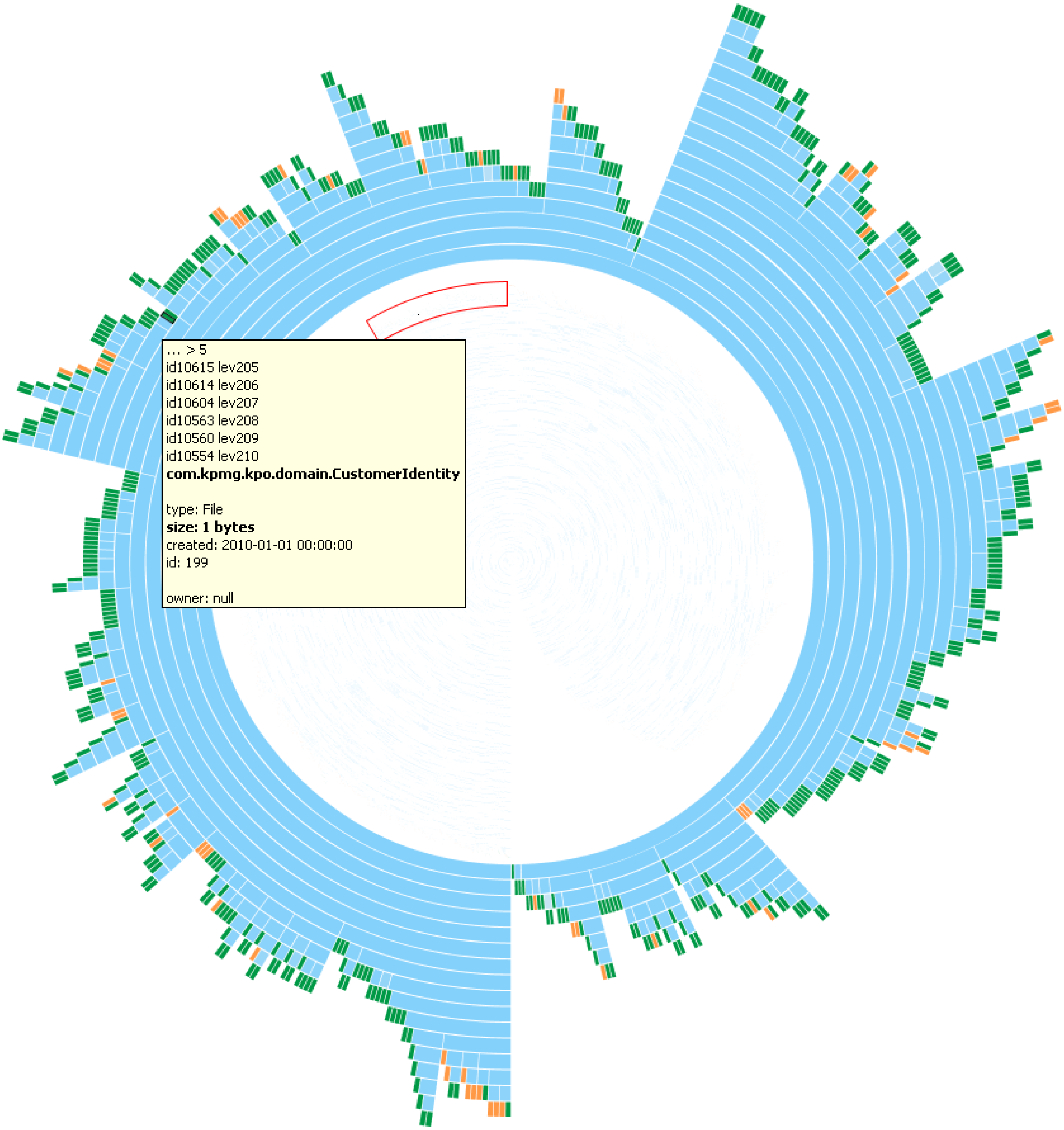

### 10.4.6 Hyperbolic tree (plane)

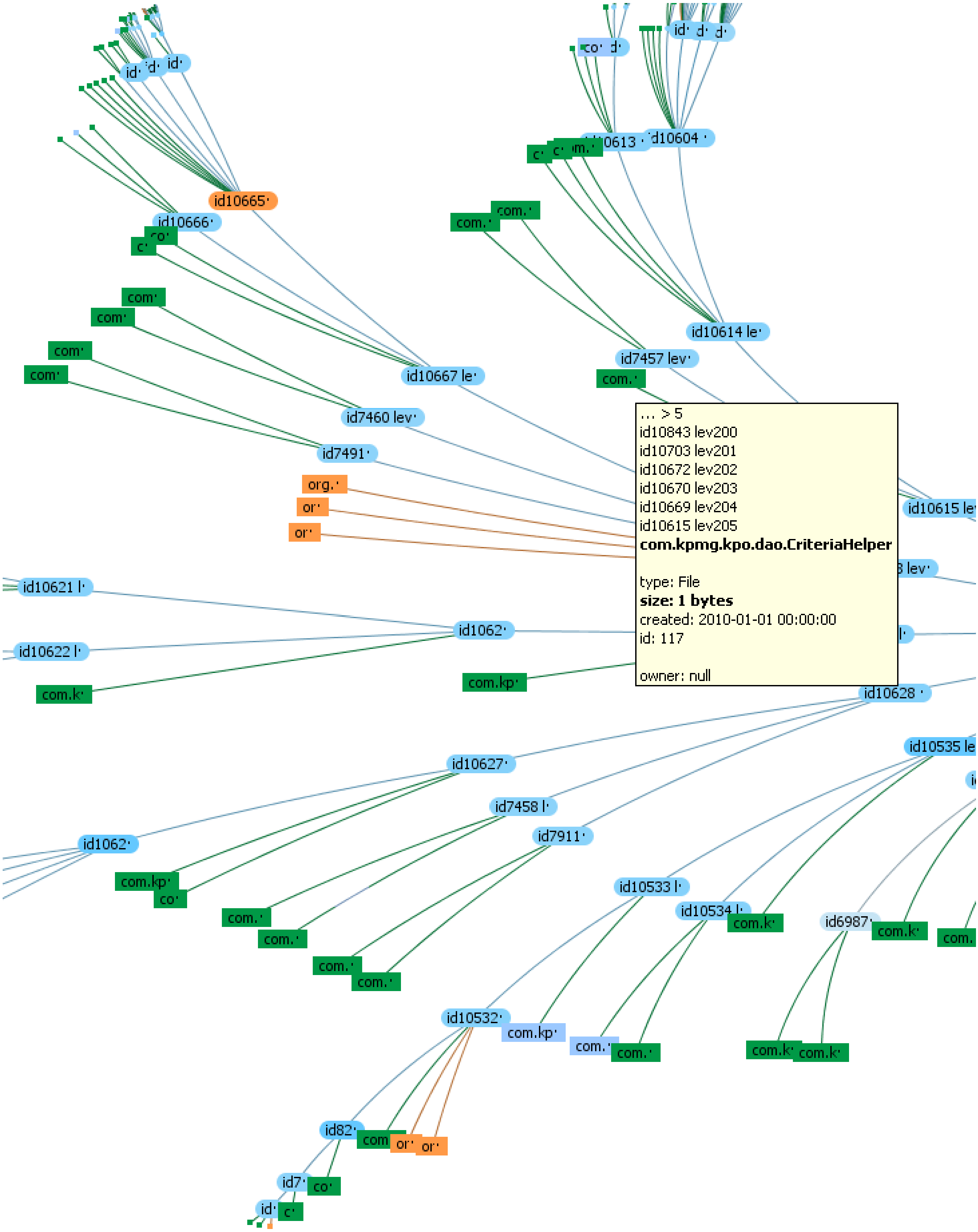

### 10.4.7 Circular TreeMap

#### 10.4.7.1 View on the whole program

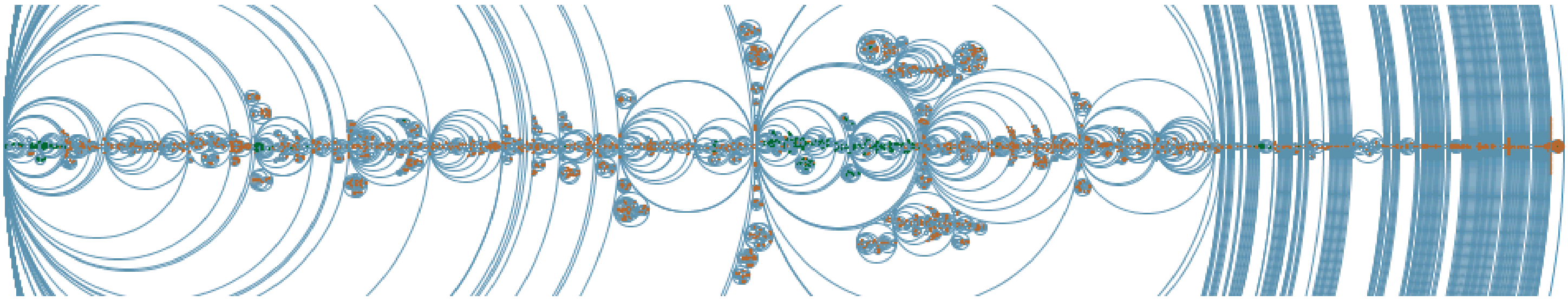

#### 10.4.7.2 Parts of the architecture

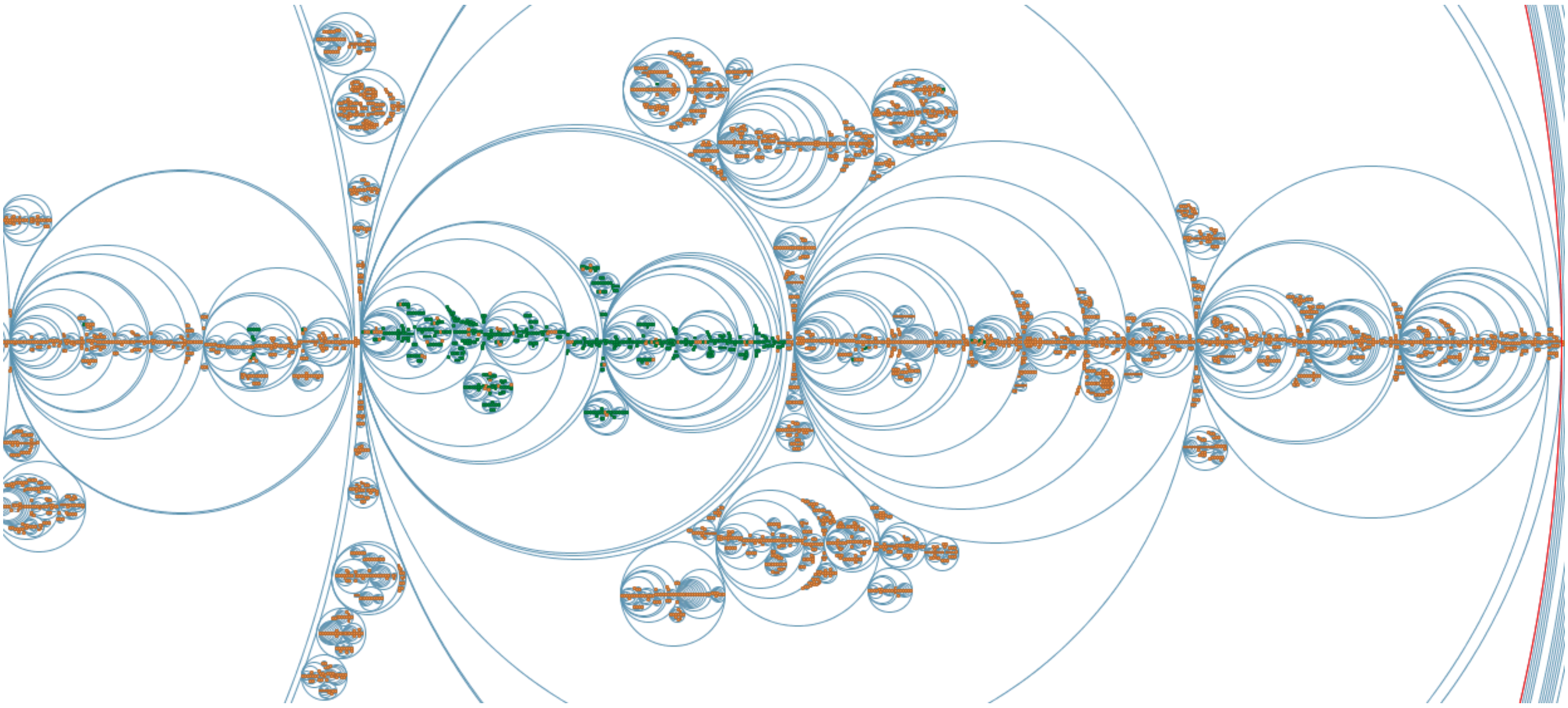

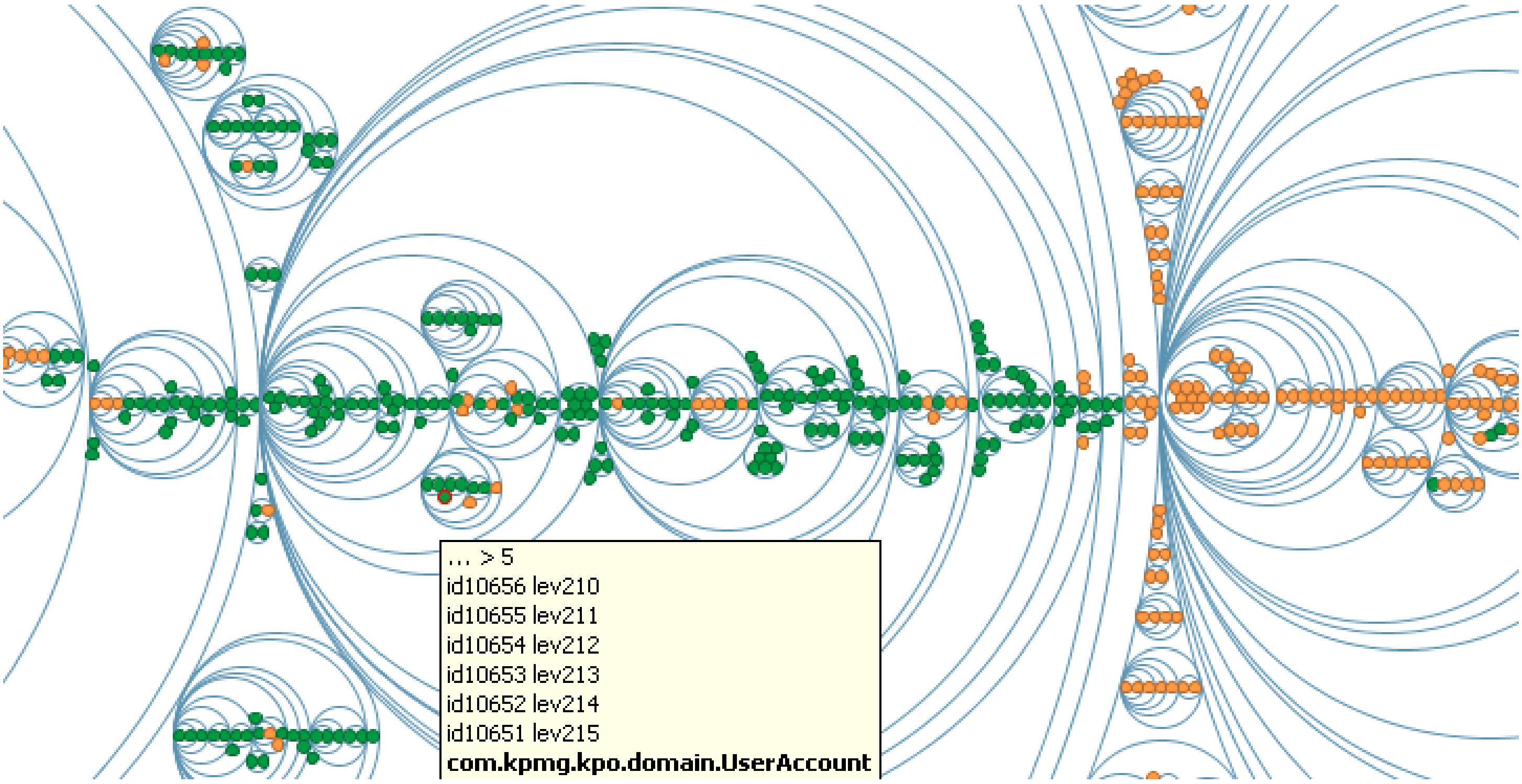

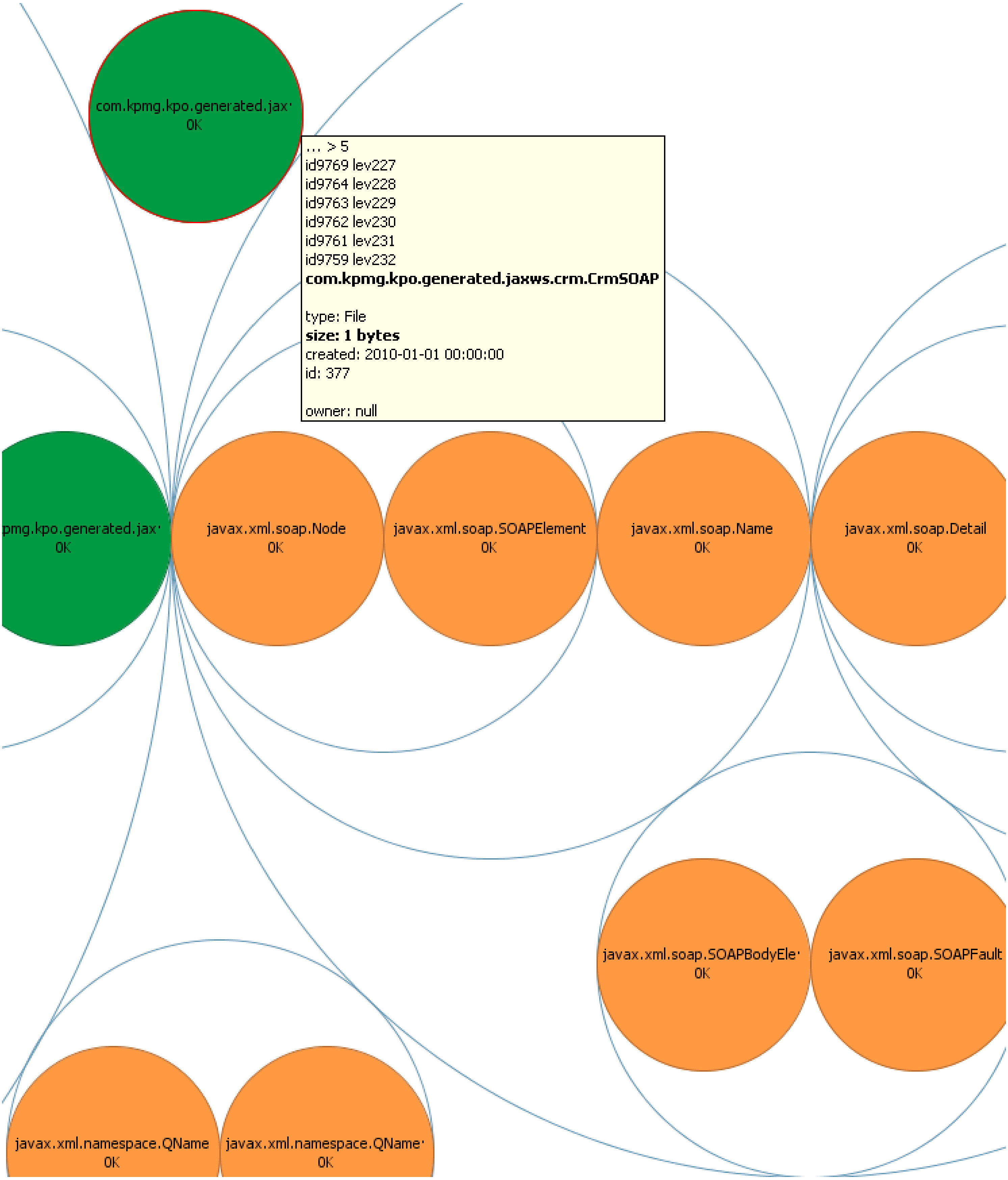

### 10.4.8 H3 Sphere Layout

## 10.4.8.1 Near class CrmSOAP

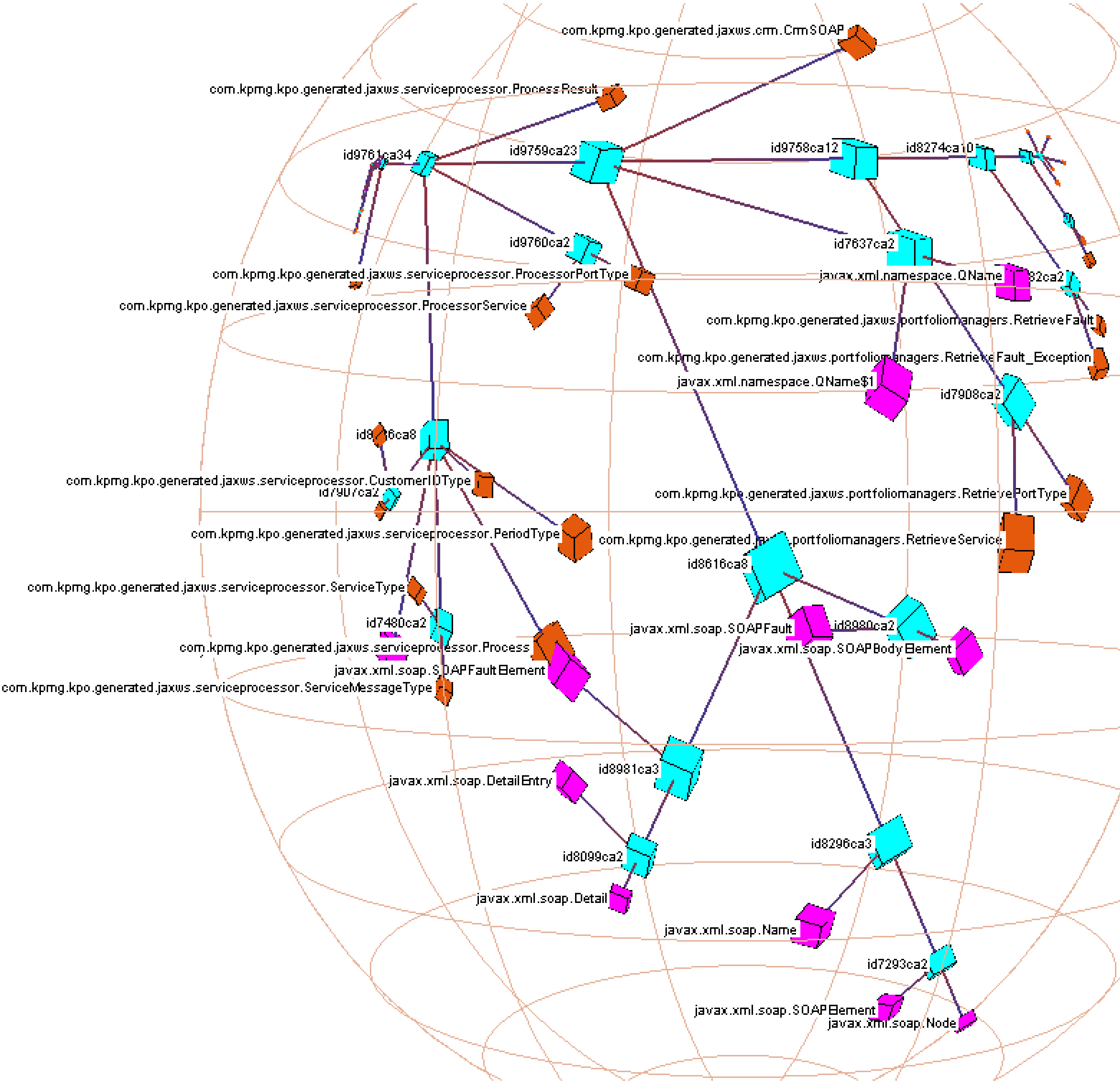

### 10.4.8.2 Classes that act together

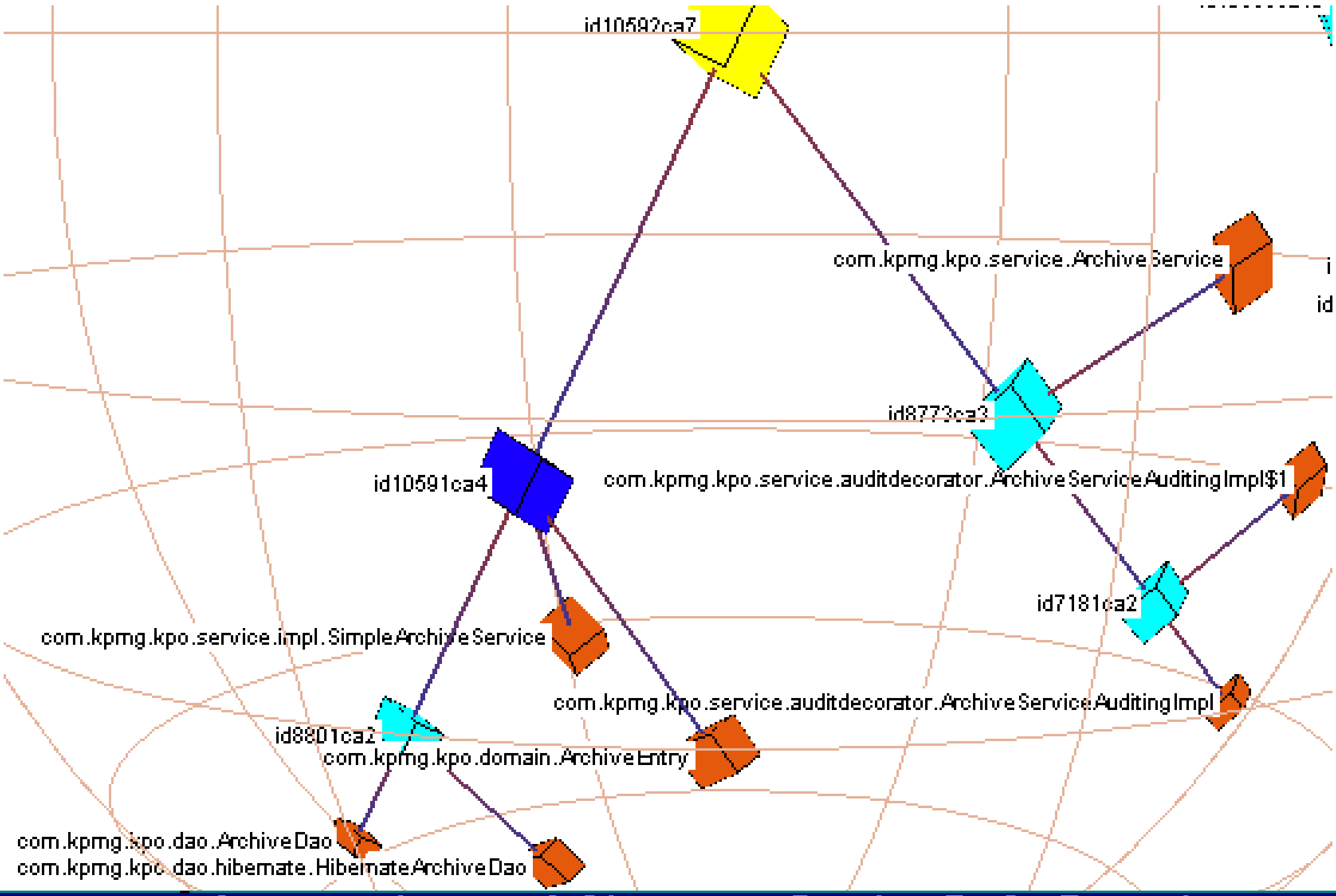

```
          com.kpmg.service.portfoliomanager.RetrieveFault_Exception
        796 0 0 VeryBig ; 0 heads
        com.kpmg.service.portfoliomanager.PortfolioManagersType
        802 0 0 VeryBig ; 0 heads
        com.kpmg.service.portfoliomanager.RetrieveResult
      793 0 0 VeryBig ; 0 heads
      com.kpmg.service.portfoliomanager.CustomerIDType
      794 0 0 VeryBig ; 0 heads
      com.kpmg.service.portfoliomanager.ObjectFactory
      800 0 0 VeryBig ; 0 heads
      com.kpmg.service.portfoliomanager.RetrievePortfolioManagers
      801 0 0 VeryBig ; 0 heads
      com.kpmg.service.portfoliomanager.RetrieveResponse
    799 0 0 VeryBig ; 0 heads
    com.kpmg.service.portfoliomanager.RetrievePortType
  9767 2 0 -0.15834708966782382000 ; 1 heads 580
  Level 229 cluster under 9768, 9769, 9770, 9771, 9772, 9775, 9778, 9779, 9781
    9766 4 0 -0.15834708966782382000 ; 1 heads 666
    Level 230 cluster under 9767, 9768, 9769, 9770, 9771, 9772, 9775, 9778, 97
      8698 2 0 0.18955194091796876000 ; 1 heads 122
      Level 231 cluster under 9766, 9767, 9768, 9769, 9770, 9771, 9772, 9775,
        122 0 0 VeryBig ; 0 heads
        com.kpmg.kpo.dao.EmployeeAuthorisationDao
        162 0 0 VeryBig ; 0 heads
        com.kpmg.kpo.dao.hibernate.HibernateEmployeeAuthorisationDao
      9765 2 0 -0.15834708966782382000 ; 1 heads 313
      Level 231 cluster under 9766, 9767, 9768, 9769, 9770, 9771, 9772, 9775,
        312 0 0 VeryBig ; 0 heads
        com.kpmg.kpo.exception.portfoliomanagers.GetPortfolioManagersException
        313 0 0 VeryBig ; 0 heads
        com.kpmg.kpo.exception.portfoliomanagers.UnknownCustomerException
      314 0 0 VeryBig ; 0 heads
      com.kpmg.kpo.exception.portfoliomanagers.UnknownSystemException
      666 0 0 VeryBig ; 0 heads
      com.kpmg.kpo.service.impl.EmployeeAuthorisationServiceImpl
    580 0 0 VeryBig ; 0 heads
    com.kpmg.kpo.service.EmployeeAuthorisationService
  803 0 0 VeryBig ; 0 heads
  com.kpmg.service.portfoliomanager.RetrieveService_
8521 2 0 0.29990281372070315000 ; 1 heads 7652
Level 228 cluster under 9769, 9770, 9771, 9772, 9775, 9778, 9779, 9781, 9782,
```

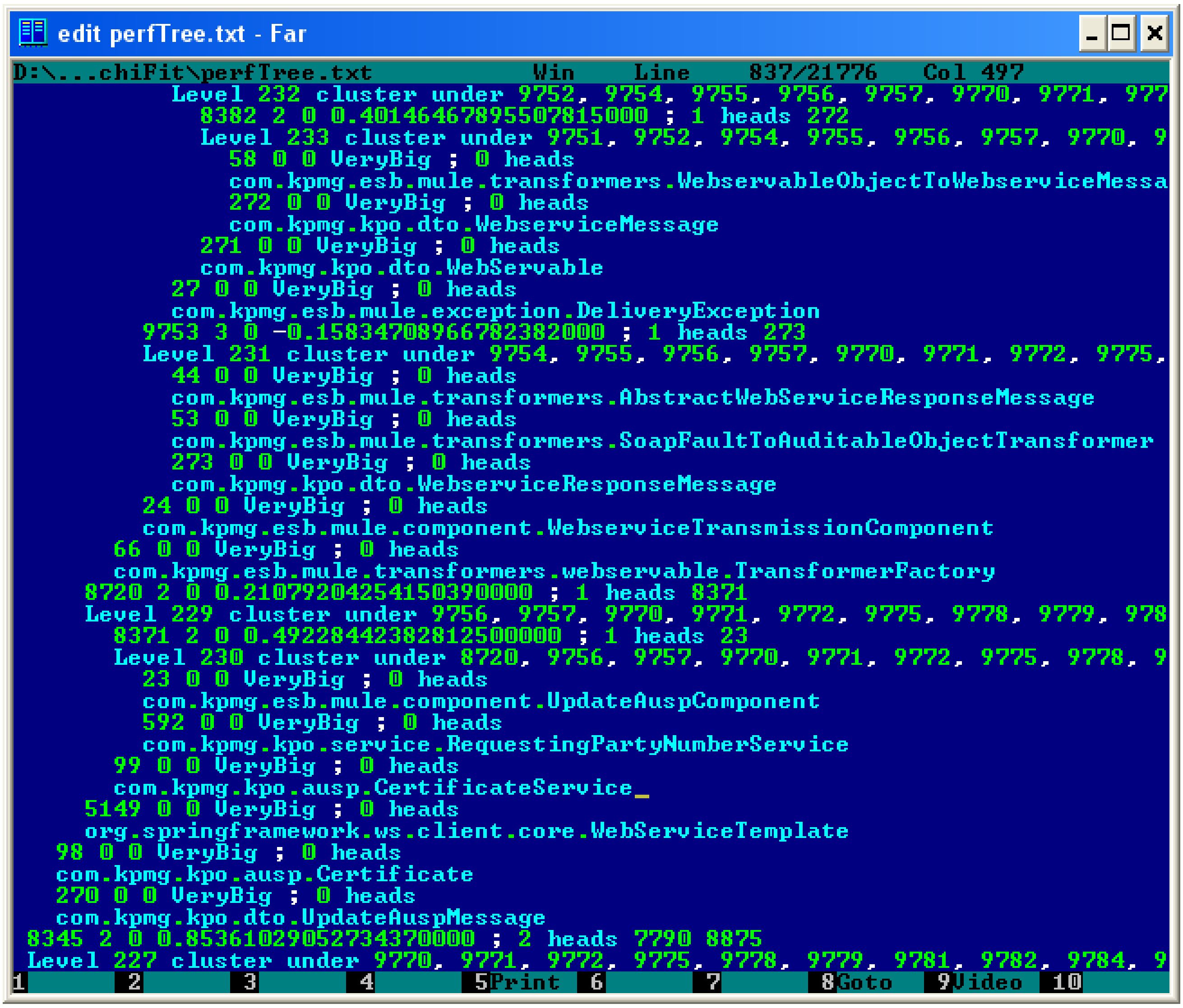

```
edit perfTree.txt - Far
D:\...chiFit\perfTree.txt          Win     Line     837/21776    Col 497
           Level 232 cluster under 9752, 9754, 9755, 9756, 9757, 9770, 9771, 977
             8382 2 0 0.4014646789550781500 ; 1 heads 272
             Level 233 cluster under 9751, 9752, 9754, 9755, 9756, 9757, 9770, 9
               58 0 0 VeryBig ; 0 heads
               com.kpmg.esb.mule.transformers.WebservableObjectToWebserviceMessa
               272 0 0 VeryBig ; 0 heads
               com.kpmg.kpo.dto.WebserviceMessage
             271 0 0 VeryBig ; 0 heads
             com.kpmg.kpo.dto.WebServable
           27 0 0 VeryBig ; 0 heads
           com.kpmg.esb.mule.exception.DeliveryException
         9753 3 0 -0.1583470896678238200 ; 1 heads 273
         Level 231 cluster under 9754, 9755, 9756, 9757, 9770, 9771, 9772, 9775,
           44 0 0 VeryBig ; 0 heads
           com.kpmg.esb.mule.transformers.AbstractWebServiceResponseMessage
           53 0 0 VeryBig ; 0 heads
           com.kpmg.esb.mule.transformers.SoapFaultToAuditableObjectTransformer
           273 0 0 VeryBig ; 0 heads
           com.kpmg.kpo.dto.WebserviceResponseMessage
         24 0 0 VeryBig ; 0 heads
         com.kpmg.esb.mule.component.WebserviceTransmissionComponent
       66 0 0 VeryBig ; 0 heads
       com.kpmg.esb.mule.transformers.webservable.TransformerFactory
     8720 2 0 0.2107920425415039000 ; 1 heads 8371
     Level 229 cluster under 9756, 9757, 9770, 9771, 9772, 9775, 9778, 9779, 978
       8371 2 0 0.4922844238281250000 ; 1 heads 23
       Level 230 cluster under 8720, 9756, 9757, 9770, 9771, 9772, 9775, 9778, 9
         23 0 0 VeryBig ; 0 heads
         com.kpmg.esb.mule.component.UpdateAuspComponent
         592 0 0 VeryBig ; 0 heads
         com.kpmg.kpo.service.RequestingPartyNumberService
       99 0 0 VeryBig ; 0 heads
       com.kpmg.kpo.ausp.CertificateService_
     5149 0 0 VeryBig ; 0 heads
     org.springframework.ws.client.core.WebServiceTemplate
   98 0 0 VeryBig ; 0 heads
   com.kpmg.kpo.ausp.Certificate
   270 0 0 VeryBig ; 0 heads
   com.kpmg.kpo.dto.UpdateAuspMessage
 8345 2 0 0.8536102905273437000 ; 2 heads 7790 8875
 Level 227 cluster under 9770, 9771, 9772, 9775, 9778, 9779, 9781, 9782, 9784, 9
1      2      3      4      5Print 6      7      8Goto  9Video 10
```

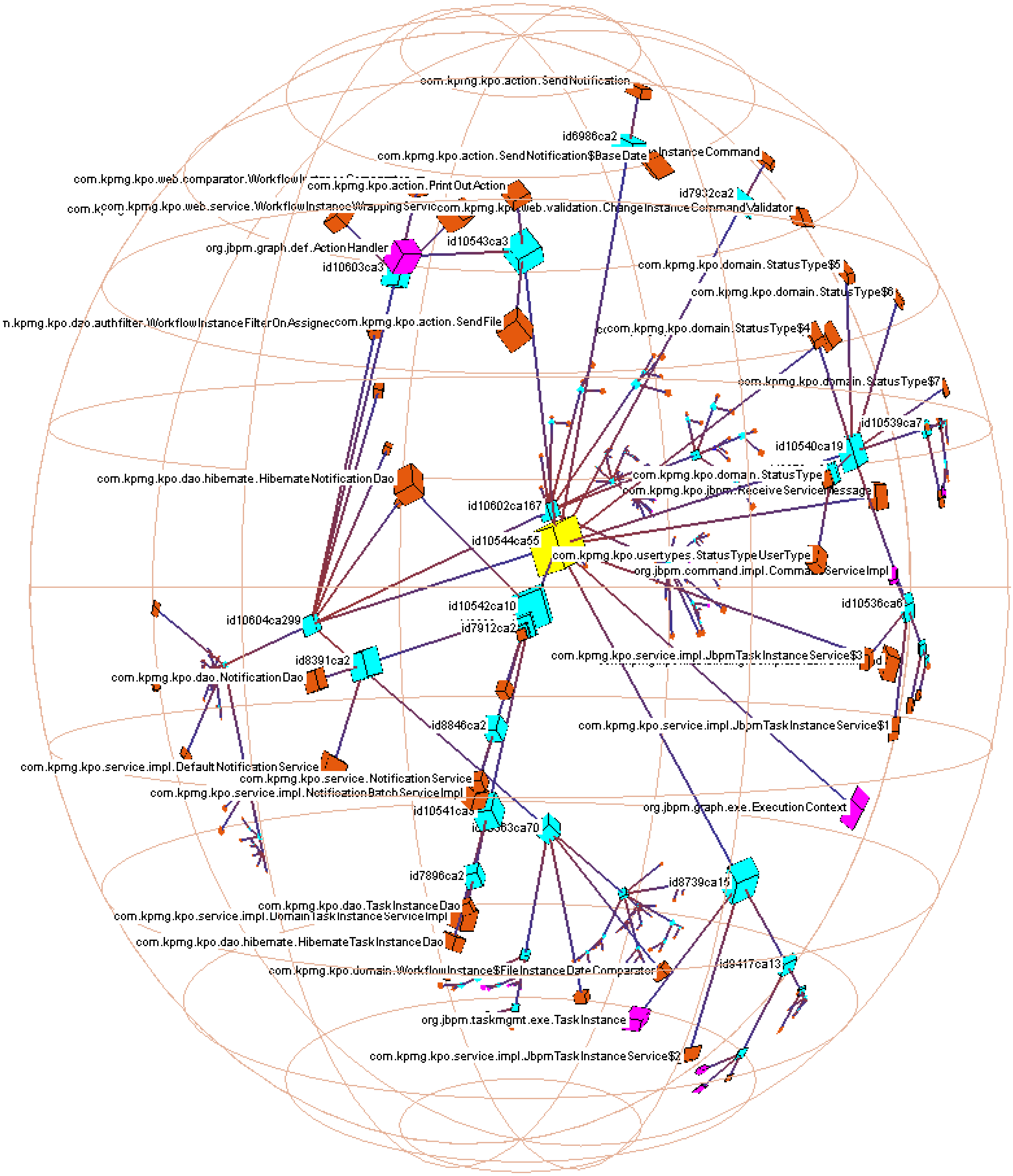
com.kpmg.kpo.action.SendNotification
id6986ca2
com.kpmg.kpo.action.SendNotification$BaseDate
com.kpmg.kpo.web.comparator.WorkflowI
com.kpmg.kpo.action.PrintOutAction
com.kpmg.kpo.web.service.WorkflowInstanceWrappingServic
com.kpmg.kpo.web.validation.ChangeInstanceCommandValidator
id7932ca2
org.jbpm.graph.def.ActionHandler
id10543ca3
id10603ca3
com.kpmg.kpo.domain.StatusType$5
com.kpmg.kpo.domain.StatusType$6
com.kpmg.kpo.domain.StatusType$4
com.kpmg.kpo.action.SendFile
com.kpmg.kpo.domain.StatusType$7
id10539ca7
id10540ca19
com.kpmg.kpo.dao.hibernate.HibernateNotificationDao
com.kpmg.kpo.domain.StatusType
id10602ca167
id10544ca55
com.kpmg.kpo.usertypes.StatusTypeUserType
org.jbpm.command.impl.CommandServiceImpl
id10536ca6
id10542ca10
id10604ca299
id7912ca2
com.kpmg.kpo.service.impl.JbpmTaskInstanceService$3
id8391ca2
com.kpmg.kpo.dao.NotificationDao
id8846ca2
com.kpmg.kpo.service.impl.JbpmTaskInstanceService$1
com.kpmg.kpo.service.impl.DefaultNotificationService
com.kpmg.kpo.service.NotificationService
com.kpmg.kpo.service.impl.NotificationBatchServiceImpl
org.jbpm.graph.exe.ExecutionContext
id7896ca2
id8739ca15
com.kpmg.kpo.dao.TaskInstanceDao
com.kpmg.kpo.service.impl.DomainTaskInstanceServiceImpl
com.kpmg.kpo.dao.hibernate.HibernateTaskInstanceDao
id9417ca13
com.kpmg.kpo.domain.WorkflowInstance$FileInstanceDateComparator
org.jbpm.taskmgmt.exe.TaskInstance
com.kpmg.kpo.service.impl.JbpmTaskInstanceService$2

# 11 References

[GoHu1961] Gomory, Hu (1961). Multi-terminal network flows. *In J. of the Society for Industrial and Applied Mathematics, Vol. 9, No. 4 (Dec, 1961), pp. 551-570, 1961*

[MaQu1967] MacQueen (1967). Some methods for classification and analysis of multivariate observations. *In Proceedings of the Fifth Berkeley Symposium on Mathematical Statistics and Probability, 1, pp. 281-297, 1967*

[Dini1970] Dinic (1970). Algorithm for Solution of a Problem of Maximum Flow in Networks with Power Estimation. *In Soviet Math. Doklady, 11: pp.1277-1280, 1970*

[AHU1983] Aho, Hopcroft, Ullman (1983). Data Structures and Algorithms. *Addison-Wesley Longman Publishing Co, 427 pages, 1983*

[Mull1990] Muller, Uhl (1990). Composing Subsystem Structures using (k, 2)-partite Graphs. *In Proc. Of the International Conference on Software Maintenance, pp.12-19, 1990*

[Gus1990] Gusfield (1990). Very simple methods for all pairs network flow analysis. *In SIAM J. Comput. 19 , pp. 143-155, 1990*

[Kau1990] Kaufman, Rousseeuw (1990). Finding groups in data. An introduction to cluster analysis. *In Wiley Series in Probability and Mathematical Statistics. Applied Probability and Statistics, New York: Wiley, 1990*

[Hag1992] Hagen, Kahng (1992). New spectral methods for ratio cut partitioning and clustering. *In IEEE Trans. On CAD, 11, pp.1074-1085, 1992*

[Gar1993] Garlan, Shaw (1993). An Introduction to Software Architecture http://www.cs.cmu.edu/afs/cs/project/vit/ftp/pdf/intro_softarch.pdf

[Bac1996] Bacon, Sweeney (1996). Fast static analysis of C++ virtual function calls. *In Proceedings of OOPSLA, pp.324-341, 1996*

[Diw1996] Diwan, Moss, McKinley (1996). Simple and effective analysis of statically-typed object-oriented programs. *In Proceedings of OOPSLA, pp. 292-305, 1996*

[Che1997] Cherkassky, Goldberg (1997). On implementing push-relabel method for the maximum flow problem. *In Algorithmica, Vol.19, No 4, pp.390-410, Springer, 1997*

[Brin1998] Brin, Page (1998). Anatomy of a Large-Scale Hypertextual Web Search Engine. *In Proc. 7$^{th}$ International World Wide Web Conference, 1998*

[Gol1998] Goldberg, Rao (1998). Beyond the flow decomposition barrier. *In Journal of the ACM, Vol 45-5, p.783-797, 1998*

[Knu1998] Knuth (1998). The Art of Computer Programming, .. *ISBN:0201485419, Vol. 1-3, 2$^{nd}$ edition, Addison-Wesley, 1998*

[Man1998] Mancoridis, Mitchell, Rorres, Chen, Gansner (1998). Using automatic clustering to produce high-level system organizations of source code. *In Proc. Of IEEE 6$^{th}$ International Workshop on Program Comprehension, 1998*

[Bunch1999] Mancoridis, Mitchell, Chen, Gansner (1999). Bunch: A Clustering Tool for the Recovery and Maintenance of Software System Structures. *In Proc. Of the International Conference on Software Maintenance, pp. 50-62, 1999*

[HITS1999] Kleinberg (1999). Authoritative sources in a hyperlinked environment. *J. of the ACM, Vol. 46, No. 5, pp.604-632, 1999*

[GoTs2001] Goldberg, Tsioutsiouliklis (2001). Cut Tree Algorithms: An Experimental Study. *In J. of Algorithms 38, pp. 51-83, 2001*

[Jain1999] Jain, Murty, Flynn (1999). Data Clustering: A Review. *In ACM Computing Surveys, 31, No.3,pp. 264-323, 1999*

[Kri1999] Krikhaar (1999). Software Architecture Reconstruction. *PhD thesis. University of Amsterdam, 1999*

[Man1999] Manning, Schutze (1999). Foundations of Statistical Natural Language Processing. *The MIT Press, 1999*

[MoJo1999] Tzerpos, Holt (1999). MoJo: A distance metric for software clusterings. *In Proc. Of the 6$^{th}$ Working Conference on Reverse Engineering, pp. 187-193, 1999*

[Soot1999] Vallée-Rai, Hendren, Sundaresan, Lam, Gagnon, Co (1999). Soot – a Java Bytecode Optimization Framework. *In Proceedings of IBM Centre for Advanced Studies Conference, CASCON 1999*

[Sun1999] Sundaresan (1999). Practical techniques for virtual call resolution in Java. *Master's thesis, McGill University, 1999*

[Kwo2000] Kwok (2000). Notes on VTA implementation is Soot. *Technical Note, Sable Research Group, McGill University, 2000*

[Rays2000] Rayside, Reuss, Hedges, Kontogiannis (2000). The Effect of Call Graph Construction Algorithms for Object-Oriented Programs on Automatic Clustering. *In Proc. Of the 8th International Workshop on Program Comprehension, IEEE, 2000*

[Riv2000] Riva (2000). Reverse Architecting: an Industrial Experience Report. *In Proceedings of the 7th Working Conference on Reverse Engineering (WCRE2000), 2000*

[Shi2000] Shi, Malik (2000). Normalized cuts and image segmentation. *In IEEE Transactions on Pattern Analysis and Machine Learning, 22, No.8, pp.888-905, 2000*

[ACDC2000] Tzerpos, Holt (2000). "ACDC: An algorithm for comprehension-driven clustering", *In Proc. Of the 7th Working Conference on Reverse Engineering, pp. 258-267, 2000*

[Ding2001] Ding, Zha, Gu, Simon (2001). A Min-max Cut Algorithm for Graph Partitioning and Data Clustering. *In Proc. of International Conference on Data Mining, IEEE, pp. 107-114, 2001*

[Gine2002] Girvan, Newman (2002). Community structure in social and biological networks. *In Proceedings of the National Academy of Sciences, 99, 7821-7826, 2002*

[Bass2003] Bass, Clements, Kazman (2003). Software Architecture In Practice. *Second Edition. Boston: Addison-Wesley,pp. 21-24, 2003*

[CLR2003] Cormen, Leiserson, Rivest (2003). Introduction to Algorithms. *MIT Press / McGraw Hill, 4th printing, 2003*

[Let2003] Lethbridge, Singer, Forward (2003). How software engineers use documentation: the state of the practice. In *IEEE Software special issue: the State of the Practice of Software Engineering, Vol. 20, No. 6, pp. 35-39, 2003*

[Lho2003] Lhotak, Hendren (2003). Scaling Java points-to analysis using Spark. *In Proc. Of the 12th International Conference on Compiler Construction, pp.153-169, 2003*

[END2004] Shtern, Tzerpos (2004). A framework for the Comparison of Nested Software Decompositions. *In Proc. Of the 11th Working Conference on Reverse Engineering, pp.284-292, 2004*

[Fla2004] Flake, Tarjan, Tsioutsiouliklis (2004). Graph Clustering and Minimum Cut Trees. *In Internet Mathematics, Vol. 1, No. 4, pp. 385-408, 2004*

[HaLe2004] Hamou-Lhadj, Lethbridge (2004). Reasoning about the Concept of Utilities. *In Proc. Of the ECOOP International Workshop on Practical Problems of Programming in the Large, LNCS, Vol. 3344, pp.10-22, 2004*

[Pal2005] Palla, Derényi, Farkas, Vicsek. (2005) Uncovering the overlapping community structure of complex networks in nature and society. *In Nature 435, 814.* http://www.cfinder.org/

[Wen2005] Wen, Tzerpos (2005). Software Clustering based on Omnipresent Object Detection. *In Proc of the 13th International Workshop on Program Comprehension, pp. 269-278, 2005*

[Bab2006] Babenko, Goldberg (2006). Experimental Evaluation of a Parametric Flow Algorithm. *Technical Report, Microsoft Research, 2006*

[Bie2006] De Bie, Cristianini (2006). Fast SDP Relaxations of Graph Cut Clustering, Transduction, and Other Combinatorial Problems. *In Journal of Machine Learning Research 7, pp. 1409-1436, 2006*

[Dong2006] Dong, Zhuang, Chen, Tai (2006). A hierarchical clustering algorithm based on fuzzy graph connectedness. *In Fuzzy Sets and Systems, Vol. 157, No. 13, pp. 1760-1774, Elsevier, 2006*

[HaLe2006] Hamou-Lhadj, Lethbridge (2006). Summarizing the Content of Large Traces to Facilitate the Understanding of the Behavior of a Software System. *In Proc. Of the 14th IEEE International Conference on Program Comprehension (ICPC`06), pp. 181-190, 2006*

[Andre2007] Andreopolos, An, Tzerpos, Wang (2007). Clustering large software systems at multiple layers. *In Information and Software Technology 49, No 3, pp.244-254, 2007*

[Kuhn2007] Kuhn, Ducasse, Girba (2007). Semantic Clustering: Identifying Topics in Source Code. *In J. on Information and Software Technology, Vol. 49, No 3, pp. 230-243*

[Bink2007] Binkley (2007). Source Code Analysis: A Road Map. *In Future of Software Engineering, IEEE, 2007*

[Czi2007] Czibula, Serban (2007). Hierarchical Clustering for Software Systems Restructuring. *In INFOCOMP Journal of Computer Science, 6, No. 4, p. 43-51, Brasil, 2007*

[Fre2007] Frey, Dueck (2007). Clustering by Passing Messages Between Data Points. *In Science, Vol 315, No. 5814, pp. 972-976, 2007* http://www.psi.toronto.edu/index.php?q=affinity%20propagation

[Maqb2007] Maqbool, Babri (2007). Hierarchical Clustering for Software Architecture Recovery. *In IEEE Transactions on Software Ingineering, Vol. 33, No. 11, pp. 759-780, 2007* http://shannon2.uwaterloo.ca/~soveisgh/software/hierarchical%20clustering.pdf

[Ra2007] Rattigan, Maier, Jensen (2007). Graph clustering with network structure indices. *In Proc. of 24th International Conference on Machine Learning (ICML '07), Vol. 227, pp. 783-390, 2007*. http://videolectures.net/icml07_rattigan_gcns/

[Rou2007] Roubtsov, Telea, Holten (2007). SQuAVisiT: A Software Quality Assessment and Visualization Toolset. *In Proceedings 7th International Conference on Source Code Analysis and Manipulation, pp.155-156, IEEE, 2007*

[Sch2007] Schaeffer (2007). Graph Clustering. Survey. *Computer Science Review, 1(1):27-64, Elsevier , 2007.*

[UpMJ2007] Shtern, Tzerpos (2007). Lossless Comparison of Nested Software Decompositions. *In WCRE: Proc. Of the 14th Working Conference on Reverse Engineering, pp.249-258, IEEE, 2007*

[Czi2008] Czibula, Czibula (2008). A Partitional Clustering Algorithm for Improving The Structure of Object-Oriented Software Systems. *Univ. Babes-Bolyai, Informatica, Vol LIII-2, 2008*

[EiNi2008] Einarsson, Nielsen (2008). A Survivor's Guide to Java Program Analysis with Soot. *BRICS, Department of Computer Science, University of Aarhus, Denmark, 2008*

[Nara2008] Narayan, Gopinath, Varadarajan (2008). Structure and Interpretation of Computer Programs. *In Proc. Of the 2nd IFIP/IEEE International Symposium on Theoretical Aspects of Software Engineering, pp.73-80, 2008*

[Pin2008] Pinzger, Graefenhain, Knab, Gall (2008). A Tool for Visual Understanding of Source Code Dependencies. *In Proc. Int'l Conf. Program Comprehension, pages 254-259. IEEE Computer Society, 2008*

[Roha2008] Rohatgi, Hamou-Lhadj, Rilling (2008). An Approach for Mapping Features to Code Based on Static and Dynamic Analysis. *In Proc. Of the 16th IEEE International Conference on Program Comprehension (ICPC), pp. 236-241 2008*

[Ser2008] Serban, Czibula (2008). Object-Oriented Software Systems Restructuring through Clustering. *In Proc. Of 9th Intl. Conference on Artificial Intelligence and Soft Computing, ICAISC, pp. 693-704,2008.*

[Kli2009] Klint, Storm, Vinju (2009). Rascal: a Domain Specific Language for Source Code Analysis and Manupulation. *In Proc. Of 9th International Working Conference on Source Code Analysis and Manipulation, IEEE, pp. 168-177, 2009*

[Pate2009] Patel, Hamou-Lhadj, Rilling (2009). Software Clustering Using Dynamic Analysis and Static Dependencies. *In Proc. of the 13th European Conference on Software Maintenance and Reengineering (CSMR'09), Architecture-Centric Maintenance of Large-Scale Software Systems, pp. 27-36, 2009*

[Kosc2009] Koschke (2009). Architecture Reconstruction: Tutorial on Reverse Engineering to the Architectural Level. *In ISSSE 2006-2008, LNCS 5413, pp. 140-173, 2009.*

[Giv2009] Givoni, Frey (2009). A Binary Variable Model for Affinity Propagation. *In Neural Comput., 21, No. 6, pp. 589-600, 2009*

[Pir2009] Pirzadeh, Alawneh, Hamou-Lhadj (2009). Quality of the Source Code for Design and Architecture Recovery Techniques: Utilities are the Problem. *In Ninth International Conference on Quality Software, pp. 465-469, 2009*

[Stan2009] Odysseus Software (2009). Tool STAN - Structure Analysis for Java. White paper, www.stan4j.com

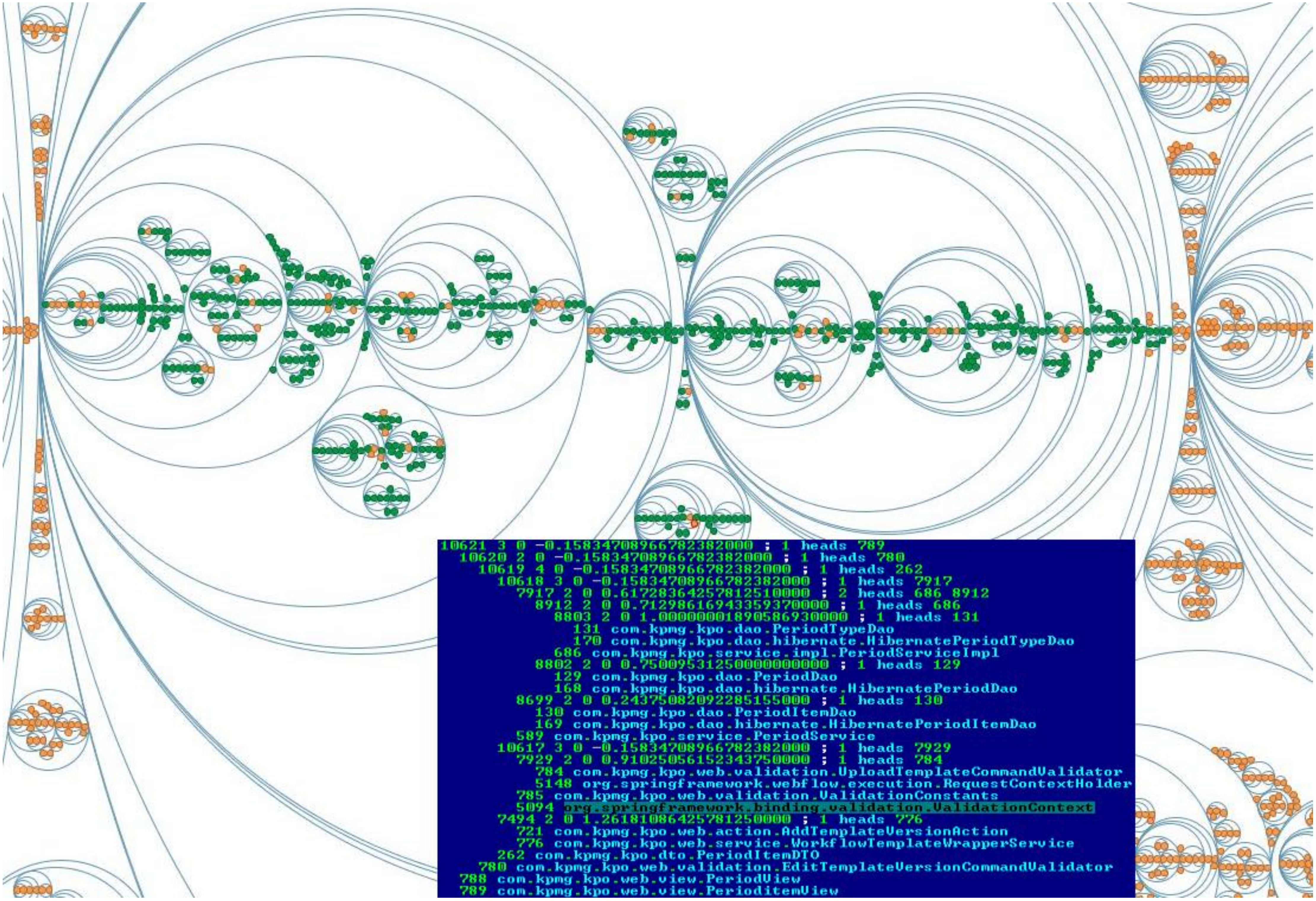
10621 3 0 -0.1583470896678238200 ; 1 heads 789
10620 2 0 -0.1583470896678238200 ; 1 heads 780
10619 4 0 -0.1583470896678238200 ; 1 heads 262
10618 3 0 -0.1583470896678238200 ; 1 heads 7917
7917 2 0 0.6172836425781251000 ; 2 heads 686 8912
8912 2 0 0.7129861694335937000 ; 1 heads 686
8803 2 0 1.0000000189058693000 ; 1 heads 131
131 com.kpmg.kpo.dao.PeriodTypeDao
170 com.kpmg.kpo.dao.hibernate.HibernatePeriodTypeDao
686 com.kpmg.kpo.service.impl.PeriodServiceImpl
8802 2 0 0.7500953125000000000 ; 1 heads 129
129 com.kpmg.kpo.dao.PeriodDao
168 com.kpmg.kpo.dao.hibernate.HibernatePeriodDao
8699 2 0 0.2437508209228515500 ; 1 heads 130
130 com.kpmg.kpo.dao.PeriodItemDao
169 com.kpmg.kpo.dao.hibernate.HibernatePeriodItemDao
589 com.kpmg.kpo.service.PeriodService
10617 3 0 -0.1583470896678238200 ; 1 heads 7929
7929 2 0 0.9102505615234375000 ; 1 heads 784
784 com.kpmg.kpo.web.validation.UploadTemplateCommandValidator
5148 org.springframework.webflow.execution.RequestContextHolder
785 com.kpmg.kpo.web.validation.ValidationConstants
5094 org.springframework.binding.validation.ValidationContext
7494 2 0 1.2618108642578125000 ; 1 heads 776
721 com.kpmg.kpo.web.action.AddTemplateVersionAction
776 com.kpmg.kpo.web.service.WorkflowTemplateWrapperService
262 com.kpmg.kpo.dto.PeriodItemDTO
780 com.kpmg.kpo.web.validation.EditTemplateVersionCommandValidator
788 com.kpmg.kpo.web.view.PeriodView
789 com.kpmg.kpo.web.view.PerioditemView